\documentclass[10pt,journal,compsoc]{IEEEtran}

\usepackage{epsfig}
\usepackage{graphicx, subfigure}
\usepackage{amsmath}
\usepackage{amssymb}
\usepackage{multirow}
\usepackage{bm,makecell,amsthm}
\usepackage{url}
\usepackage{algorithm}
\usepackage{algorithmic}
\usepackage{array}
\usepackage{fixltx2e}
\usepackage{dblfloatfix}
\usepackage{mathrsfs}
\usepackage{url,color}

\begin{document}

\renewcommand{\algorithmicrequire}{\textbf{Input:}}   
\renewcommand{\algorithmicensure}{\textbf{Output:}}  

\title{A Generalized Framework for Edge-preserving and Structure-preserving Image Smoothing}

\author{Wei~Liu, Pingping~Zhang, Yinjie Lei, Xiaolin~Huang, Jie~Yang and Michael Ng

\IEEEcompsocitemizethanks{\IEEEcompsocthanksitem Wei Liu and Michael Ng are with the Department of Mathematics, The University of Hong Kong, Hong Kong, China. Email: liuweicv@hku.hk, mng@maths.hku.hk

\IEEEcompsocthanksitem Pingping Zhang is with the School of Artificial Intelligence, Dalian University of Technology, Dalian, Liaoning, 116024, China. Email: zhpp@dlut.edu.cn

\IEEEcompsocthanksitem Yinjie Lei is with the School of Electronics and Information Engineering, Sichuan University, Chengdu, Sichuan, 610065, China. Email: yinjie@scu.edu.cn

\IEEEcompsocthanksitem Xiaolin Huang (corresponding author) and Jie Yang (corresponding author) are with the Institute of Image Processing and Pattrn Recognition \& Institute of Medical Robotics, Shanghai Jiao Tong University, Shanghai, 200240, China. Email: \{xiaolinhuang, jieyang\}@sjtu.edu.cn

\IEEEcompsocthanksitem This research is partly supported by National Key R$\&$D Program of China (No. 2019YFB1311503) and NSFC, China (No: 61876107, U1803261, 61977046), Shanghai Municipal Science and Technology Major Project (2021SHZDZX0102), the Fundamental Research Funds for the Central Universities (No. DUT20RC(3)083), HKRGC GRF (No. 12200317, 12300218, 12300519, 17201020).
}
}


\IEEEtitleabstractindextext{

\begin{abstract}
Image smoothing is a fundamental procedure in applications of both computer vision and graphics. The required smoothing properties can be different or even contradictive among different tasks. Nevertheless, the inherent smoothing nature of one smoothing operator is usually fixed and thus cannot meet the various requirements of different applications. In this paper, we first introduce the truncated Huber penalty function which shows strong flexibility under different parameter settings. A generalized framework is then proposed with the introduced truncated Huber penalty function. When combined with its strong flexibility, our framework is able to achieve diverse smoothing natures where contradictive smoothing behaviors can even be achieved. It can also yield the smoothing behavior that can seldom be achieved by previous methods, and superior performance is thus achieved in challenging cases. These together enable our framework capable of a range of applications and able to outperform the state-of-the-art approaches in several tasks, such as image detail enhancement, clip-art compression artifacts removal, guided depth map restoration, image texture removal, etc. In addition, an efficient numerical solution is provided and its convergence is theoretically guaranteed even the optimization framework is non-convex and non-smooth. A simple yet effective approach is further proposed to reduce the computational cost of our method while maintaining its performance. The effectiveness and superior performance of our approach are validated through comprehensive experiments in a range of applications. Our code is available at \emph{\tt\url{https://github.com/wliusjtu/Generalized-Smoothing-Framework}}.
\end{abstract}

}

\maketitle

\IEEEdisplaynontitleabstractindextext


\section{Introduction}
\label{SecIntroduction}

\IEEEPARstart{T}{he} key challenge of many tasks in both computer vision and graphics can be attributed to image smoothing. At the same time, the required smoothing properties can vary dramatically for different tasks. In this paper, depending on the required smoothing properties, we roughly classify a large number of applications into four groups.

Applications in the first group require the smoothing operator to smooth out small details while preserving strong edges, and the amplitudes of these strong edges can be reduced but the edges should be neither blurred nor sharpened. Representatives in this group are image detail enhancement and HDR tone mapping \cite{farbman2008edge, fattal2007multiscale, he2013guided}. Blurring edges can result in halos while sharpening edges will lead to gradient reversals \cite{farbman2008edge}.

\begin{figure}
  \centering
  \setlength{\tabcolsep}{0.5mm}
  \begin{tabular}{cc}
  \includegraphics[width=0.48\linewidth]{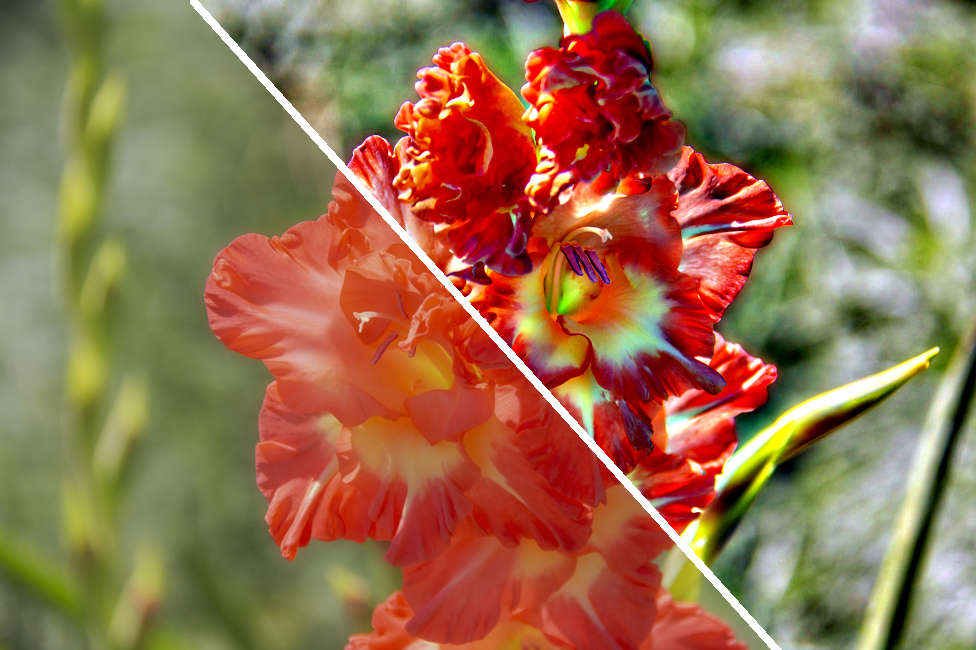}&
  \includegraphics[width=0.48\linewidth]{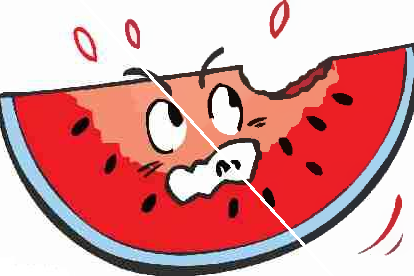}\\
  (a) & (b)\\

  \includegraphics[width=0.48\linewidth]{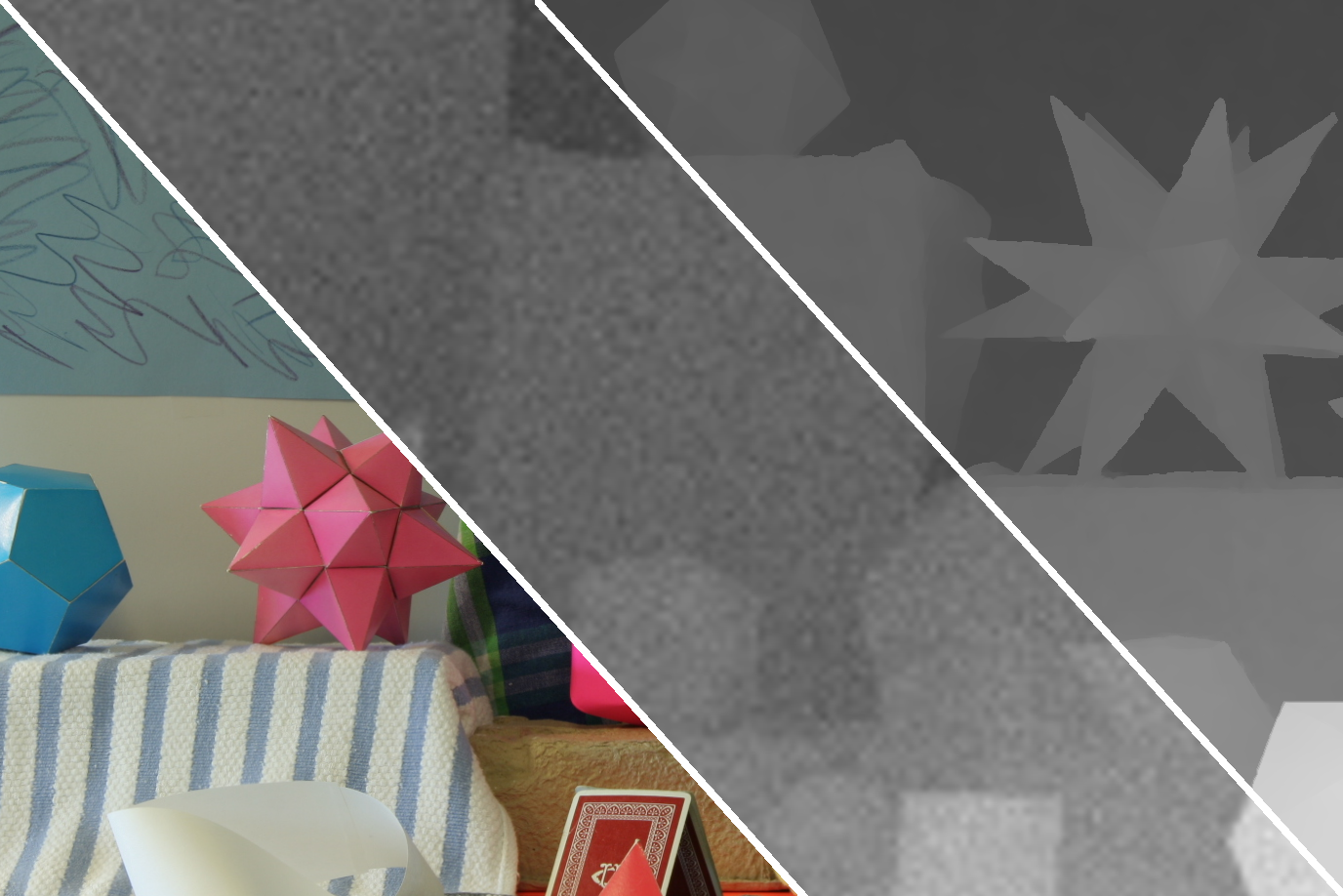}&
  \includegraphics[width=0.48\linewidth]{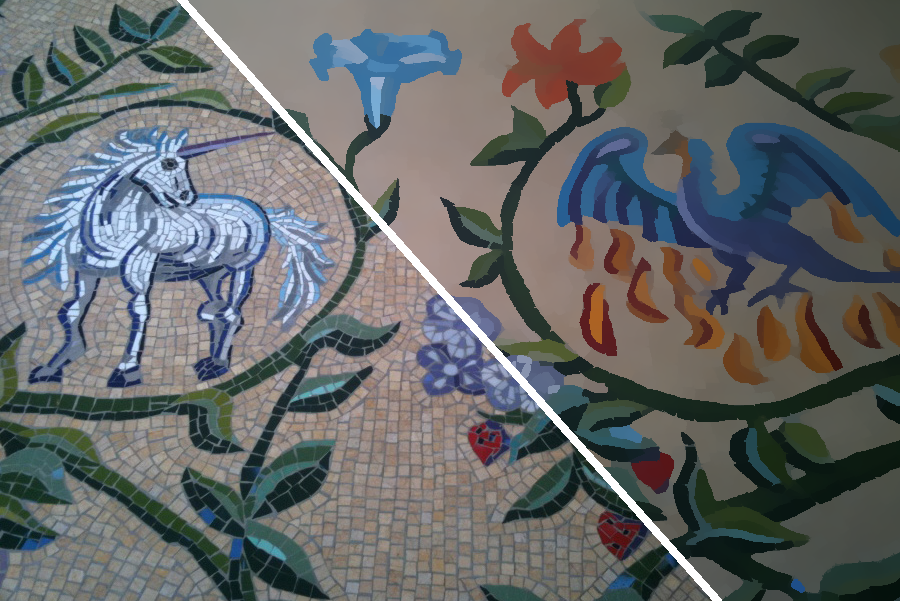}\\
  (c) & (d)\\
  \end{tabular}
  \caption{Our method is capable of (a) image detail enhancement, (b) clip-art compression artifacts removal, (c) guided depth map upsampling and (d) image texture removal. These applications are representatives of edge-preserving and structure-preserving image smoothing tasks which require contradictive smoothing properties.}\label{FigCover}
\end{figure}

The second group includes tasks like clip-art compression artifacts removal \cite{nguyen2015fast, xu2011image, wang2006deringing}, image abstraction and pencil sketch production \cite{xu2011image}. In contrast to the ones in the first group, these tasks require to smooth out small details while sharpening strong edges. This is because edges can be blurred in the compressed clip-art image and they need to be sharpened when the image is recovered, an example is illustrated in Fig.~\ref{FigCover}(b). Sharper edges can produce better visual quality in image abstraction and pencil sketch. At the same time, the amplitudes of strong edges are not allowed to be reduced in these tasks.

\begin{figure*}
\centering
\setlength{\tabcolsep}{0.5mm}
\begin{tabular}{cccc}
\includegraphics[width=0.22\linewidth]{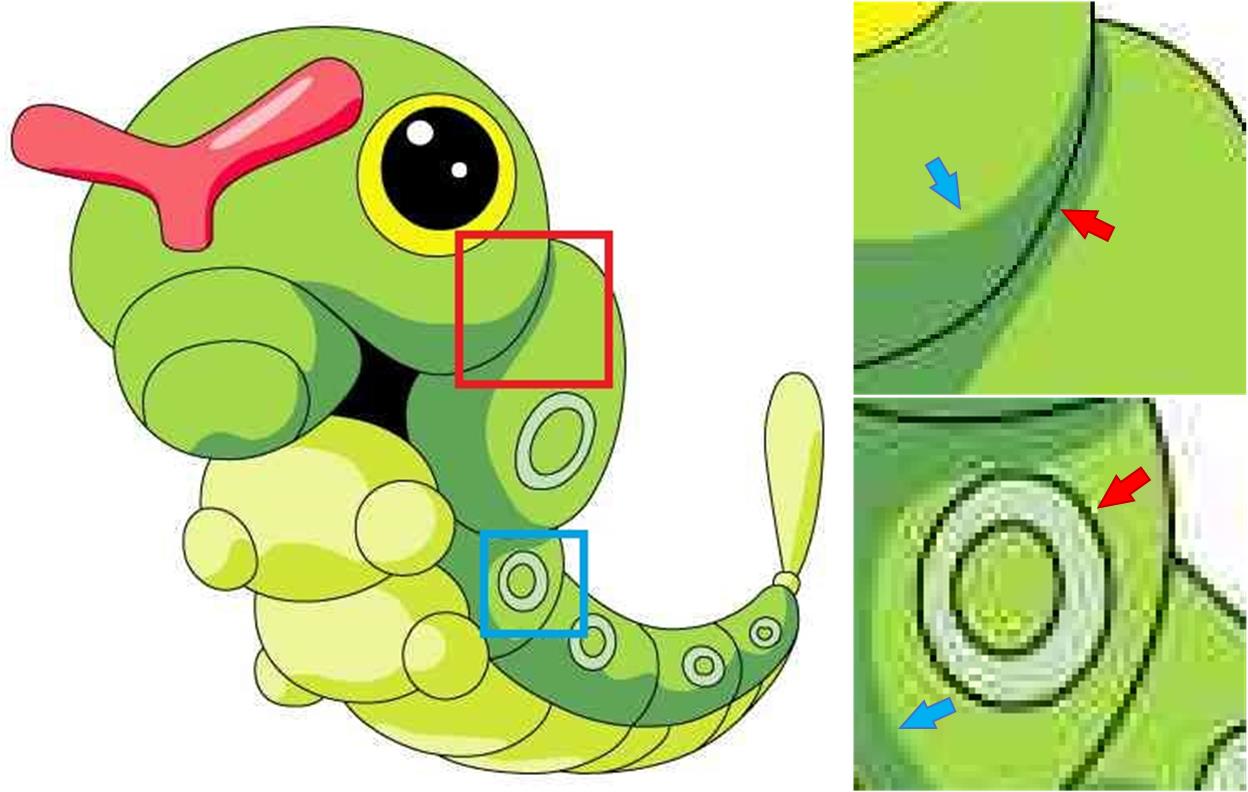}&
\includegraphics[width=0.22\linewidth]{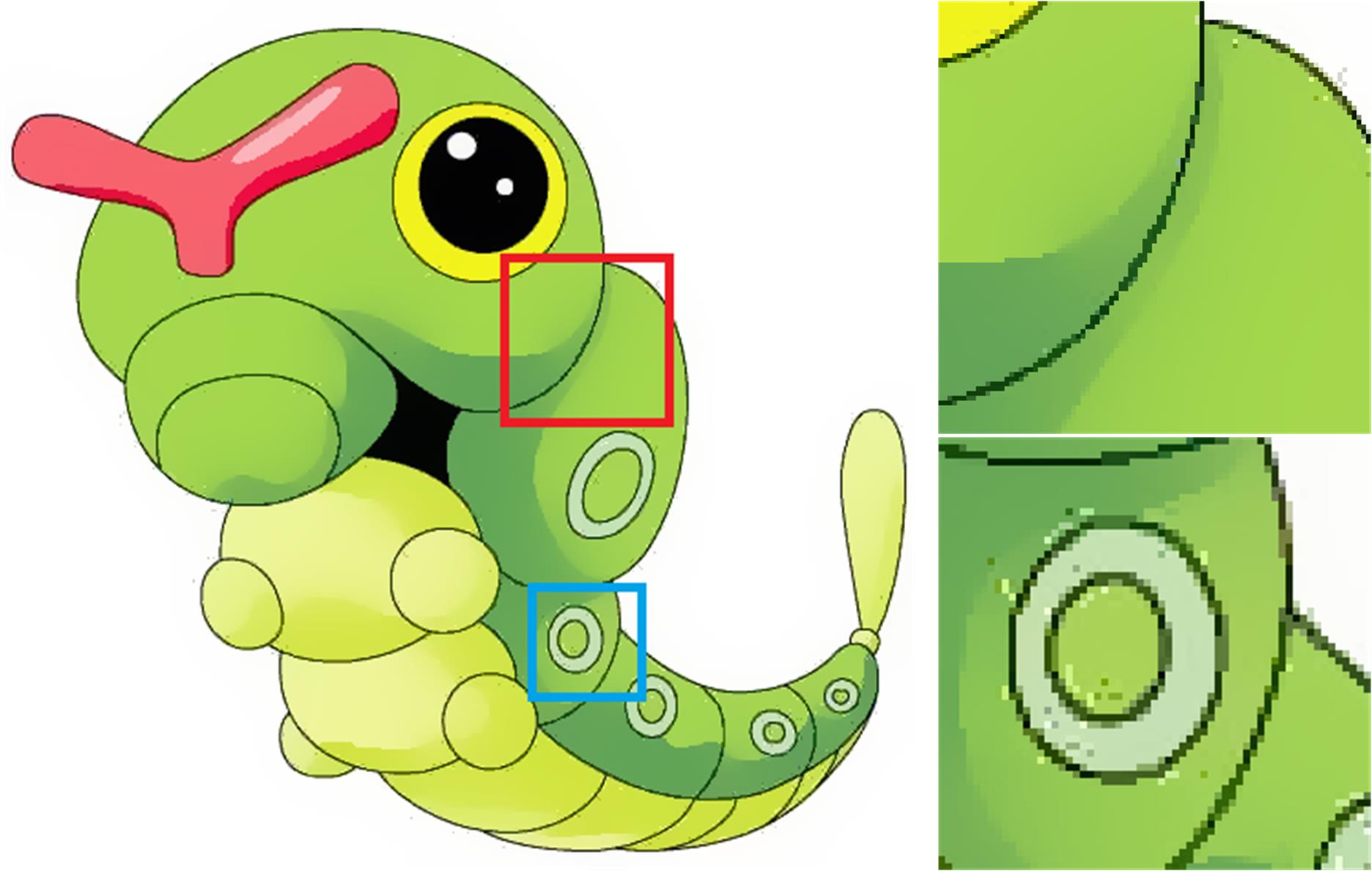}&
\includegraphics[width=0.22\linewidth]{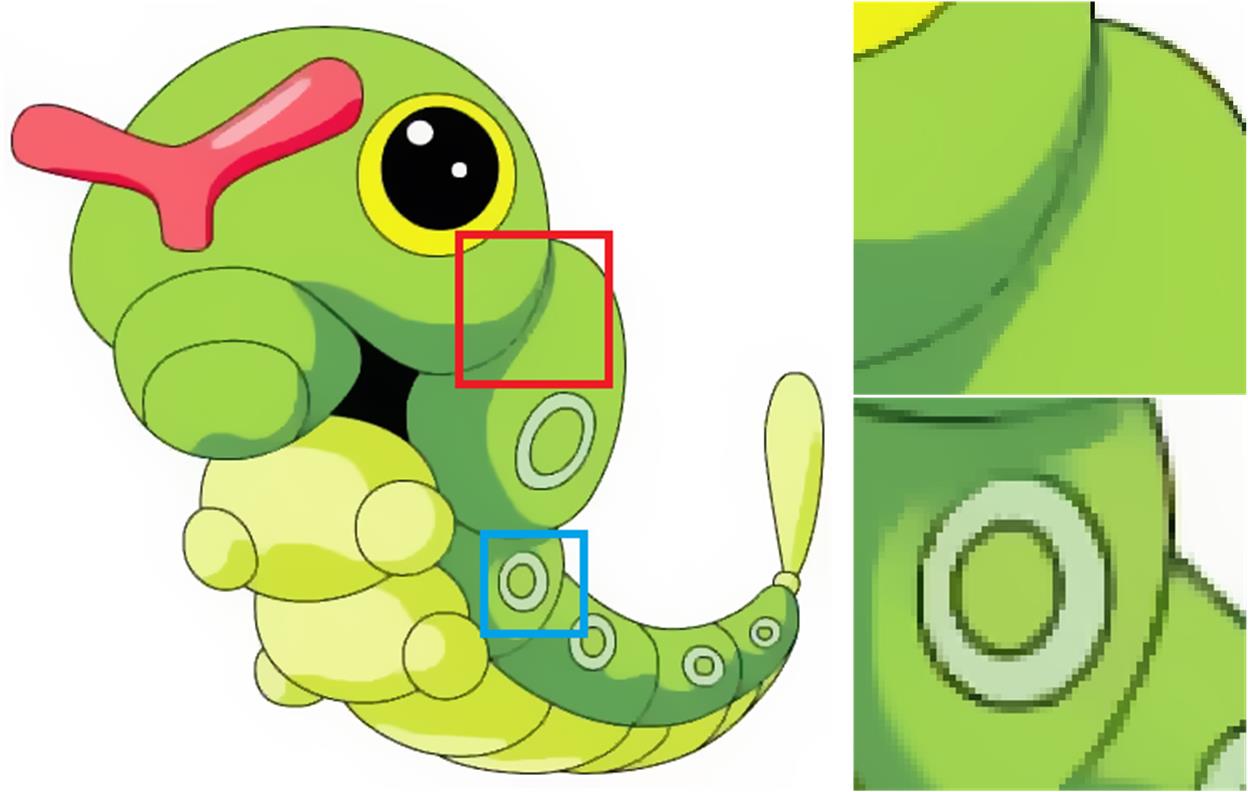}&
\includegraphics[width=0.22\linewidth]{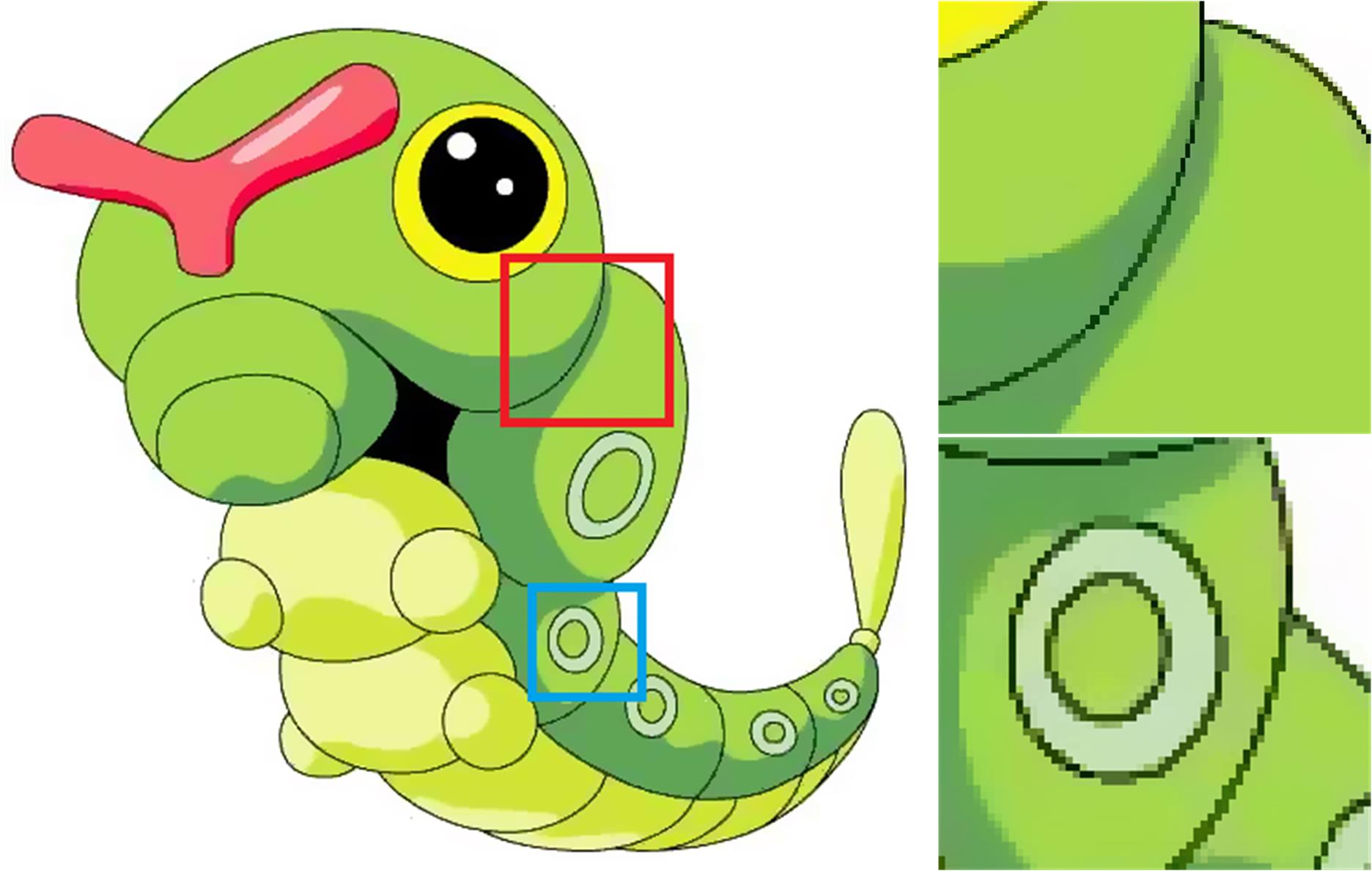}\\
(a) input & (b) SD filter & (c) RTV & (d) ours
\end{tabular}
\caption{Clip-art compression artifacts removal. (a) Input compressed JPEG image. Smoothing result of (b) edge-preserving smoother SD filter \cite{ham2018robust}, (c) structure-preserving smoother RTV smoothing \cite{xu2012structure} and (d) our method of the simultaneous edge-preserving and structure-preserving mode. Pay attention to the difference between the ``black lines'' labeled with the red arrows (small structures with strong edges) and  the ``shades'' labeled with the blue arrows (large structures with weak edges) in different results. }\label{FigEdgeAndStructureAwareSmooth}
\end{figure*}

Guided image filtering, such as guided depth map upsampling \cite{park2011high, ferstl2013image, liu2017robust} and flash/no flash filtering \cite{kopf2007joint, petschnigg2004digital}, is categorized into the third group. The structure inconsistency between the guidance image and the target image, which can cause blurring edges and texture copy artifacts in the smoothed image \cite{ham2018robust, liu2017robust}, should be properly handled by the specially designed smoothing operator. They also need to sharpen edges in the smoothed image due to the reason that low-quality capture of depth maps and the noise in the no-flash images can lead to blurred edges, see Fig.~\ref{FigCover}(c) for example.

Tasks in the fourth group require to smooth the image in a scale-aware manner, e.g., image texture removal \cite{xu2012structure, zhang2014rolling, cho2014bilateral}. This kind of tasks require to smooth out small structures even when they contain strong edges, while large structure should be properly preserved even the edges are weak, Fig.~\ref{FigCover}(d) shows an example. This is totally different from that in the above three groups where they all aim at preserving strong edges.

Generally, the smoothing procedures in the first to the third groups are usually considered as \emph{edge-preserving image smoothing} since they try to preserve salient edges, while the smoothing processes in the fourth group are classified as \emph{structure-preserving image smoothing} because they aim at preserving salient structures.

A diversity of smoothing operators have been proposed for various tasks in the literature. Generally, each of them is designed to meet the requirements of certain applications, and its inherent smoothing nature is usually fixed. Therefore, there is seldom any smoothing operator that can meet all the smoothing requirements of the above four groups. For example, the $L_0$ norm smoothing \cite{xu2011image} can sharpen strong edges and is suitable for clip-art compression artifacts removal, however, this will lead to gradient reversals in image detail enhancement and HDR tone mapping. The weighted least squares (WLS) smoothing \cite{farbman2008edge} performs well in image detail enhancement and HDR tone mapping, but it is not capable of sharpening edges. In a higher view, edge-preserving smoothing operators are also not directly applicable to structure-preserving tasks. Thus, designing a smoothing operator that is capable of these different smoothing properties still remains a challenging problem.

Besides the challenge mentioned above, there are also challenging cases that cannot be well handled by either edge-preserving smoothing or structure-preserving smoothing. Fig.~\ref{FigEdgeAndStructureAwareSmooth} illustrates an example of clip-art compression artifacts removal. In this example, the heavy compression artifacts lead to some small structures with large-amplitude edges. At the same time, both the ``black lines'' labeled with the red arrows (small structures with strong edges) and the ``shades'' labeled with the blue arrows (large structures with weak edges) need to be preserved, and the edges should also be sharpened. The challenges of this case are twofold. On the one hand, if we remove the artifacts in an edge-preserving manner, then the weak edges around the ``shades'' will also be smoothed as their amplitudes are small. Fig.~\ref{FigEdgeAndStructureAwareSmooth}(b) illustrates the result of the static/dynamic (SD) filter \cite{ham2018robust} which is an edge-preserving smoother. The edges of the ``black lines'' are sharpened in the result, but the edges of the ``shades'' are blurred while some large-amplitude artifacts still retain in the result. On the other hand, if the artifacts are removed with a structure-preserving smoother, the ``black lines'' may also be removed as their structures are small. Fig.~\ref{FigEdgeAndStructureAwareSmooth}(c) shows the result of relative total variation (RTV) smoothing \cite{xu2012structure} which is a structure-preserving smoother. Although the ``shades'' are preserved, some ``black lines'' are also removed together with the artifacts. Besides, the edges are not sharpened in the result. In contrast, besides the various edge-preserving and structure-preserving smoothing properties mentioned in previous paragraphs, our method is also able to yield simultaneous edge-preserving and structure-preserving smoothing behavior which is seldom achieved by previous approaches. This blended smoothing property can enjoy the advantages of both edge-preserving smoothing and structure-preserving smoothing. Fig.~\ref{FigEdgeAndStructureAwareSmooth}(d) shows the result of our method where both the ``black lines'' and the ``shades'' are preserved, and the edges are also sharpened. As we will show in Sec.~\ref{SecPropertyAnalysis}, this blended simultaneous edge-preserving and structure-preserving smoothing property can act as a promising alternative for handling the tasks in the second group and the third group for better performance.

In this paper, we propose a new smoothing operator. In contrast to most of the smoothing operators in the literature, it can achieve various smoothing behaviors and is able to handle all the challenges mentioned above. The main contributions of this paper are as follows:

\begin{itemize}
  \item[1.] We introduce the \emph{truncated Huber penalty} function which has seldom been used in image smoothing. By varying the parameters, it shows strong flexibility.

  \item[2.] A generalized framework is proposed with the truncated Huber penalty function. When combined with the strong flexibility of the truncated Huber penalty function, our model can achieve various smoothing behaviors. We show that it is able to handle the tasks in the four groups mentioned above. Besides, our model is also able to achieve simultaneous edge-preserving and structure-preserving smoothing property which can yield superior performance over previous methods in challenging cases (e.g., Fig.~\ref{FigEdgeAndStructureAwareSmooth}). All these are seldom achieved by previous smoothing operators in the literature. We also show that our method is able to achieve state-of-the-art performance in many tasks.

  \item[3.] An efficient numerical solution to the proposed optimization framework is provided. Its convergence is theoretically guaranteed even the framework is non-convex and non-smooth. A simple yet effective approach is further proposed to reduce the computational cost of our method while maintaining its performance.
\end{itemize}

This manuscript is the extension of its conference version \cite{liu2020generalized} with the following differences: (1) We provide more detailed analysis of the introduced penalty function in Sec.~\ref{SecTruncatedHuber} and the proposed model in Sec.~\ref{SecPropertyAnalysis}. (2) The concept of ``dilated neighborhood'' is further introduced in Sec.~\ref{SecDilatedNeighbor} to reduce the computational cost of our method while maintaining its performance. (3) We show more applications and experimental results in Sec.~\ref{SecExperiments}. More quantitative evaluation is also provided to validate the effectiveness of our method.

The rest of this paper are organized as follows: Sec.~\ref{SecRelatedWork} describes the related work of our method. Sec.~\ref{SecOurMethod} is devoted to our approach including the introduction of the truncated Huber penalty function, the definition of our model, its numerical solution and further property analysis of the model. Our method is compared against many state-of-the-art approaches in a range of applications in Sec.~\ref{SecExperiments}. We draw the conclusion and summarize the limitations of this paper in Sec.~\ref{SecConclusion}.

\section{Related Work}
\label{SecRelatedWork}

Image smoothing has been a well-studied research filed. Tremendous smoothing operators have been proposed in recent decades. In terms of edge-preserving smoothing, bilateral filter (BLF) \cite{tomasi1998bilateral} is the early work that has been used in various tasks such as image detail enhancement \cite{fattal2007multiscale}, HDR tone mapping \cite{durand2002fast}, etc. However, it is prone to produce results with gradient reversals and halos \cite{farbman2008edge}. Gastal et~al. \cite{gastal2012adaptive} proposed adaptive manifold filter (AMF) as an alternative of BLF to handle high-dimension data such RGB color images. They further introduced domain transform filter (DTF) \cite{gastal2011domain} for fast image processing. Similar to BLF, these approaches also share similar problems of resulting in gradient reversals and halos. Guided filter (GF) \cite{he2013guided} can produce results free of gradient reversals but halos can still exist. The WLS smoothing \cite{farbman2008edge} forms image smoothing as a global optimization problem. It has been one of the well-known milestones due to its superior performance in handling the gradient reversals and halos. However, as noted by Hessel et~al. \cite{hessel2018quantitative}, the WLS smoothing is prone to result in compartmentalization artifacts. In contrast, our method provides a promising alternative that can properly eliminate the compartmentalization artifacts as well as gradient reversals and halos. The $L_0$ norm smoothing proposed by Xu et~al. \cite{xu2011image} is able to smooth out weak edges while sharpening strong edges, which can be applied to the tasks in the second group. Its shortage is that low-amplitude structures in the image can also be eliminated. To handle the structure inconsistency problem, Shen et~al. \cite{shen2015mutual} proposed to perform mutual-structure joint filtering. They also explored the relation between the guidance image and the target image via optimizing a scale map \cite{shen2015multispectral}, however, additional processing was adopted for structure inconsistency handling. Ham et~al. \cite{ham2018robust} proposed to handle the structure inconsistency by combining a static guidance weight with a Welsch's penalty \cite{holland1977robust} regularized smoothness term, which led to a static/dynamic (SD) filter. Gu et~al. \cite{gu2017learning} presented a weighted analysis representation model for guided depth map enhancement. They also proposed to smooth images by layer decomposition, and different sparse representation models were adopted for different layers \cite{gu2017joint}.

\begin{figure*}[!t]
\centering
\setlength{\tabcolsep}{0.25mm}
\begin{tabular}{cccc}
\includegraphics[width=0.22\linewidth]{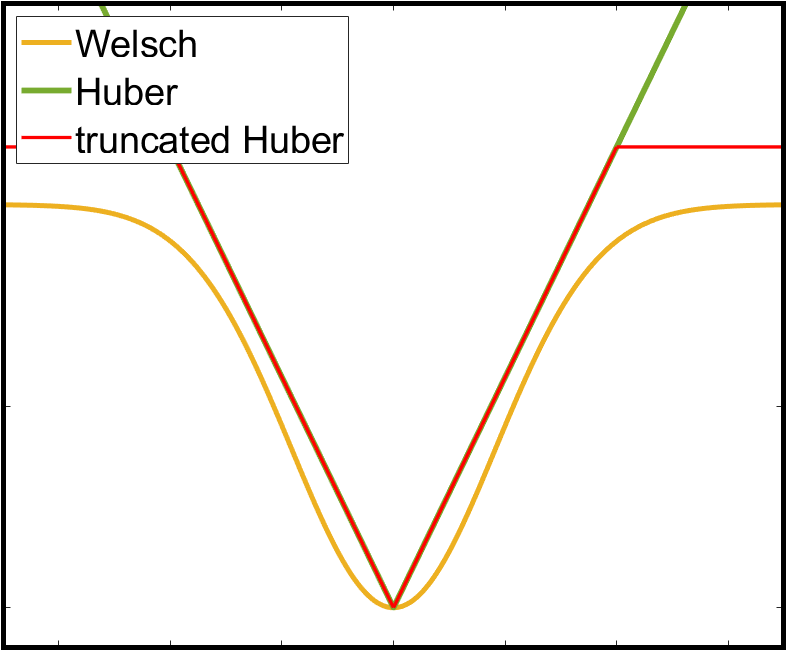}&
\includegraphics[width=0.219\linewidth]{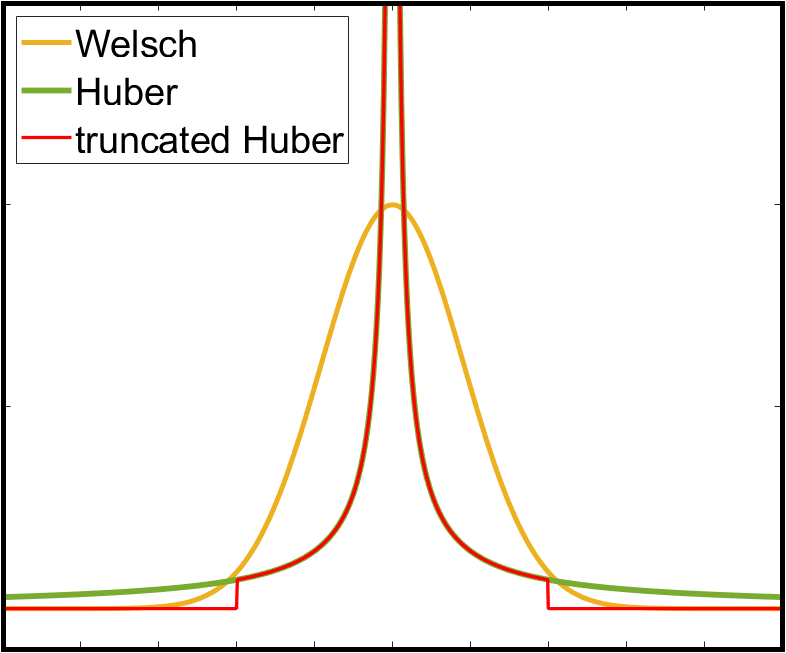}&
\includegraphics[width=0.22\linewidth]{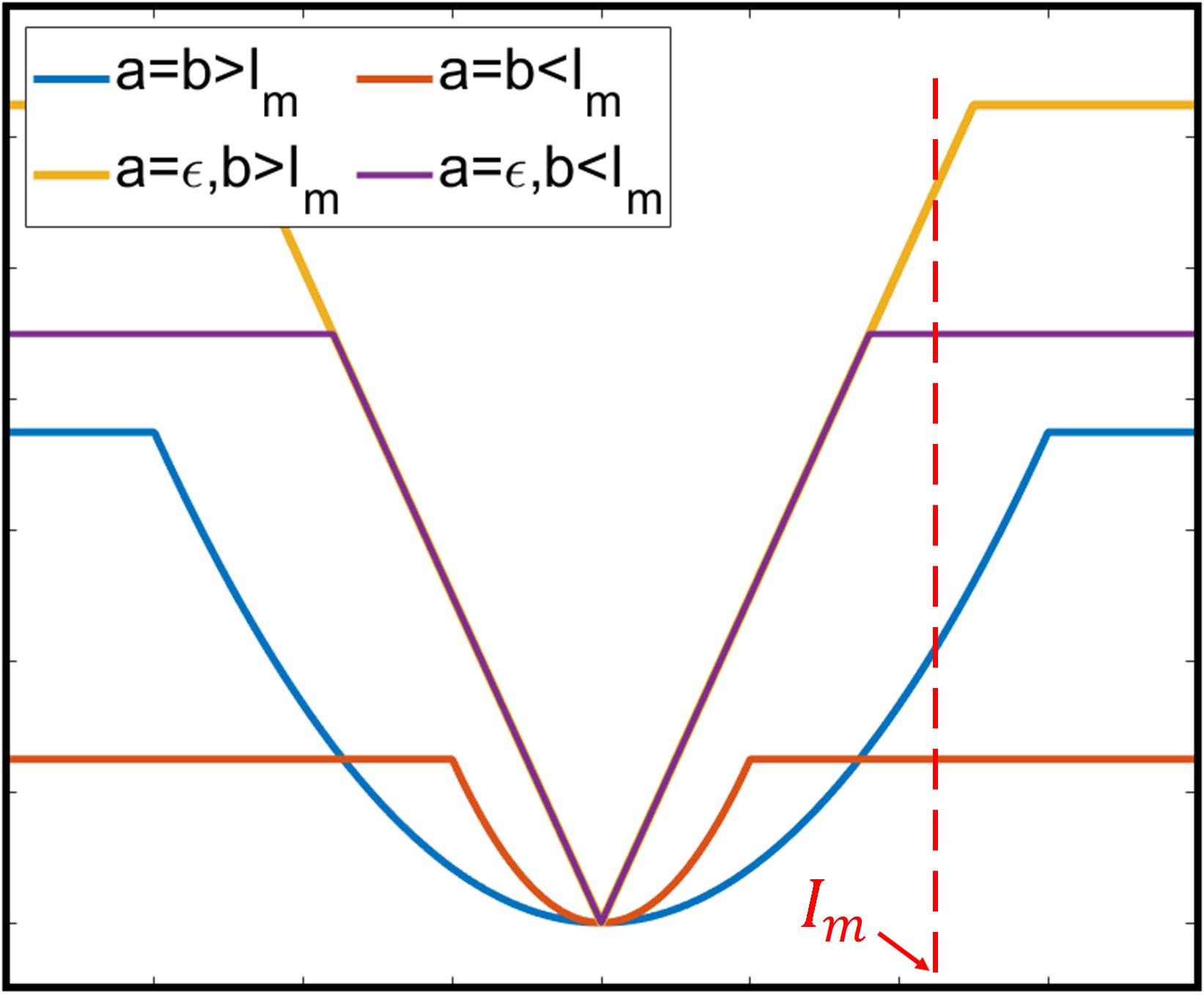}&
\includegraphics[width=0.2205\linewidth]{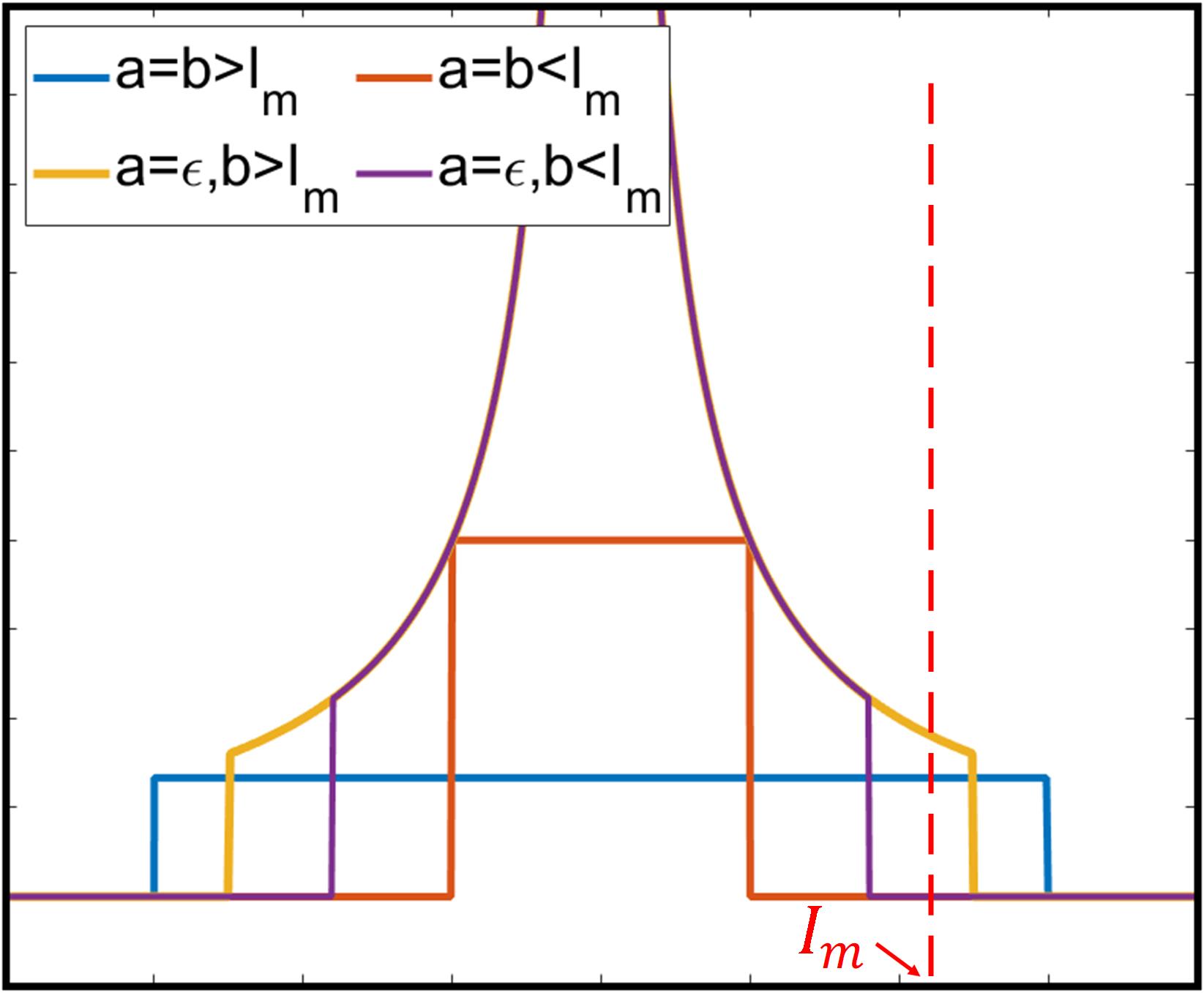}\\
 (a) penalty function & (b) edge stopping function & (c) penalty function & (d) edge stopping function\\
\end{tabular}
\caption{Plots of (a) different penalty functions and (c) the truncated Huber penalty function under different parameter settings. Their corresponding edge stopping functions are plotted in (b) and (d).}\label{FigPenaltyComp}
\end{figure*}

\begin{figure*}[!t]
\centering
\setlength{\tabcolsep}{0.25mm}
\begin{tabular}{cccccc}
\includegraphics[width=0.16\linewidth]{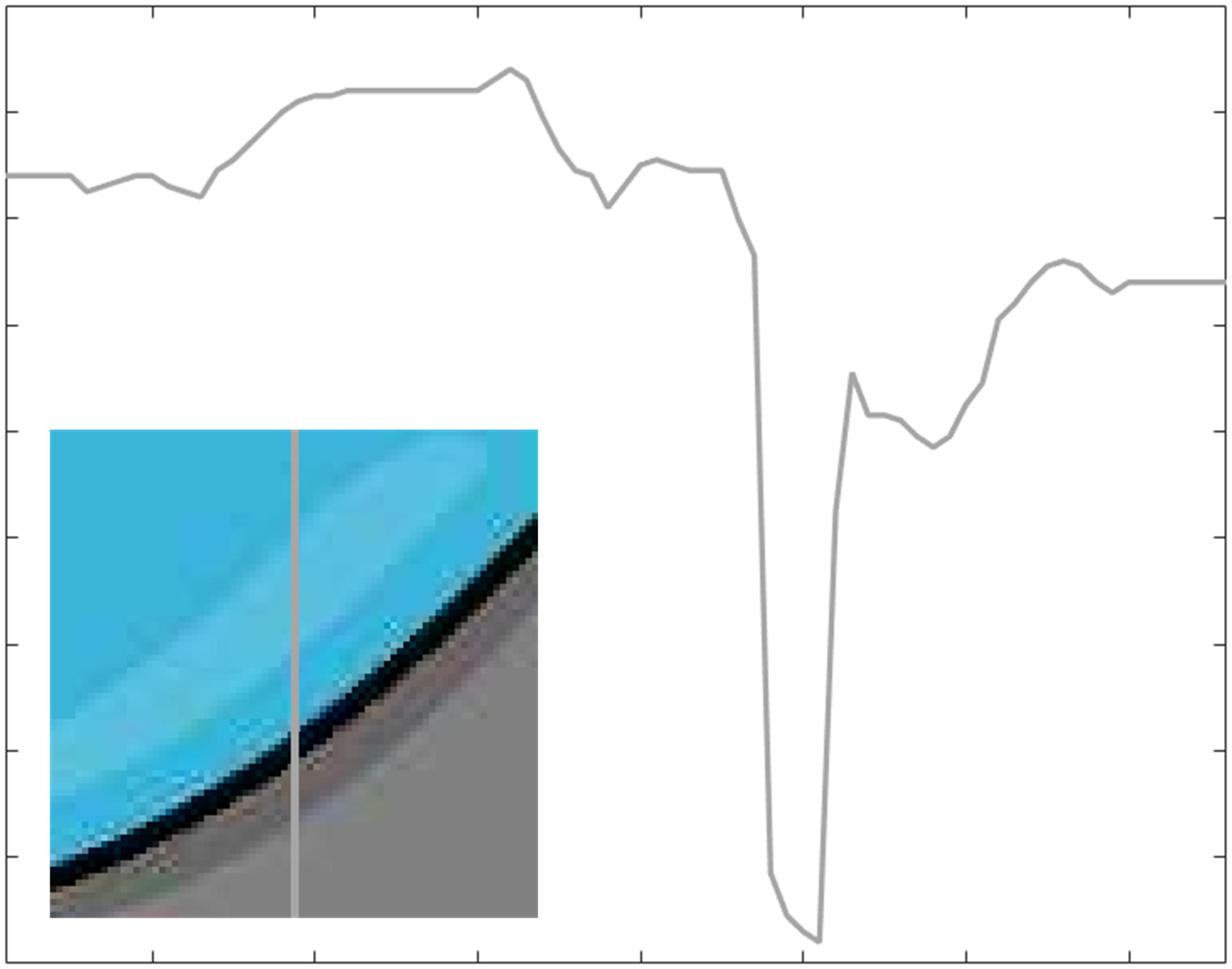}&
\includegraphics[width=0.16\linewidth]{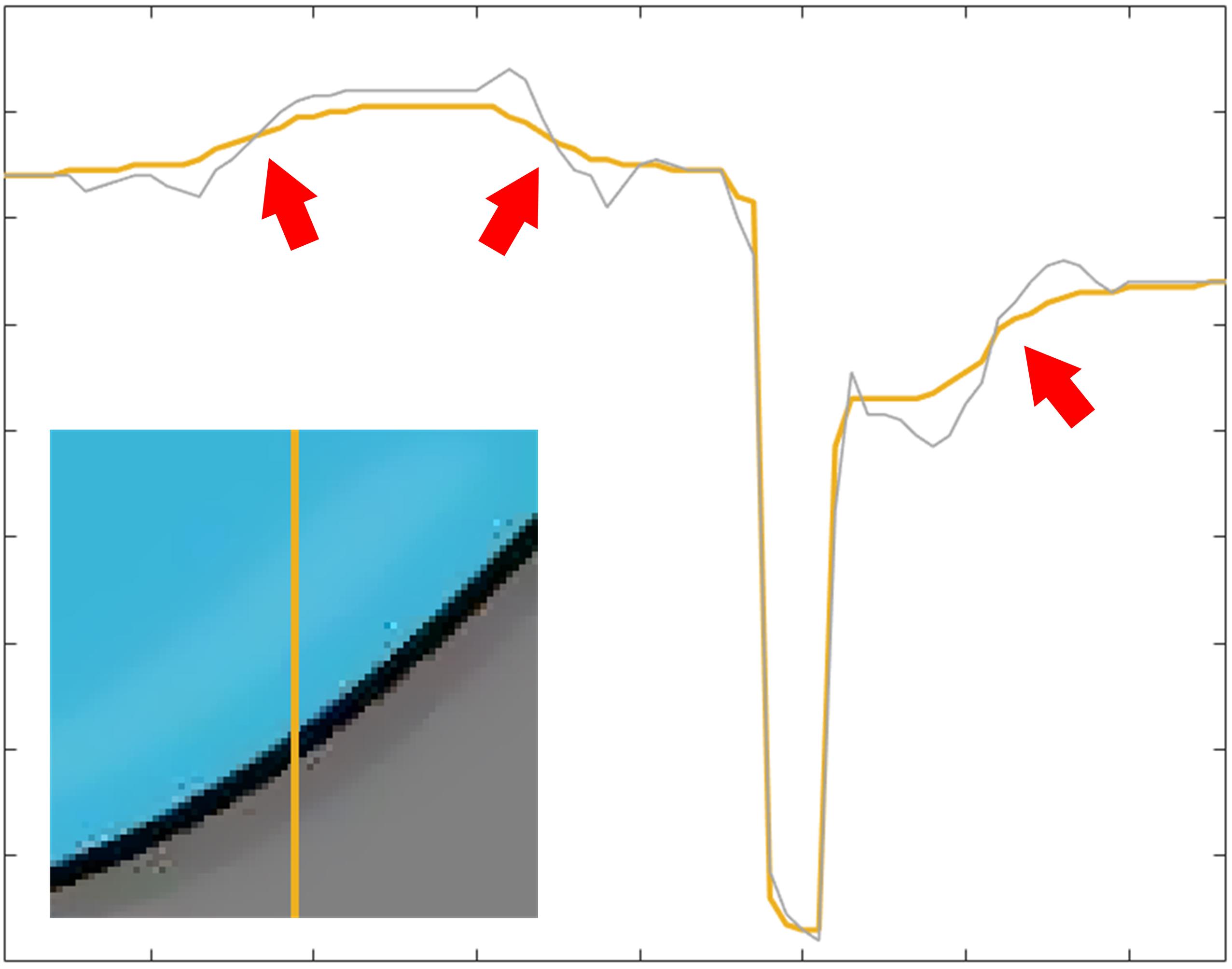}&
\includegraphics[width=0.16\linewidth]{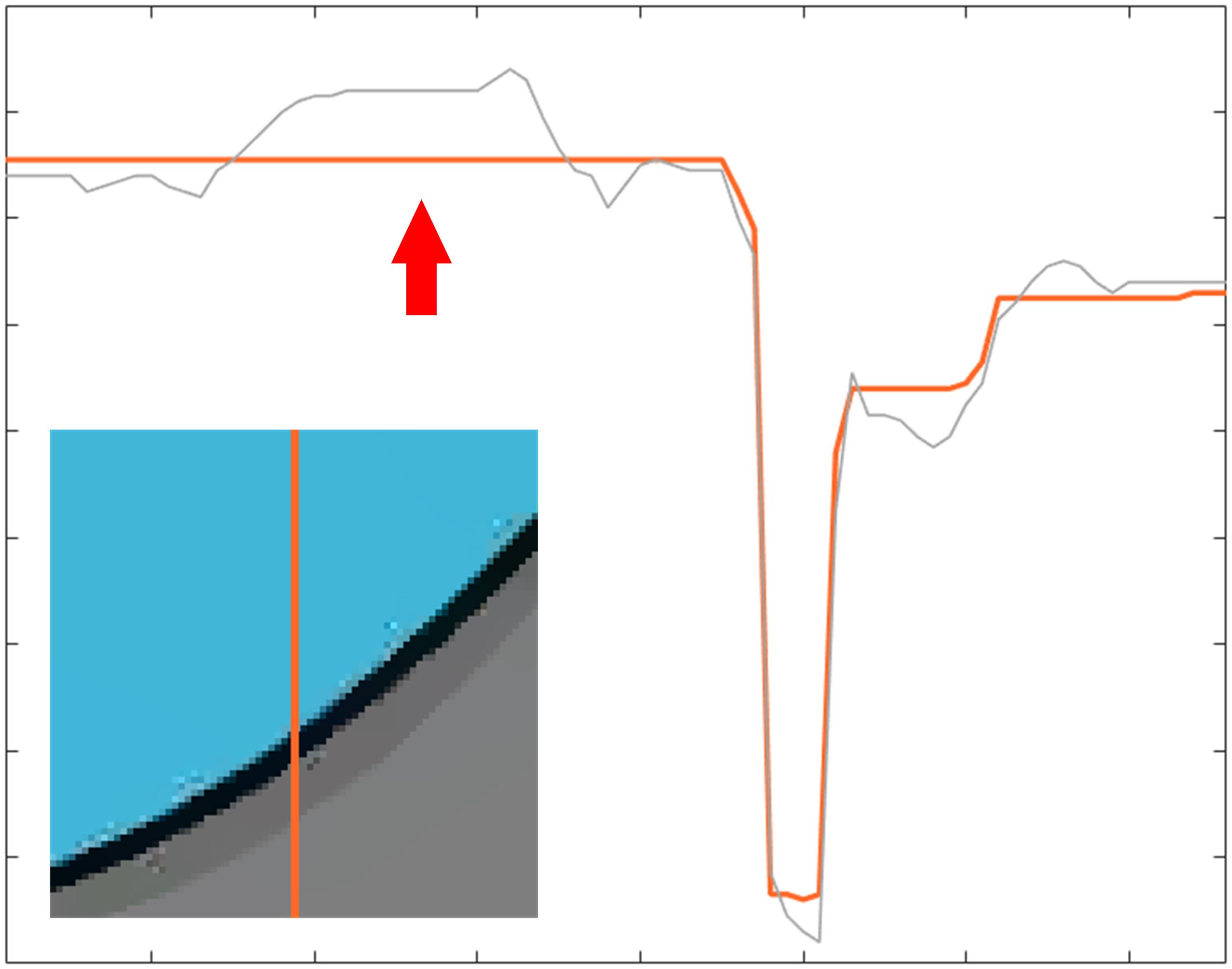}&
\includegraphics[width=0.16\linewidth]{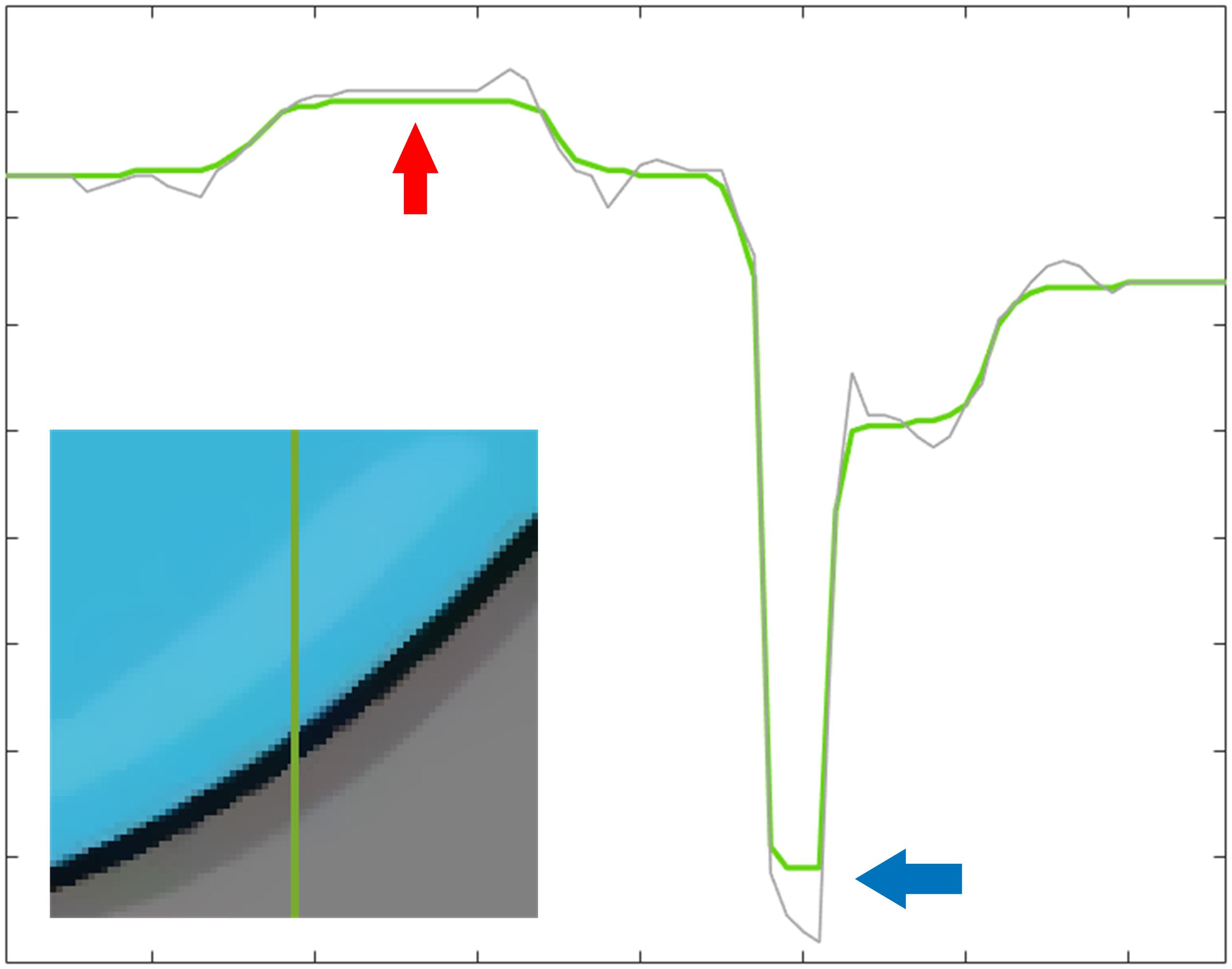}&
\includegraphics[width=0.16\linewidth]{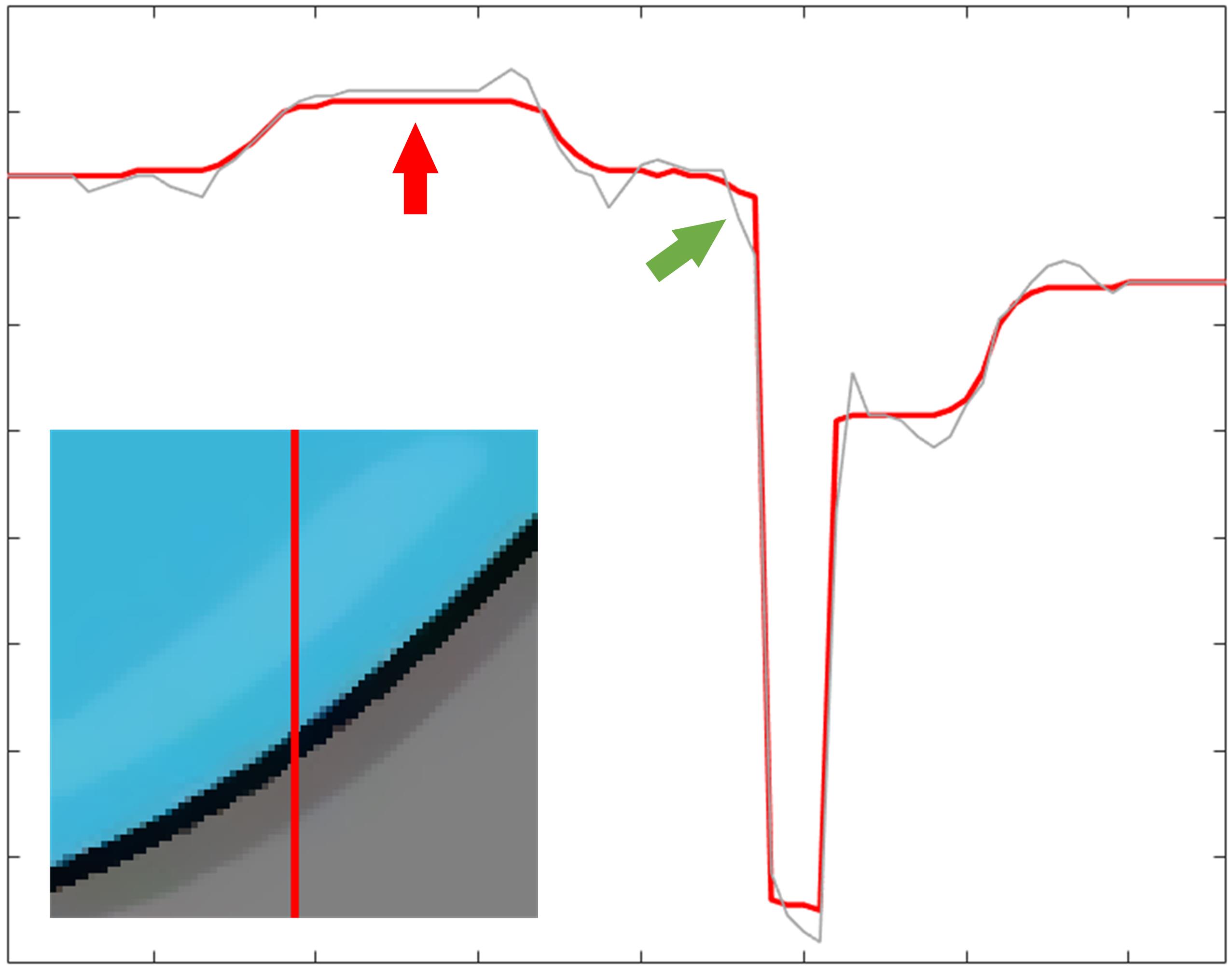}&
\includegraphics[width=0.16\linewidth]{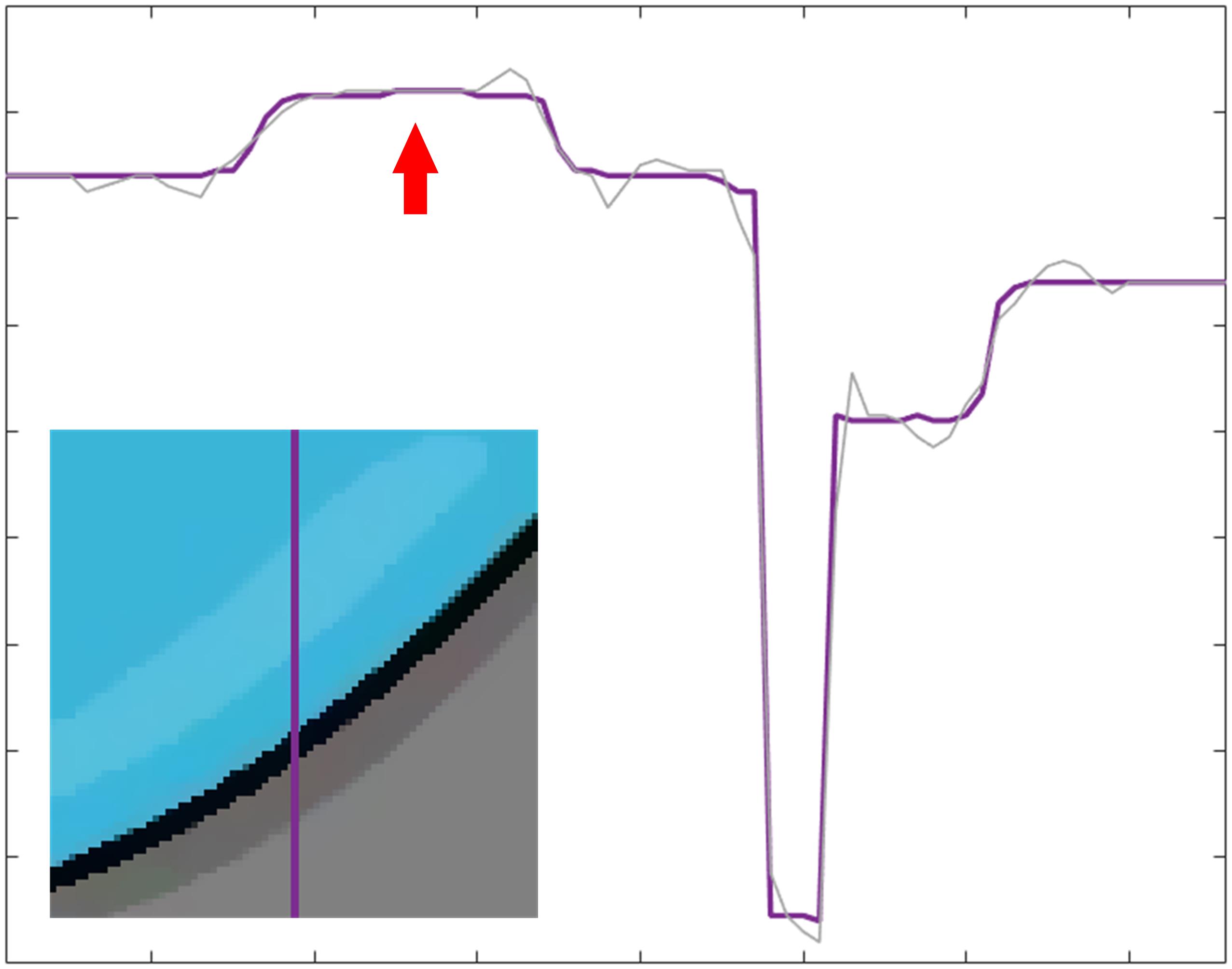}\\
 (a) input & (b) SD filter & (c) $L_0$ norm & (d) ours (EP-2) & (e) ours (EP-2) & (f) ours (EP$\&$SP)\\
\end{tabular}
\caption{1D illustration of real smoothing results.  The column labeled with the line in each image is plotted. (a) Input. Smoothing result of (b) SD filter \cite{ham2018robust} which adopts the Welsch's penalty function for regularization, (c)  $L_0$ norm smoothing \cite{xu2011image} which approximates $L_0$ norm with a series of truncated $L_2$ norms for regularization, (d) our method of the EP-2 mode which adopts the truncated Huber penalty in Eq.~(\ref{EqTruncatedHuber}) ($a=\epsilon, b>I_m$) for regularization, (e) our method of the EP-2 mode which adopts the truncated Huber penalty in Eq.~(\ref{EqTruncatedHuber}) ($a=\epsilon, b<I_m$) for regularization and (f) our method of the EP$\&$SP mode which adopts the truncated Huber penalty in Eq.~(\ref{EqTruncatedHuber}) ($a=\epsilon, b<I_m$) for both regularization and data fidelity.}\label{FigNormMotivation}
\end{figure*}

In terms of structure-preserving smoothing, Zhang et~al. \cite{zhang2014rolling} proposed to smooth structures of different scales with a rolling guidance filter (RGF). Cho et~al. \cite{cho2014bilateral} modified the original BLF with local patch-based analysis of texture features and obtained a bilateral texture filter (BTF) for image texture removal. Karacan et~al. \cite{karacan2013structure} proposed to smooth image textures by making use of region covariances that captured local structure and textural information. Xu et~al. \cite{xu2012structure} adopted the relative total variation (RTV) as a prior to regularize the texture smoothing procedure. Chen et~al. \cite{chan2005aspects} proved that the TV-$L_1$ model \cite{chan2005aspects, nikolova2004variational} could smooth images in a scale-aware manner, which is ideal for structure-preserving smoothing, e.g., image texture removal \cite{aujol2006structure, buades2010fast}.

Most of the approaches mentioned above are limited to a few applications because their inherent smoothing natures are usually fixed. In contrast, our method proposed in this paper can have strong flexibility in achieving various smoothing behaviors, which enables wider applications of our method than most of them. In addition, our method is also able to achieve the smoothing behavior that is seldom achieved by previous approaches. This makes our method able to better handle the challenges in many tasks. We show that our method can show superior performance over these methods in several applications that they are specially designed for.

In recent years, a large majority of deep learning based approaches have also been proposed \cite{chen2017fast, gharbi2017deep, gharbi2015transform, xu2015deep, isola2017image, liu2016learning}. These methods adopt neural network architectures to learn the smoothing behaviors of different existing filters. However, their main drawback is that different models need to be trained separately for different filters. Even for a given filter, different models also need to be trained separately for different parameter settings of the filter. Thus, the deep learning based methods are not easy to tune the parameters. Chen et~al. \cite{chen2017fast} and Fan et~al. \cite{fan2018decouple} tried to make their methods tunable of the parameters, however, their methods only explored the tunability of one parameter while the other parameters need to be fixed.

\section{Our Approach}
\label{SecOurMethod}

The first challenge raised in Sec.~\ref{SecIntroduction} is that we need a switch which can control whether to sharpen edges or not. This can be achieved by adopting a proper penalty function to regularize the output. However, a penalty function with such kind of ``switch'' is seldom introduced in the literature. A widely utilized one to sharpen edges is the Welsch's penalty function \cite{holland1977robust} which was also adopted in the recent proposed SD filter \cite{ham2018robust} to produce sharp edges. The shortage of the Welsch's penalty function is that it is close to $L_2$ norm when the input is small, as illustrated in Fig.~\ref{FigPenaltyComp}(a). This means it cannot properly preserve week edges. An example is shown in Fig.~\ref{FigNormMotivation}(b) where the week edges labeled with the red arrows are largely smoothed. The gradient $L_0$ norm smoothing proposed by Xu et~al. \cite{xu2011image} is another well-known method that is able to sharpen salient edges. However, as they utilize a series of truncated $L_2$ norms to gradually approximate the $L_0$ norm, weak edges can seldom be preserved in their results, as shown in Fig.~\ref{FigNormMotivation}(c).

\subsection{Truncated Huber Penalty Function}
\label{SecTruncatedHuber}
Beside there is seldom a ``switchable'' penalty function, the current widely used penalty functions for sharpening edges do not work well to preserve weak edges as mentioned above. To overcome these shortages, we introduce the truncated Huber penalty function which is defined as:
\begin{eqnarray}\label{EqTruncatedHuber}
{ h_T(x)=\left\{\begin{array}{r}
   h(x), \ \ \ \ \ |x|\leq b\\
   b -\frac{a}{2}, \ \ \ \ |x|>b
  \end{array}\right.
  \text{s.t.} \ \ \ a \leq b,
}
\end{eqnarray}
where $a,b$ are constants. $h(\cdot)$ is the Huber penalty function \cite{huber1964robust} defined as:
\begin{eqnarray}\label{EqHuber}
{
  h(x)=\left\{\begin{array}{r}
   \frac{1}{2a}x^2, \ \ \ \ \ \ |x|<a\\
   |x|-\frac{a}{2}, \ \ \ |x|\geq a
  \end{array}\right.,
}
\end{eqnarray}
$h_T(\cdot)$ and $h(\cdot)$ are plotted in Fig.~\ref{FigPenaltyComp}(a) with $a=\epsilon$ where $\epsilon$ is a sufficient small value (e.g., $\epsilon=10^{-3}$).

The truncation in Eq.~(\ref{EqTruncatedHuber}) enables $h_T(\cdot)$ to sharpen edges. This can be better understood through the concept of edge stopping functions \cite{perona1990scale, black1998robust}. The edge stopping function of a penalty function $\rho(x)$ is defined as $\varphi(x)=\frac{\rho'(x)}{x}$ where $\rho'(x)$ is the derivative of $\rho(x)$ with respect to $x$. A larger value of $\varphi(x)$ means that $\rho(x)$ has larger penalty on the input $x$ and vice versa. This concept is not directly applicable to $h_T(x)$ because it is not differentiable at $x=b$. However, if we define the value of $h'_T(b)$  as its left limit and the value of $h'_T(-b)$  as its right limit, then we have the edge stopping function of $h_T(x)$ as:
\begin{eqnarray}\label{EqTruncatedHuberEdgeStop}
{
  \varphi_T(x)=\frac{h'_T(x)}{x}=\left\{\begin{array}{r}
                                           \frac{1}{a},    \ \ \ \ \ \ \ \ \ \  |x|<a \\
                                           \frac{1}{x},    \ \ \ \ a\leq |x|\leq b\\
                                           0,                    \ \ \ \ \ \ \ \ \ \ \  |x|>b
                                         \end{array}\right..
 }
\end{eqnarray}
Eq.~(\ref{EqTruncatedHuberEdgeStop}) indicates that the parameter $b$ in $h_T(x)$ can act as a threshold: for the input $x$ larger than $b$, it will directly not be penalized while the input smaller than $b$ remains penalized. $\varphi_T(x)$ is illustrated in Fig.~\ref{FigPenaltyComp}(b). If we use $h_T(x)$ to regularize the gradients of the output image, its penalty behaviour can enable the salient edges of the input image to be sharpened. This property also shares a similar mechanism with the Welsch's penalty function. The edge stopping function of the Welsch's penalty function rapidly reduces (at an exponential rate) to the values close to zero when the input increases, as shown in Fig.~\ref{FigPenaltyComp}(b). Similarly, the value of the edge stopping function of $h_T(\cdot)$  directly decreases to zero when the input is larger than $b$.

Based on the above analysis, we first show that $h_T(\cdot)$ in Eq.~(\ref{EqTruncatedHuber}) is a switchable penalty function where the value of $b$ can properly control whether to sharpen edges or not. Given the input intensity values are within $[0, I_m]$, then the amplitude of any edge will fall in $[0, I_m]$. If we set $b>I_m$, $h_T(\cdot)$ will be actually the same as $h(\cdot)$ because the second condition in Eq.~(\ref{EqTruncatedHuber}) can never be met, then $h_T(\cdot)$ will be an edge-preserving penalty function that does not sharpen edges. Conversely, when we set $b<I_m$, the truncation in $h_T(\cdot)$ will be activated. This can lead to penalty on weak edges without penalizing strong edges, and the strong edges are thus sharpened. In this way, $b$ can act as a switch to decide whether $h_T(\cdot)$ can sharpen strong edges or not. Besides the switchable property mentioned above, $h_T(\cdot)$ can also properly preserve weak edges if we set $a=\epsilon$ to a sufficient small value. This is because $h_T(\cdot)$ will be close to the $L_1$ norm for the small input when $a$ is sufficient small, and $L_1$ norm shows good edge-preserving property for weak edges. We show examples in Fig.~\ref{FigNormMotivation}(d) and (e) where we use $h_T(\cdot)$ with $a=\epsilon$ for regularization, and the weak edges are better preserved than that in Fig.~\ref{FigNormMotivation} (b) and (c). In addition, we set $b>I_m$ in Fig.~\ref{FigNormMotivation} (d) where the salient edges are not sharpened, and we set $b<I_m$ in Fig.~\ref{FigNormMotivation} (e) where the salient edges are sharpened, see the comparison between the regions labeled with the blue arrow and the green arrow.

Generally, $h_T(\cdot)$ can show strong flexibility which enable it to yield different penalty behaviours under different parameter settings. If we set $a=\epsilon, b>I_m$, $h_T(\cdot)$ will be close to the $L_1$ norm in this case, and thus it will be an edge-preserving penalty function that does not sharpen edges. We can also set $a=\epsilon, b<I_m$ to enable $h_T(\cdot)$ to sharpen edges. Similarly, by setting $a=b>I_m$ and $a=b<I_m$, $h_T(\cdot)$ can be easily switched between the $L_2$ norm and the truncated $L_2$ norm. We should note out that the Welsch's penalty function does not enjoy such kind of flexibility. Different cases of $h_T(\cdot)$ and their corresponding edge stopping functions are illustrated in Fig.~\ref{FigPenaltyComp}(c) and (d).

\subsection{Model}
\label{SecModel}

Given an input image $f$ and a guidance image $g$, the smoothed output image $u$ is the solution that gives the minimum to the following objective function:
\begin{equation}\label{EqObjFun}
\begin{array}{l}
  E_u(u)=\sum_{i}\sum_{j\in N_d(i)}\omega^s_{i,j}h_T(u_i-f_j)\\
    \ \ \ \ \ \ \ \ \ + \lambda\sum_{i}\sum_{j\in N_s(i)}\omega^s_{i,j}\omega^g_{i,j}h_T(u_i-u_j),
\end{array}
\end{equation}
where $h_T$ is defined in Eq.(\ref{EqTruncatedHuber}); $N_d(i)$ is the $(2r_d+1)\times (2r_d+1)$ square patch centered at $i$; $N_s(i)$ is the $(2r_s+1)\times (2r_s+1)$ square patch centered at $i$ excluding pixel $i$; $\lambda$ is a parameter that controls the overall smoothing strength. The first term of Eq.~(\ref{EqObjFun}) is denoted as the data term and the second term of Eq.~(\ref{EqObjFun}) is denoted as the smoothness term in this paper. To be clear, we adopt $\{a_d,b_d\}$ and $\{a_s,b_s\}$ to denote the parameters of $h_T(\cdot)$ in the data term and the smoothness term, respectively. $\omega^s_{i,j}$ is a Gaussian spatial kernel defined as:
\begin{equation}
  \omega^s_{i,j}=\exp\left(\frac{|i-j|^2}{2\sigma^2}\right),
\end{equation}
we simply fix $\sigma=r_d$ for the data term and $\sigma=r_s$ for the smoothness term in all of our experiments. The guidance weight $\omega^g_{i,j}$ is defined as:
\begin{equation}\label{EqGuidanceWeight}
{
  \omega^g_{i,j}=\frac{1}{|g_i-g_j|^\alpha + \delta},
}
\end{equation}
where $\alpha$ determines the sensitivity to the edges in $g$ which can be the input image, i.e., $g=f$. $|\cdot|$ represents the absolute value. $\delta$ is a small constant which is set as $\delta=10^{-3}$.

The adoption of $h_T(\cdot)$ enables our model in Eq.~(\ref{EqObjFun}) to enjoy strong flexibility. As we will show in Sec.~\ref{SecPropertyAnalysis}, under different parameter settings, our model is able to achieve different smoothing behaviors, and it is thus capable of various tasks that require different smoothing properties. In addition, our model can also yield simultaneous edge-preserving and structure-preserving smoothing that is seldom achieved by previous methods, which enables our model to achieve better performance in challenging cases.

\subsection{Numerical Solution}
\label{SecNumericalSolution}

Our model in Eq.~(\ref{EqObjFun}) is not only non-convex but also non-smooth, which arises from the adopted $h_T(\cdot)$. Commonly used approaches \cite{lanckriet2009convergence, nikolova2005analysis, wang2008new, zhang2004surrogate, zhang2017nonconvex, zheng2020globally} for solving non-convex optimization problems are thus not applicable. To tackle this problem, we need to first rewrite $h_T(\cdot)$ in a new equivalent form. By defining $\nabla^d_{i,j}=u_i-f_j$ and $\nabla^s_{i,j}=u_i-u_j$, we have:
\begin{equation}\label{EqRelationWithHuberL0}
{
h_T(\nabla^\ast_{i,j})=\min_{l^\ast_{i,j}}\left\{h(\nabla^\ast_{i,j}-l^\ast_{i,j})+(b_\ast-\frac{a_\ast}{2})|l^\ast_{i,j}|_0\right\},
}
\end{equation}
where $\ast\in\{d,s\}$, $|l^\ast_{i,j}|_0$ is the $L_0$ norm of $l^\ast_{i,j}$. The minimum of the right side of Eq.~(\ref{EqRelationWithHuberL0}) is obtained at:
\begin{eqnarray}\label{EqTruncatedHuberMinCondition}
{
  l^\ast_{i,j}=\left\{\begin{array}{r}
   0, \ \ \ \ \ \ \ \ |\nabla^\ast_{i,j}|\leq b_\ast\\
   \nabla^\ast_{i,j}, \ \ \ |\nabla^\ast_{i,j}|>b_\ast
  \end{array}\right.
  , \ \ \ast\in\{d,s\}.
}
\end{eqnarray}
The detailed proof of Eq.~(\ref{EqRelationWithHuberL0}) and Eq.~(\ref{EqTruncatedHuberMinCondition}) is provided in Appendix A. These two equations also theoretically validate our analysis in Sec.~\ref{SecTruncatedHuber} and Fig.~\ref{FigPenaltyComp}(c): we have $|\nabla^\ast_{i,j}|\in[0, I_m]$ if the intensity values are within $[0, I_m]$. Then if $b>I_m$, based on Eq.~(\ref{EqRelationWithHuberL0}) and Eq.~(\ref{EqTruncatedHuberMinCondition}), we will always have $h_T(\nabla^\ast_{i,j})=h(\nabla^\ast_{i,j})$ which means $h_T(\cdot)$ degrades to $h(\cdot)$.

We can then define a new energy function as:
\begin{equation}\label{EqObjFunAuxUL}
{
\begin{array}{r}
   E_{ul}(u, l^d, l^s)=\sum\limits_{i,j}\omega^s_{i,j}\left(h(\nabla^{d}_{i,j} - l^d_{i,j}) + (b_d-\frac{a_d}{2})|l^d_{i,j}|_0 \right)\\
   \ \ \ \ \ \ + \lambda\sum\limits_{i,j}\omega_{i,j}\left(h(\nabla^{s}_{i,j} - l^s_{i,j}) + (b_s-\frac{a_s}{2})|l^s_{i,j}|_0 \right)
\end{array}
},
\end{equation}
where $\omega_{i,j}=\omega^s_{i,j}\omega^g_{i,j}$. Based on Eq.~(\ref{EqRelationWithHuberL0}) and Eq.~(\ref{EqTruncatedHuberMinCondition}), we then have:
\begin{equation}\label{EqEnergyRelation1}
{
E_u(u)=\min_{l^\ast}E_{ul}(u, l^\ast),\ \ast\in\{d,s\}.
}
\end{equation}

Given Eq.~(\ref{EqTruncatedHuberMinCondition}) as the optimum condition of Eq.~(\ref{EqEnergyRelation1}) with respect to $l^\ast$, optimizing $E_{ul}(u, l^d, l^s)$ with respect to $u$ only involves the Huber penalty function $h(\cdot)$. The problem can thus be optimized through the half-quadratic (HQ) optimization technique \cite{geman1995nonlinear, nikolova2005analysis}. More specifically, a variable $\mu^\ast (\ast\in\{d,s\})$ and a function $\psi(\mu^\ast_{i,j})$ with respect to $\mu^\ast$ exist such that:
\begin{equation}\label{EqMultHQ}
\small
{
h(\nabla^\ast_{i,j} - l^\ast_{i,j})=\min_{\mu^\ast_{i,j}}\left\{\mu^\ast_{i,j}(\nabla^\ast_{i,j}-l^\ast_{i,j})^2 + \psi(\mu^\ast_{i,j}) \right\},\ast\in\{d,s\}
}
\end{equation}
where the optimum is yielded at:
\begin{eqnarray}\label{EqMultHQCondition}
{
  \mu^\ast_{i,j}=\left\{\begin{array}{r}
   \frac{1}{2a_\ast}, \ \ \ \ \ \ \ \ \ \ \ \ \ \  |\nabla^\ast_{i,j} - l^\ast_{i,j}|< a_\ast\\
   \frac{1}{2|\nabla^\ast_{i,j} - l^\ast_{i,j}|}, \ \  \   |\nabla^\ast_{i,j} - l^\ast_{i,j}| \geq a_\ast
  \end{array}\right.
  , \ \ \ast\in\{d,s\}.
}
\end{eqnarray}
The detailed proof of Eq.~(\ref{EqMultHQ}) and Eq.~(\ref{EqMultHQCondition}) is provided in Appendix B. Then we can further define a new energy function as:
\begin{small}
\begin{equation}\label{EqObjFunAuxULMu}
{
\begin{array}{l}
   E_{ul\mu}(u, l^d, l^s, \mu^d, \mu^s)= \\
   \ \ \ \ \ \ \ \ \sum\limits_{i,j}\omega^s_{i,j}\left(\mu^d_{i,j}(\nabla^{d}_{i,j} - l^d_{i,j})^2 + \psi(\mu^d_{i,j}) + (b_d-\frac{a_d}{2})|l^d_{i,j}|_0 \right)\\
   \ \  +\lambda\sum\limits_{i,j}\omega_{i,j}\left(\mu^s_{i,j}(\nabla^{s}_{i,j} - l^s_{i,j})^2 + \psi(\mu^s_{i,j}) + (b_s-\frac{a_s}{2})|l^s_{i,j}|_0 \right)
\end{array}
}.
\end{equation}
\end{small}
 Based on Eq.~(\ref{EqMultHQ}) and Eq.~(\ref{EqMultHQCondition}), we then have:
\begin{equation}\label{EqEnergyRelation2}
{
  E_{ul}(u, l^\ast)=\min\limits_{\mu^\ast}E_{ul\mu}(u, l^\ast, \mu^\ast),\ \ast\in\{d,s\}.
}
\end{equation}

Given Eq.~(\ref{EqMultHQCondition}) as the optimum condition of $\mu^\ast$ in Eq.~(\ref{EqEnergyRelation2}), optimizing $E_{ul\mu}(u, l^d, l^s, \mu^d, \mu^s)$ with respect to $u$ only involves the $L_2$ norm penalty function, which has a closed-form solution. However, since the optimum conditions in Eq.~(\ref{EqTruncatedHuberMinCondition}) and Eq.~(\ref{EqMultHQCondition}) both involve $u$, therefore, the final solution $u$ can only be obtained in an iterative manner. Assuming we have got $u^k$, then $(l^\ast)^{k}$ and $(\mu^\ast)^{k} (\ast\in\{s,d\})$ can be updated through Eq.~(\ref{EqTruncatedHuberMinCondition}) and Eq.~(\ref{EqMultHQCondition}) with $u^k$, respectively. Finally, $u^{k+1}$ is obtained with:
\begin{equation}\label{EqIterativeSolution}
{
   u^{k+1}=\underset{u}{\arg\min}E_{ul\mu}\left(u, (l^\ast)^k, (\mu^\ast)^k\right), \ \ast\in\{d,s\},
}
\end{equation}
Eq.(\ref{EqIterativeSolution}) has a close-form solution as:
\begin{equation}\label{EqCloseFormSolution}
{
   u^{k+1}=\left(\mathcal{A}^k - 2\lambda\mathcal{W}^k\right)^{-1}\left(D^k + 2\lambda S^k\right),
}
\end{equation}
where $\mathcal{W}^k$ is an affinity matrix with $\mathcal{W}^k_{i,j}=\omega_{i,j}(\mu^s_{i,j})^k$, $\mathcal{A}^k$ is a diagonal matrix with $\mathcal{A}^k_{ii}=\sum_{j\in N_d(i)}\omega^s_{i,j}(\mu^d_{i,j})^k + 2\lambda\sum_{j\in N_s(i)}\omega_{i,j}(\mu^s_{i,j})^k$, $D^k$ is a vector with $D^k_i=\sum_{j\in N_d(i)}\omega^s_{i,j}(\mu^d_{i,j})^k(f_j+(l^d_{i,j})^k)$ and $S^k$ is also a vector with $S^k_i=\sum_{j\in N_s(i)}\omega_{i,j}(\mu^s_{i,j})^k(l^s_{i,j})^k$.

The above optimization procedure monotonically decreases the value of $E_u(u)$ in each step, and its convergence is theoretically guaranteed. Given $u^k$ in the $k$th iteration and $\ast\in\{s,d\}$, then for any $u$, we have:
\begin{equation}\label{EqEnergyRelationTruncation}
{
  E_u(u)\leq E_{ul}(u, (l^\ast)^k),\ E_u(u^k)=E_{ul}(u^k, (l^\ast)^k),
}
\end{equation}
\begin{eqnarray}\label{EqEnergyRelationHQ}
{\left\{
\begin{array}{l}
  E_{ul}(u,(l^\ast)^k)\leq E_{ul\mu}(u, (l^\ast)^k, (\mu^\ast)^k)\\
  E_{ul}(u^k,(l^\ast)^k)=E_{ul\mu}(u^k, (l^\ast)^k, (\mu^\ast)^k)
\end{array}\right. .
}
\end{eqnarray}
Given $(l^\ast)^k$ has been updated through Eq.~(\ref{EqTruncatedHuberMinCondition}), Eq.~(\ref{EqEnergyRelationTruncation}) is based on Eq.~(\ref{EqEnergyRelation1}) and Eq.~(\ref{EqRelationWithHuberL0}). After $(\mu^\ast)^k$ has been updated through Eq.~(\ref{EqMultHQCondition}), Eq.~(\ref{EqEnergyRelationHQ}) is based on Eq.~(\ref{EqEnergyRelation2}) and Eq.~(\ref{EqMultHQ}). We now have:
\begin{equation}\label{EqEnergyDecrease1}
{
\begin{array}{l}
  E_{ul}(u^{k+1},(l^\ast)^k)\leq E_{ul\mu}(u^{k+1}, (l^\ast)^k, (\mu^\ast)^k)\\
  \leq E_{ul\mu}(u^{k}, (l^\ast)^k, (\mu^\ast)^k)=E_{ul}(u^{k},(l^\ast)^k)
\end{array},
}
\end{equation}
the first and the second inequalities follow from Eq.~(\ref{EqEnergyRelationHQ}) and Eq.~(\ref{EqIterativeSolution}), respectively. We finally have:
\begin{small}
\begin{equation}\label{EqEnergyDecrease2}
{
  E_u(u^{k+1})\leq E_{ul}(u^{k+1}, (l^\ast)^k)\leq E_{ul}(u^{k}, (l^\ast)^k)=E_{u}(u^k),
}
\end{equation}
\end{small}
the first and the second inequalities follow from Eq.~(\ref{EqEnergyRelationTruncation}) and Eq.~(\ref{EqEnergyDecrease1}), respectively. Since the value of $E_u(u)$ is bounded from below, Eq.~(\ref{EqEnergyDecrease2}) thus indicates that the convergence of our iterative scheme is theoretically guaranteed.

\begin{algorithm}[t]
\caption {A Generalized Framework for Edge-preserving and Structure-preserving Image Smoothing}\label{Alg}
\begin{algorithmic}[1]
\REQUIRE
Input image $f$, guide image $g$, iteration number $N$, parameter $\lambda, \alpha, a_\ast, b_\ast, r_\ast$, $u^0\leftarrow f$, with $\ast\in\{d,s\}$\\

\FOR{$k=0:N$}
\STATE With $u^k$, compute $(\nabla^\ast_{i,j})^k$, update $(l^\ast_{i,j})^k$ according to Eq.~(\ref{EqTruncatedHuberMinCondition})
\STATE With $(l^\ast_{i,j})^k$, update $(\mu^\ast_{i,j})^k$ according to Eq.~(\ref{EqMultHQCondition})
\STATE With $(l^\ast_{i,j})^k$ and $(\mu^\ast_{i,j})^k$, solve for $u^{k+1}$ according to Eq.~(\ref{EqIterativeSolution}) (or Eq.~(\ref{EqCloseFormSolution}))
\ENDFOR
\ENSURE
Smoothed image $u^{N+1}$
\end{algorithmic}
\end{algorithm}
\begin{table*}
 \newcommand{\tabincell}[2]{\begin{tabular}{@{}#1@{}}#2\end{tabular}}
  \centering
  \caption{Parameter settings for different tasks. $\epsilon$ refers to a small constant, e.g., $\epsilon=10^{-3}$; $I_m$ denotes the maximum intensity value of the input image $f$; $g=f$ means the guidance image is the same as the input image to be smoothed; $g=g$ denotes that the guidance image is different from the input image to be smoothed. $SP$ and $EP$ are short for $structure-preserving$ and $edge-preserving$, respectively.}\label{TabParameter}
  \resizebox{1\textwidth}{!}
  {
  \begin{tabular}{c|ccccccccc|c|c}
    \Xhline{1.2pt}
     mode & $g$ & $\alpha$ & $a_d$ & $b_d$ & $a_s$ & $b_s$ & $r_d$ & $r_s$ & $N$ & Properties & Applications\\
    \Xhline{1.2pt}
    SP-1                & $f$ & 0.5 & $\epsilon$ & $>I_m$ & $\epsilon$ & $>I_m$ & 1 & 1 & 10 & structure-preserving & tasks in the fourth group\\
    \hline
    SP-2                & $f$ & 0.2 & $\epsilon$ & $>I_m$ & $\epsilon$ & $>I_m$ & 1 & 1 & 1 & structure-preserving & tasks in the first group \\
    \hline
    EP-1                & $f$ & 1.2 & $>I_m$ & $>I_m$ & $>I_m$ & $>I_m$ & 0 & 1 & 1 & edge-preserving, do not sharpen edges& tasks in the first group\\
    \hline
    EP-2                & $f/g$ & 0.5 & $>I_m$ & $>I_m$ & $\epsilon$ & $<I_m$ & 0 & $\geq$1 & 10 & edge-preserving, sharpen edges & tasks in the second group and the third group \\
    \hline
    EP$\&$SP    & $f/g$ & 0.5 & $\epsilon$ & $<I_m$ & $\epsilon$ & $<I_m$ & $\geq$1 & $\geq$1 & 10 & \tabincell{c}{simultaneous edge-preserving and \\  structure-preserving, sharpen edges} & tasks in the second group and the third group\\
    \Xhline{1.2pt}
  \end{tabular}
  }
\end{table*}

\begin{figure*}
\centering
\setlength{\tabcolsep}{0.1mm}
\begin{tabular}{ccccccc}
\includegraphics[width=0.1422\linewidth]{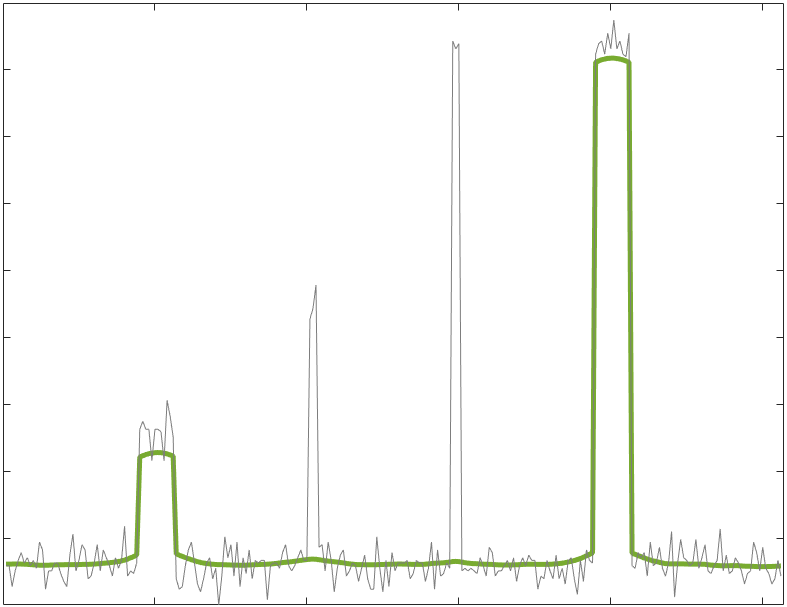}&
\includegraphics[width=0.1418\linewidth]{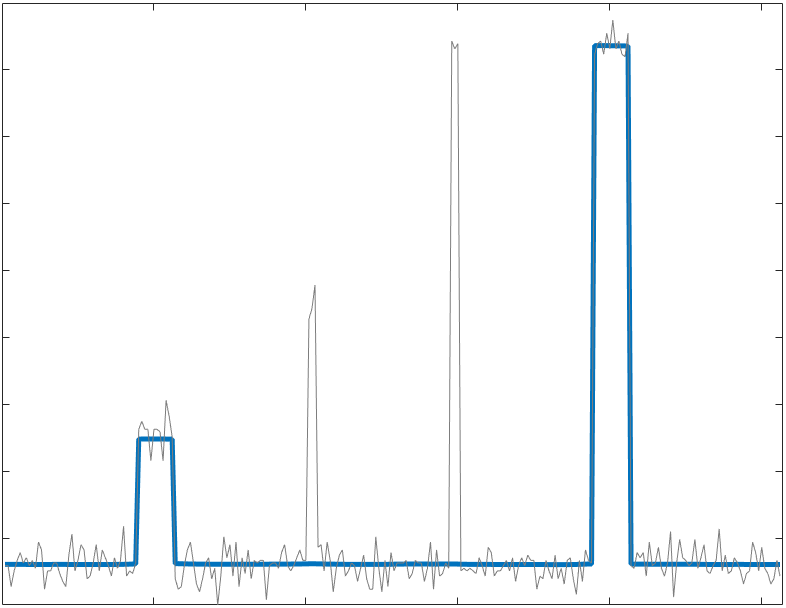} &
\includegraphics[width=0.1422\linewidth]{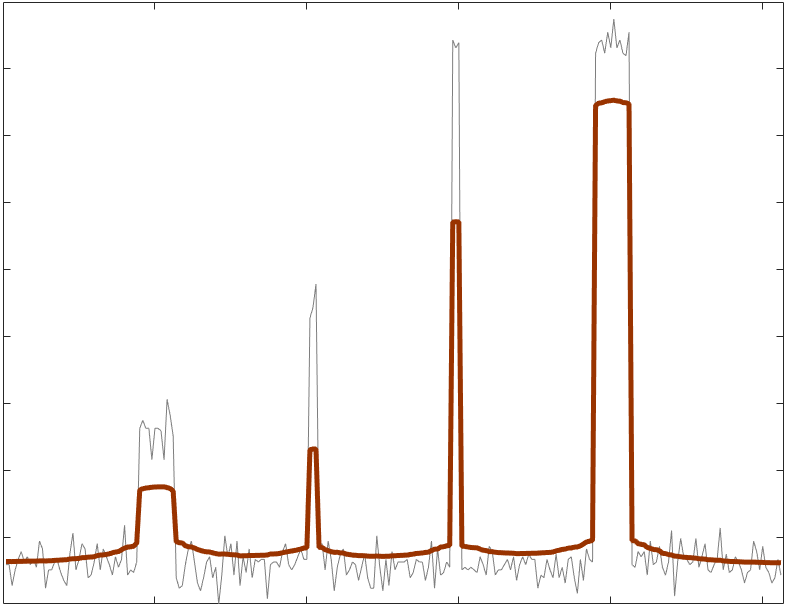}&
\includegraphics[width=0.1418\linewidth]{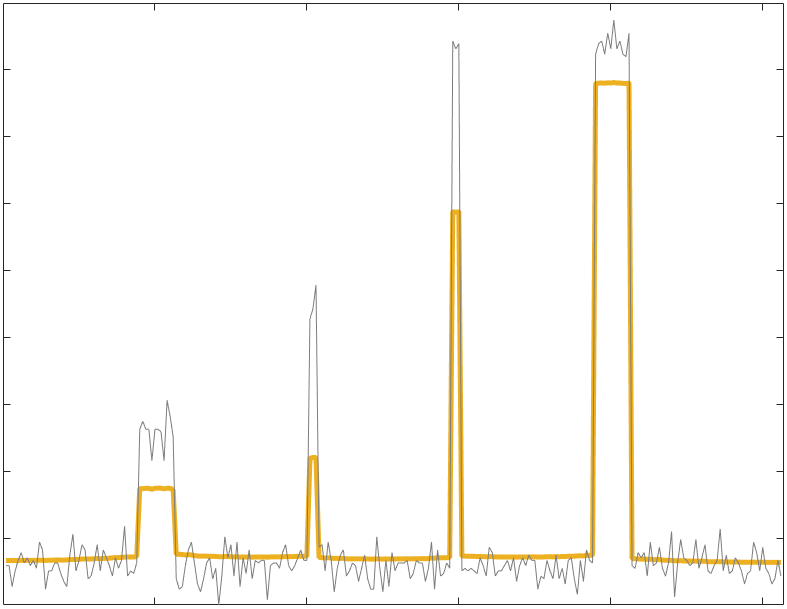}&
\includegraphics[width=0.142\linewidth]{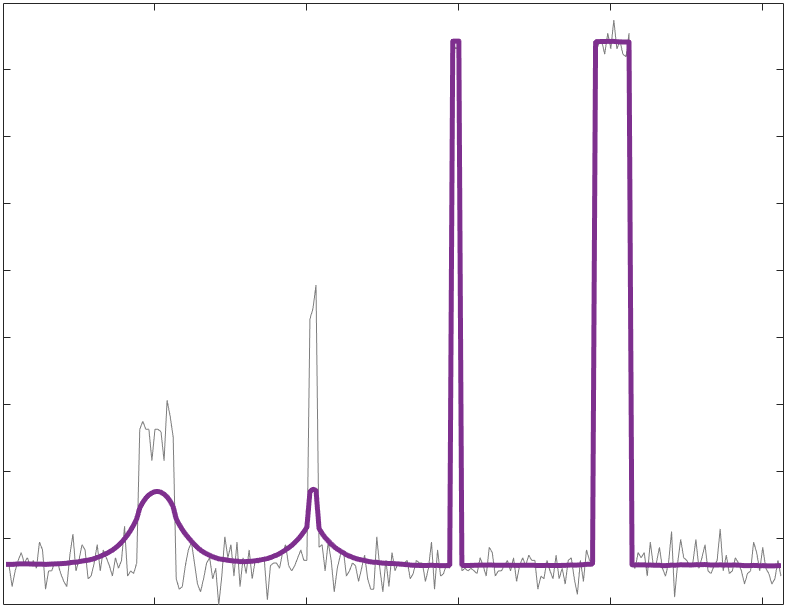}&
\includegraphics[width=0.1414\linewidth]{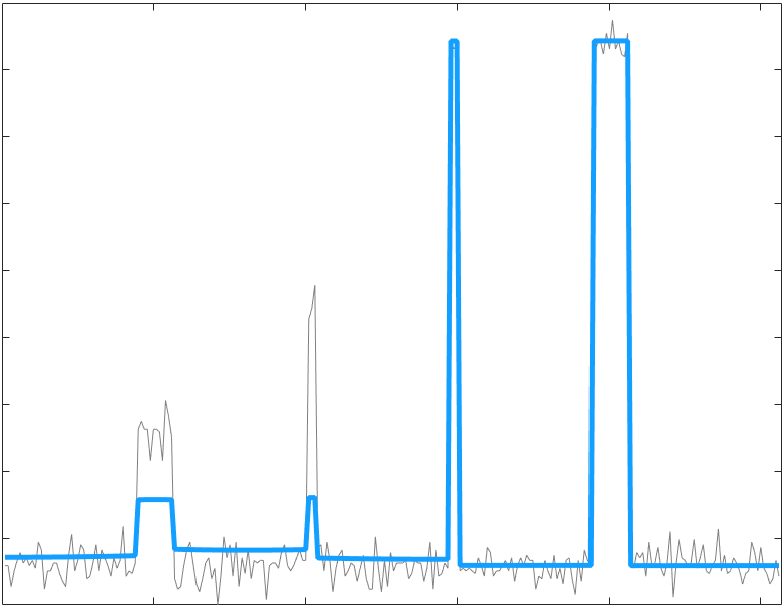}&
\includegraphics[width=0.1418\linewidth]{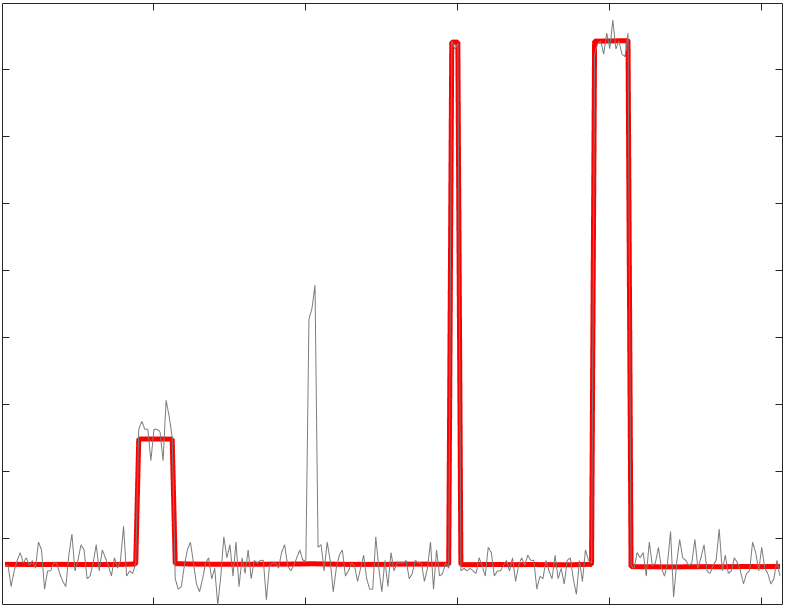}\\

(a) TV-$L_1$ & (b) ours (SP-1) & (c) WLS & (d) ours (SP-2) & (e) SD filter & (f) ours (EP-2) & (g) ours (EP$\&$SP)
\end{tabular}
\caption{1D signal with structures of different scales and amplitudes. Smoothing result of (a) TV-$L_1$ smoothing \cite{buades2010fast}, (b) our method of the SP-1 mode, (c) WLS \cite{farbman2008edge} (our method of the EP-1 mode), (d) our method of the SP-2 mode, (e) SD filter \cite{ham2018robust}, (f) our method of the EP-2 mode and (g) our method of the EP$\&$SP mode.}\label{Fig1DComp}
\end{figure*}
\begin{figure}
  \centering
  \setlength{\tabcolsep}{0.25mm}
  \begin{tabular}{ccc}
  \includegraphics[width=0.32\linewidth]{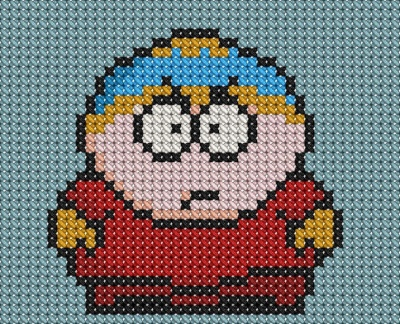} &
  \includegraphics[width=0.32\linewidth]{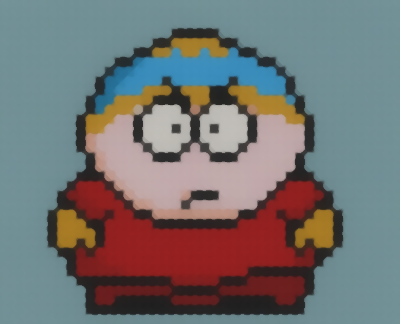} &
  \includegraphics[width=0.32\linewidth]{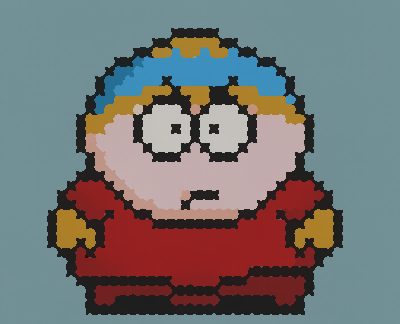}\\

  (a) input & (b) TV-$L_1$ & (c) ours (SP-1)
  \end{tabular}
  \caption{Comparison of image texture removal results. (a) Input image. Result of (b) TV-$L_1$  smoothing\cite{buades2010fast}, (e) our method of the SP-1 mode.}\label{FigMyVsTVL1}
\end{figure}

The above optimization procedure is iteratively performed $N$ times to get the final smoothed output $u^N$. $N$ can vary for different applications, and some tasks do not need to iterate the above procedure until it converges. These will be detailed in Sec.~\ref{SecPropertyAnalysis}. In all our experiments, we set $u^0=f$, which is able to produce promising results in each application. Our optimization procedure is summarized in Algorithm~\ref{Alg}.

\subsection{Property Analysis}
\label{SecPropertyAnalysis}

Under different parameter settings, the strong flexibility of $h_T(\cdot)$ makes our model able to achieve various smoothing behaviors. First, we show that some classical approaches can be viewed as special cases of our model. For example, by setting $a_d=b_d>I_m, a_s=\epsilon,b_s>I_m,\alpha=0,r_d=0,r_s=1$, our model is a close approximation of the TV model \cite{rudin1992nonlinear} which is a representative edge-preserving smoother. If we set $g=f, a_d=b_d=a_s=b_s>I_m,\alpha=1.2,r_d=0,r_s=1$, our model will be the WLS smoother \cite{farbman2008edge} which performs well in handling gradient reversals and halos in image detail enhancement and HDR tone mapping. With parameters $a_d=\epsilon,b_d>I_m,a_s=\epsilon,b_s>I_m,\alpha=0,r_d=0,r_s=1$, our model is a close approximation of the TV-$L_1$ model \cite{aujol2006structure, buades2010fast} which is classical for structure-preserving smoothing.

\begin{figure*}[!ht]
  \centering
  \setlength{\tabcolsep}{0.25mm}
  \begin{tabular}{ccccc}
  \includegraphics[width=0.198\linewidth]{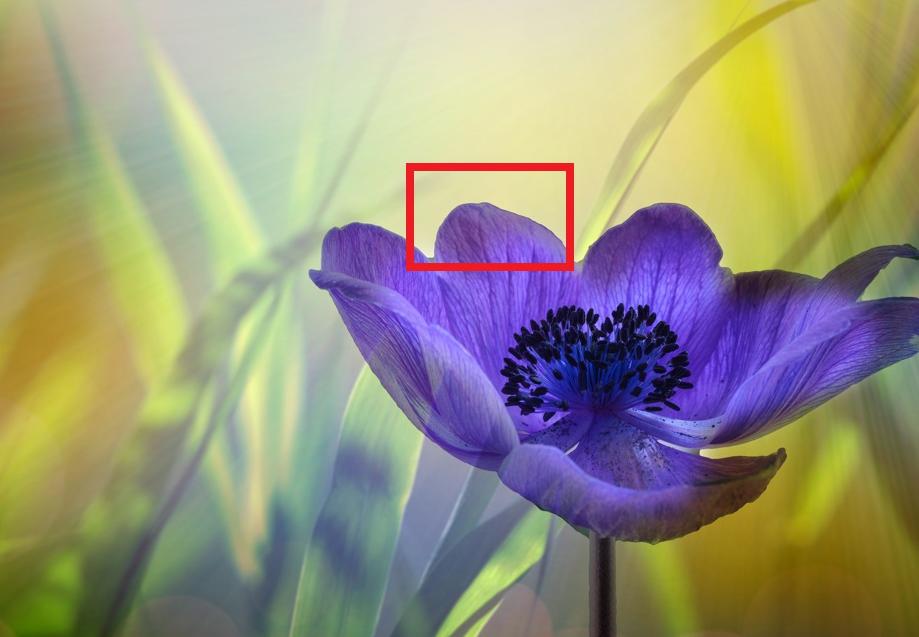} &
  \includegraphics[width=0.198\linewidth]{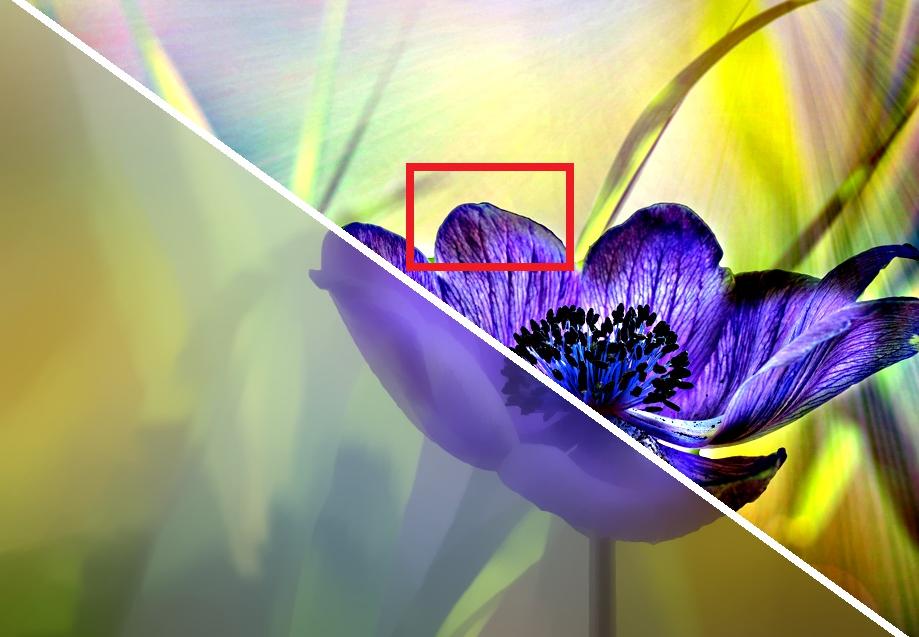} &
  \includegraphics[width=0.198\linewidth]{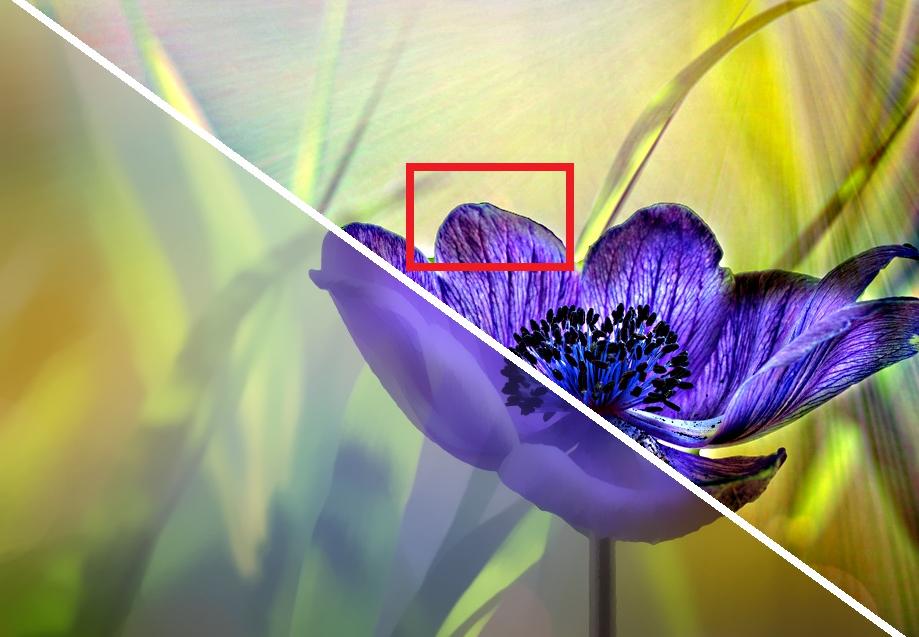} &
  \includegraphics[width=0.1945\linewidth]{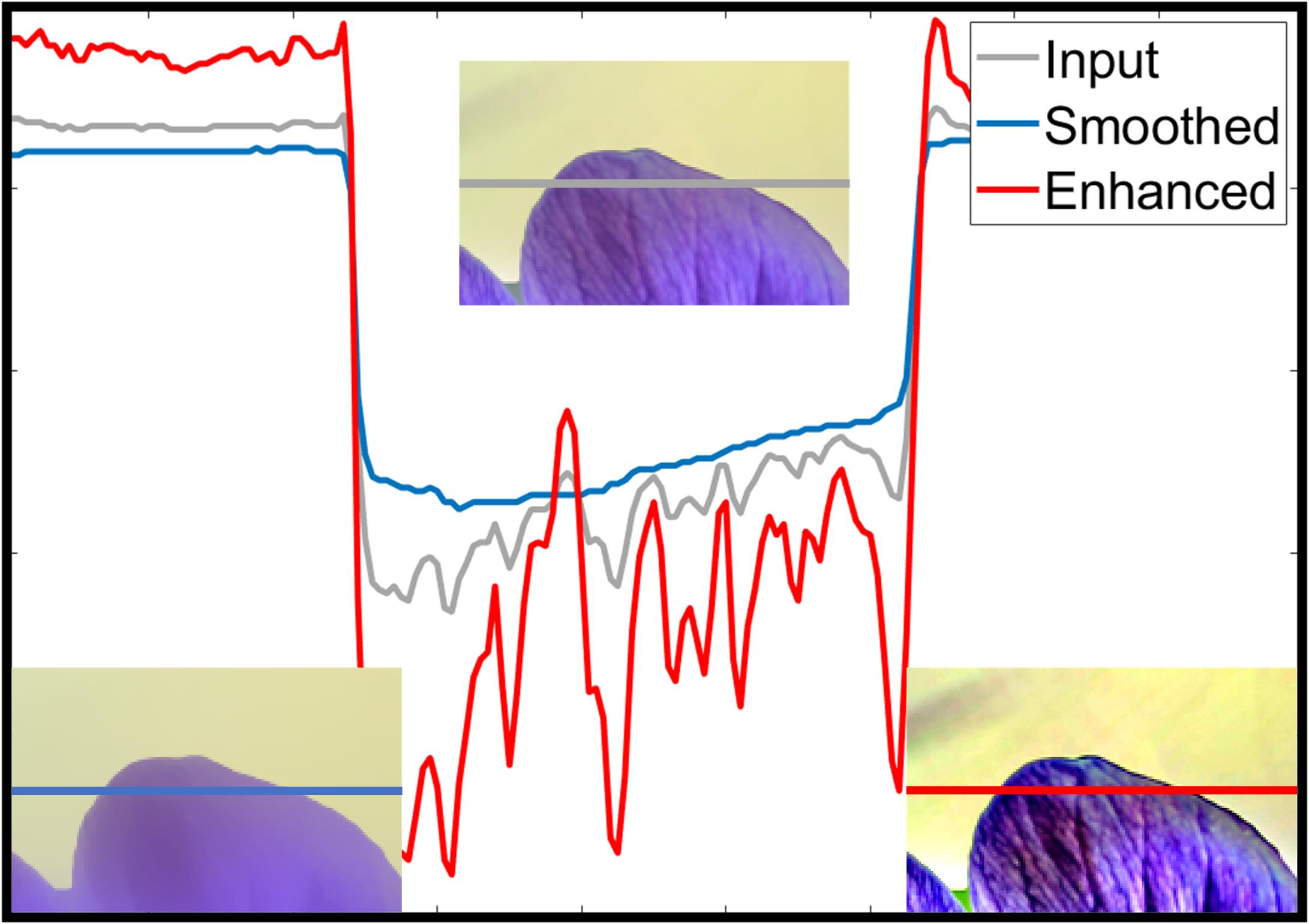}&
  \includegraphics[width=0.1945\linewidth]{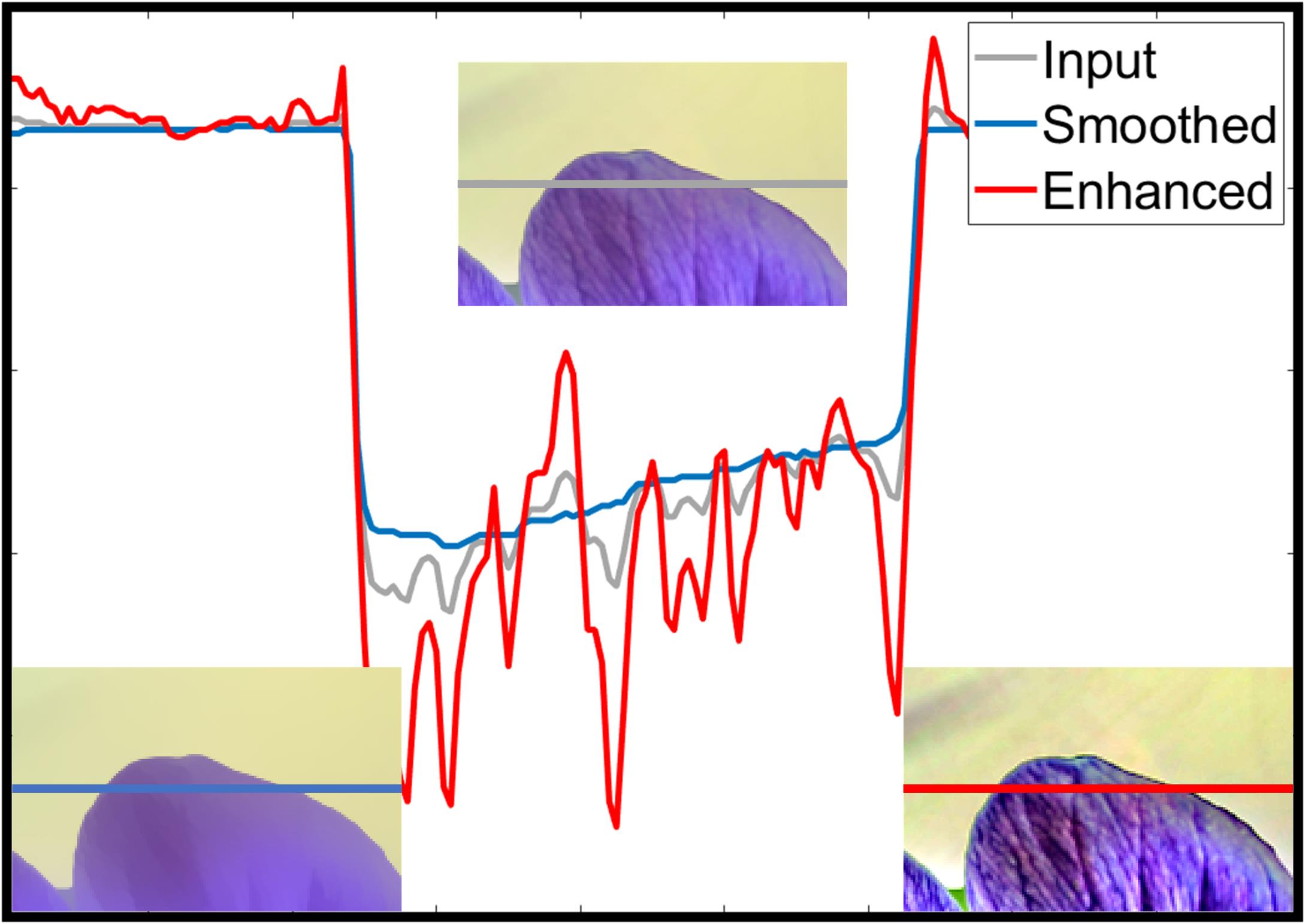}\\

  \includegraphics[width=0.198\linewidth]{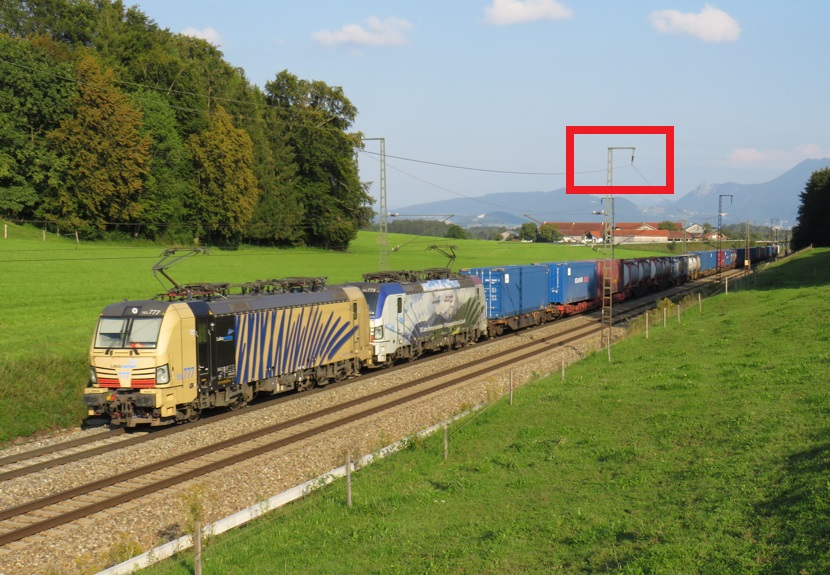} &
  \includegraphics[width=0.198\linewidth]{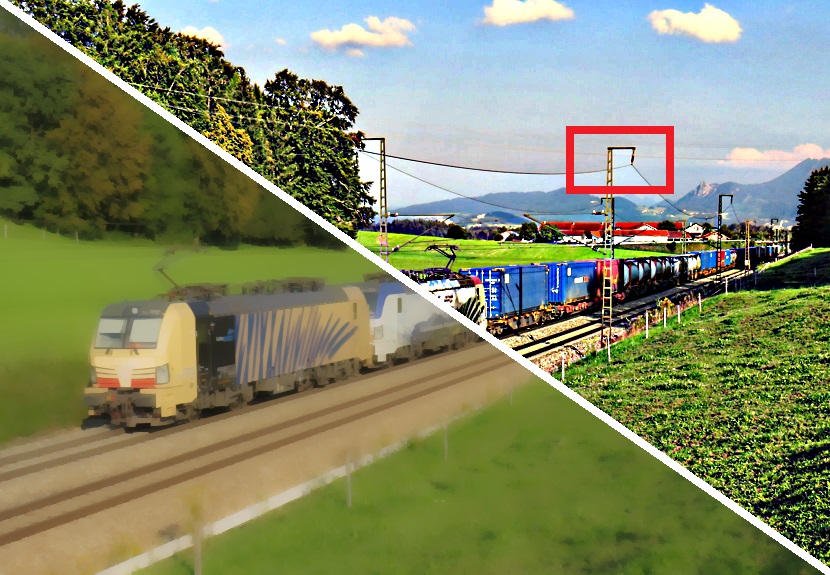} &
  \includegraphics[width=0.198\linewidth]{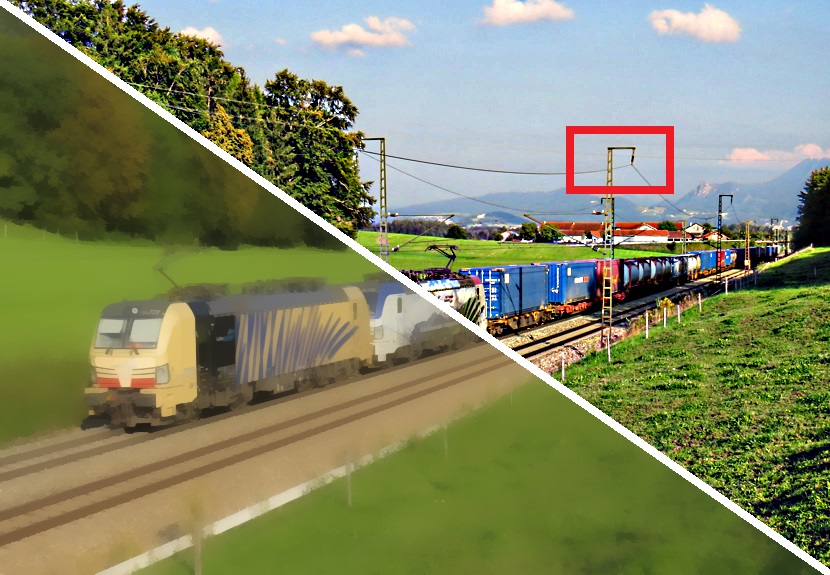} &
  \includegraphics[width=0.1945\linewidth]{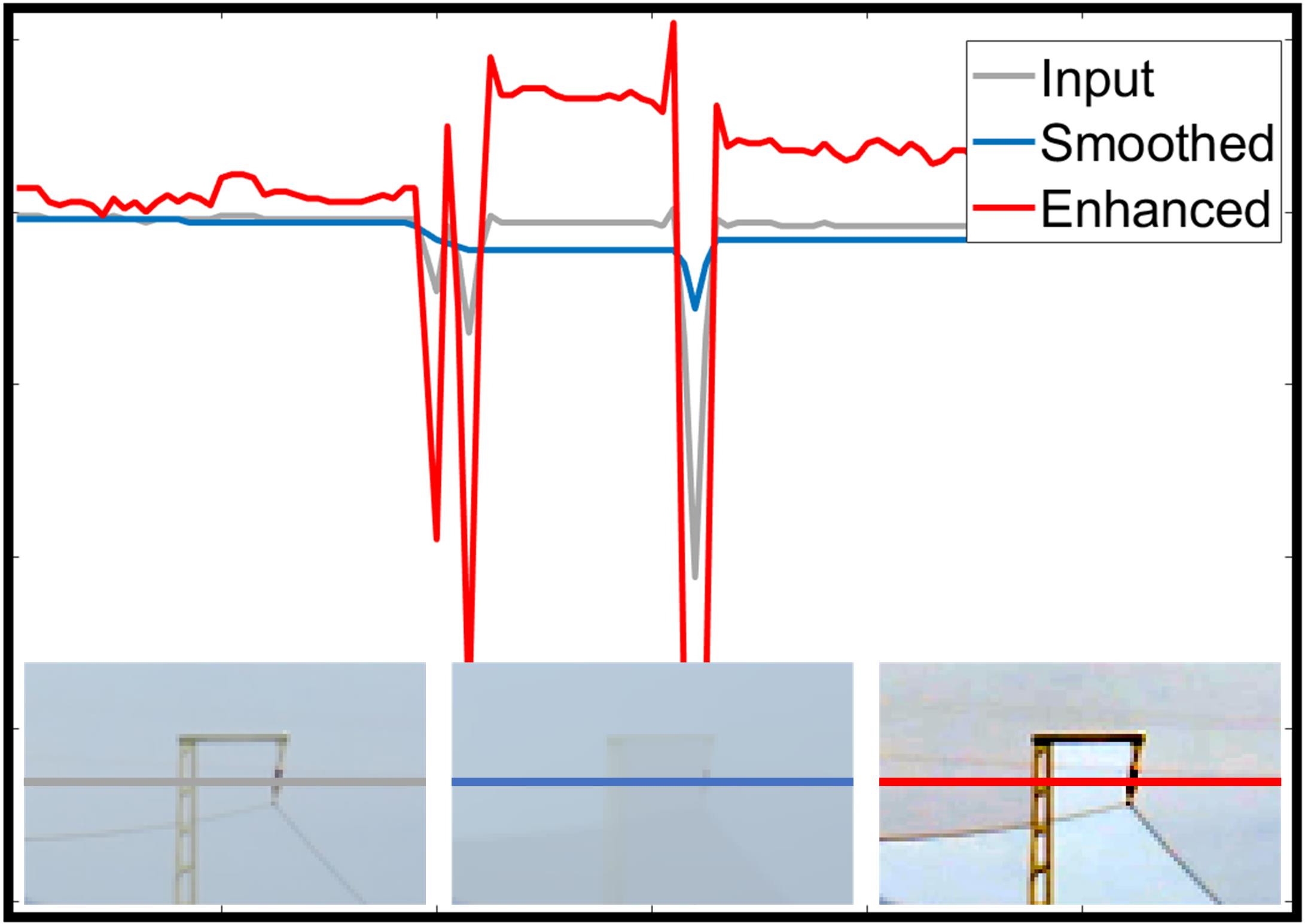}&
  \includegraphics[width=0.1945\linewidth]{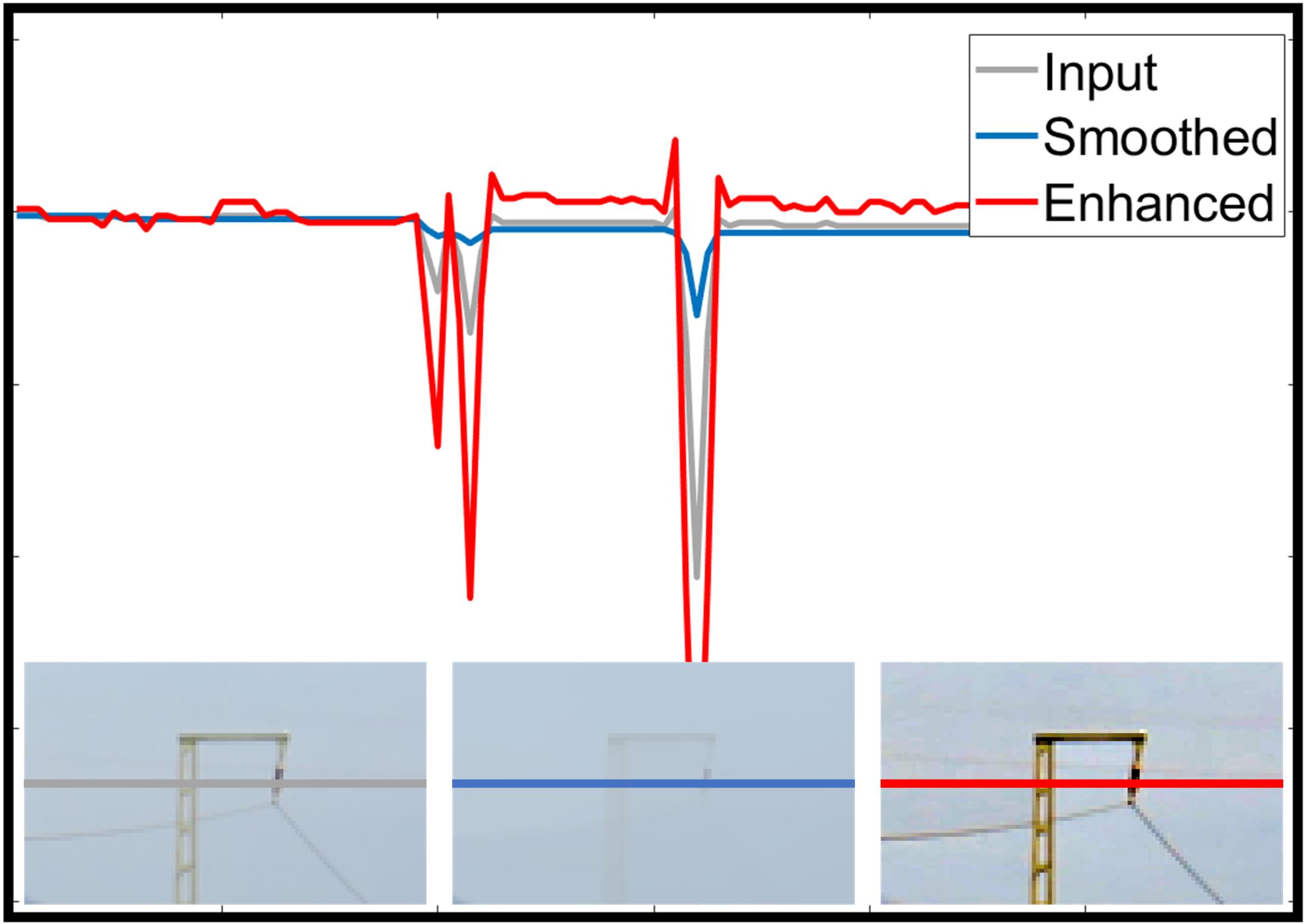}\\
  (a) input & (b) WLS/ours (EP-1) & (c) ours (SP-2) & (d) WLS/ours (EP-1) & (e) ours (SP-2)\\
  \end{tabular}
  \caption{Comparison of intensity shift in Image detail enhancement. (a) Input image. Result of (b) WLS \cite{farbman2008edge} (our method of the EP-1 mode) and (c) our method of the SP-2 mode. 1D plot of the highlighted region in the result of (d) WLS \cite{farbman2008edge} (our method of the EP-1 mode) and (e) our method of the SP-2 mode.}\label{FigMyVsWLS}
\end{figure*}
\begin{figure}
  \centering
  \setlength{\tabcolsep}{0.5mm}
  \begin{tabular}{cc}
  \includegraphics[width=0.49\linewidth]{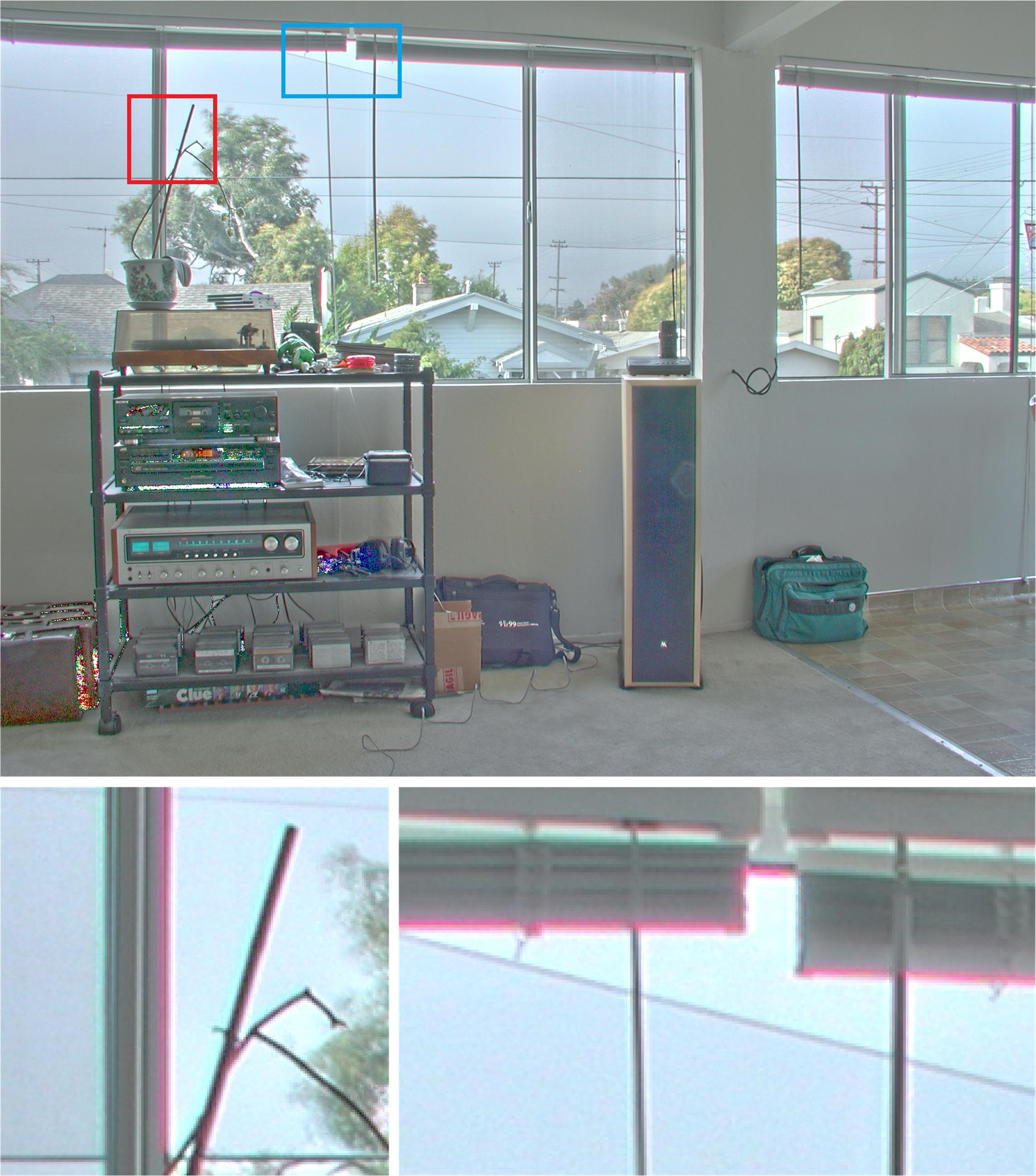} &
  \includegraphics[width=0.49\linewidth]{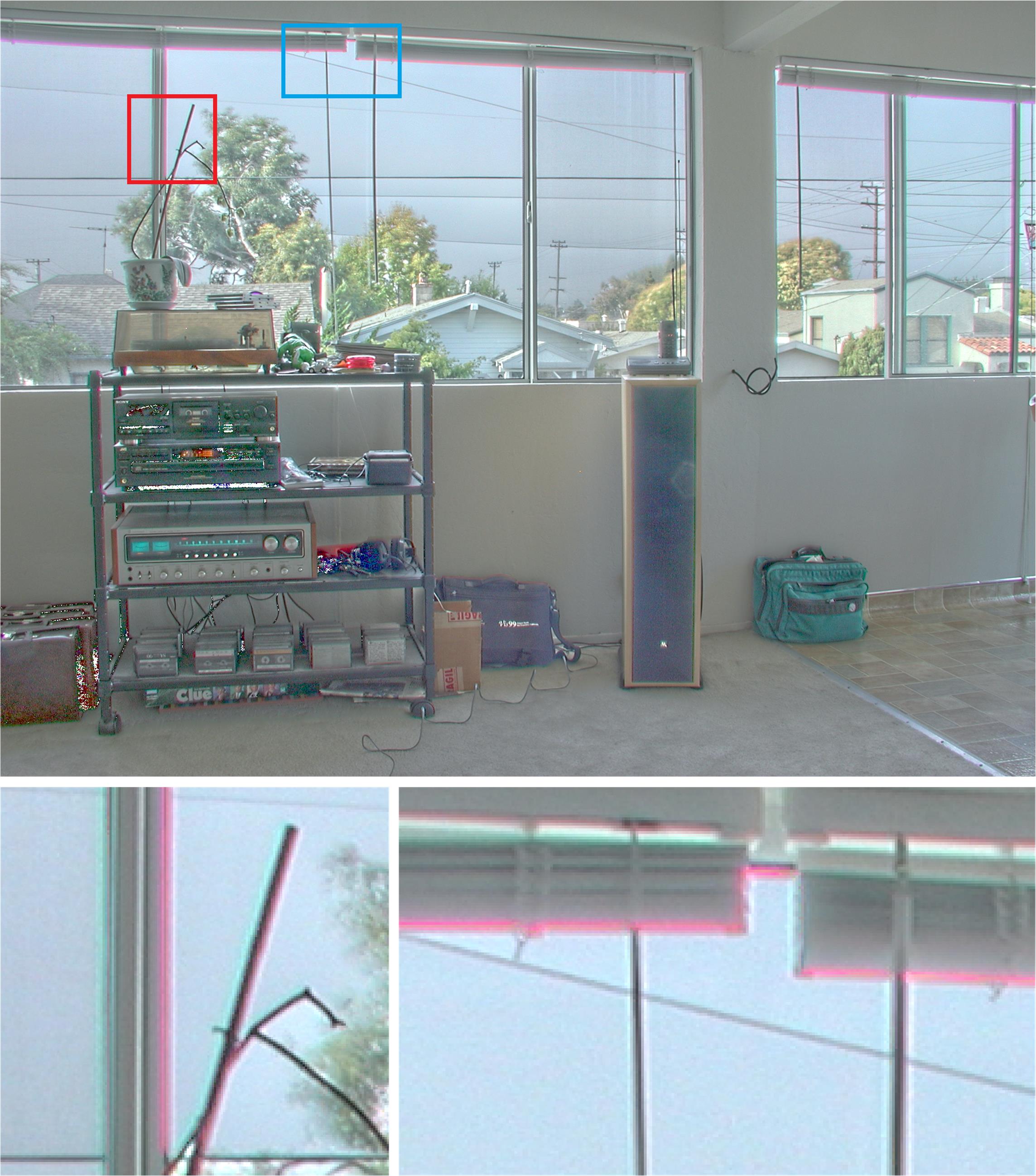} \\

  (a) WLS/ours (EP-1)& (b) ours (SP-2)\\
  \end{tabular}
  \caption{Comparison of intensity shift in HDR tone mapping. Result of (a) WLS \cite{farbman2008edge} (our method of the EP-1 mode) and (b) our method of the SP-2 mode.}\label{FigMyVsWLS_HDR}
\end{figure}
\begin{figure*}[!ht]
  \centering
  \setlength{\tabcolsep}{0.5mm}
  \begin{tabular}{cccc}
  \includegraphics[width=0.288\linewidth]{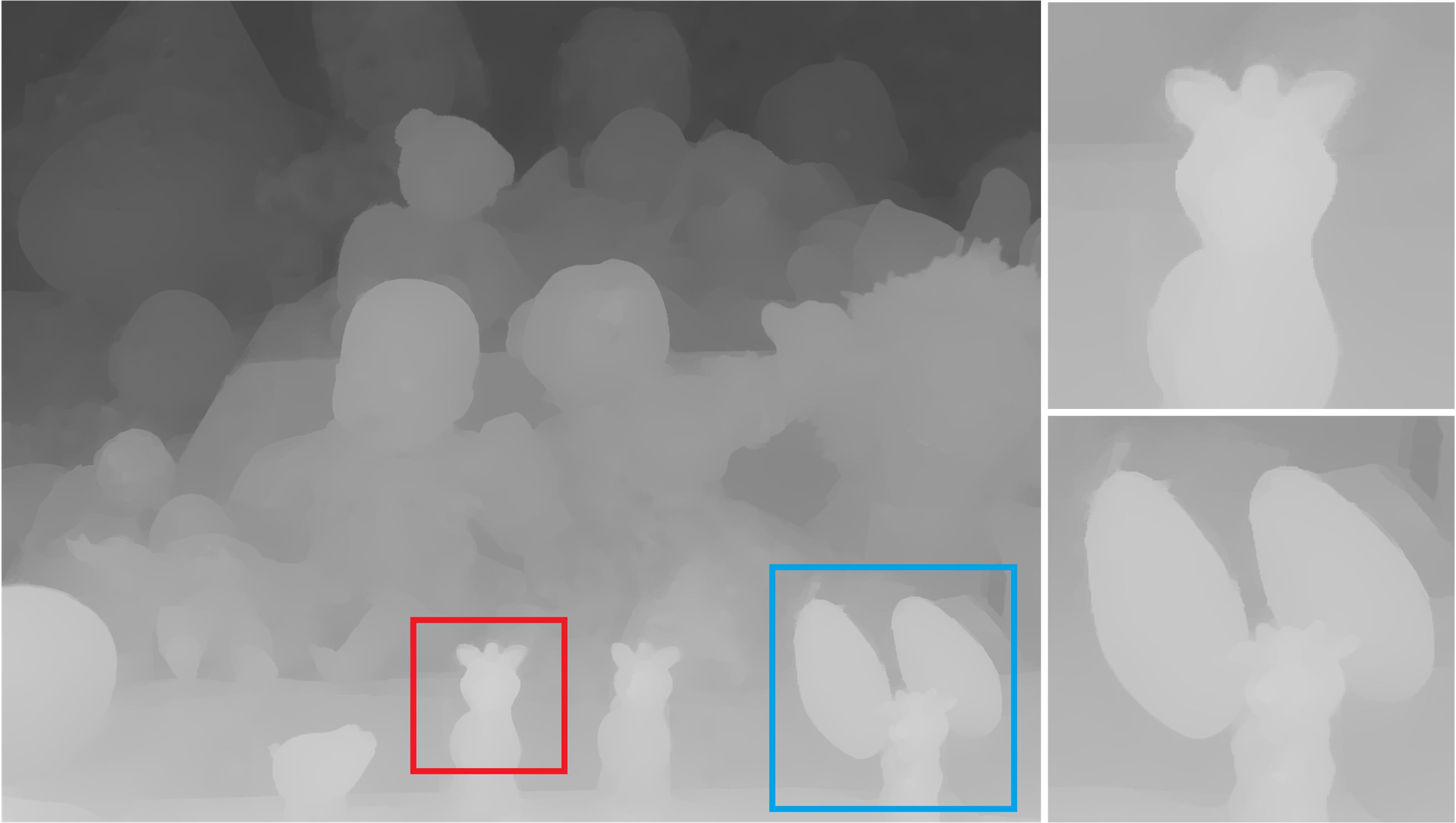} &
  \includegraphics[width=0.288\linewidth]{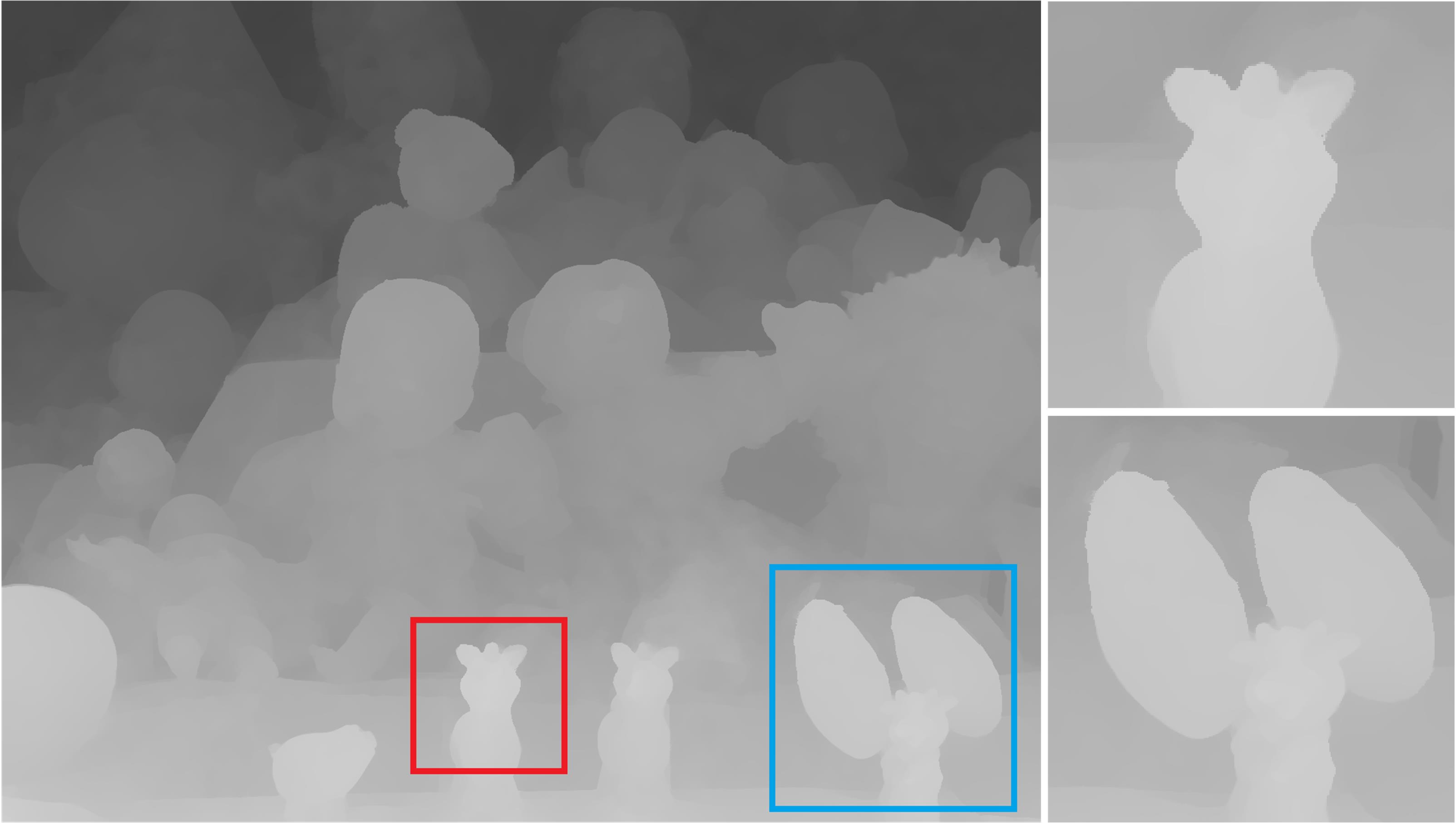} &
  \includegraphics[width=0.205\linewidth]{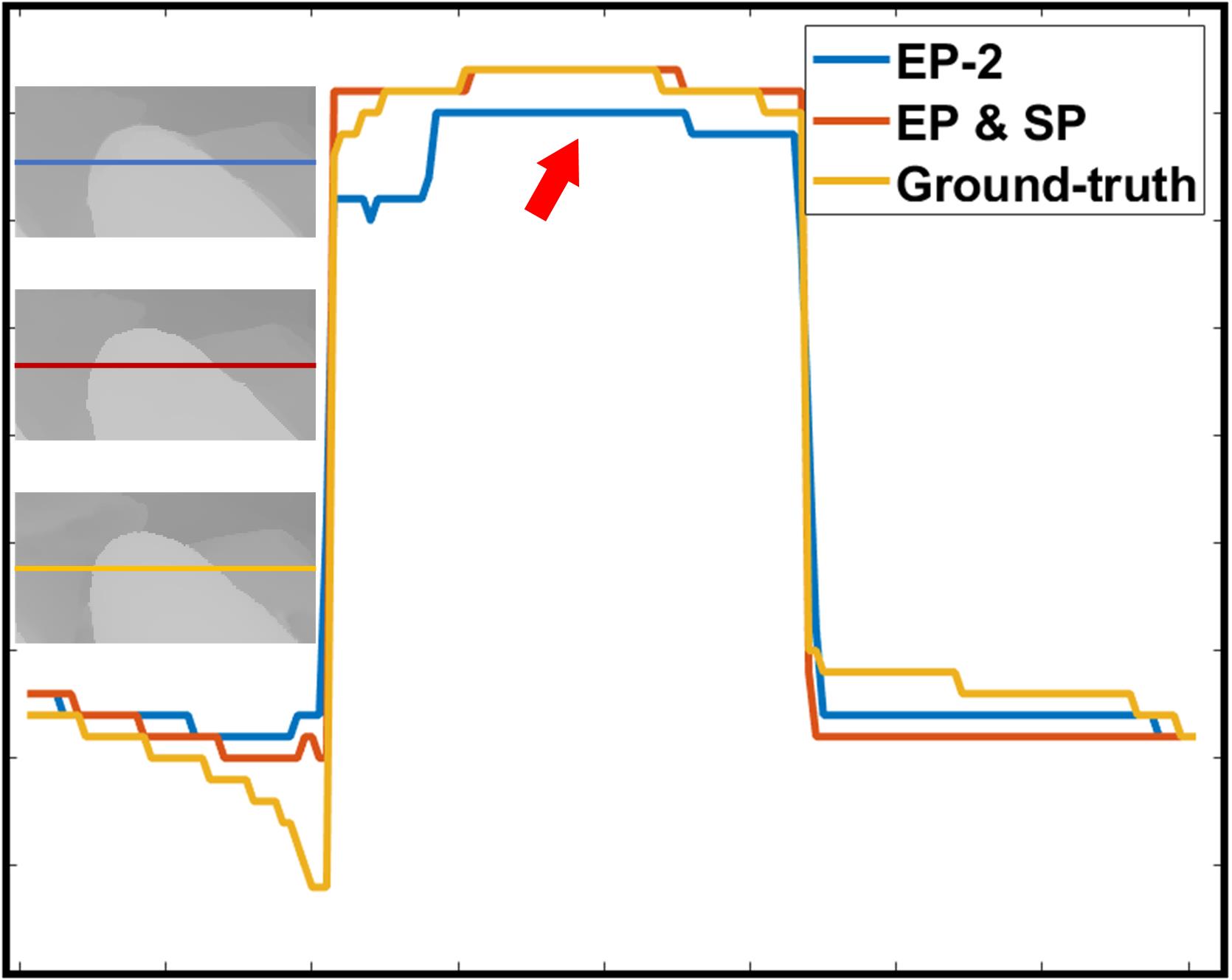} &
  \includegraphics[width=0.2\linewidth]{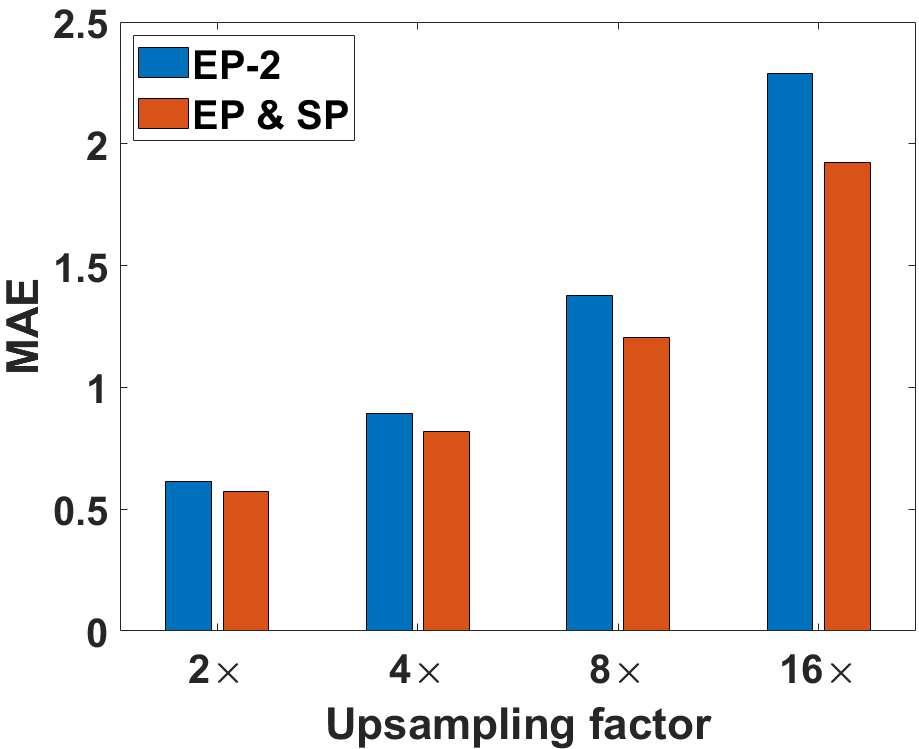} \\

  (a) ours (EP-2) & (b) ours (EP$\&$SP) & (c) 1D plot & (d) MAE comparison
  \end{tabular}
  \caption{Comparison between the EP-2 mode and the EP$\&$SP mode of our method. 8$\times$ guided depth map upsampling result of (a) our method of the EP-2 mode and (b) our method of the EP$\&$SP mode. (c) 1D plots of the labeled regions. (d) MAE comparison of different upsampling factors.}\label{FigMyVsTV}
\end{figure*}
\begin{figure*}
  \centering
  \setlength{\tabcolsep}{0.25mm}
  \begin{tabular}{cccc}
     \includegraphics[width=0.2655\linewidth]{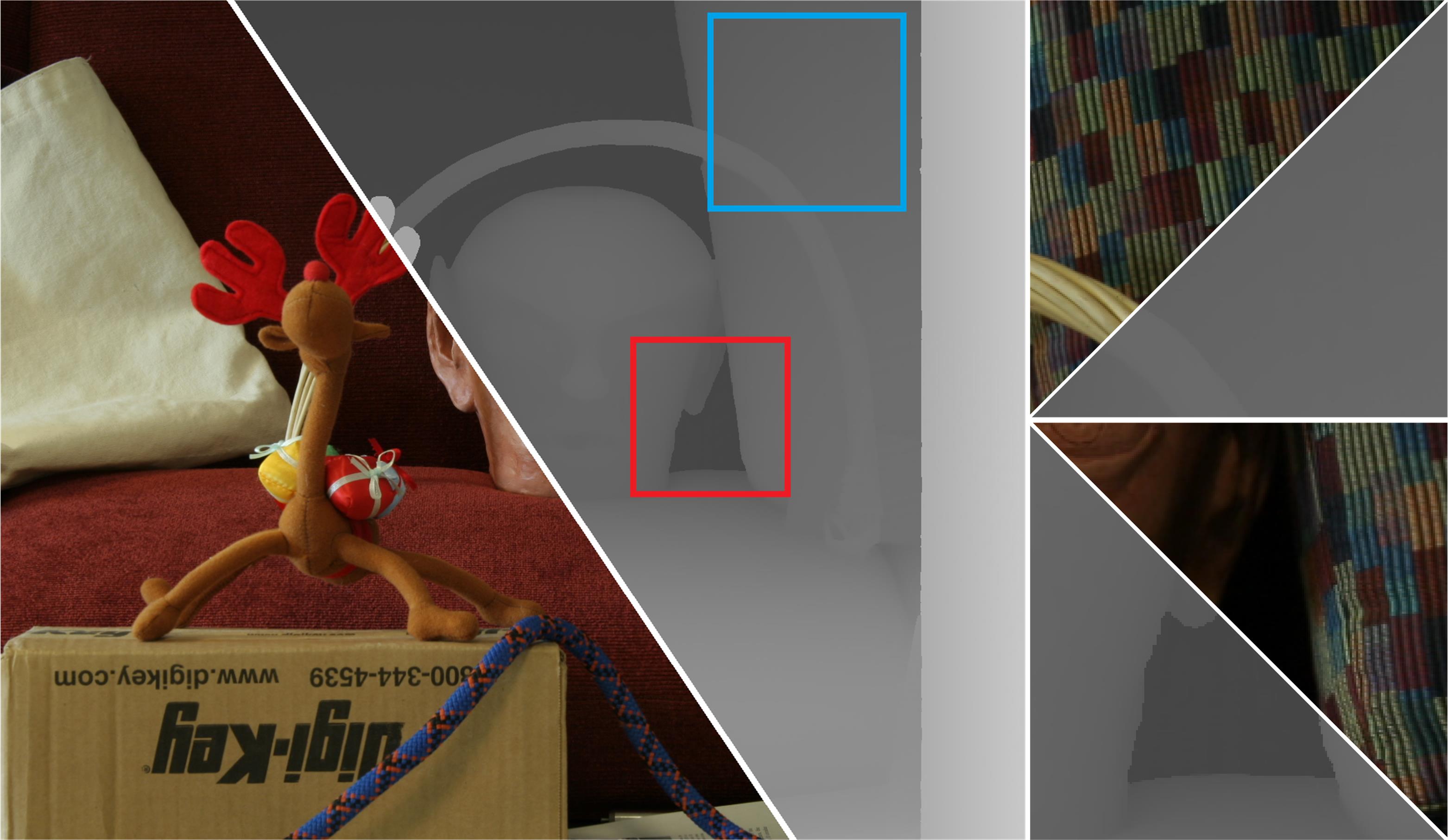} &
     \includegraphics[width=0.2655\linewidth]{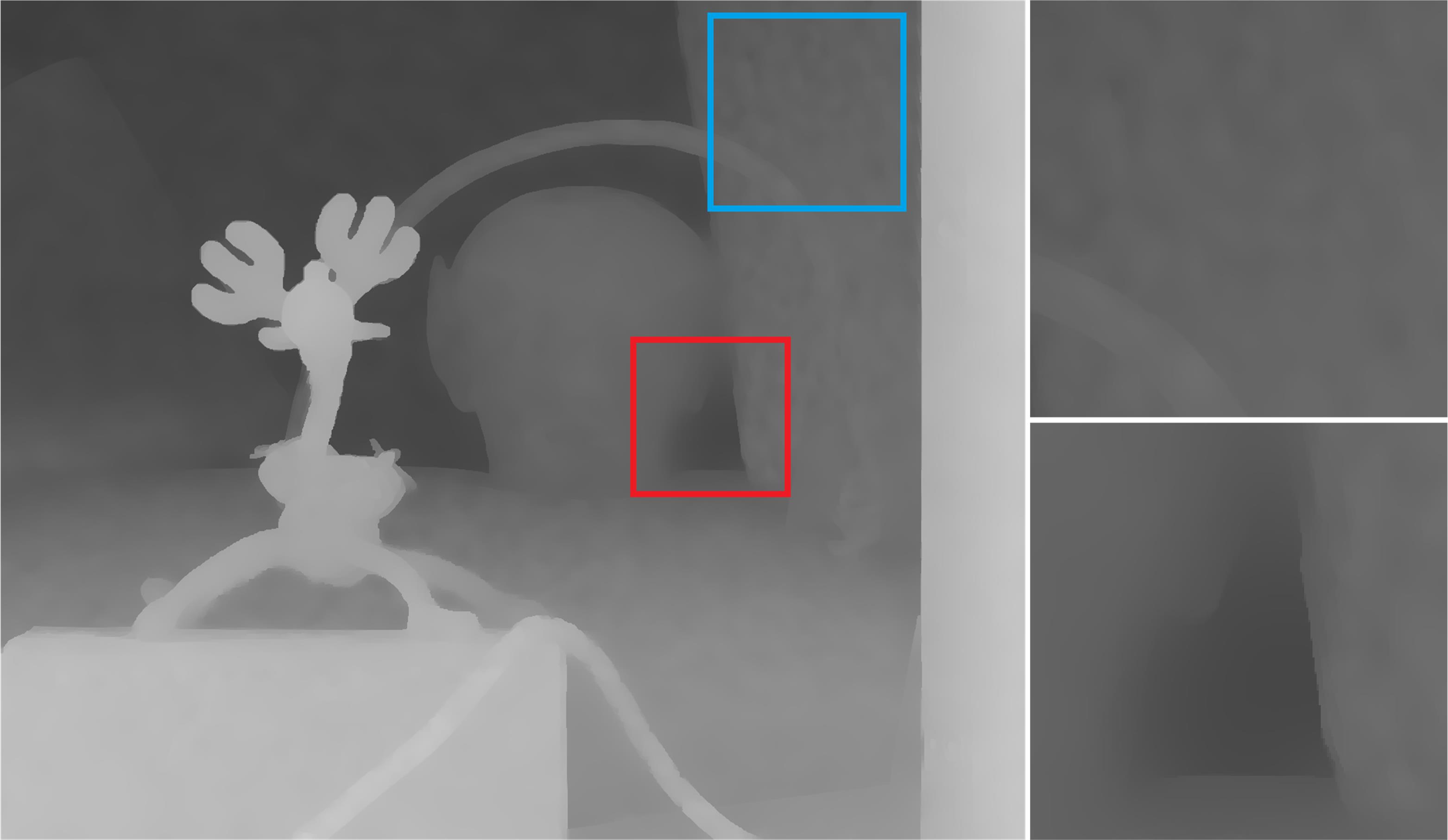} &
     \includegraphics[width=0.2655\linewidth]{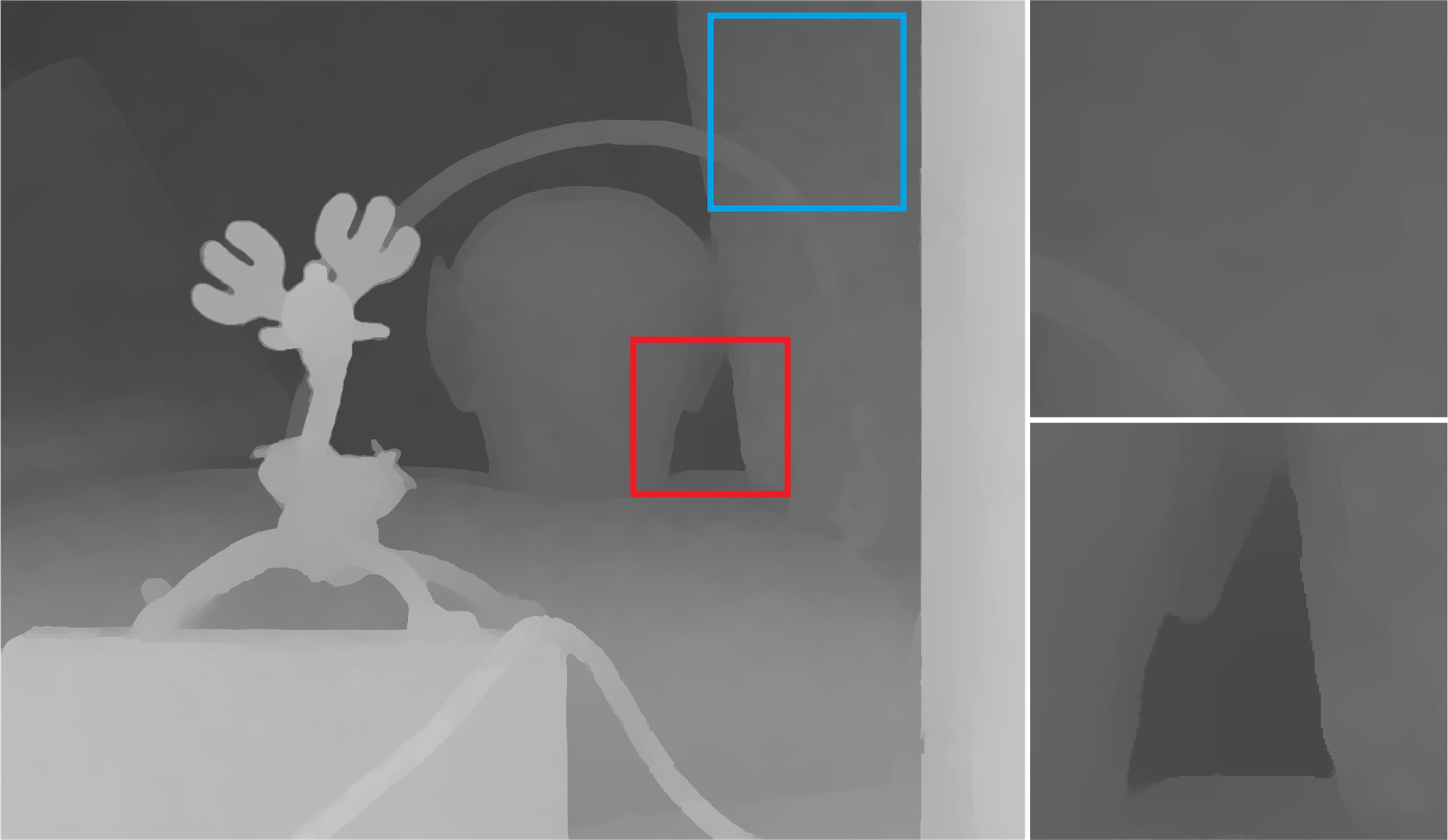} &
     \includegraphics[width=0.191\linewidth]{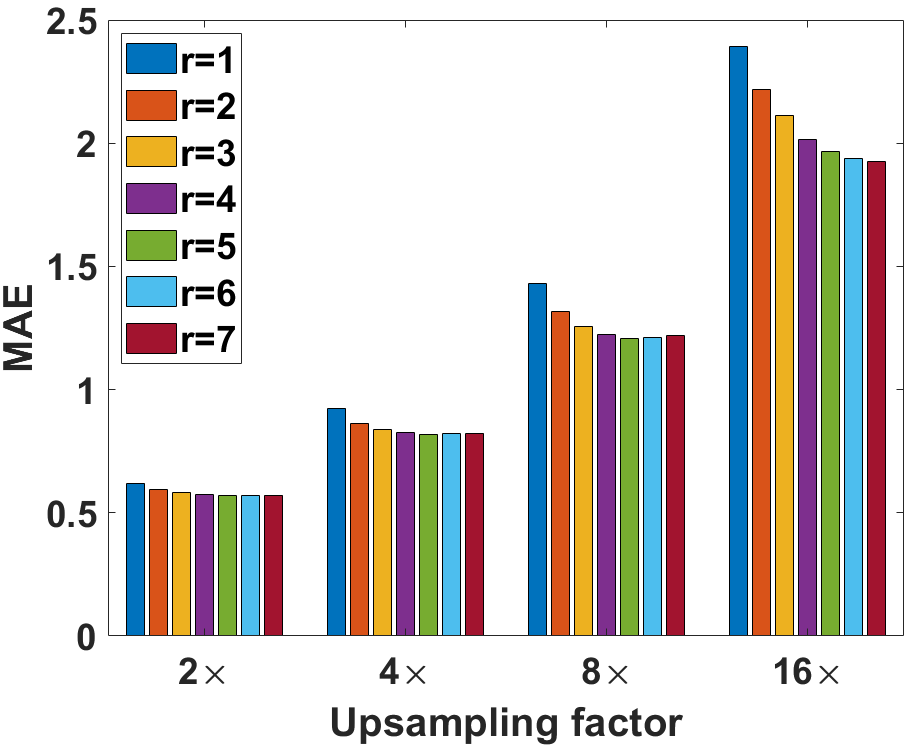}\\

     (a) color image/depth & (b) $r_d=r_s=1, \lambda=1.4$ & (c) $r_d=r_s=5, \lambda=0.5$ & (d) MAE comparison
  \end{tabular}
  \caption{Comparison of different neighborhood radiuses in our model in terms of 8$\times$ guided depth map upsampling. (a) Guidance image and ground-truth depth map. Result of our method of the EP$\&$SP mode with (b) $r_d=r_s=1,\lambda=1.4$ and (c) $r_d=r_s=5,\lambda=0.5$. $r$ represents the value of $r_d=r_s=r$. (d) MAE comparison under different neighborhood radiuses.}\label{FigRadiusLargeVsSmall}
\end{figure*}

\subsubsection{Tasks in the Fourth Group}
Besides the special cases mentioned above, our model can also achieve other smoothing behaviors under different parameter settings and better results can be further produced. To simplify the analysis in the following subsections, we first start with the tasks in the fourth group which require structure-preserving smoothing. For these tasks, the parameters are set as $a_d=\epsilon,b_d>I_m,a_s=\epsilon,b_s>I_m,r_d=r_s,\alpha=0.5,g=f$. In this case, our model has the following two advantages: first, the setting $a_d=\epsilon,b_d>I_m,a_s=\epsilon,b_s>I_m$ enables our model to have the structure-preserving property similar to that of the TV-$L_1$ model; second, the guidance weight with $\alpha=0.5,g=f$ can make our model to produce sharper edges in the results than the TV-$L_1$ model. We illustrate this with 1D smoothing results in Fig.~\ref{Fig1DComp}(a) and (b). Fig.~\ref{FigMyVsTVL1}(b) and (c) further show a comparison of image texture removal results. As shown in the figures, both the TV-$L_1$ model and our model can properly remove the small textures, however, edges in our result are much sharper than that in the result of the TV-$L_1$ model. We fix $r_d=r_s=1$. The value of $\lambda$ depends on the texture size. Larger $\lambda$ can lead larger structures to be removed. The iteration number is set as $N=10$. Our model under this parameter setting is denoted as the $\emph{SP-1 mode}$ of our method in Tab.~\ref{TabParameter}.

\subsubsection{Tasks in the First Group}

When dealing with image detail enhancement and HDR tone mapping in the first group, one way is to set the parameters so that our model can perform WLS smoothing \cite{farbman2008edge}, which is denoted the $\emph{EP-1 mode}$ of our method in Tab.~\ref{TabParameter}. The WLS smoothing is a strong baseline in the literature that performs well in handling gradient reversals and halos. However, it can also produce intensity shift in the results \cite{he2013guided, liu2020real}. On the one hand, the intensity shift can be adopted to enhance the image contrast, which is appealing in image detail enhancement, as shown in the first row of Fig.~\ref{FigMyVsWLS} (b) and (d). On the other hand, the intensity shift can lead to compartmentalization artifacts in some cases \cite{hessel2018quantitative, liu2020real}. One example of image detail enhancement is illustrated in the second row of Fig.~\ref{FigMyVsWLS} (b) and (d). Fig.~\ref{FigMyVsWLS_HDR}(a) further shows an example of HDR tone mapping.

The reason of the intensity shift in WLS smoothing (or our method of the EP-1 mode) can be obtained through the following analysis. In fact, if we set $a_d=b_d>I_m, a_s=\epsilon,b_s>I_m,r_d=0,r_s=1,\alpha=0.2,g=f$ in our model, the first iteration of Algorithm~\ref{Alg} ($N=1$) can also be considered to be equivalent to WLS smoothing. Our model under this parameter setting can be approximately viewed as a ``weighted TV'' model. At the same time, it is known that the TV model can lead to intensity shift \cite{chan2005aspects}, and this is why the WLS smoothing can result in intensity shift in the results. In contrast, the TV-$L_1$ model does not cause intensity shift \cite{chan2005aspects}. Thus, we can set the parameters of our model as follows: $a_d=\epsilon,b_d>I_m,a_s=\epsilon,b_s>I_m,r_d=r_s=1,\alpha=0.2,g=f$, which can be approximately viewed as a ``weighted TV-$L_1$'' model. Similarly, the first iteration of Algorithm~\ref{Alg} ($N=1$) can be used to smooth the input image. Note that we set $r_d=1$ instead of $r_d=0$ because our experiments show that $r_d=1$ can achieve better performance when we set $N=1$. We denote our model under this parameter setting as the $\emph{SP-2 mode}$ of our method in Tab.~\ref{TabParameter}. Our experiments show that, with a large value of $\lambda$, the amplitudes of different structures will decrease at different rates, i.e., the amplitudes of small structures can have a larger decrease than the large ones, as illustrated in Fig.~\ref{Fig1DComp}(d). At the same time, edges are neither blurred nor sharpened. These properties are desirable for image detail enhancement and HDR tone mapping. Compared with the result of WLS smoothing in Fig.~\ref{Fig1DComp}(c), edges in the result produced by our method of the SP-2 mode in Fig.~\ref{Fig1DComp}(d) are better preserved (see the bottom of the 1D signals). Fig.~\ref{FigMyVsWLS_HDR}(b) and the second row of Fig.~\ref{FigMyVsWLS}(c) and (e) further show the results of HDR tone mapping  and  image detail enhancement obtained with our method of the SP-2 mode. As shown in the figures, the compartmentalization artifacts are properly alleviated. However, we should point out that our method of the SP-2 mode is not prone to enhance image contrast, as shown in the first row of Fig.~\ref{FigMyVsWLS}(c) and (e). In this case, it is inferior to the WLS smoothing (our method of the EP-1 mode). In addition, $\lambda$ in the SP-2 mode of our method is usually much larger than that in the SP-1 mode, for example, the results in Fig.~\ref{FigMyVsWLS}(c) is generated with $\lambda=20$.

Image detail enhancement and HDR tone mapping in the first group have long been considered as edge-preserving tasks. In fact, most methods proposed for these tasks in the literature are edge-preserving smoothers \cite{farbman2008edge, he2013guided, xu2011image, tomasi1998bilateral, gastal2012adaptive, gastal2011domain, fattal2009edge}, our method of the EP-1 mode (or WLS smoothing \cite{farbman2008edge}) is also in the spirit of edge-preserving smoothing. In contrast to previous methods, our method of the SP-2 mode provides a promising alternative that handles these tasks in a structure-preserving manner and shows better performance.

\subsubsection{Tasks in the Second Group and the Third Group}

To sharpen edges that is required by the tasks in the second group and the third group, based on the analysis in Sec.~\ref{SecTruncatedHuber}, we can set $a_s=\epsilon, b_s<I_m$ in the smoothness term. For the data term, we can set the parameters as $a_d=b_d>I_m$, i.e., $L_2$ norm penalty in the data term. This results in an edge-preserving smoother, which is denoted as the $\emph{EP-2 mode}$ of our method in Tab.~\ref{TabParameter}. This kind of parameter setting for sharpening edges is in the same spirit to the SD filter \cite{ham2018robust} where a Welsch's penalty function is adopted in the smoothness term. However, as analyzed in Sec.~\ref{SecTruncatedHuber}, our truncated Huber penalty function can better preserve weak edges than the Welsch's penalty function. This can be validated through our analysis in Sec.~\ref{SecTruncatedHuber} and the comparison of the 1D smoothing results shown in Fig.~\ref{Fig1DComp}(e) and (f).

Our method of the EP-2 mode can better preserve edges than the SD filter, however, the weak edges in the large structure are also penalized, as shown in the left part of Fig.~\ref{Fig1DComp}(f). If we only need to remove the small structure in the middle left of Fig.~\ref{Fig1DComp}(f) with the rest structures properly preserved, then this will be a quite challenging case, and there are seldom existing smoothers that can properly handle this. The challenge lies on the fact that we need to both preserve large structures with weak edges and small structures with strong edges at the same time. On the one hand, if we perform edge-preserving smoothing, the large structure on the left will be penalized because the corresponding edges are weak, as shown in Fig.~\ref{Fig1DComp}(e) and (f). On the other hand, if we adopt structure-preserving smoothing, then the small structure in the middle right will be removed due to its small structure size, as shown in Fig.~\ref{Fig1DComp}(a) and (b).

In fact, we can also set the parameters of the data term as $a_d=\epsilon, b_d<I_m$. The parameter setting $a_d=a_s=\epsilon$ makes our model enjoy the structure-preserving property. At the same time, $b_s<I_m$ enables our model not to penalize large-amplitude edges, which is the edge-preserving property. In this way, our model has the simultaneous edge-preserving and structure-preserving property, which is denoted as the $\emph{EP}\&\emph{SP mode}$ of our method in Tab.~\ref{TabParameter}. The truncation $b_d<I_m$ in the data term can help our model to be robust against the outliers in the input image, for example, the noise in the no-flash image and the low-quality depth map. With the simultaneous edge-preserving and structure-preserving smoothing nature, our model is able to handle the challenging case mentioned above. Fig.~\ref{Fig1DComp}(g) shows the 1D smoothing result of our method of the EP$\&$SP mode. This challenging case is also of practical importance such as the example of clip-art compression artifacts removal shown in Fig.~\ref{FigEdgeAndStructureAwareSmooth}. Fig.~\ref{FigMyVsTV} further shows an example of guided depth map upsampling in the third group. Our method of the EP$\&$SP mode shows better performance than our method of the EP-2 mode in both edge-preserving property and quantitative evaluation, as shown in Fig.~\ref{FigMyVsTV}(c) and (d), respectively. Mean absolute errors (MAE) are adopted as the evaluation metric in Fig.~\ref{FigMyVsTV}(d). Another example can also be found in Fig.~\ref{FigNormMotivation} where our method of the EP$\&$SP mode can better preserve weak edges than our method of the EP-2 mode. Please refer to the regions labeled with the red arrows in Fig.~\ref{FigNormMotivation}(d) and (e). We adopt the EP$\&$SP mode of our method for all the tasks in the second group and the third group hereafter. We further fix $\alpha=0.5, r_d=r_s, N=10$. We empirically set $b_d=b_s=0.05I_m\sim0.2I_m$ and $r_d=r_s\geq1$ depending on the applied task and the input noise level.

The structure inconsistency issue in the third group can also be easily handled by our model. Note that $\mu_{i,j}^s$ in Eq.~(\ref{EqObjFunAuxULMu}) is computed with the smoothed image in each iteration, as formulated in Eq.~(\ref{EqMultHQCondition}). It thus can reflect the inherent natures of the smoothed image. The guidance weight $\omega_{i,j}$ can provide additional structural information from the guidance image $g$. This means that $\mu_{i,j}^s$ and $\omega_{i,j}$ can complement each other. In fact, the equivalent guidance weight of Eq.~(\ref{EqObjFunAuxULMu}) in each iteration is $\mu_{i,j}^s\omega_{i,j}$, which can reflect the property of both the smoothed image and the guidance image. In this way, it can properly handle the structure inconsistency problem to avoid blurring edges and texture copy artifacts. Similar ideas were also adopted in previous approaches \cite{ham2018robust, liu2017robust}.

\begin{figure*}
  \centering
  \setlength{\tabcolsep}{0.25mm}
  \begin{tabular}{cccc}
     \includegraphics[width=0.19\linewidth]{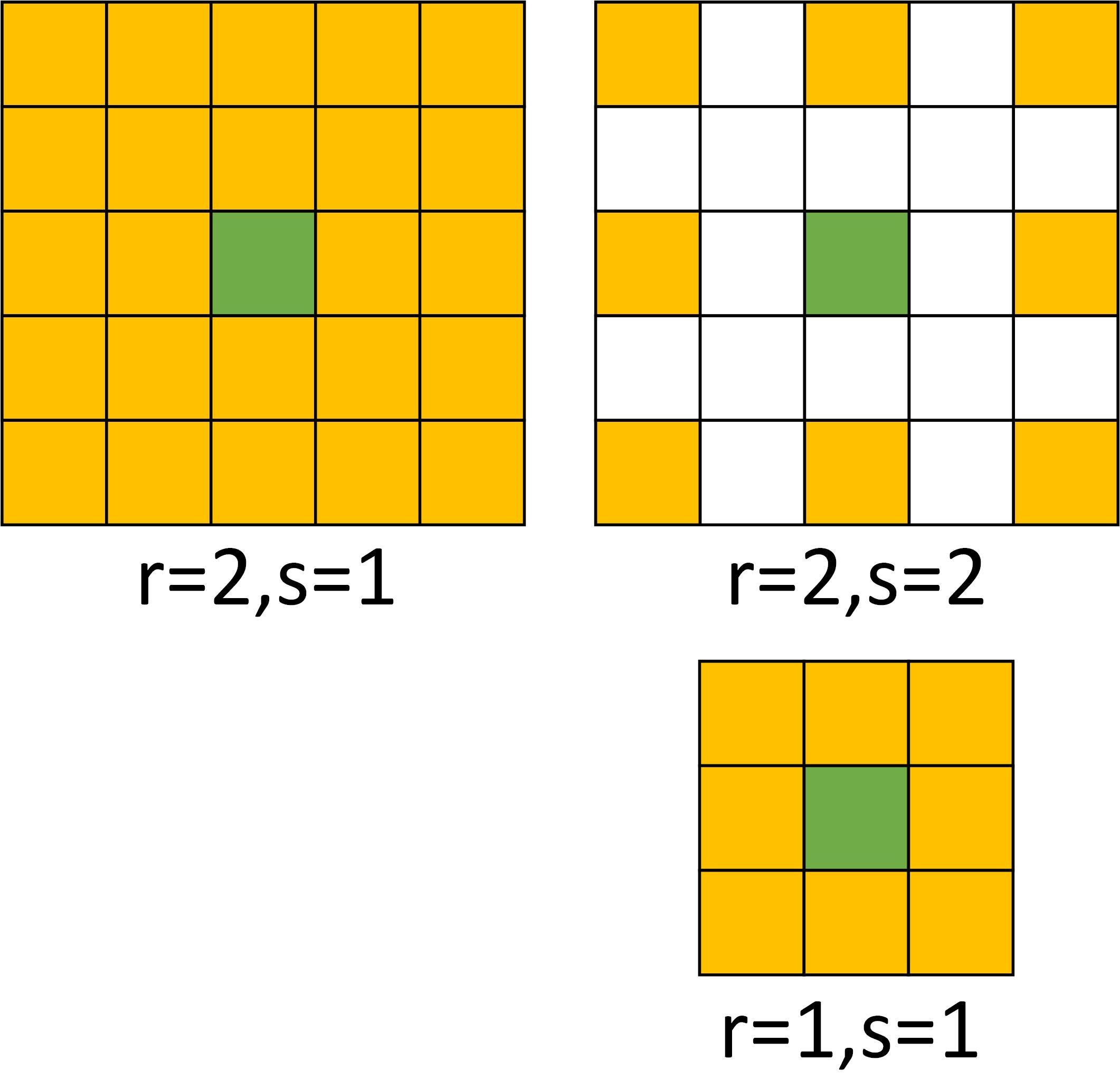} &
     \includegraphics[width=0.35\linewidth]{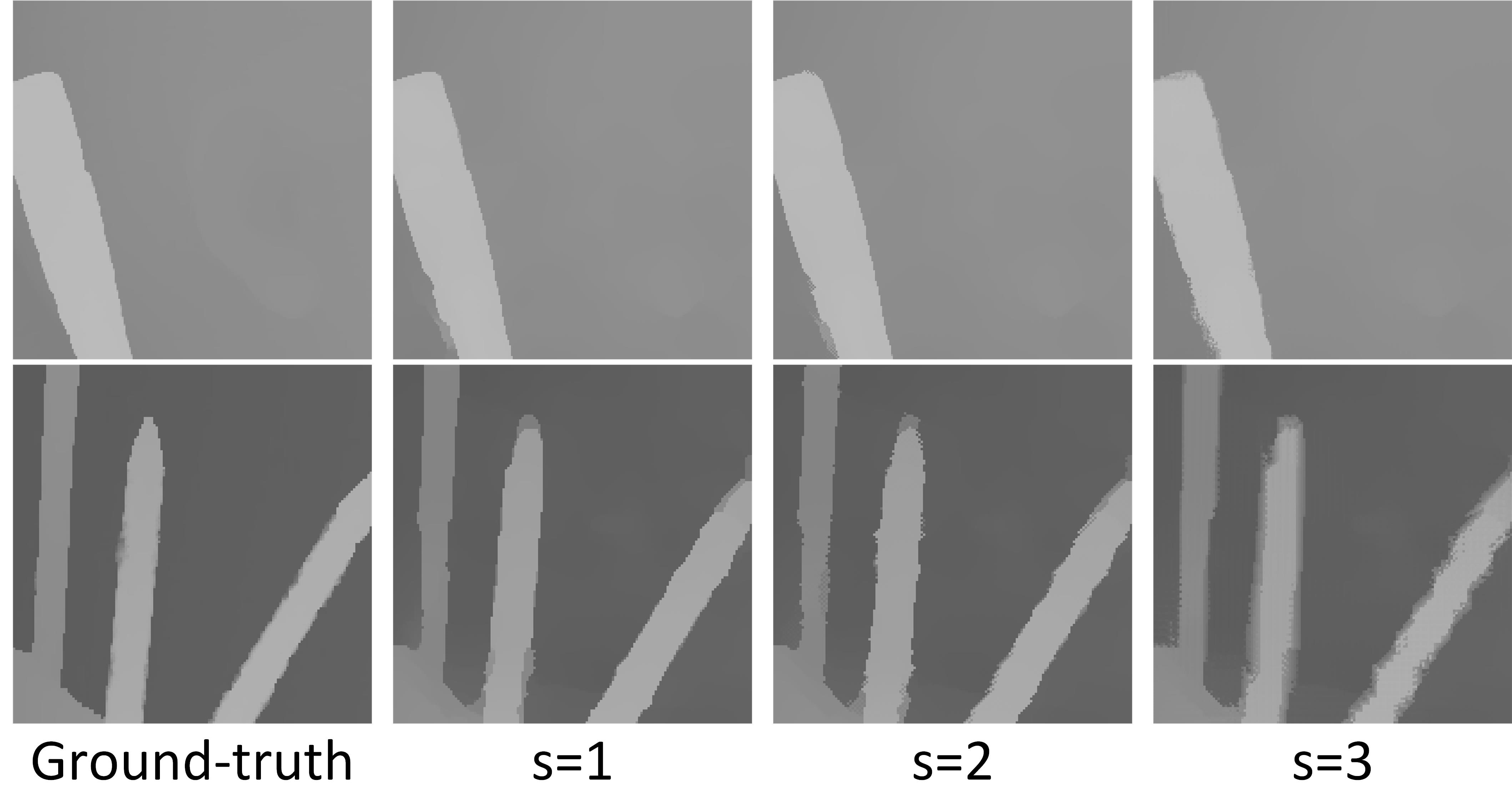} &
     \includegraphics[width=0.225\linewidth]{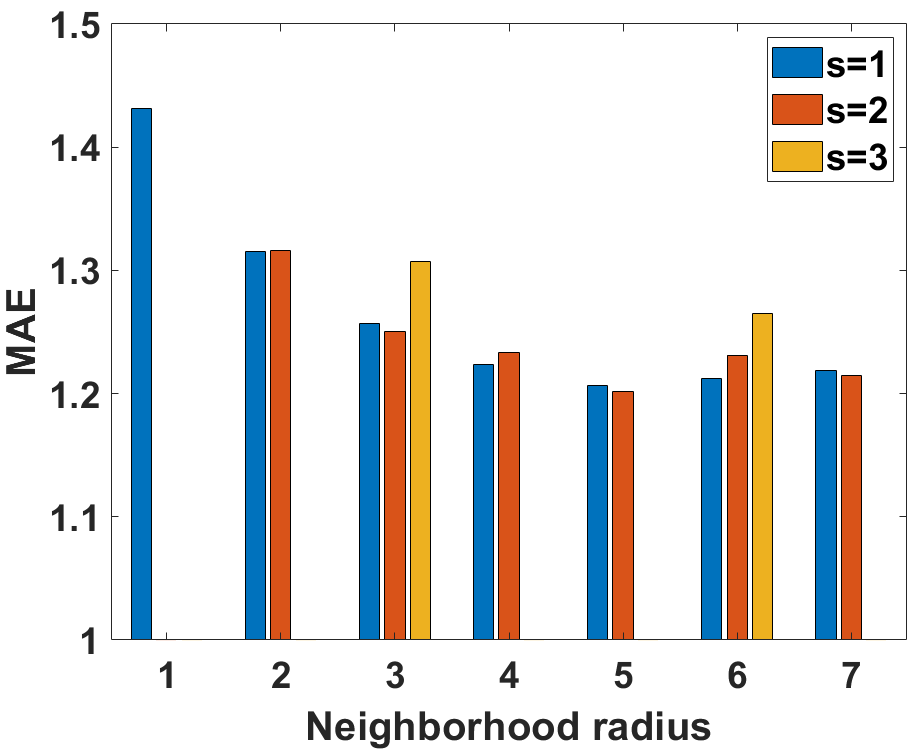} &
     \includegraphics[width=0.225\linewidth]{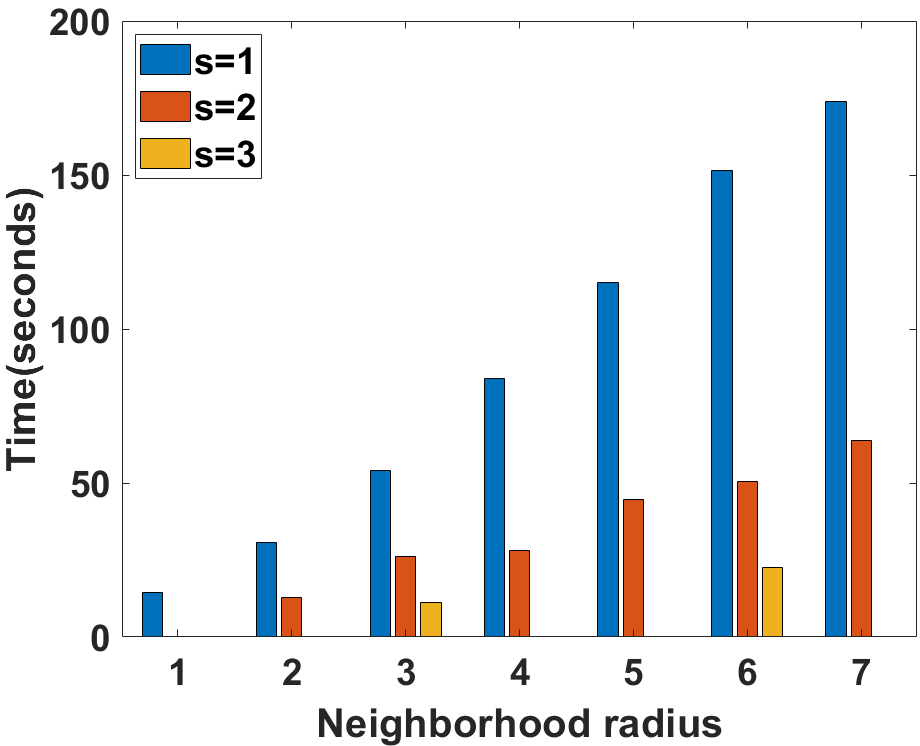}\\

     (a) {\footnotesize dilated neighborhood} & (b) upsampled depth map & (c) MAE comparison & (d) time comparison
  \end{tabular}
  \caption{Comparison of different strides in the dilated neighborhood of our model. (a) Illustration of the dilated neighborhoods with different radiuses and strides. (b) Guided depth map upsampling results obtained with different strides. (c) MAE comparison of 8$\times$ umsampling results obtained with different neighborhood radiuses and strides. (d) Computational cost comparison between different neighborhood radiuses and strides. The size of the input image is $1088\times1376$. All the results in (b)$\sim$(d) are obtained with our method of the EP$\&$SP mode.}\label{FigSparseNeighbor}
\end{figure*}
\begin{figure}
  \centering
  \setlength{\tabcolsep}{0.25mm}
  \begin{tabular}{c}
     \includegraphics[width=0.8\linewidth]{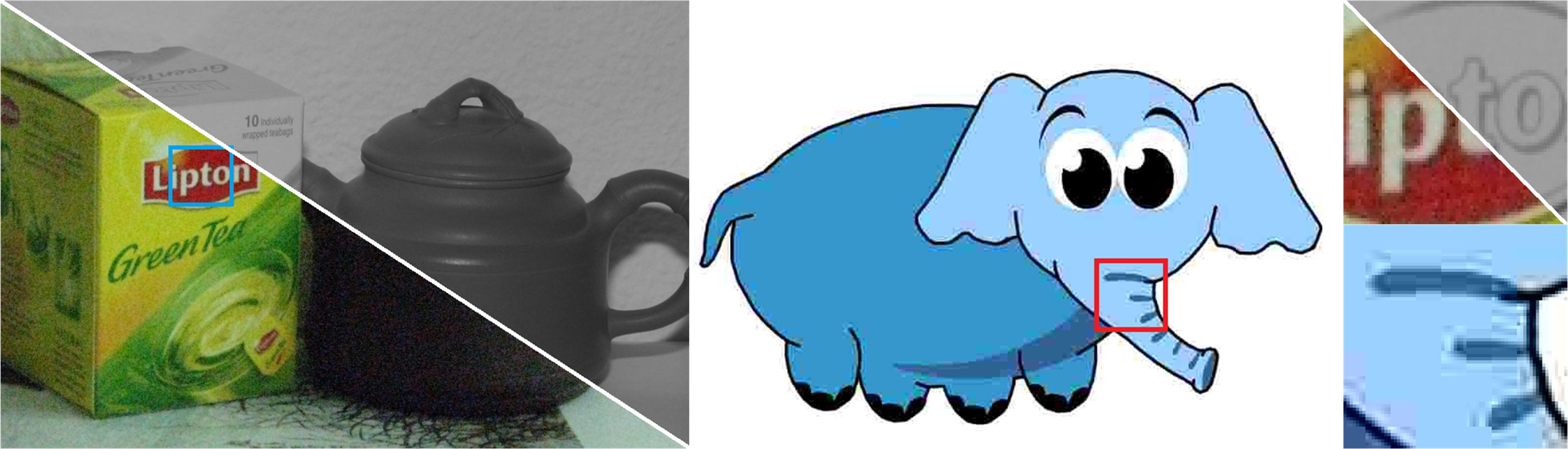} \\
     (a) input
  \end{tabular}
  \begin{tabular}{ccc}
     \includegraphics[width=0.26\linewidth]{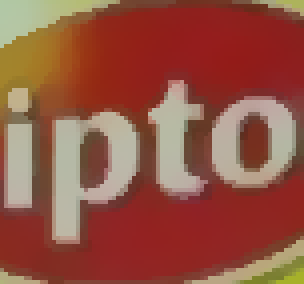} &
     \includegraphics[width=0.26\linewidth]{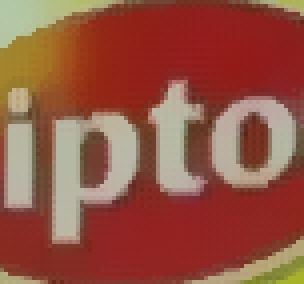} &
     \includegraphics[width=0.26\linewidth]{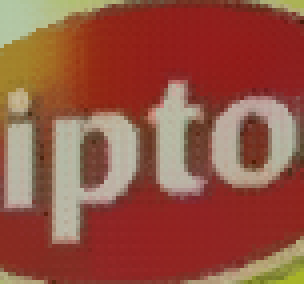} \\
     \includegraphics[width=0.26\linewidth]{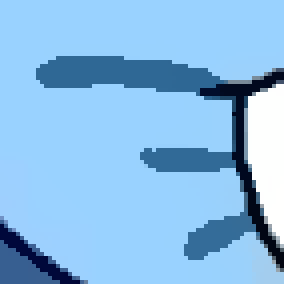} &
     \includegraphics[width=0.26\linewidth]{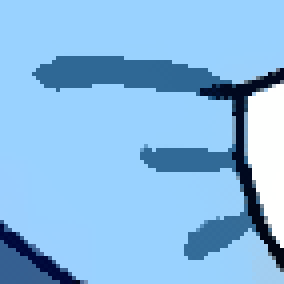} &
     \includegraphics[width=0.26\linewidth]{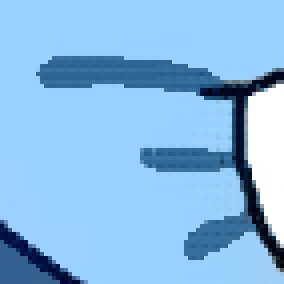} \\
     (b) s=1 & (c) s=2 & (d) s=3\\
  \end{tabular}
  \caption{Comparison of different strides in the dilated neighborhood of our model. (a) Input of flash/no flash filtering on the left and clip-art compression artifacts removal on the right. Smoothing results of different strides are highlighted in (b)$\sim$(d). }\label{FigSparseNeighborOtherApp}
\end{figure}

\subsection{Dilated Non-local Neighborhood for Computational Cost Reduction}
\label{SecDilatedNeighbor}

In the EP$\&$SP mode of our method, we set the neighborhood radius as $r_d=r_s=1\sim7$. When the input noise level is high, we usually adopt a non-local neighborhood system where the neighborhood radius is larger than 1. This is different from most previous methods that adopt a 4-connected/8-connected neighborhood system \cite{ham2018robust} or only $x$-axis and $y$-axis gradients \cite{farbman2008edge, xu2011image, xu2012structure, liu2020real} in the smoothness term. Both the neighborhood radius and $\lambda$ in our model can be used to control the smoothing strength. We show that a larger neighborhood radius with a smaller $\lambda$ can yield better performance than a smaller neighborhood with a larger $\lambda$ especially for heavy input noise. Fig.~\ref{FigRadiusLargeVsSmall} shows examples of guided depth map upsampling where a larger neighborhood can better preserve edges and smooth noise than a smaller neighborhood. The performance improvement is especially noticeable for large upsampling factors (heavy input noise) as shown in Fig.~\ref{FigRadiusLargeVsSmall}(d).

A larger neighborhood can yield better performance, however, this is also achieved at the expense of higher computational cost, as illustrated in Fig.~\ref{FigSparseNeighbor}(d). This is due to the reason that with the increasing of the neighborhood radius, more neighbor pixels are involved, which leads to the increase of the computational cost. Inspired by the dilated convolution \cite{yu2015multi, long2015fully} in the recent deep learning technique, we can adopt a \emph{dilated neighborhood} to reduce the computational cost. The dilated neighborhood of pixel $i$ is defined as:
\begin{equation}\label{EqDilatedNeighbor}
\small
   N^D(i)=\{j| j=i - r + t\ast s; t=0, 1,\cdots,\lfloor\frac{2*r + 1}{s}\rfloor \},
\end{equation}
where $s$ is denoted as the \emph{stride} of $N^D(i)$ and $r$ is the neighborhood radius. When $s=1$, $N^D(i)$ is the same as the original neighborhood defined in our model. Fig.~\ref{FigSparseNeighbor}(a) illustrates examples of the dilated neighborhood. As shown in the figure, the parameter setting of $r=2,s=2$ has the same long range connection between neighbor pixels as $r=2,s=1$ does, however, the number of involved pixels is greatly reduced, which is the same as that of $r=1,s=1$. With much fewer neighbor pixels involved, the computational cost is thus reduced. According to our experiments, for a range of neighborhood radiuses, $s=2$ is able to achieve very close performance to $s=1$ with the computational cost greatly reduced, as shown in Fig.~\ref{FigSparseNeighbor}(b)$\sim$(d). Further increasing the value of the stride can lead to blurring edges and noticeable performance drop, as shown in Fig.~\ref{FigSparseNeighbor}(b) and (c). Note that the measured time is for our method of the EP$\&$SP mode where the iteration number is $N=10$. For other modes of our method with the iteration number of $N=1$, the required computing time is only around $\frac{1}{10}$ of that shown in Fig.~\ref{FigSparseNeighbor}(d). More examples of other tasks are illustrated in Fig.~\ref{FigSparseNeighborOtherApp}.

\section{Applications and Experimental Results}
\label{SecExperiments}

Our method is applied to various tasks in the first to the fourth groups to validate the effectiveness. Comparisons with the state-of-the-art approaches in each application are also presented. When adopting our method for image smoothing, the intensity values of the input image are first normalized into range $[0, 1]$ before the smoothing, they are then normalized back to their original range for quantitative evaluation. For the parameters that are not listed in the caption of our result in each figure, their values are fixed and shown in Tab.~\ref{TabParameter}. For all the compared methods, we adopt either their default parameters or the ones proposed in their papers during the smoothing, and the ones that achieve better performance are used to produce results for quantitative comparison.

\begin{table*}[!ht]
\centering
\caption{Quantitative evaluation of HDR tone mapping results. The best results are in \textbf{bold}. The second best results are \underline{underlined}.}\label{TabHDR}
\newcommand{\tabincell}[2]{\begin{tabular}{@{}#1@{}}#2\end{tabular}}
\resizebox{1.005\textwidth}{!}
{
\begin{tabular}{c|ccccccccccccc}
\Xhline{1.2pt}

& AMF \cite{gastal2012adaptive} & BLF \cite{tomasi1998bilateral} & DTF-NC \cite{gastal2011domain} & DTF-IC \cite{gastal2011domain} & DTF-RF \cite{gastal2011domain} & EAW \cite{fattal2009edge} & FGS \cite{min2014fast} & GF \cite{he2013guided} & $L_0$ norm \cite{xu2011image} & SG-WLS \cite{liu2017semi} & SWF \cite{yin2019side} & \tabincell{c}{Ours (EP-1)\\ /WLS \cite{farbman2008edge}} & Ours(SP-2)\\

\Xhline{1.2pt}

Naturalness & \underline{0.4511} & 0.4401 & 0.4231 & 0.4191 & 0.4175 & 0.3998 & 0.4408 & 0.4197 & 0.4026 &  0.4420 & 0.4207 & 0.4463 & \textbf{0.4703}\\
Fidelity & 0.8257 & 0.8227 & 0.8263 & 0.8015 & 0.8152 & 0.8338 & 0.8398 & 0.8175 & 0.8265 & 0.8425 & 0.8047 &  \textbf{0.8461} & \underline{0.8451} \\
TMQI & 0.8614 & 0.8586 & 0.8570 & 0.8488 & 0.8525 & 0.8541 & 0.8641 & 0.8536 & 0.8534 & 0.8653 & 0.8496 & \underline{0.8666} & \textbf{0.8712}\\

\Xhline{1.2pt}
\end{tabular}
}
\end{table*}

\begin{table*}[!ht]
\centering
\caption{Quantitative comparison of clip-art compression artifacts removal results. The best results are in \textbf{bold}. The second best results are \underline{underlined}.}\label{TabClipArt}

\resizebox{1\textwidth}{!}
{
 \begin{tabular}{c|ccccccccc|ccccccccc}
  \Xhline{1.2pt}
  \multicolumn{1}{c}{\multirow{2}{*}{}} & \multicolumn{9}{|c|}{{PSNR}} & \multicolumn{9}{c}{{SSIM}} \\
  \hline
   compression quality & 10 & 20 & 30 & 40 & 50 & 60 & 70 & 80 & 90 & 10 & 20 & 30 & 40 & 50 & 60 & 70 & 80 & 90\\
  \Xhline{1.2pt}

  JPEG  & 31.15 & 33.22 & 34.40 & 35.36 & 36.10 & 36.92 & 37.85 & 39.04 & 40.45 & 0.9482  &  0.9622  &  0.9685  &  0.9692  &  0.9757  &  0.9784  &  0.9796  &  0.9874  &  0.9930 \\
  Wang \cite{wang2006deringing} & 31.27 & 33.30 & 34.47 & 35.39 & 36.04 & 36.77 & 37.69 & 38.55 & 39.48 &  0.9518   & 0.9664  &  0.9725   & 0.9729   & 0.9792    & 0.9811   &  0.9823 &   0.9888   &  0.9923\\
  BTF \cite{cho2014bilateral}& \underline{32.12} &  \underline{34.39} & 35.75 & 36.73 & 37.44 & 38.18 & 38.94 & 39.68 & 40.47 &  \underline{0.9657}  &   0.9759 &   0.9789  &  0.9814 &    \underline{0.9843}  &   \underline{0.9866}   &  0.9879   &  \underline{0.9923}  &  0.9941\\
  $L_0$ norm \cite{xu2011image} & 31.16 & 32.81 & 34.86 & 35.85 & 36.50 & 37.47 & 38.26 & 39.26 & 40.31 & 0.9522  &  0.9639 &   0.9743  &  0.9796  &  0.9811    & 0.9837    & 0.9851  &  0.9886  &  0.9902\\
  Region Fusion \cite{nguyen2015fast} & 31.91 & 34.37 &  \underline{35.92} &  \underline{37.29} &  \underline{38.26} & \underline{39.37} & \underline{40.55} &  \textbf{41.98} &  \textbf{42.34} & 0.9626  &   \underline{0.9761}   &    \underline{0.9793}  &  \underline{0.9815} &   0.9833    & 0.9852   &   \underline{0.9882}   &  0.9905   &   \underline{0.9947}\\
  deep prior \cite{ulyanov2018deep} & 28.52 & 28.88 & 29.04 & 29.26 & 29.32 & 29.38 & 29.52 & 29.61 & 29.88 & 0.9401 & 0.9437 & 0.9459 & 0.9468 & 0.9483 & 0.9497 & 0.9516 & 0.9529 & 0.9544\\
  Ours &  \textbf{33.56} &  \textbf{35.63} &  \textbf{37.21} &  \textbf{38.31} &  \textbf{38.95} &  \textbf{39.91} &  \textbf{40.81} & 41.61 & 42.14 & \textbf{0.9823}  &   \textbf{0.9878}  &    \textbf{0.9883}  &   \textbf{0.9910}  &  \textbf{0.9928}   &   \textbf{0.9933} &   \textbf{0.9947} &  \textbf{0.9962}  &  \textbf{0.9973}\\
  \Xhline{1.2pt}
  \end{tabular}
 }
 \end{table*}

\begin{figure*}[!ht]
  \centering
  \setlength{\tabcolsep}{0.5mm}
  \begin{tabular}{cccc}
  \includegraphics[width=0.23\linewidth]{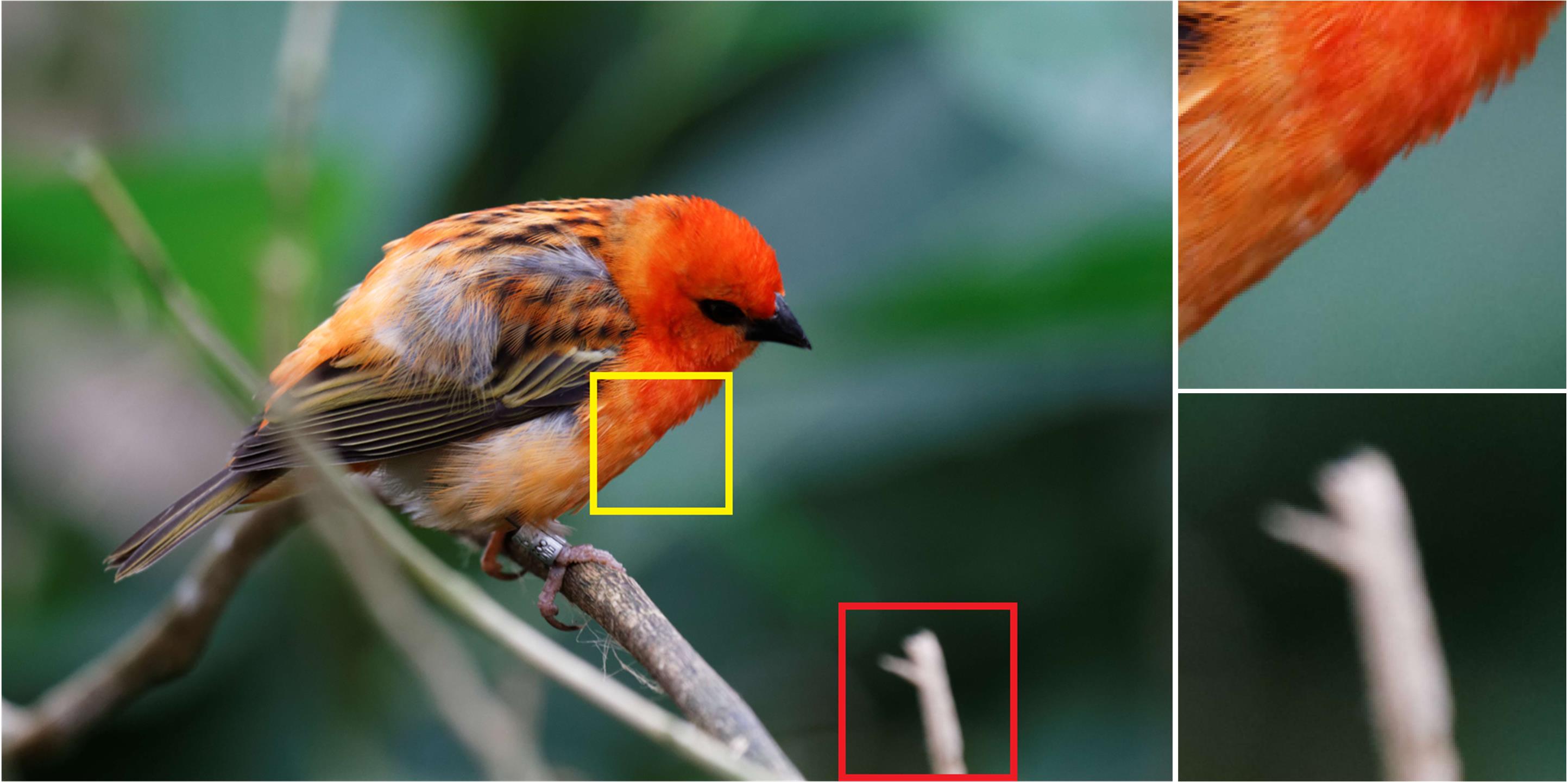} &
  \includegraphics[width=0.23\linewidth]{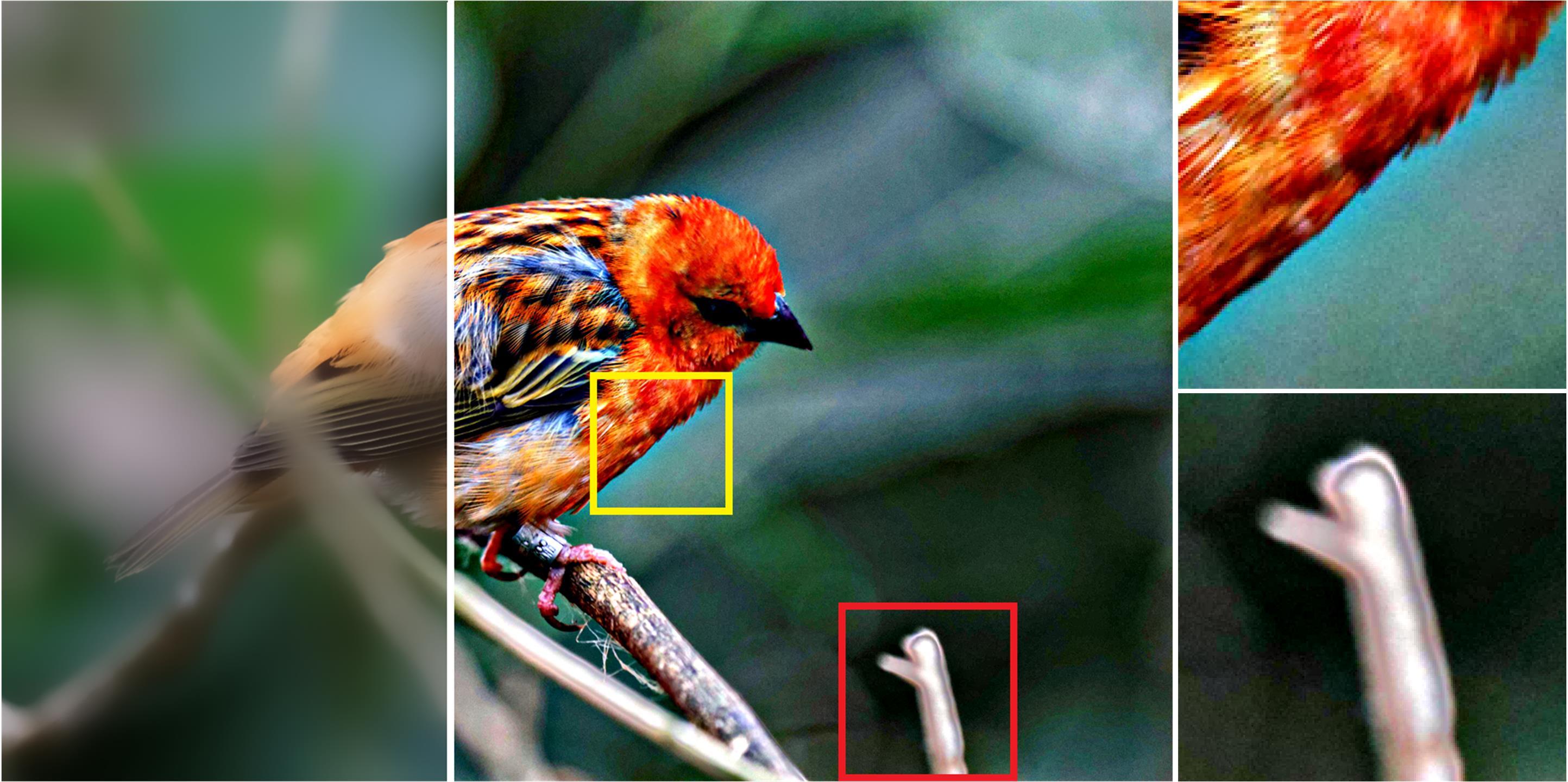} &
  \includegraphics[width=0.23\linewidth]{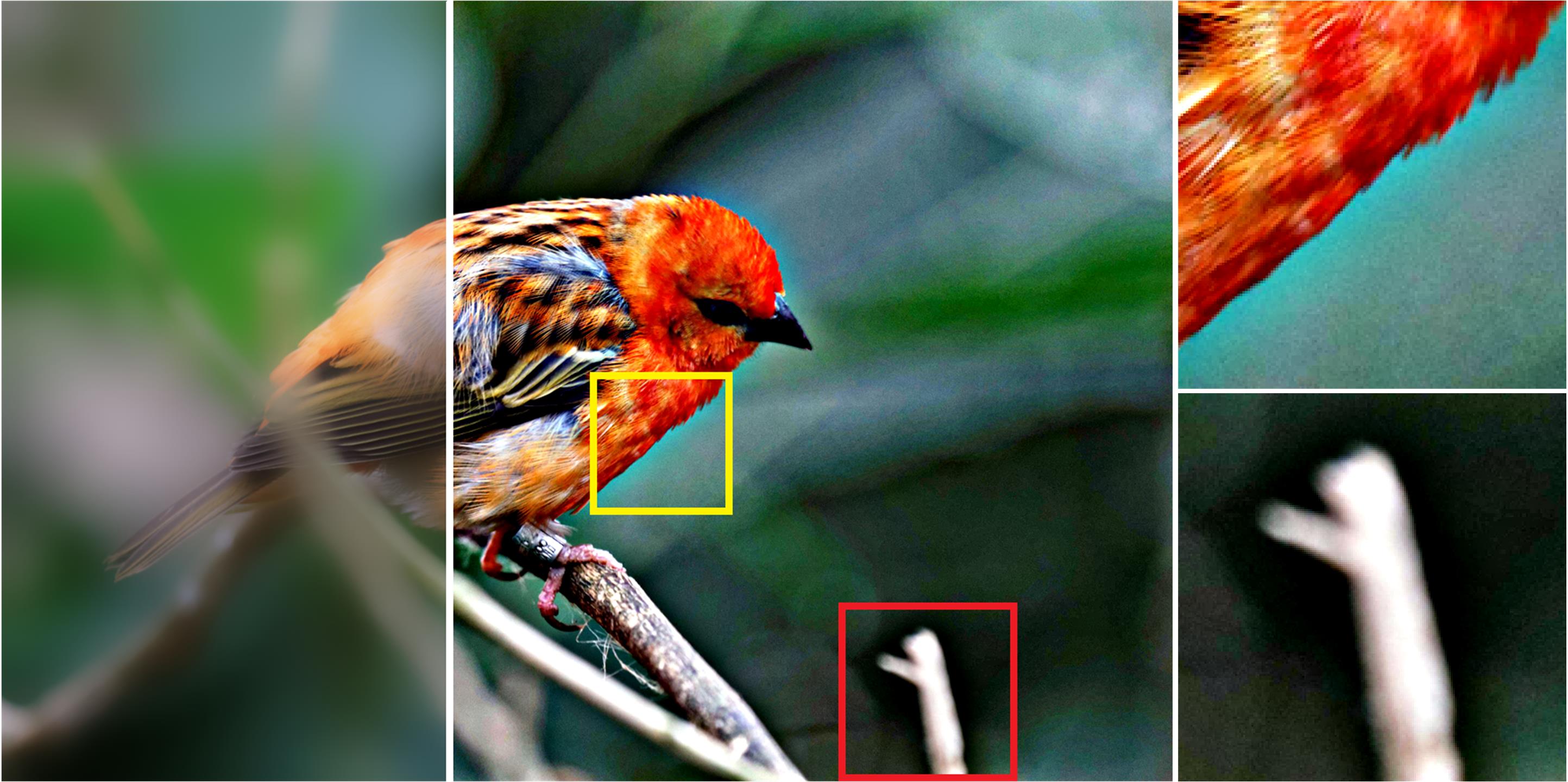} &
  \includegraphics[width=0.23\linewidth]{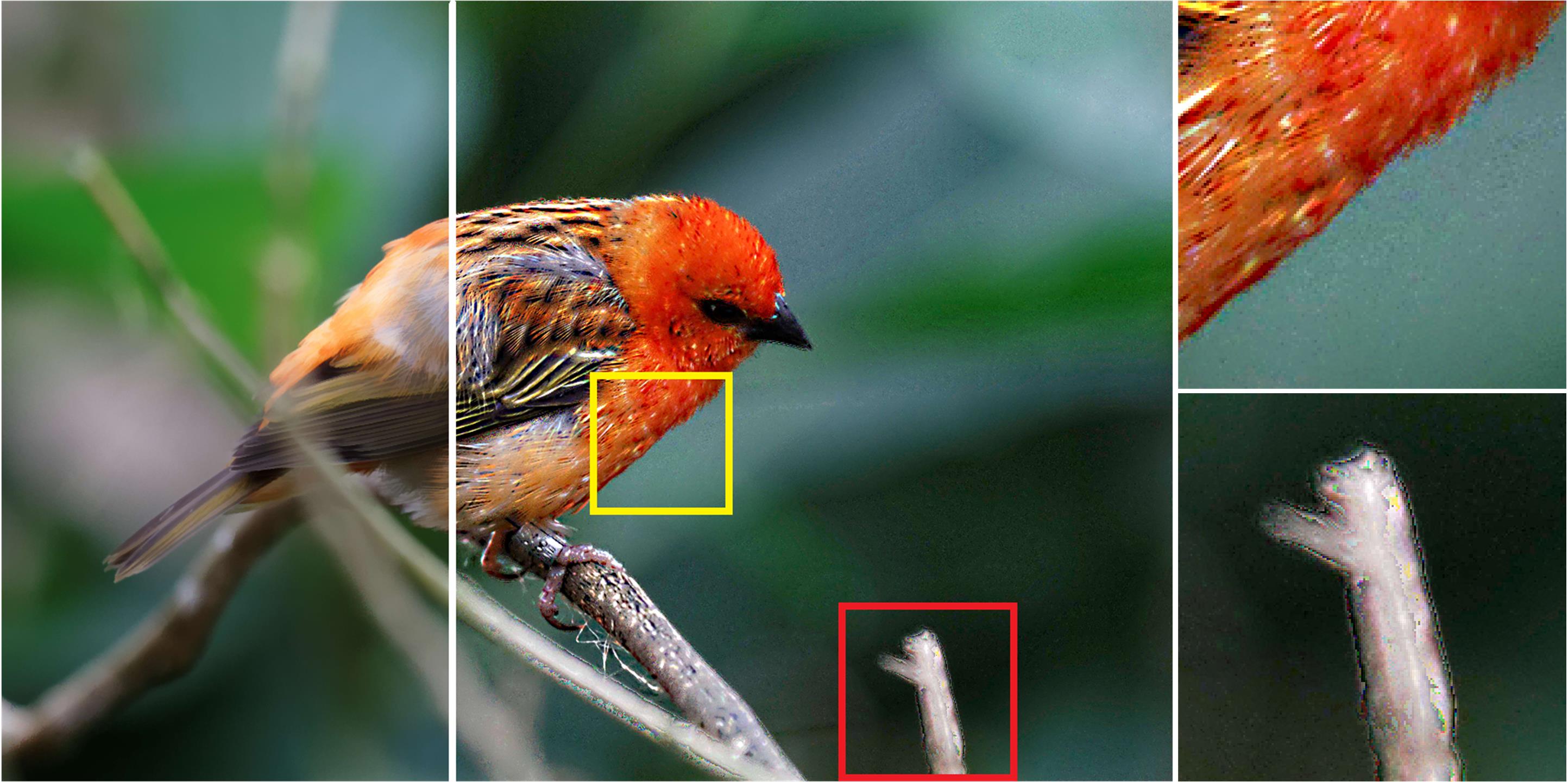} \\
  (a) input & (b) AMF & (c) GF & (d)  SWF \\

  \includegraphics[width=0.23\linewidth]{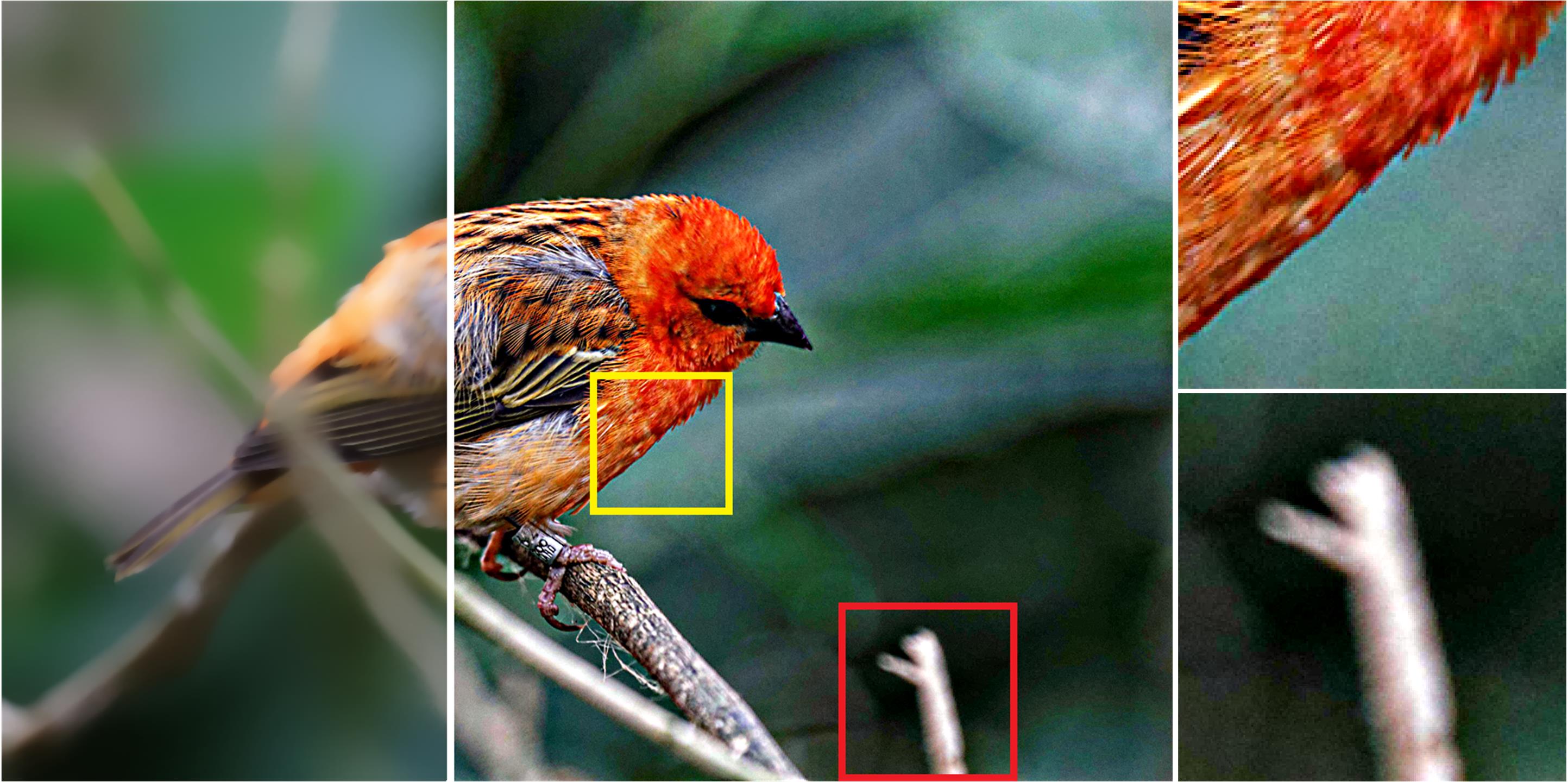} &
  \includegraphics[width=0.23\linewidth]{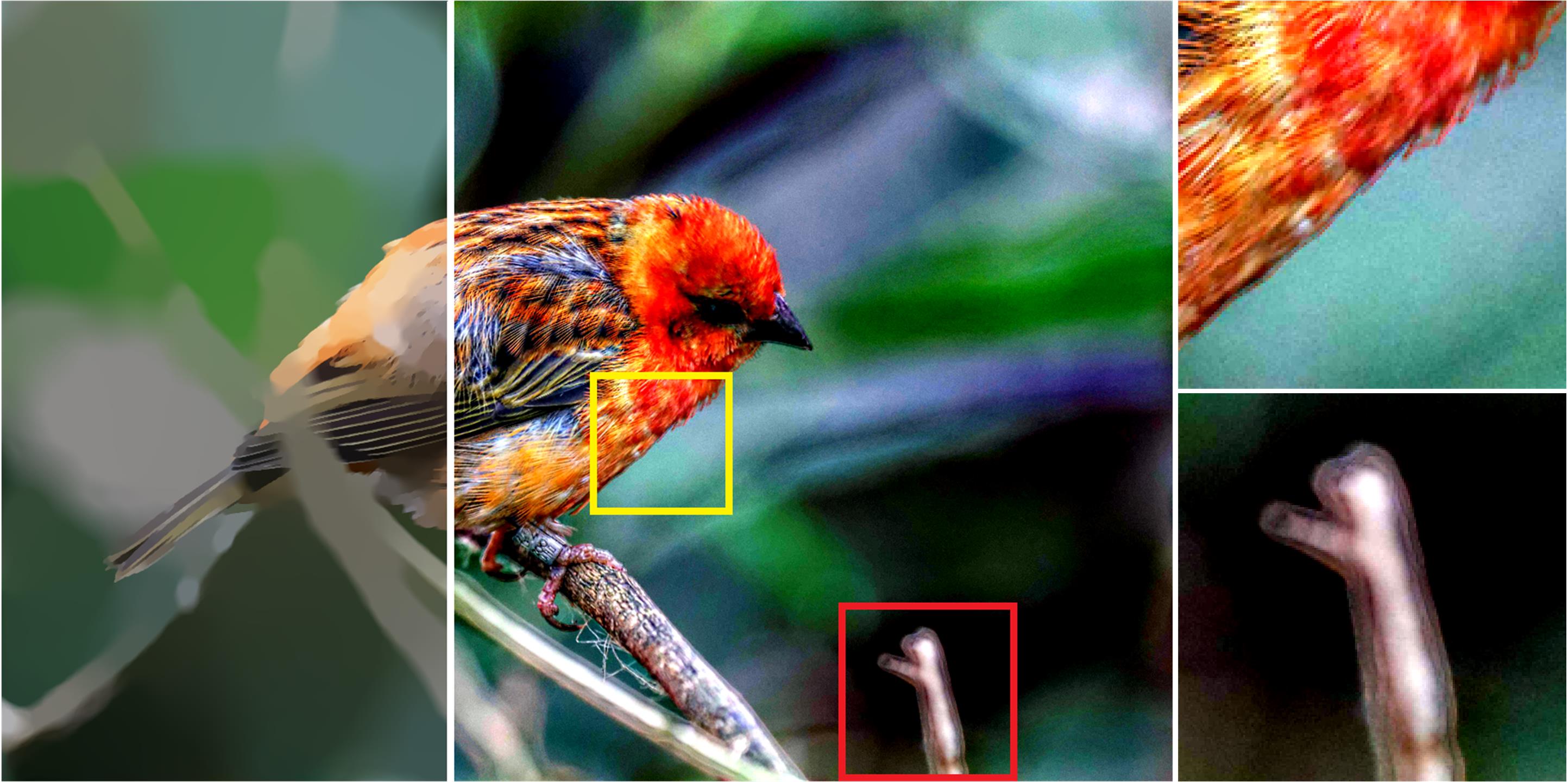} &
  \includegraphics[width=0.23\linewidth]{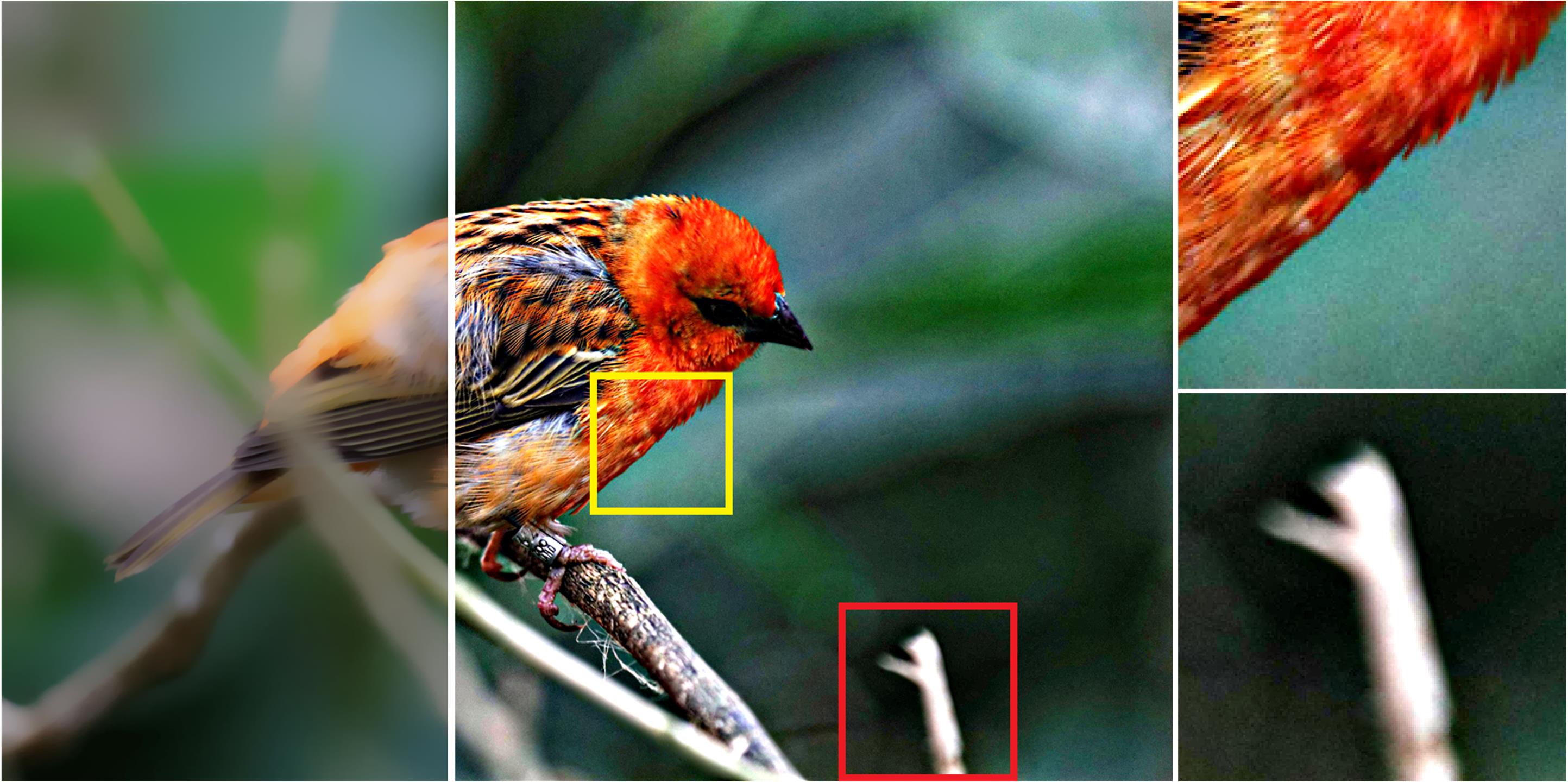} &
  \includegraphics[width=0.23\linewidth]{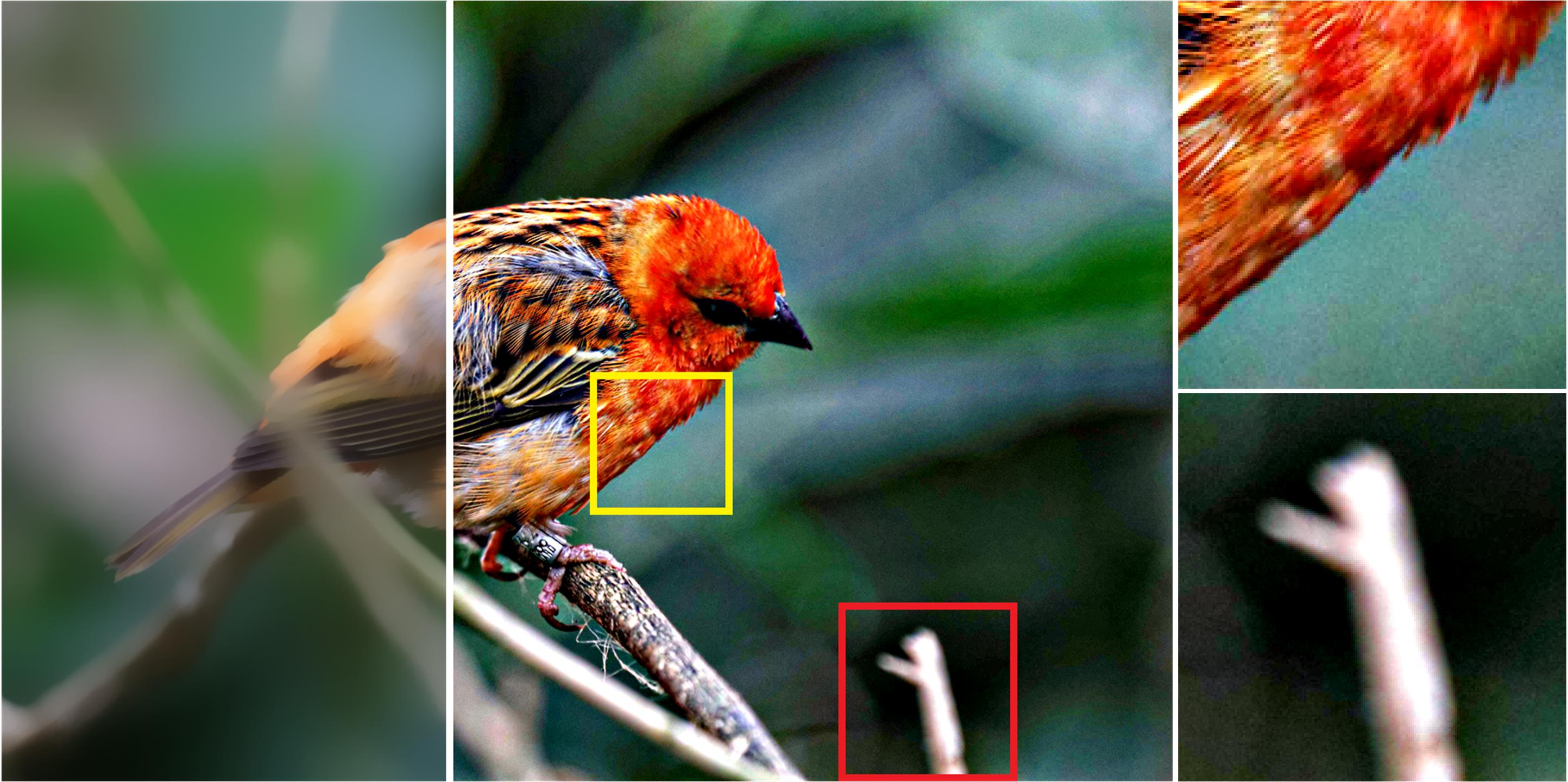}\\
   (e) SG-WLS & (f) $L_0$ norm & (g) WLS/ours(EP-1) & (h) ours(SP-2) \\

  \end{tabular}
  \caption{Image detail enhancement results of different approaches. (a) Input image. Smoothed image and 3$\times$ detail enhanced image of (b) AMF \cite{gastal2012adaptive} ($\sigma_s=20,\sigma_r=0.25$), (c) GF \cite{he2013guided} ($r=20,\epsilon=0.15^2$), (d) SWF \cite{yin2019side} ($r=7, iteration=1$), (e) SG-WLS \cite{liu2017semi} ($r=2, \tau=1, \lambda=50$), (f) $L_0$ norm smoothing \cite{xu2011image} ($\lambda=0.02$), (g) our method of the EP-1 mode/WLS \cite{farbman2008edge} ($\lambda=1,\alpha=1.2$) and (h) our method of the SP-2 mode ($\lambda=20$).}\label{FigDetailEnhancement}
\end{figure*}
\begin{figure*}[!ht]
  \centering
  \setlength{\tabcolsep}{0.5mm}
  \begin{tabular}{cccc}
  \includegraphics[width=0.23\linewidth]{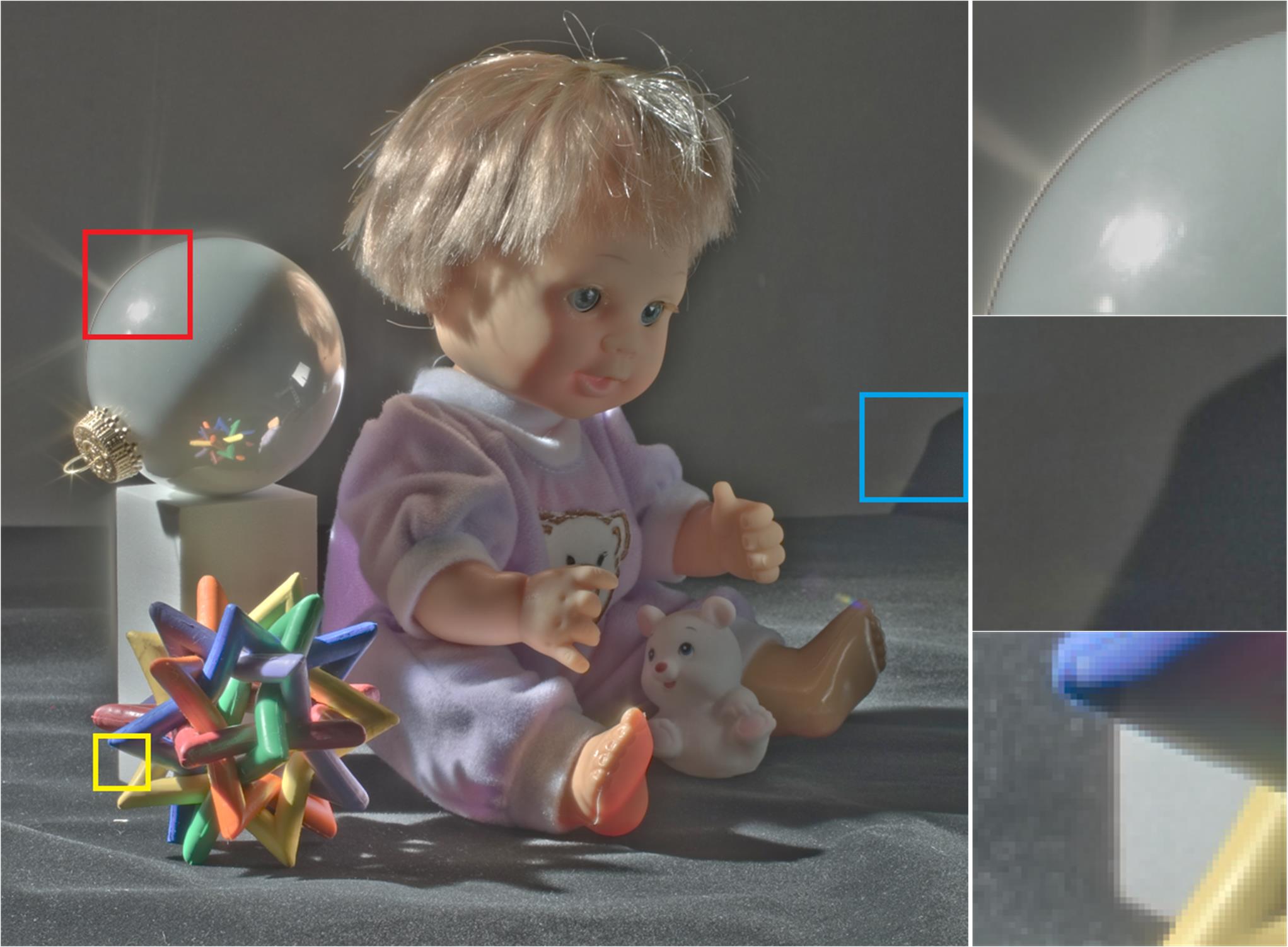} &
  \includegraphics[width=0.23\linewidth]{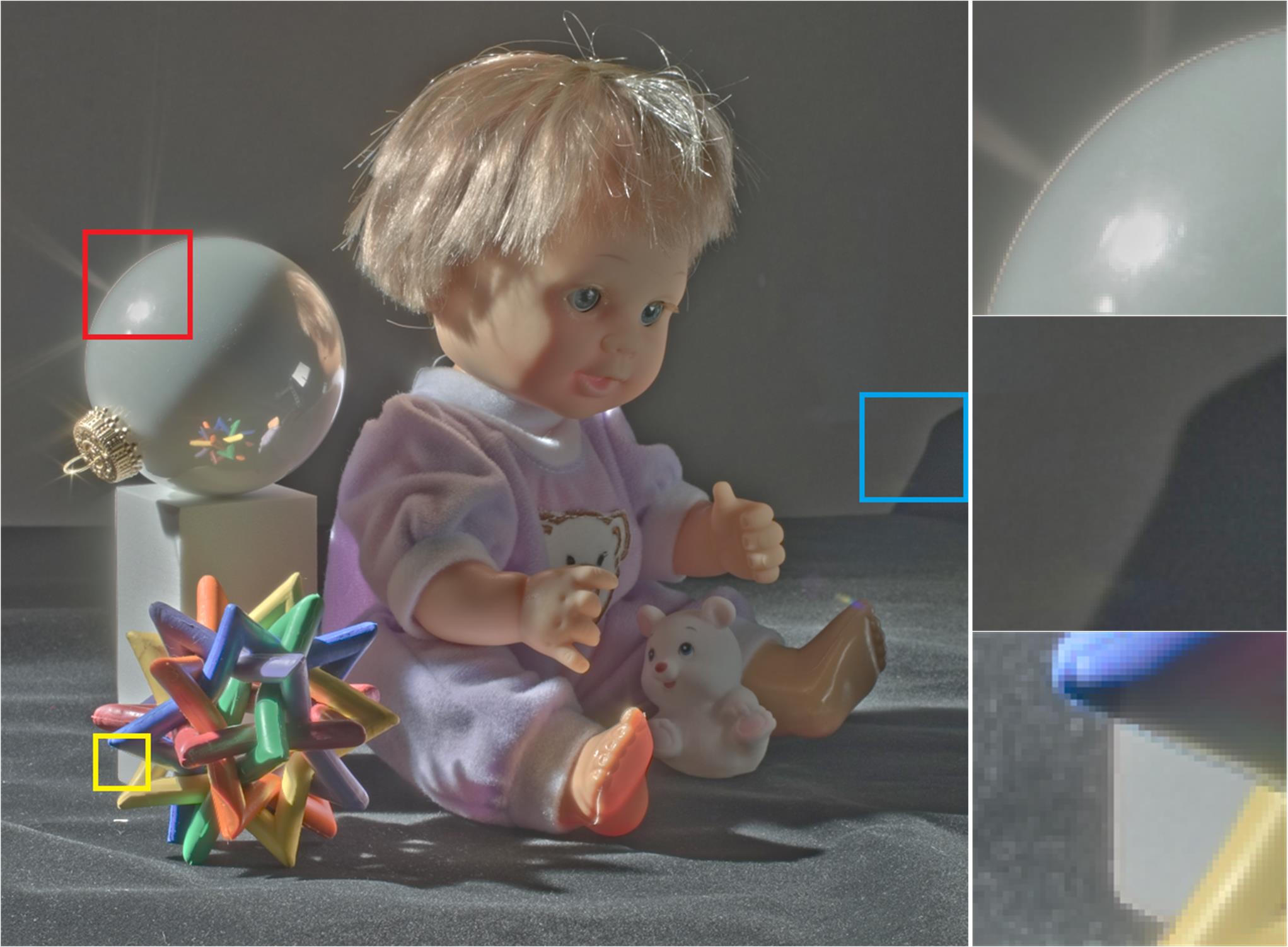} &
  \includegraphics[width=0.23\linewidth]{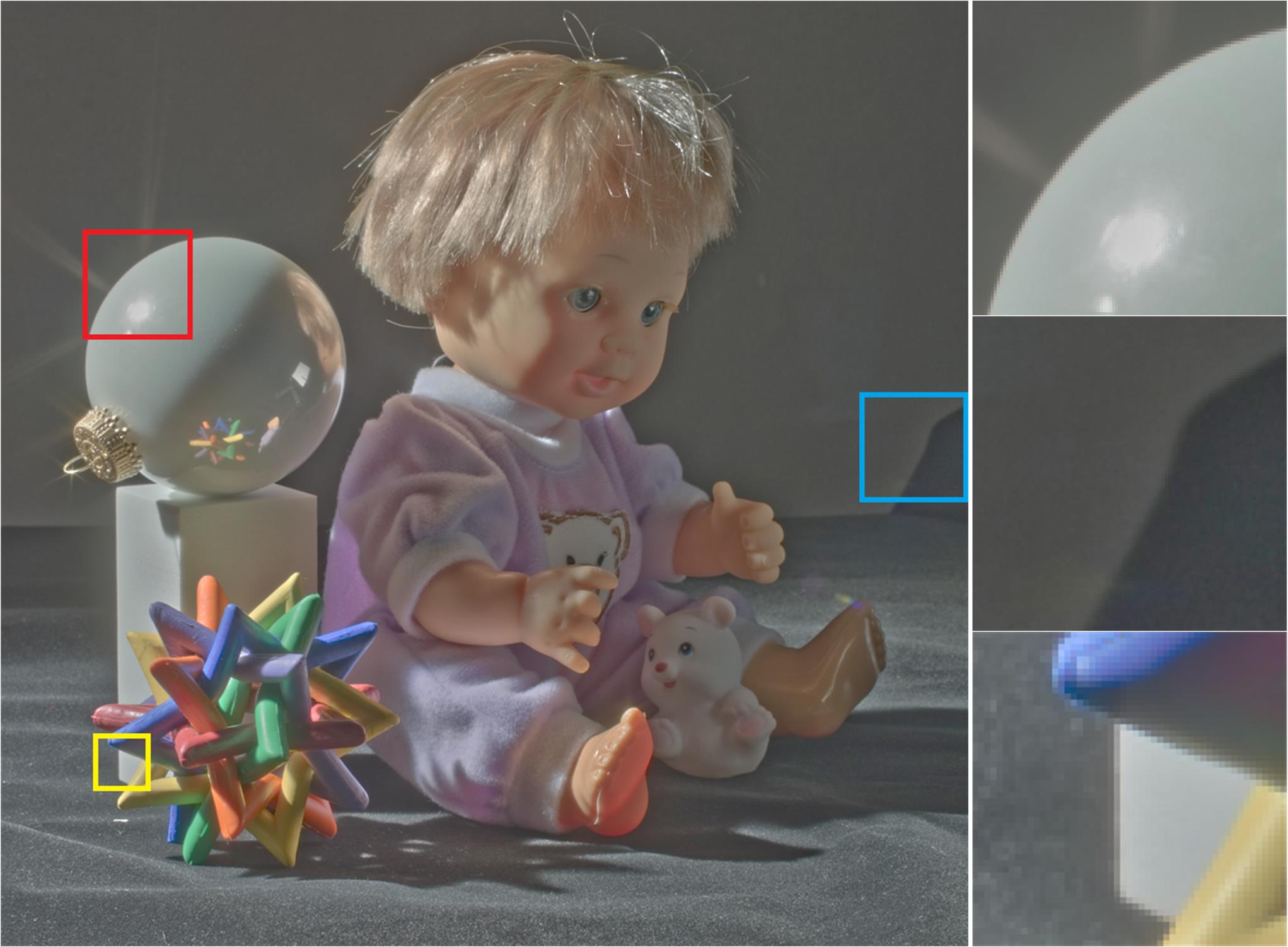} &
  \includegraphics[width=0.23\linewidth]{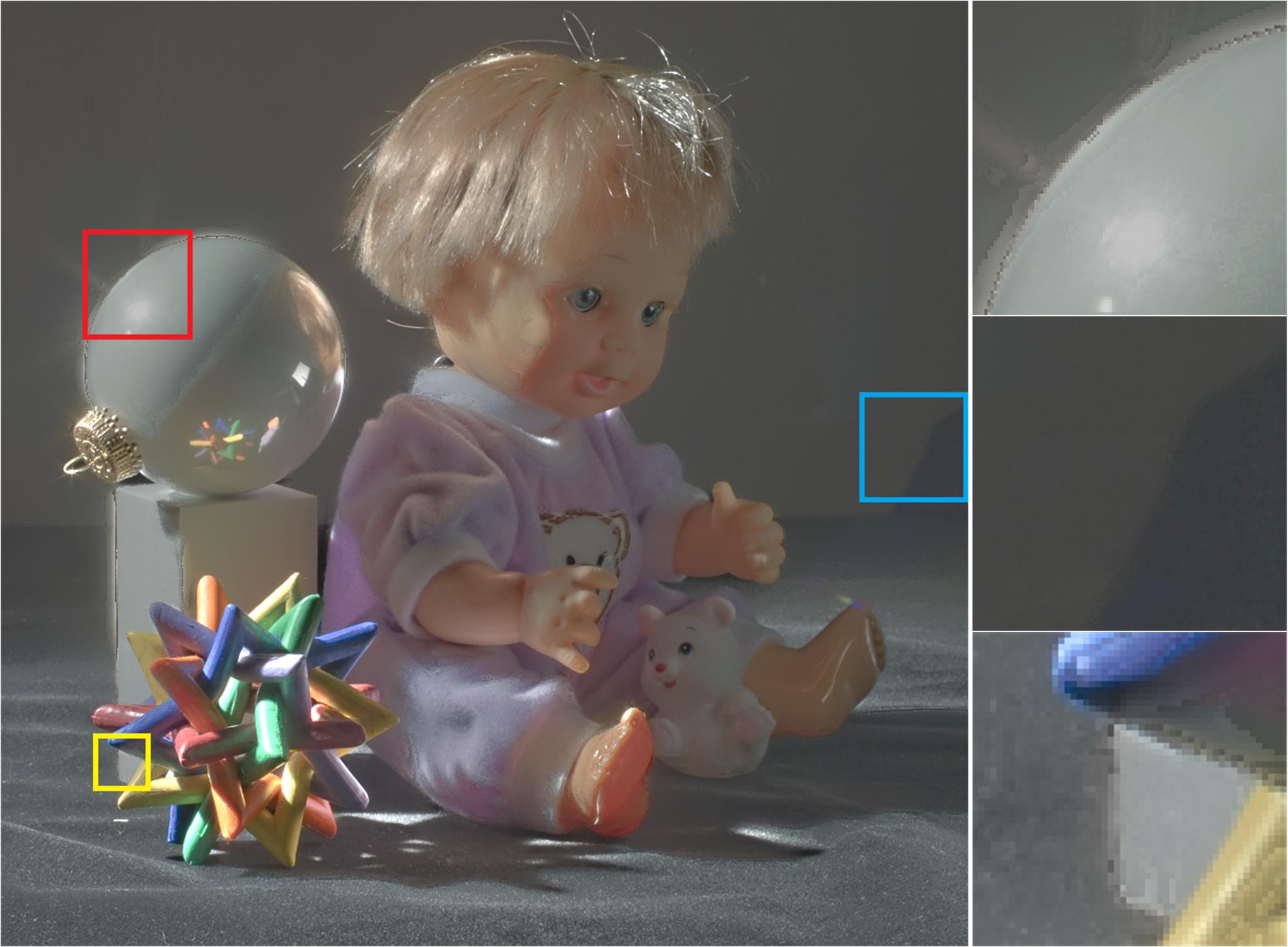} \\
  (a) BF & (b) AMF & (c) GF & (d) SWF \\

  \includegraphics[width=0.23\linewidth]{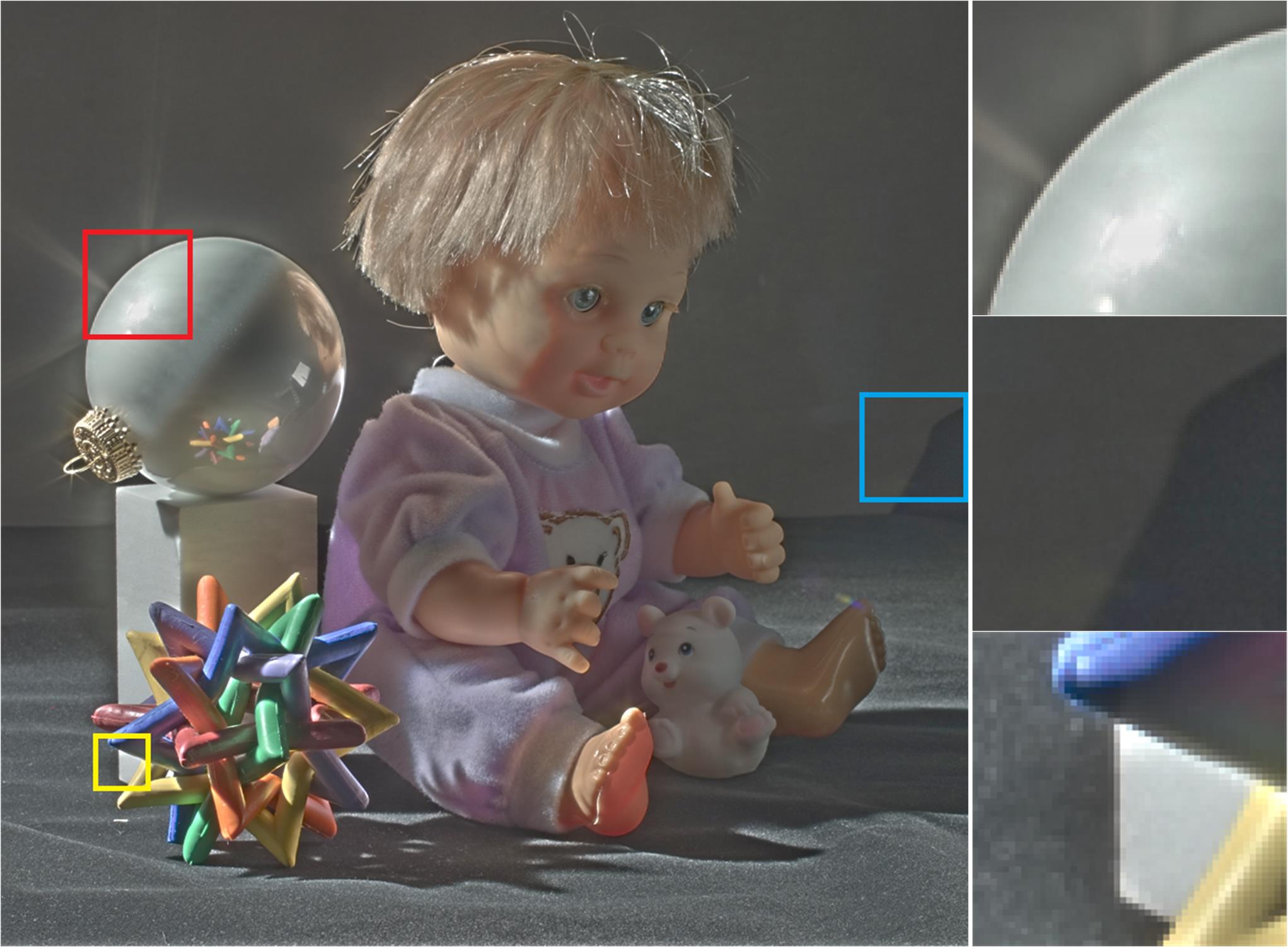} &
  \includegraphics[width=0.23\linewidth]{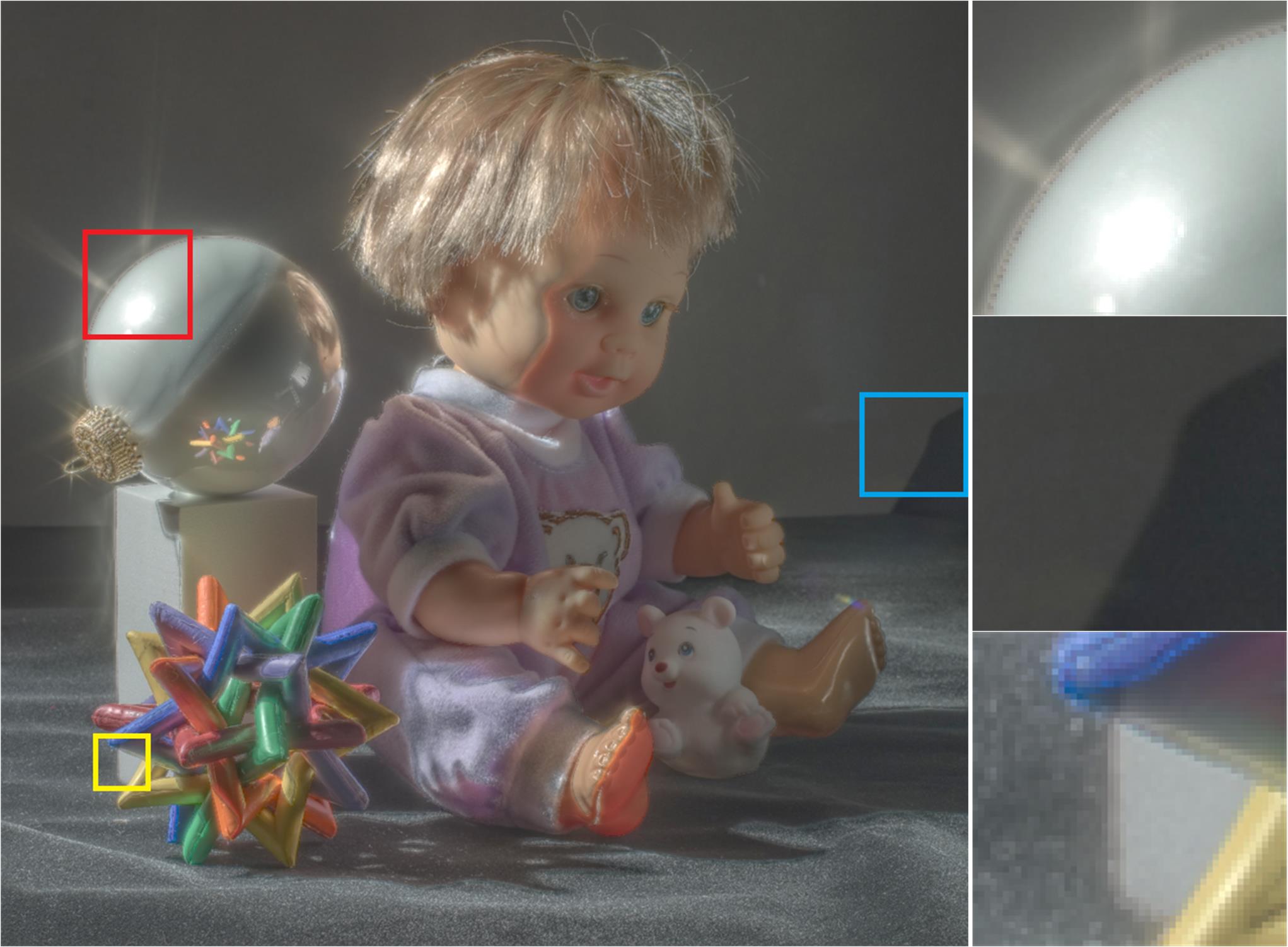} &
  \includegraphics[width=0.23\linewidth]{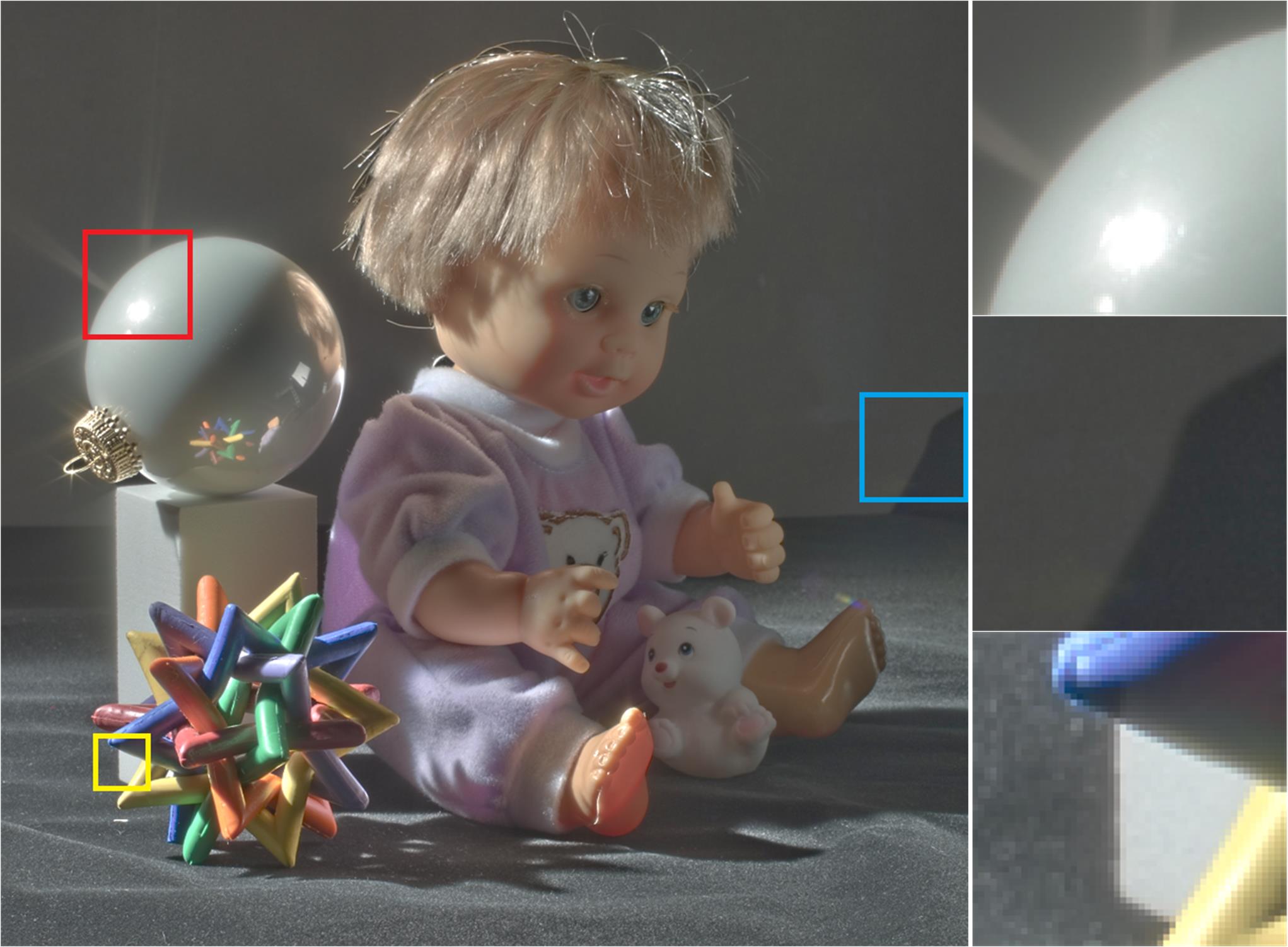} &
  \includegraphics[width=0.23\linewidth]{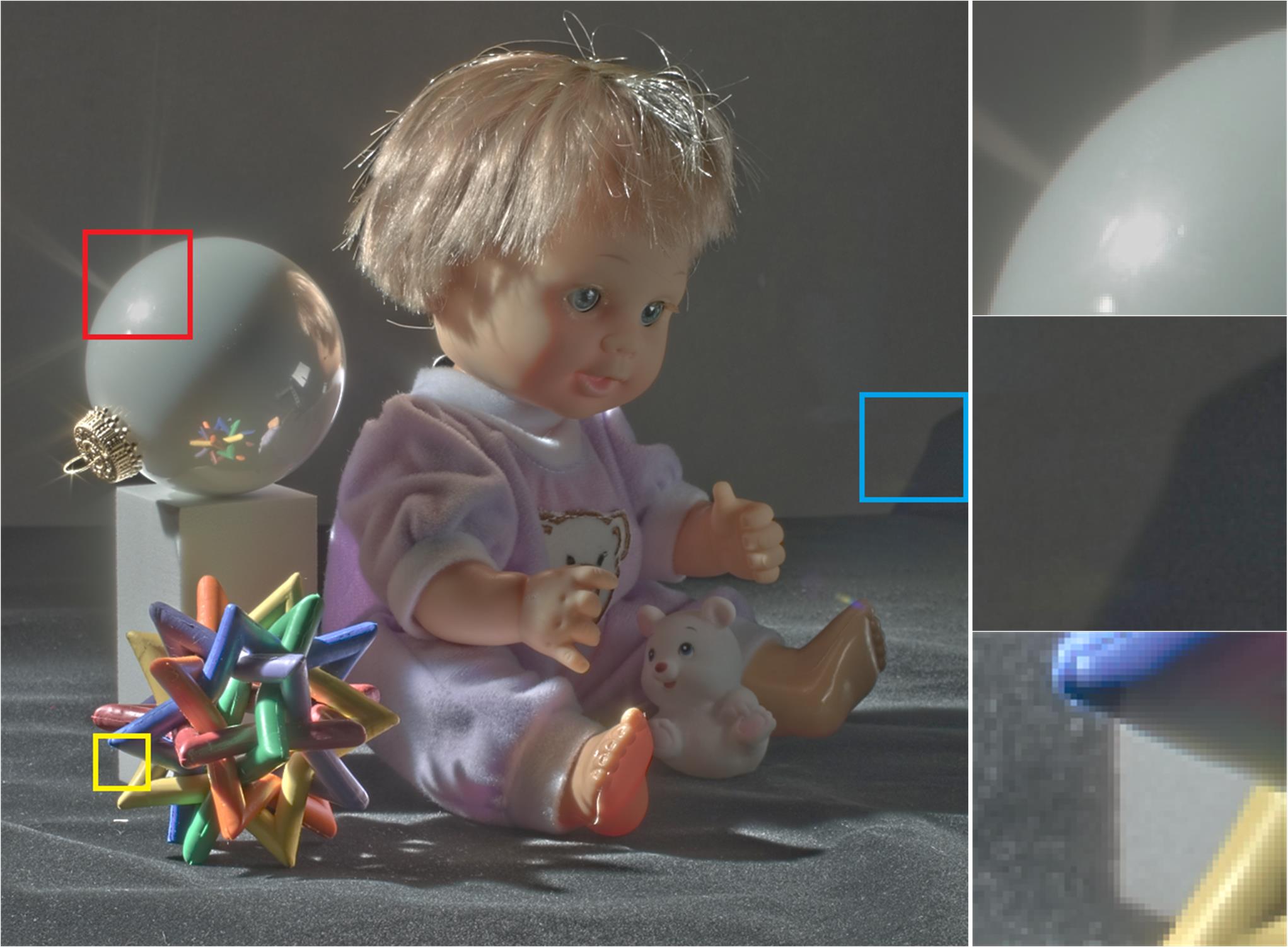}\\
  (e) SG-WLS & (f) $L_0$ norm & (g) WLS/ours(EP-1) & (h) ours(SP-2) \\
  \end{tabular}
   \caption{HDR tone mapping results of different approaches. Result of (a) BLF \cite{tomasi1998bilateral} ($\sigma_s=20,\sigma_r=0.2$), (b) AMF \cite{gastal2012adaptive} ($\sigma_s=16,\sigma_r=0.12$), (c) GF \cite{he2013guided} ($r=20,\epsilon=0.1^2$), (d) SWF \cite{yin2019side} ($r=10, iteration=2$), (e) SG-WLS \cite{liu2017semi} ($r=2,\tau=1,\lambda=75$), (f) $L_0$ norm smoothing \cite{xu2011image} ($\lambda=0.01$), (g) our method of the EP-1 mode/WLS \cite{farbman2008edge} ($\lambda=10,\alpha=1.2$)  and (h) our method of the SP-2 mode ($\lambda=200$).}\label{FigHDRToneMapping}
\end{figure*}

\subsection{Tasks in the First Group}
Image detail enhancement and HDR tone mapping are representative tasks in the first group. They both require to decompose the input image into a base layer and a detail layer. The slight difference is that the decomposition is applied to the log-luminance channel of the input HDR image while the input image is directly decomposed in image detail enhancement. The challenge of these tasks is that if the edges are sharpened by the smoothing procedure, it will result in gradient reversals, and halos will occur if the edges are blurred. We apply our method of both the EP-1 mode (equal to WLS \cite{farbman2008edge}) and the SP-2 mode to these two tasks. Except for the value of $\lambda$ which differs for different input images, the values of the other parameters are fixed as those in Tab.~\ref{TabParameter}. Fig.~\ref{FigDetailEnhancement} and Fig.~\ref{FigHDRToneMapping} show visual comparison of image detail enhancement and tone mapping results produced by different smoothers, respectively. As shown in the figures, there are either gradient reversals or halos in the results of the compared methods. Some results even contain both gradient reversals and halos, as shown in Fig.~\ref{FigDetailEnhancement}(b) and Fig.~\ref{FigHDRToneMapping}(a) and (b). In contrast, no gradient reversals and halos exist in the results of WLS smoothing and our method of the SP-2 mode. However, the tone mapping result of WLS smoothing contains slight compartmentalization artifacts caused by intensity shift, as shown in the region labeled with the yellow box in Fig.~\ref{FigHDRToneMapping}(g). This is properly eliminated in the result of our method of the SP-2 mode, shown in Fig.~\ref{FigHDRToneMapping}(h).

\begin{table*}[!ht]
\centering
\caption{Quantitative comparison on the noisy simulated ToF data. Results are evaluated with the MAE between the upsampled depth map and the ground-truth depth map. The best results are in \textbf{bold}. The second best results are \underline{underlined}.}\label{TabToFSimulated}

\resizebox{1\textwidth}{!}
{
\begin{tabular}{c|cccc|cccc|cccc|cccc|cccc|cccc}

\Xhline{1.2pt}
  \multicolumn{1}{c}{\multirow{2}{*}{}} & \multicolumn{4}{|c|}{\emph{Art}} & \multicolumn{4}{c|}{\emph{Book}} & \multicolumn{4}{c|}{\emph{Dolls}} & \multicolumn{4}{c|}{\emph{Laundry} } & \multicolumn{4}{c|}{\emph{Moebius}} & \multicolumn{4}{c}{\emph{Reindeer}}\\

  \cline{2-25} 
  & $2\times$ & $4\times$ & $8\times$ & $16\times$ & $2\times$ & $4\times$ & $8\times$ & $16\times$ & $2\times$ & $4\times$ & $8\times$ & $16\times$ & $2\times$ & $4\times$ & $8\times$ & $16\times$ & $2\times$ & $4\times$ & $8\times$ & $16\times$ & $2\times$ & $4\times$ & $8\times$ & $16\times$ \\
  \Xhline{1.2pt}

  GF\cite{he2013guided} & 1.91 & 2.23 & 3.08 & 4.87 & 0.84 & 1.12 & 1.73 & 2.82 & 0.84 & 1.11 & 1.69 & 2.71 & 1.01 & 1.31 & 2.0 & 3.33 & 0.92 & 1.19 & 1.78 & 2.84 & 1.06 & 1.32 & 1.98 & 3.31 \\

  TGV\cite{ferstl2013image} & 0.8 &  \underline{1.21} & \underline{2.01} & 4.59 & 0.61 & 0.88 & 1.21 & 2.19 & 0.66 &  \underline{0.95} & \underline{1.38} & 2.88 & {0.61} &  \textbf{0.87} &  \underline{1.36} & 3.06 & {0.57} &  \underline{0.77} & 1.23 & 2.74 & 0.61 &  \underline{0.85} & \underline{1.3} & 3.41 \\

  AR\cite{yang2014color} & 1.17 & 1.7 & 2.93 & 5.32 & 0.98 & 1.22 & 1.74 & 2.89 & 0.97 & 1.21 & 1.71 & 2.74 & 1 & 1.31 & 1.97 & 3.43 & 0.95 & 1.2 & 1.79 & 2.82 & 1.07 & 1.3 & 2.03 & 3.34\\

  SG-WLS\cite{liu2017semi} & 1.26 & 1.9 & 3.07 & - & 0.82 & 1.12 & 1.73 & - & 0.87 & 1.11 &	1.81 & - & 0.86 & 1.17 & 2 & - & 0.82 & 1.08 &1.79 & - & 0.9 & 1.32 & 2.01 & -\\

  FGI\cite{li2016fast} & 0.9 & 1.37 & 2.46 & 4.89 & 0.66 &  \underline{0.85} &  1.23 & 1.96 & 0.74 & \underline{0.95} & 1.41 &  \underline{2.13} & 0.71 & 0.99 & 1.59 & \underline{2.67} & 0.67 & 0.82 &  \underline{1.2} &  \underline{1.87} & 0.75 & 0.94 & 1.55 & 2.73\\

  SGF\cite{zhang2015segment} & 1.42 & 1.85 & 3.06 & 5.55 & 0.84 & 1.11 & 1.76 & 3.03 & 0.87 & 1.2 & 1.88 & 3.26 & 0.74 & 1.1 & 1.96 & 3.63 & 0.81 & 1.13 & 1.84 & 3.16 & 0.93 & 1.25 & 2.03 & 3.67\\

  SD Filter\cite{ham2018robust} & 1.16 & 1.64 & 2.74 & 5.52 & 0.86 & 1.1 & 1.57 & 2.68 & 1.04 & 1.27 & 1.73 & 2.76 & 0.96 & 1.25 & 1.94 & 3.54 & 0.93 & 1.14 & 1.68 & 2.75 & 1.05 & 1.31 & 1.99 & 3.43\\

  FBS\cite{barron2016fast} & 1.93 & 2.39 & 3.29 & 5.05 & 1.42 & 1.55 & 1.76 & 2.48 & 1.33 & 1.45 & 1.69 & 2.26 & 1.32 & 1.49 & 1.77 & 2.67 & 1.16 & 1.29 & 1.61 & 2.44 & 1.63 & 1.76 & 2.01 & 2.69\\

  Park et~al.\cite{park2011high} & 1.66 & 2.47 & 3.44 & 5.55 & 1.19 & 1.47 & 2.06 & 3.1 & 1.19 & 1.56 & 2.15 & 3.04 & 1.34 & 1.73 & 2.41 & 3.85 & 1.2 & 1.5 & 2.13 & 2.95 & 1.26 & 1.65 & 2.46 & 3.66 \\

  Shen et~al.\cite{shen2015mutual} & 1.79 & 2.21 & 3.2 & 5.04 & 1.34 & 1.69 & 2.25 & 3.13 & 1.37 & 1.58 & 2.05 & 2.85 & 1.49 & 1.74 & 2.34 & 3.5 & 1.34 & 1.56 & 2.09 & 2.99 & 1.29 & 1.55 & 2.19 & 3.33\\

  Gu et~al.\cite{gu2017learning} &  \underline{0.61} & 1.46 & 2.98 & 5.09 &  \textbf{0.52} & 0.95 & 1.87 & 2.98 & \underline{0.63} & 1.02 & 1.89 & 2.92 &  \textbf{0.58} & 1.14 & 2.21 & 3.58 &  \underline{0.53} & 0.96 & 1.89 & 2.99 &  \textbf{0.52} & 1.07 & 2.17 & 3.59\\

  DJF\cite{li2019joint} & - & 2.28 & 3.30 & 5.45 & - & 1.71 & 2.02 & 2.69 & - & 1.69 & 2.01 & 2.62 & - & 1.85 & 2.39 & 3.43 & - & 1.67 & 2.09 & 2.82 & - & 1.85 & 2.37 & 3.47\\

  DKN\cite{kim2020deformable} &  - & 2.30 & 2.95 & 5.25 & - & 1.90 & 2.22 & 2.41 & - & 1.85 & 2.25 & 2.88 & - & 2.01 & 2.43 & 3.35 & - & 1.91 & 2.27 & 2.93 & - & 2.01 & 2.35 & 3.28\\

 \hline
 \hline
  Ours (s=1)&  \textbf{0.60} &  \textbf{0.98} & \textbf{1.63} &  \textbf{3.06} &  \underline{0.54} &  \textbf{0.76} &  \textbf{1.13} & \textbf{1.61} & \textbf{0.62} & \textbf{0.90} &  \textbf{1.27} &  \textbf{1.87} &  \underline{0.61} &  \underline{0.91} & \underline{1.36} & \textbf{2.21} &  \textbf{0.50} &  \textbf{0.74} & \textbf{1.11} & \textbf{1.73} & \underline{0.55} &  \textbf{0.82} & \textbf{1.26} & \textbf{2.07}\\
  Ours (s=2) &  \underline{0.61} & \textbf{0.98} &  \textbf{1.63} &  \underline{3.09} & \underline{0.54} &  \textbf{0.76} &  \underline{1.14} &  \underline{1.63} & \underline{0.63} & \textbf{0.90} & \textbf{1.27} &  \textbf{1.87} & 0.62 & \underline{0.91} &  \textbf{1.35} &  \textbf{2.21} & \textbf{0.50} &  \textbf{0.74} &  \textbf{1.11} & \textbf{1.73} & 0.56 & \textbf{0.82} &  \textbf{1.26} &  \underline{2.08}\\

  \Xhline{1.2pt}
\end{tabular}
}
\end{table*}
\begin{table*}[!ht]
\centering
\caption{Quantitative comparison on real ToF data. The errors are calculated as the MAE to the measured ground-truth depth maps. The best results are in \textbf{bold}. The second best results are \underline{underlined}.}\label{TabToFReal}

\resizebox{1\textwidth}{!}
{
\begin{tabular}{c|cccccccccccccc}
  \Xhline{1.2pt}
    & GF\cite{he2013guided} & SD Filter\cite{ham2018robust} & SG-WLS\cite{liu2017semi} & Shen et~al.\cite{shen2015mutual} & Park et~al.\cite{park2011high} & TGV\cite{ferstl2013image} & AR\cite{yang2014color} & Gu et~al.\cite{gu2017learning} & SGF\cite{zhang2015segment} & FGI\cite{li2016fast} & FBS\cite{barron2016fast} & DJF\cite{li2019joint} & DKN\cite{kim2020deformable} & Ours(s=1/s=2)  \\
  \Xhline{1.2pt}
  \emph{Books} &  15.55 & 13.47 & 14.71 & 15.47 & 14.31 &  \underline{12.8} & 14.37 & 13.87 & 13.57 & 13.03 & 15.93 & 14.33 & 14.52 & { \textbf{12.45}}/12.49 \\
  \emph{Devil} &  16.1 & 15.99 & 16.24 & 16.18 & 15.36 &  \underline{14.97} & 15.41 & 15.36 & 15.74 & 15.09  & 17.21 & 15.09 & 15.14 & {14.49}/ \textbf{14.48}\\
  \emph{Shark} & 17.1 & 16.18 & 16.51 & 17.33 & 15.88 &  \underline{15.53} & 16.27 & 15.88 & 16.21 & 15.82 & 16.33 & 15.82 & 15.79 & { \textbf{15.04}}/15.09\\
  \Xhline{1.2pt}
\end{tabular}
}
\end{table*}

\begin{figure*}[!ht]
  \centering
  \setlength{\tabcolsep}{0.5mm}
  \begin{tabular}{cccc}
  \includegraphics[width=0.23\linewidth]{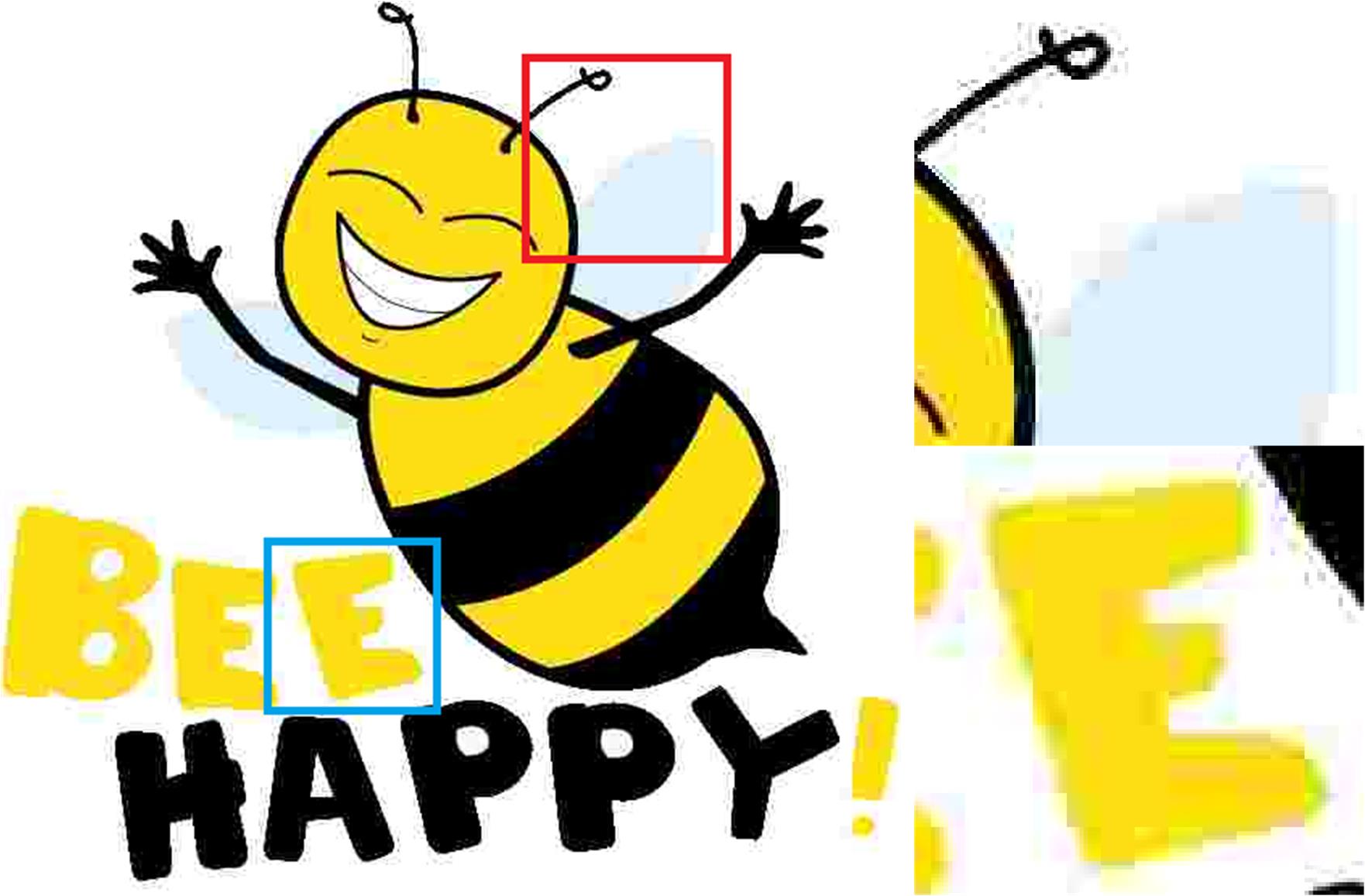} &
  \includegraphics[width=0.23\linewidth]{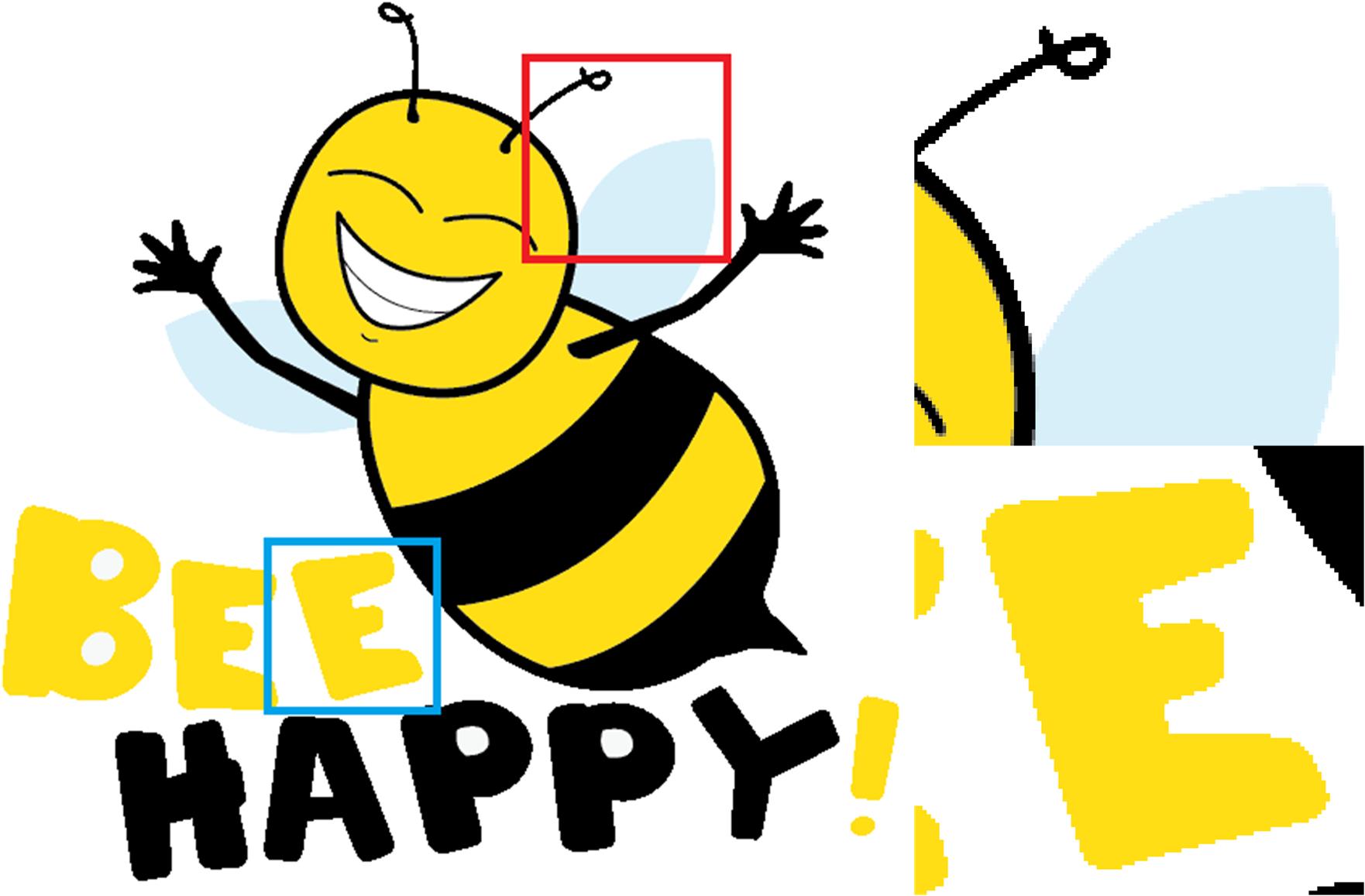} &
  \includegraphics[width=0.23\linewidth]{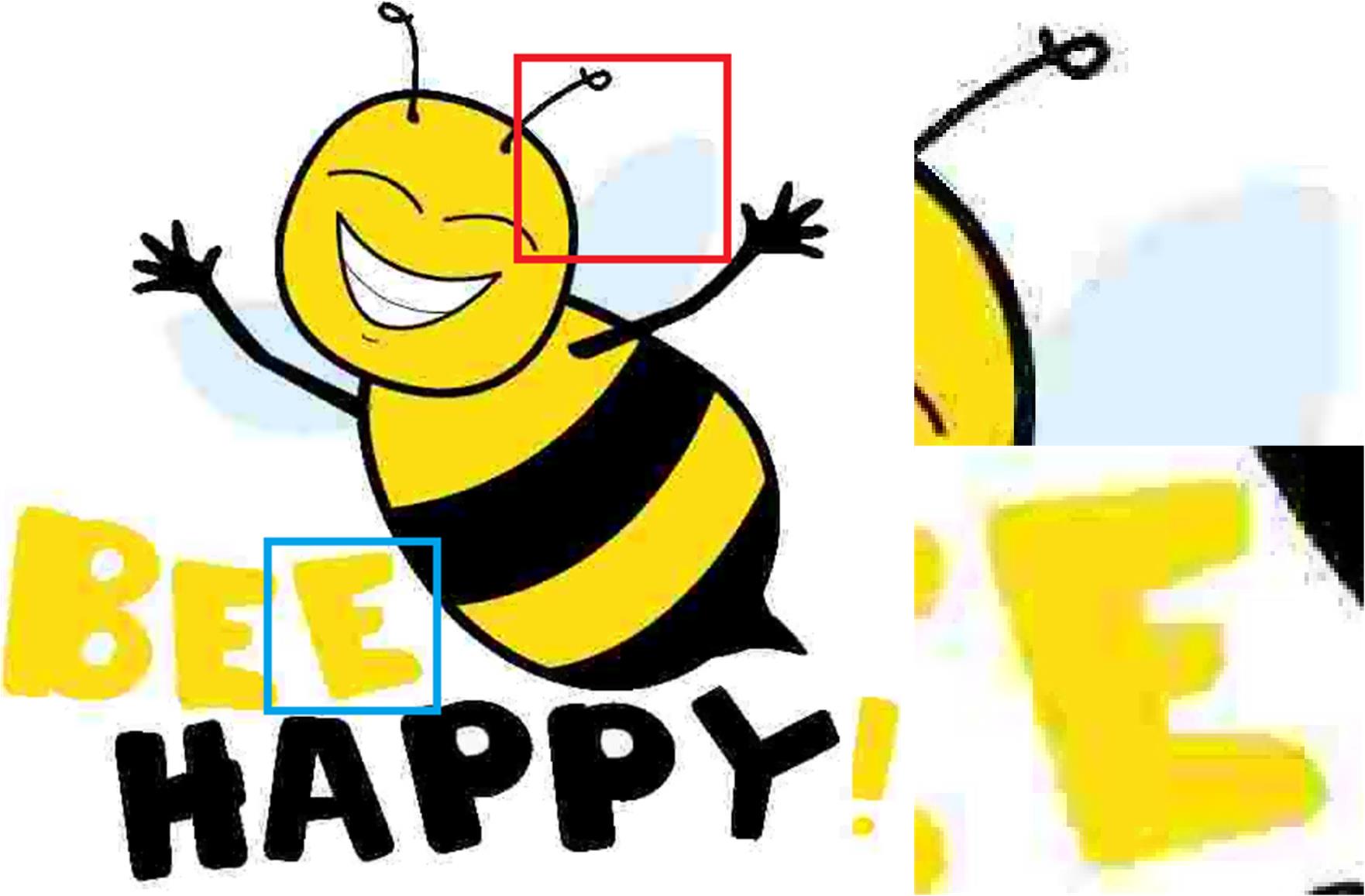} &
  \includegraphics[width=0.23\linewidth]{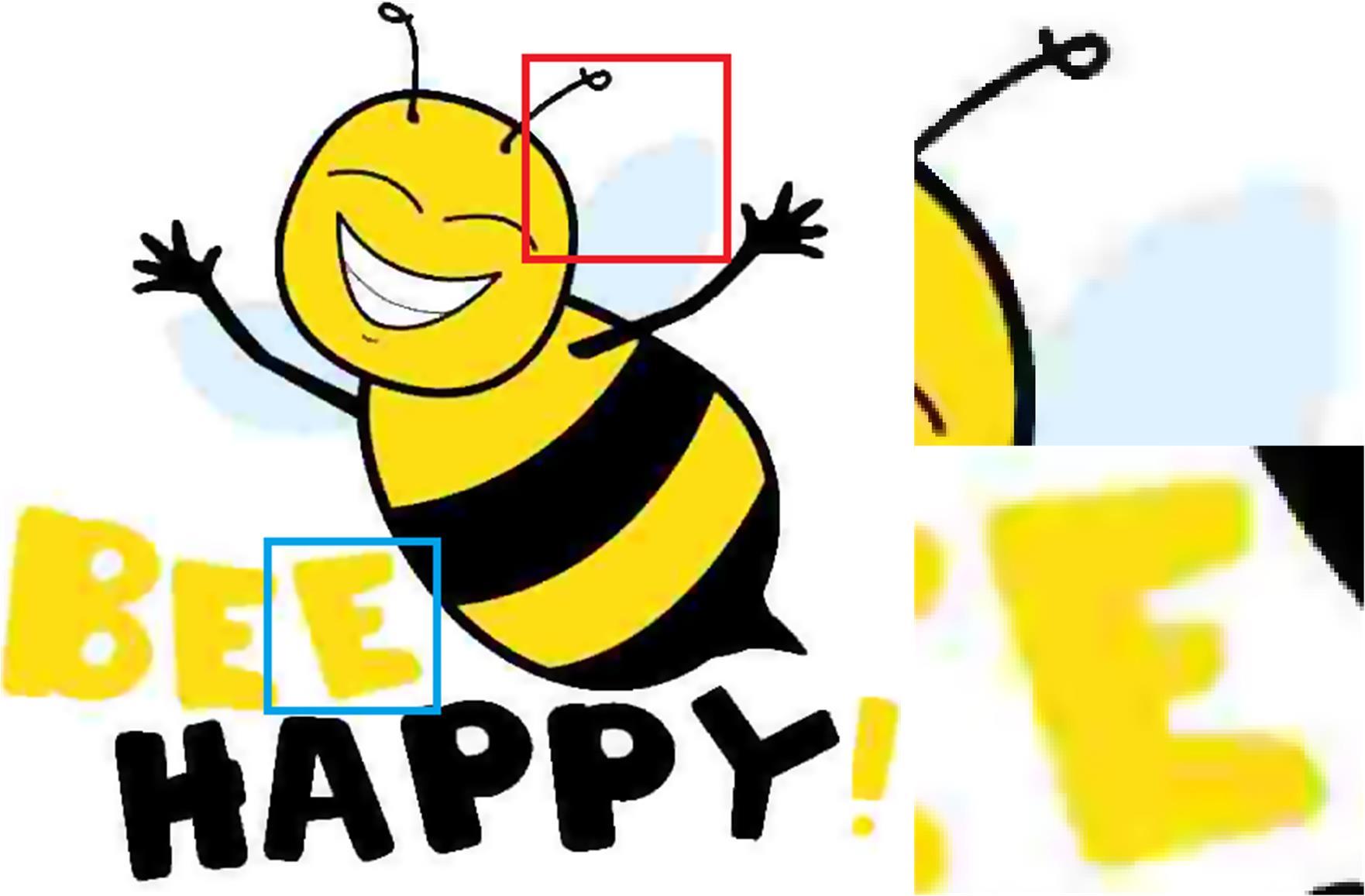} \\
  (a) JPEG input  & (b) ground-truth & (c) Wang et al. & (d) BTF\\

  \includegraphics[width=0.23\linewidth]{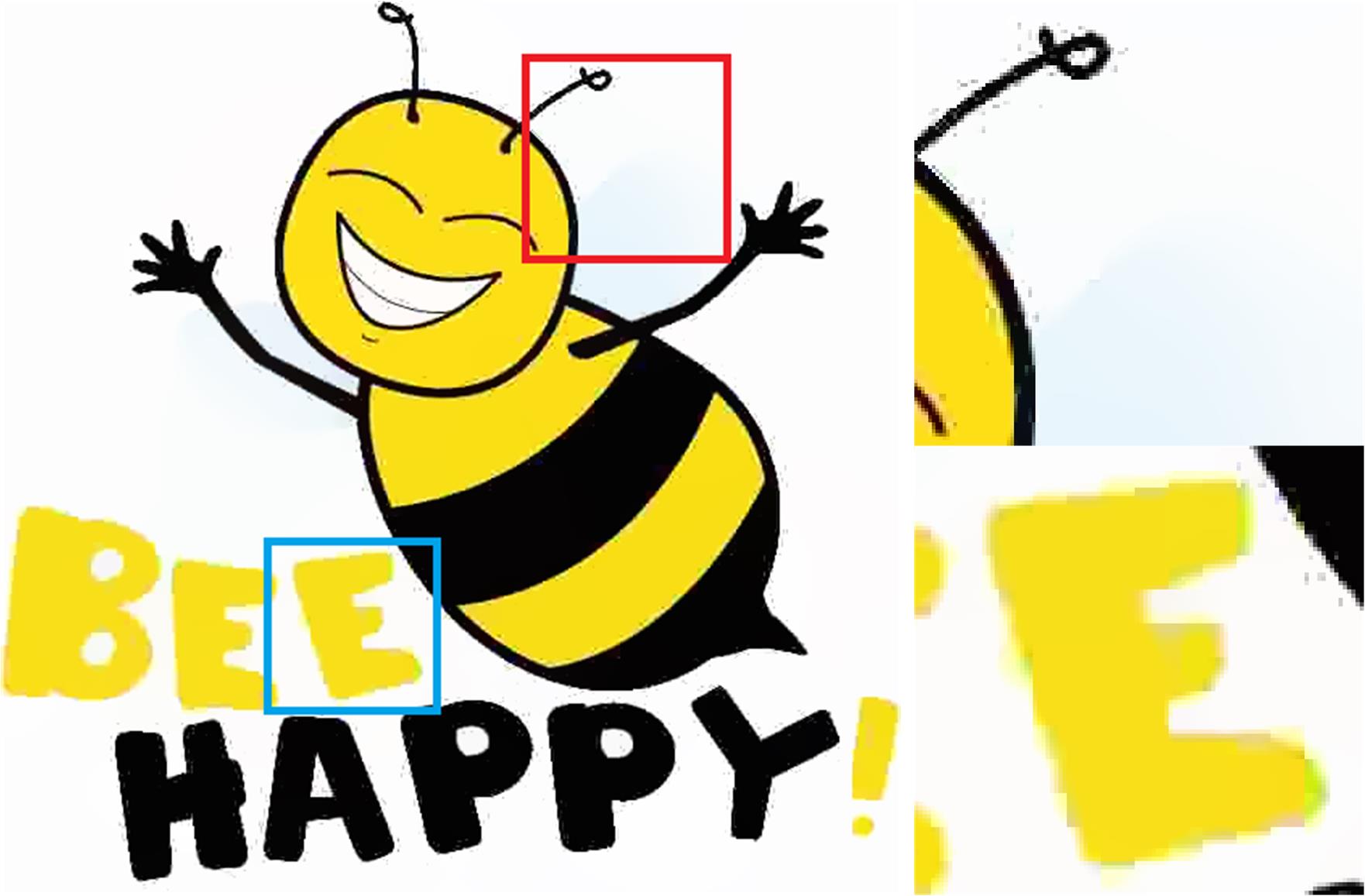} &
  \includegraphics[width=0.23\linewidth]{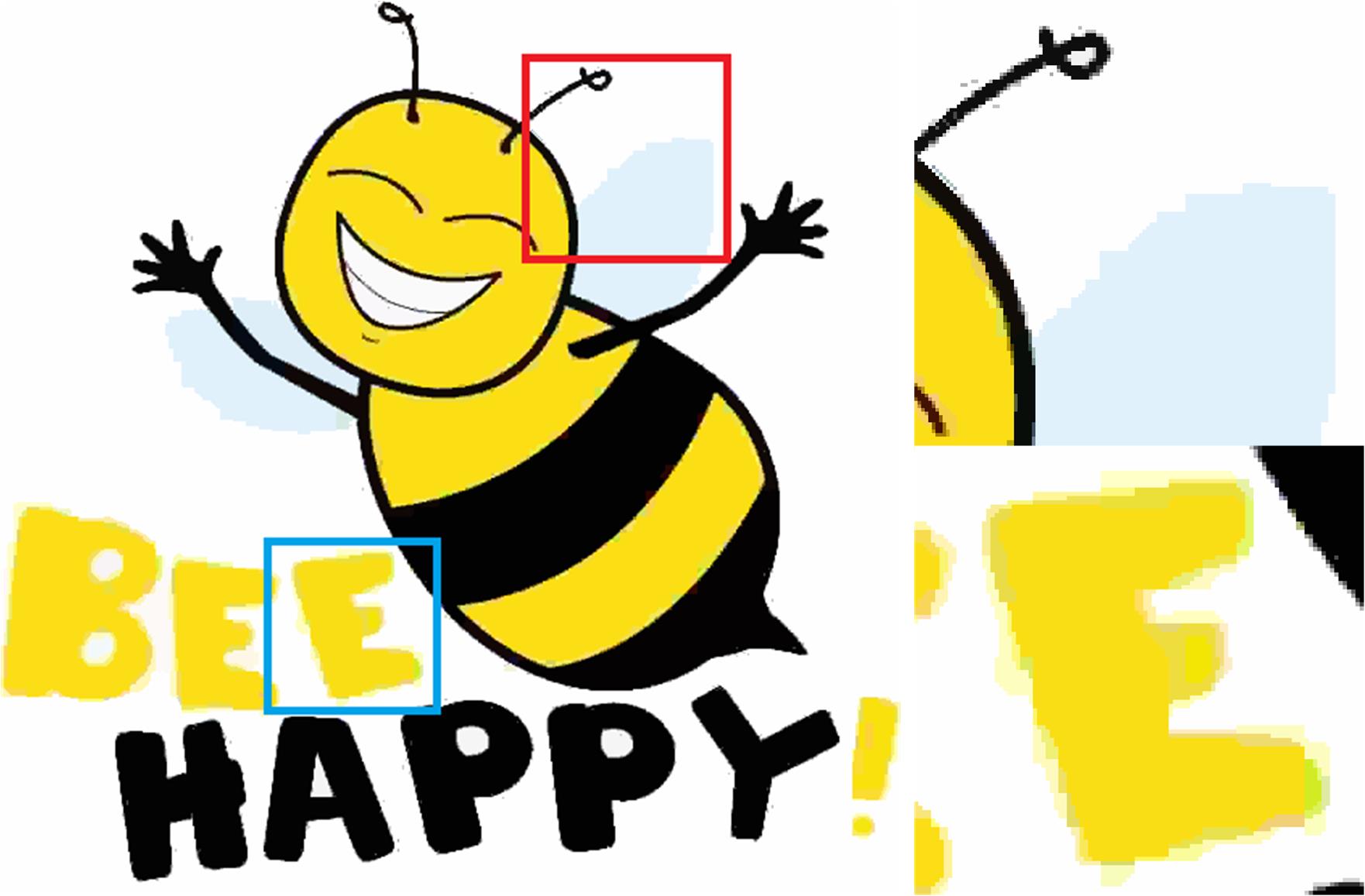} &
  \includegraphics[width=0.23\linewidth]{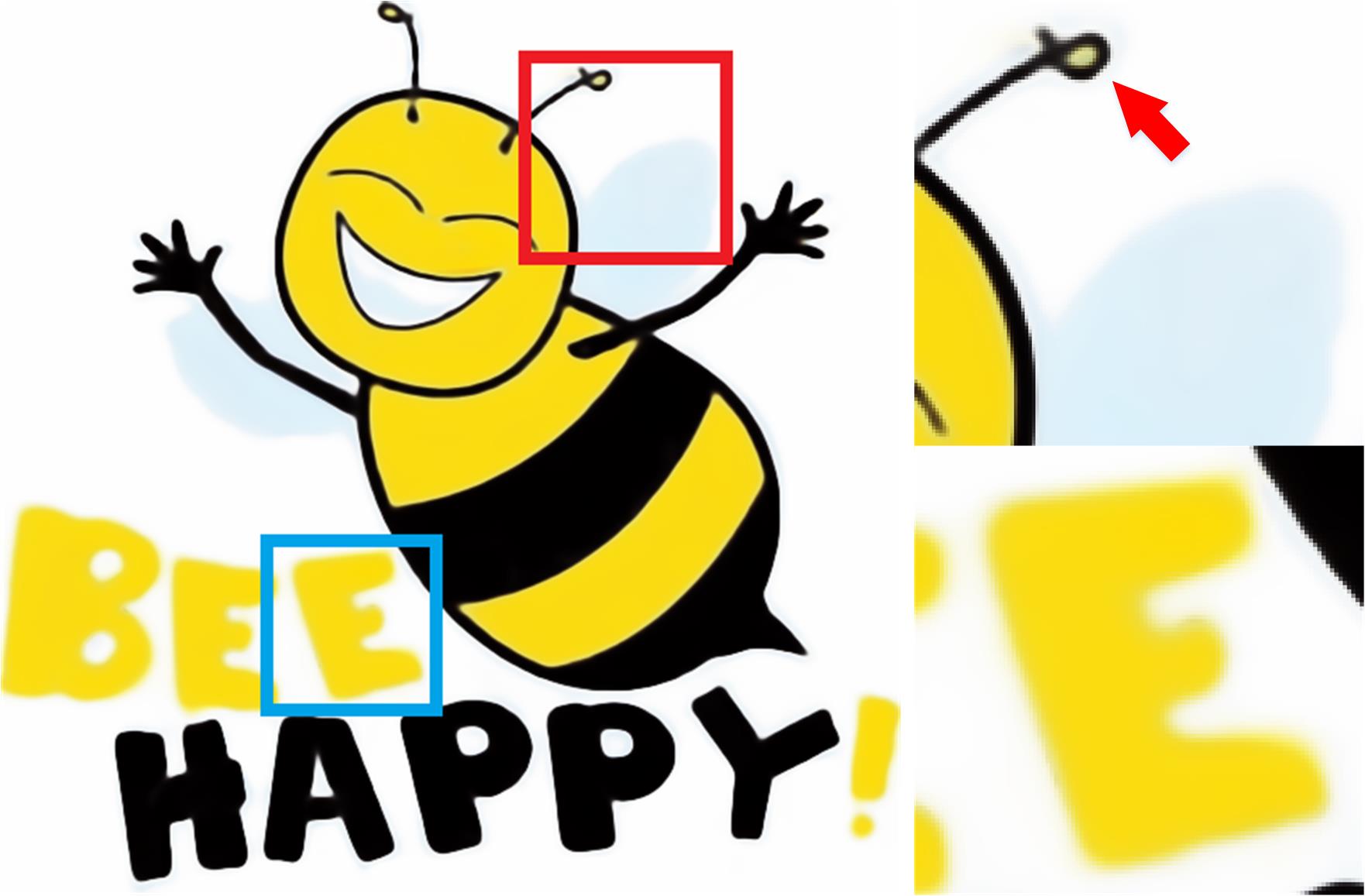} &
  \includegraphics[width=0.23\linewidth]{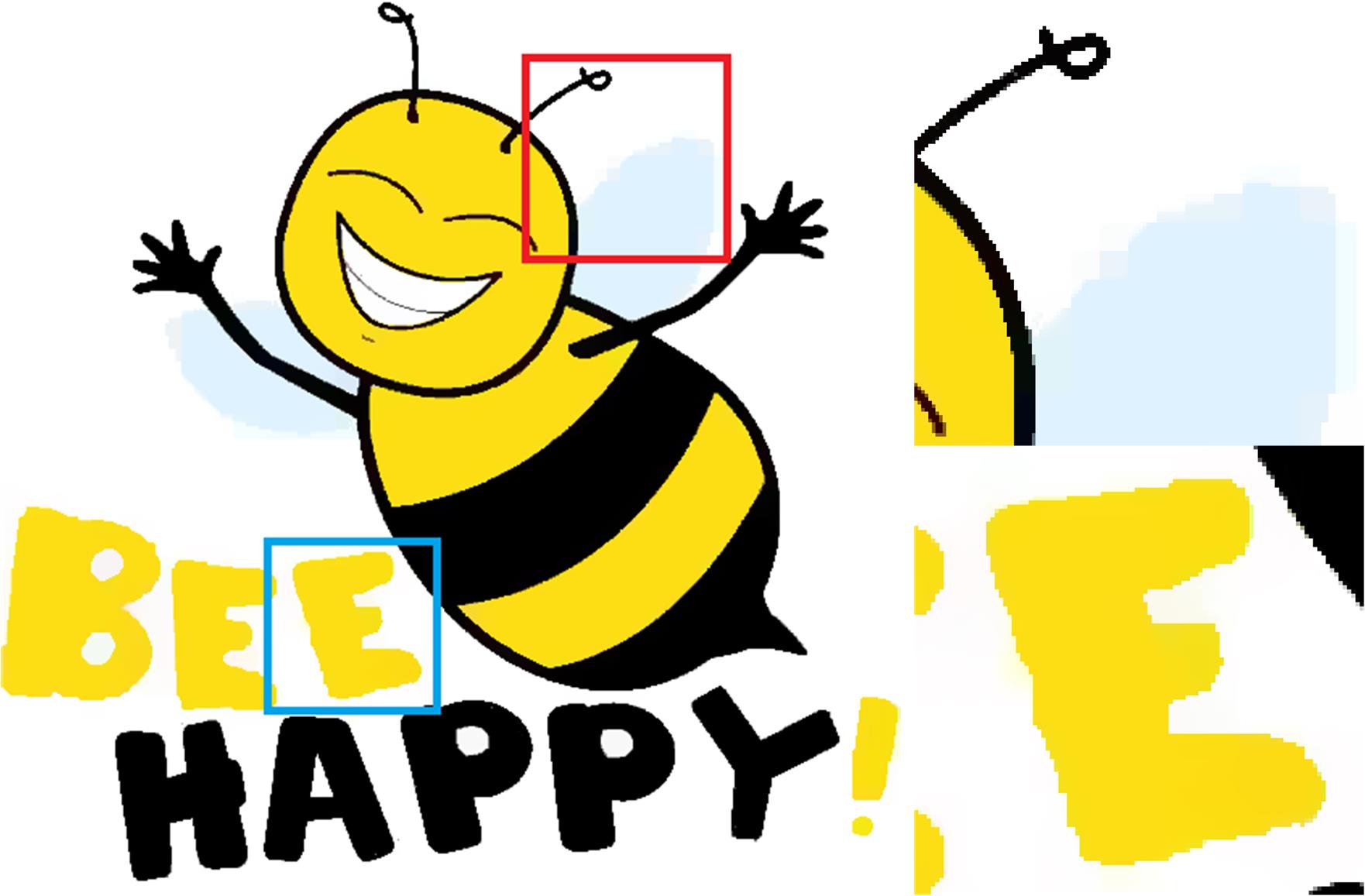}\\
   (e) $L_0$ norm & (f) region fusion & (g) deep image prior & (h) ours(EP$\&$SP)\\

  \end{tabular}
  \caption{Clip-art compression artifacts removal results of different methods. (a) Input compressed image and (b) the corresponding ground-truth image. Result of (c) the approach proposed by Wang et~al. \cite{wang2006deringing}, (d) BTF \cite{cho2014bilateral} ($k=3,n_{itr}=2$), (e) $L_0$ norm smoothing \cite{xu2011image} ($\lambda=0.01$), (f) region fusion approach \cite{nguyen2015fast} ($\lambda=0.05$), (g) deep image prior \cite{ulyanov2018deep} and (h) our method of the EP$\&$SP mode ($r_d=r_s=2,\lambda=0.4,b_d=b_s=0.15$).}\label{FigClipArt}
\end{figure*}

\begin{figure*}[!ht]
  \centering
  \setlength{\tabcolsep}{0.5mm}
  \begin{tabular}{cccc}
  \includegraphics[width=0.23\linewidth]{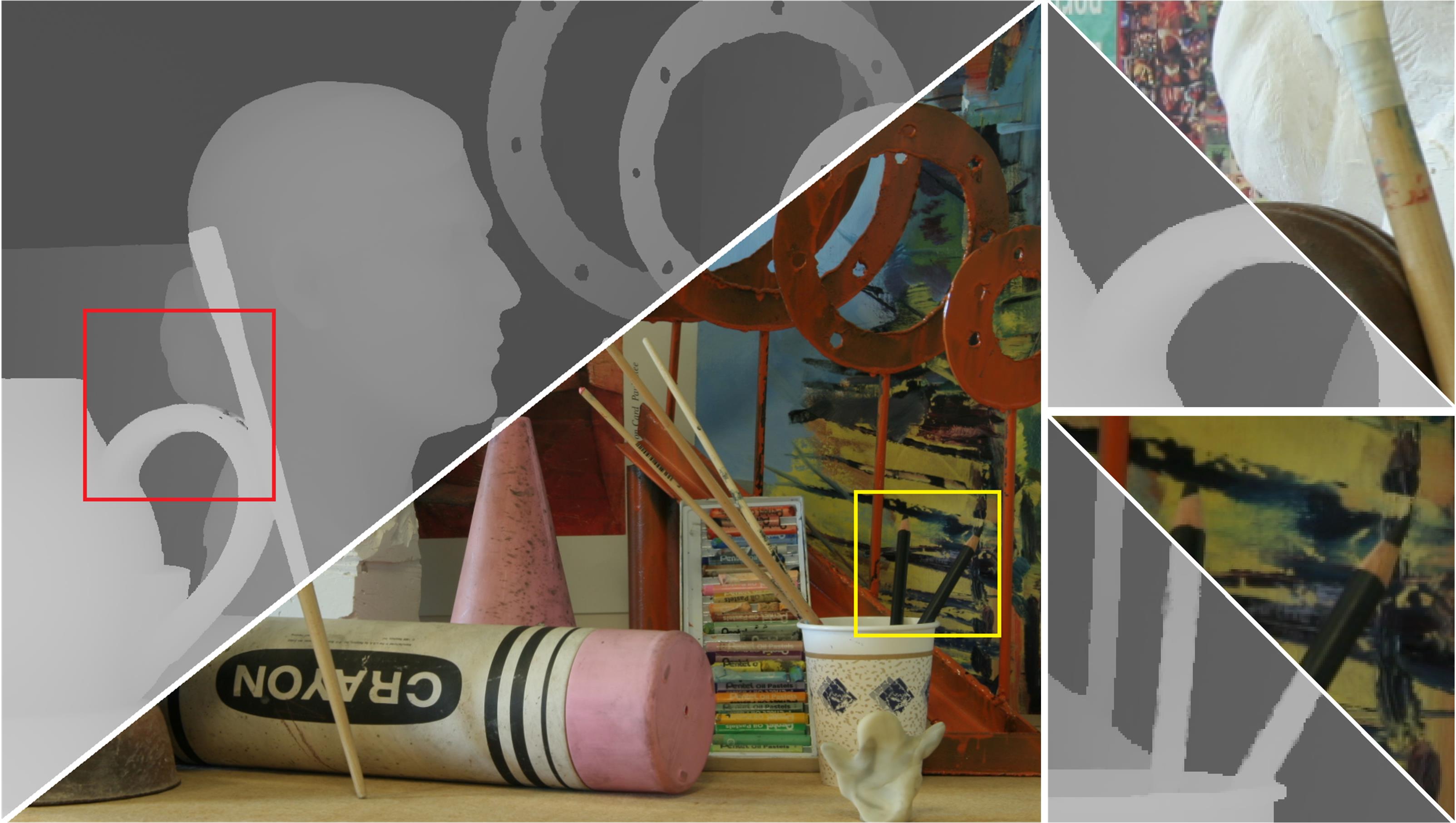} &
  \includegraphics[width=0.23\linewidth]{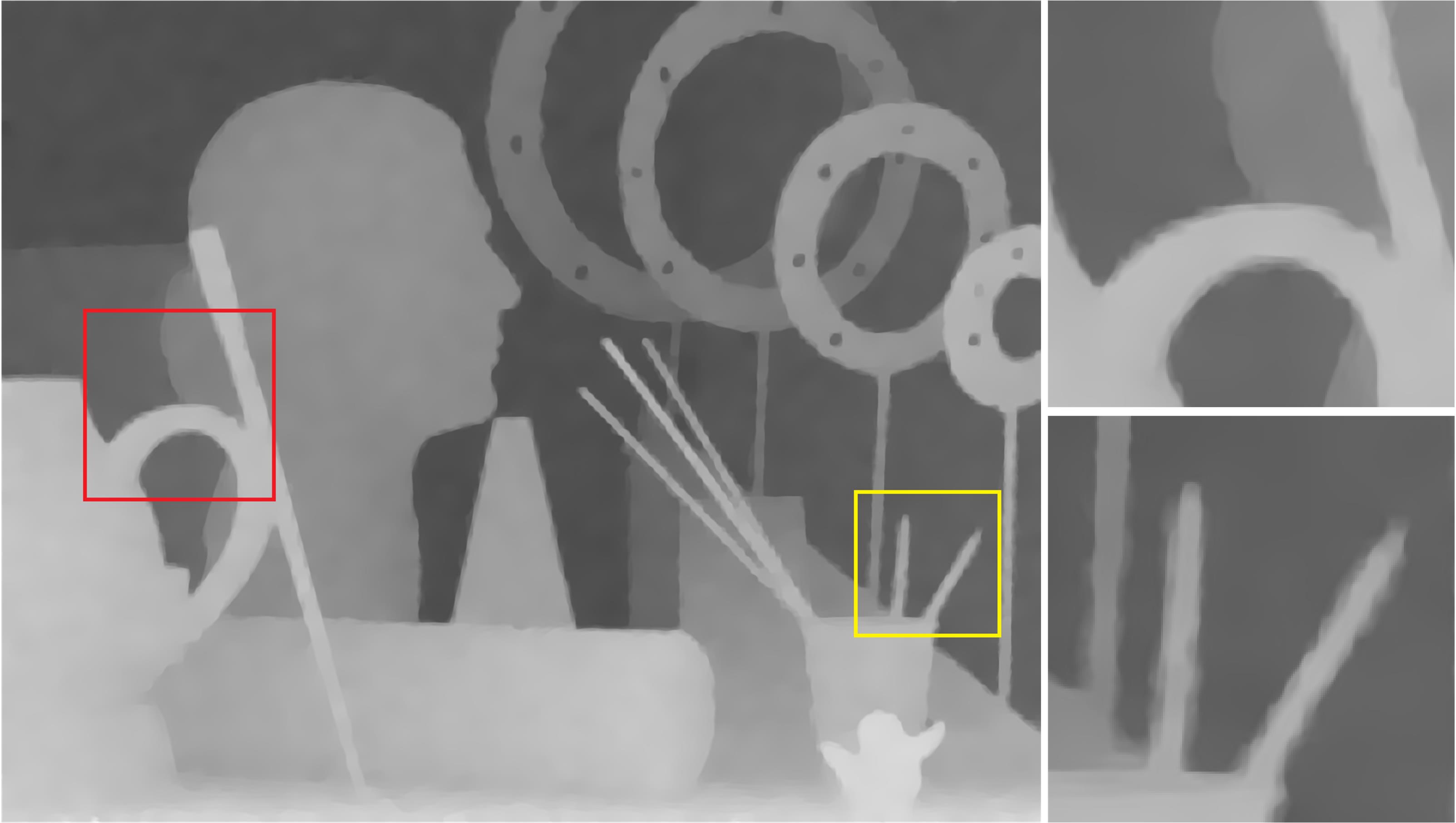} &
  \includegraphics[width=0.23\linewidth]{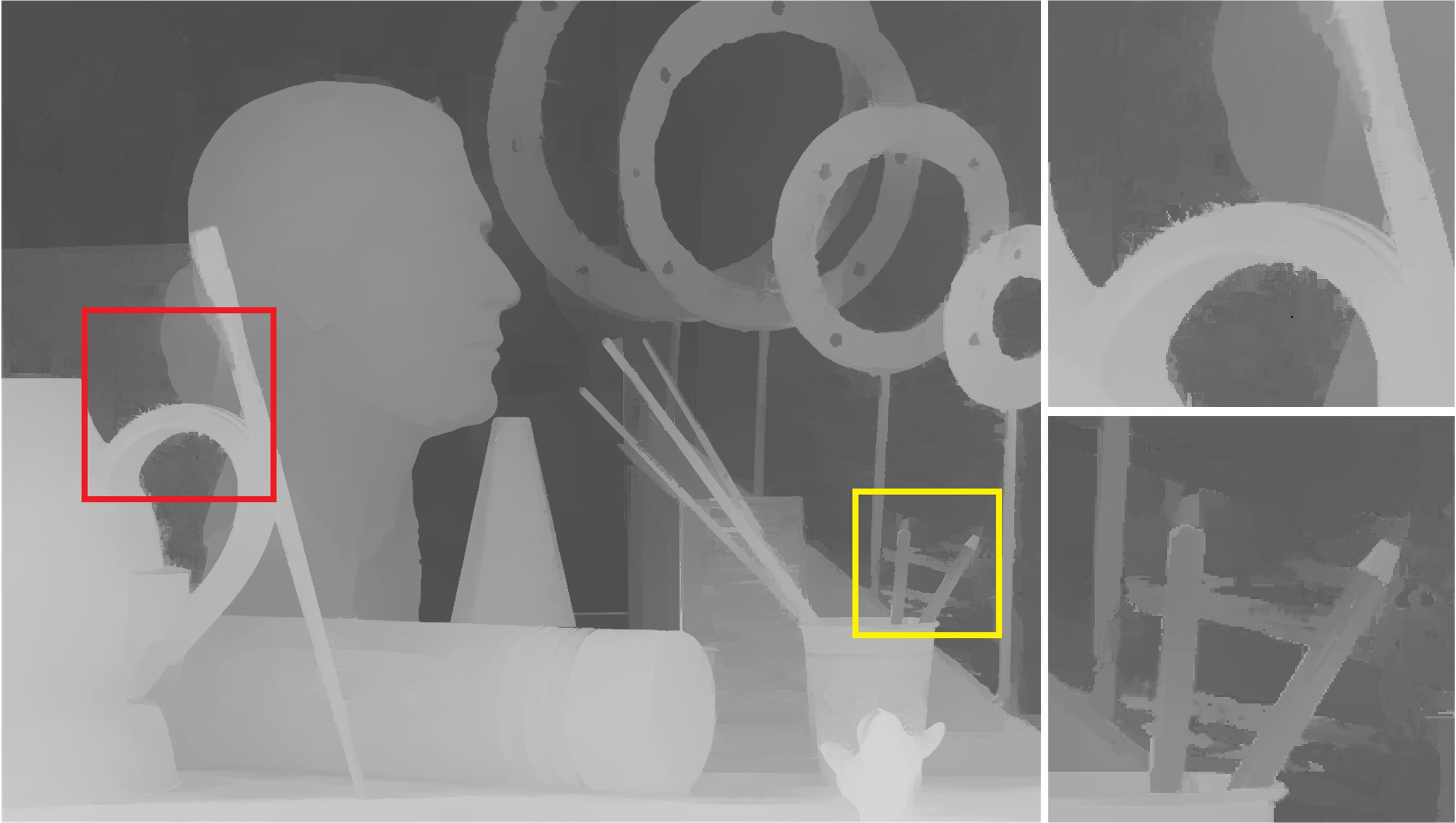} &
  \includegraphics[width=0.23\linewidth]{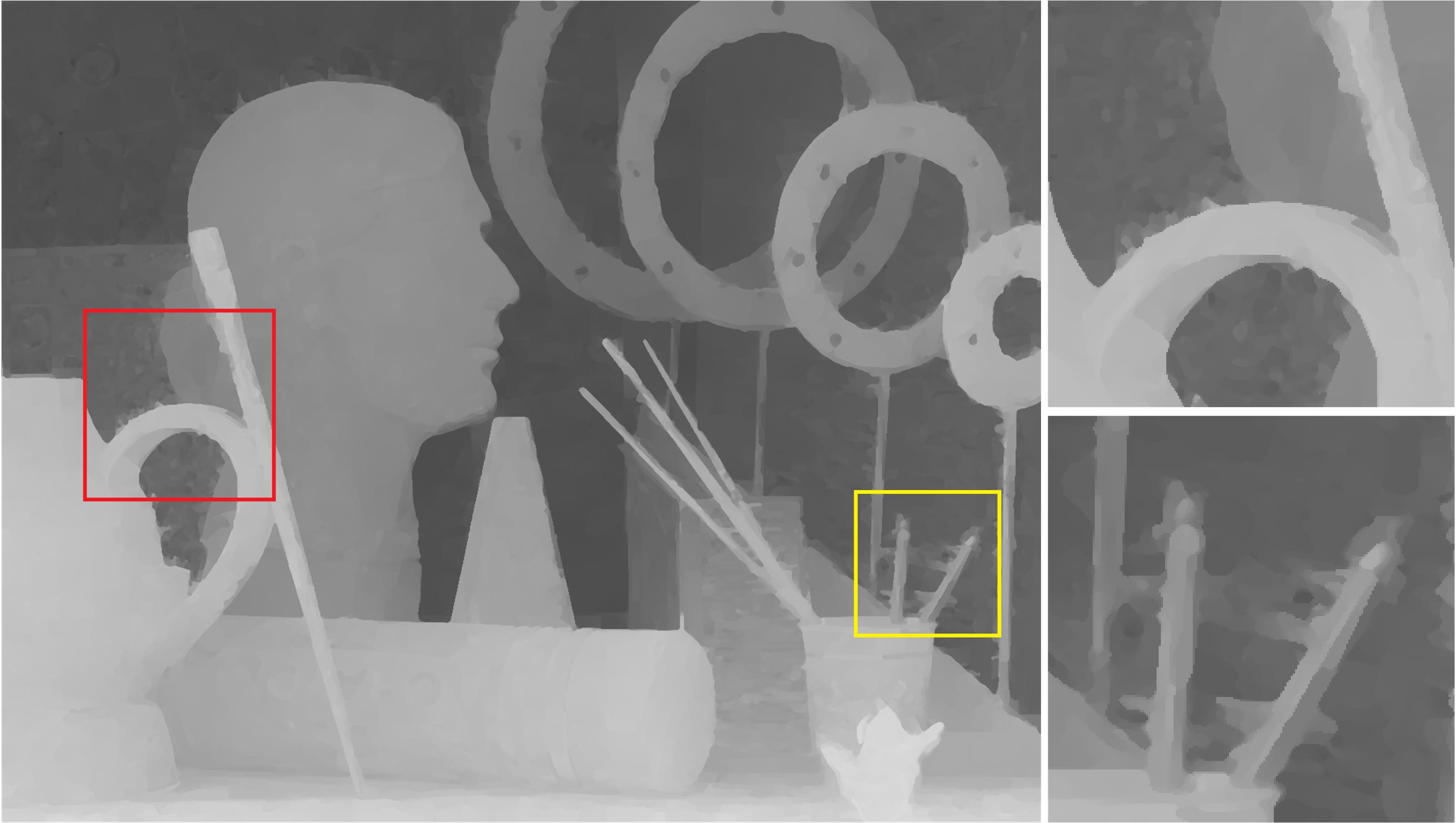} \\
  (a) color/ground-truth & (b) Gu et al. & (c) FBS & (d) SGF\\

  \includegraphics[width=0.23\linewidth]{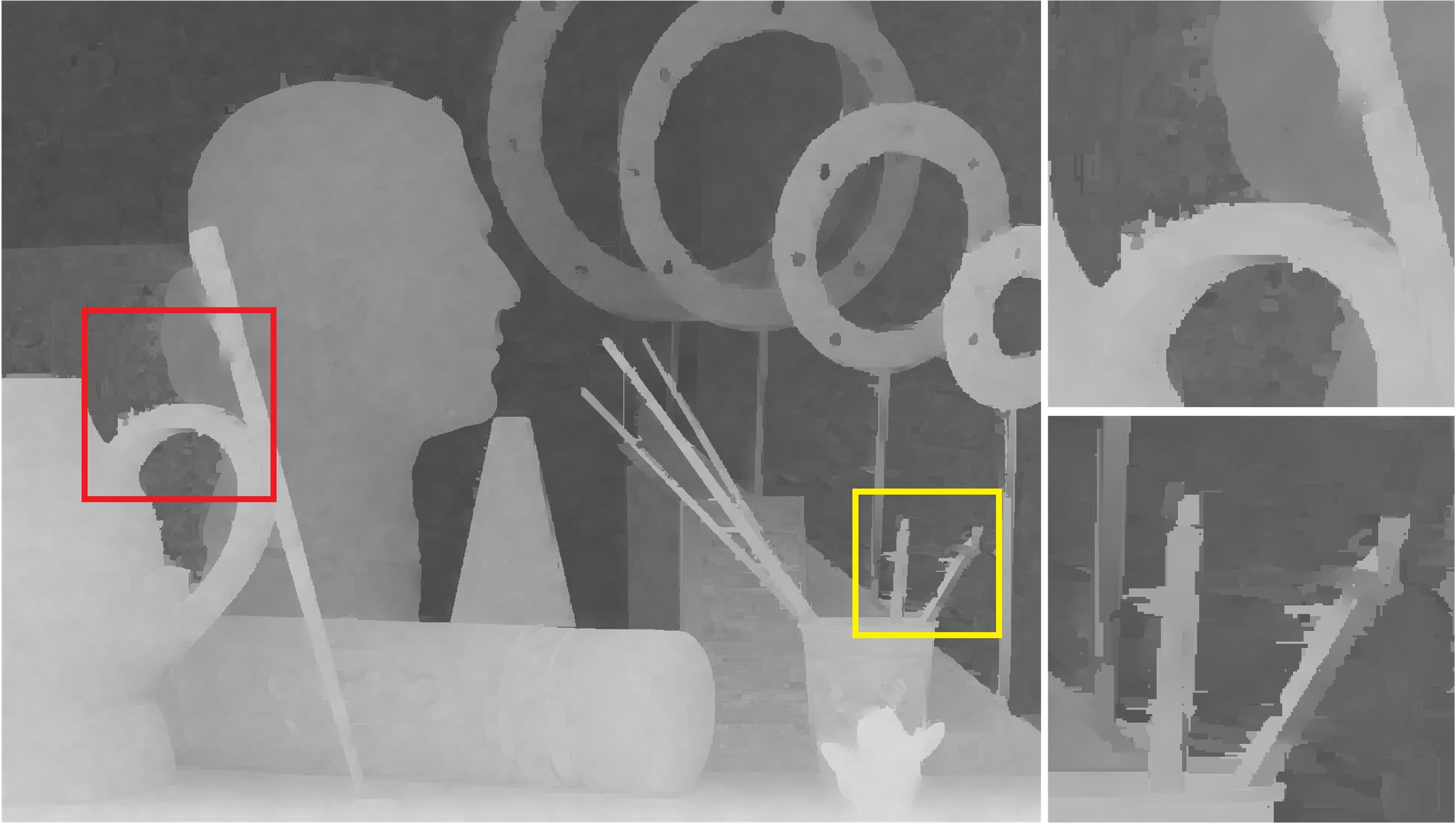} &
  \includegraphics[width=0.23\linewidth]{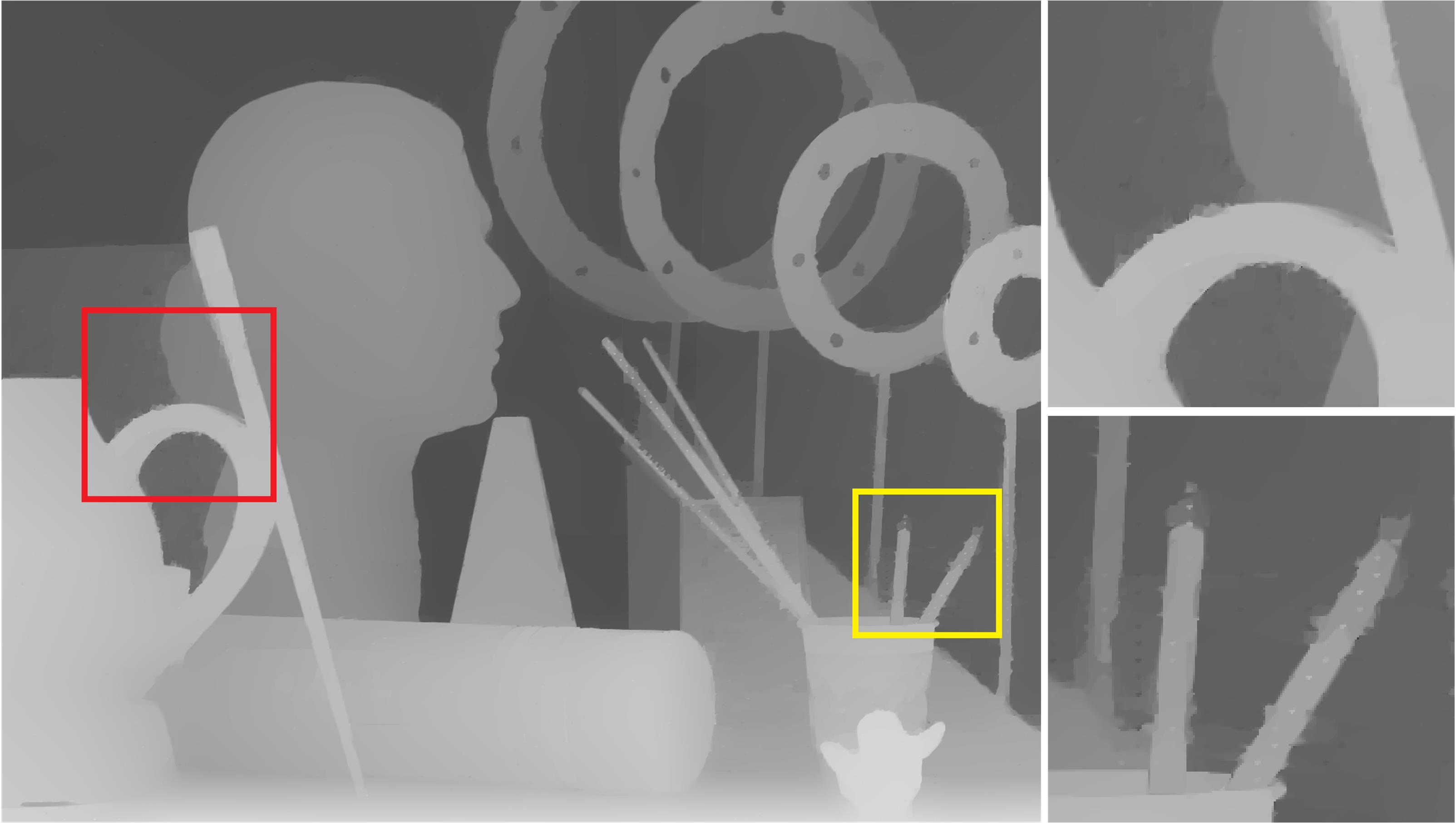} &
  \includegraphics[width=0.23\linewidth]{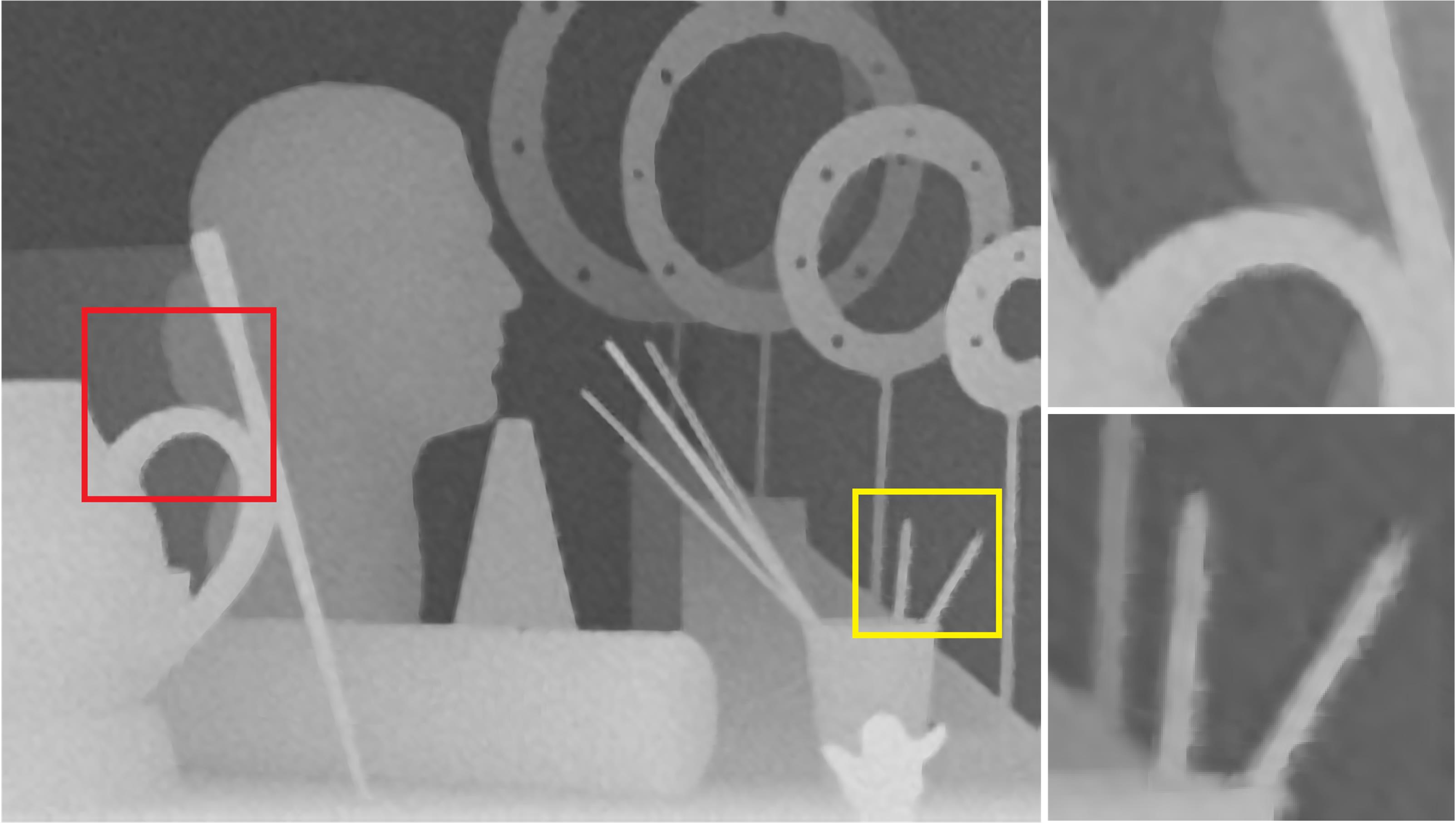} &
  \includegraphics[width=0.23\linewidth]{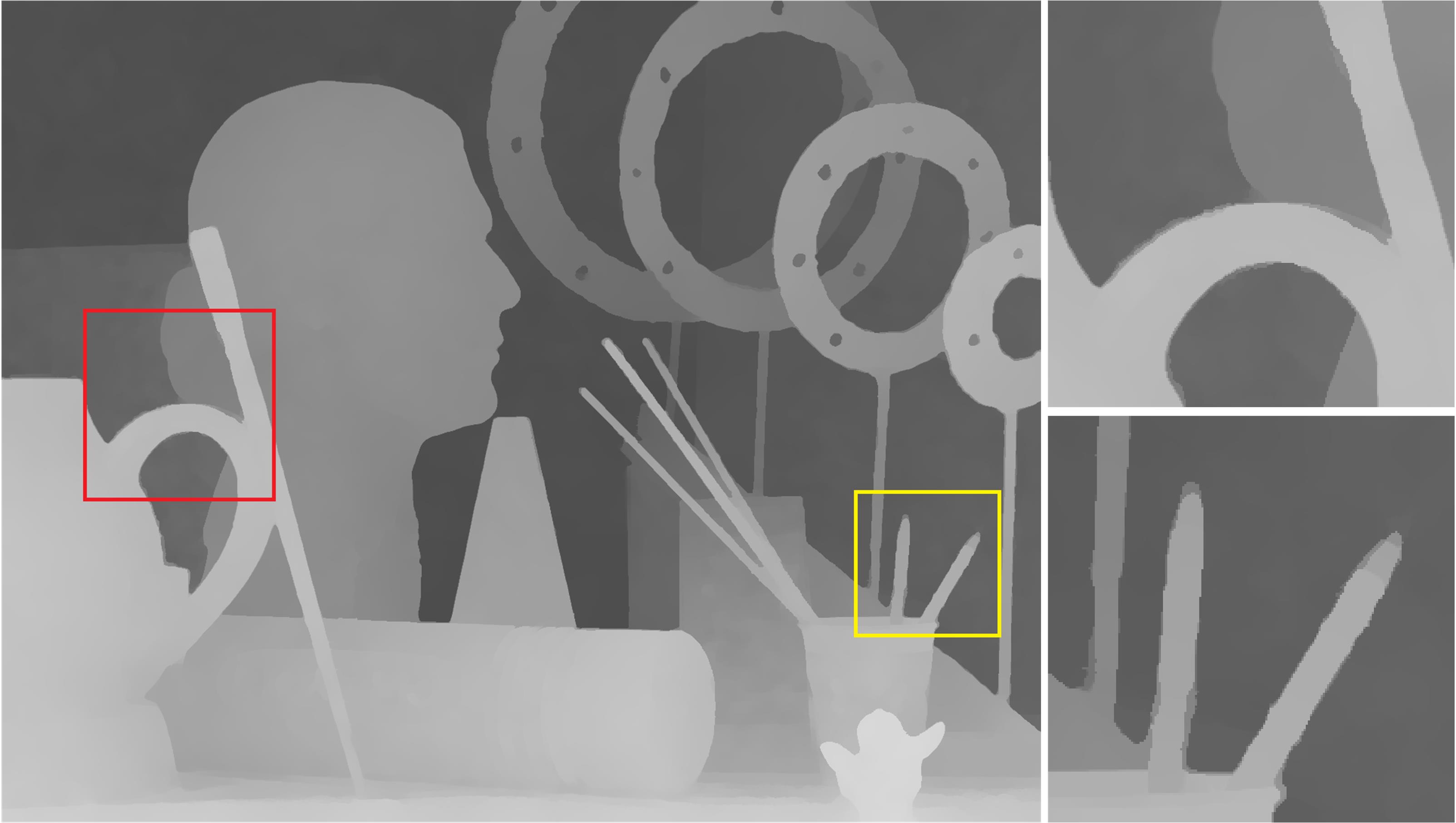} \\
  (e) SD filter & (f) TGV & (g) DJF & (h) ours(EP$\&$SP)\\
  \end{tabular}
  \caption{Guided depth map upsampling results of simulated ToF data. (a) Guidance color image and the ground-truth depth map. 8$\times$ upsampling result of (b) the learning-based approach proposed by Gu et~al. \cite{gu2017learning}, (c) FBS ($\sigma_{xy}=8,\sigma_l=4,\sigma_{uv}=3, \sigma'_{xy}=\sigma'_{rgb}=16, \lambda=4$), (d) SGF \cite{zhang2015segment} ($r=16, \sigma=0.05,\tau=20/225$), (e) SD filter \cite{ham2018robust} ($\lambda=1,\mu=500,\nu=200,k=20$), (f) TGV \cite{ferstl2013image} ($\alpha_0=9, \alpha_1=1,  w=30, \beta=10, \gamma=0.6$), (g) DJF \cite{li2019joint} and (h) our method of the EP$\&$SP mode ($r_d=r_s=5, s=1,\lambda=0.5,b_d=b_s=0.08$).}\label{FigToFSimulated}
\end{figure*}

We collect 40 HDR images to further quantitatively evaluate the performance of all the compared approaches. All the tone mapping results are evaluated in terms of tone mapping quality index (TMQI) proposed by Yeganeh et~al. \cite{yeganeh2012objective}. TMQI first evaluates the structural fidelity and the naturalness of the tone mapped images. The two measurements are then adjusted by a power function and averaged to give a final score ranging from 0 to 1. Larger values of TMQI indicate better quality of the tone mapped images, and vice versa. Tab.~\ref{TabHDR} shows the mean TMQI, structural fidelity and naturalness of the results produced by each of the compared methods. As shown in the table, our method is able to achieve the best performance in most cases.

\subsection{Tasks in the Second Group}

\begin{figure*}[!ht]
  \centering
  \setlength{\tabcolsep}{0.5mm}
  \begin{tabular}{cccc}
  \includegraphics[width=0.23\linewidth]{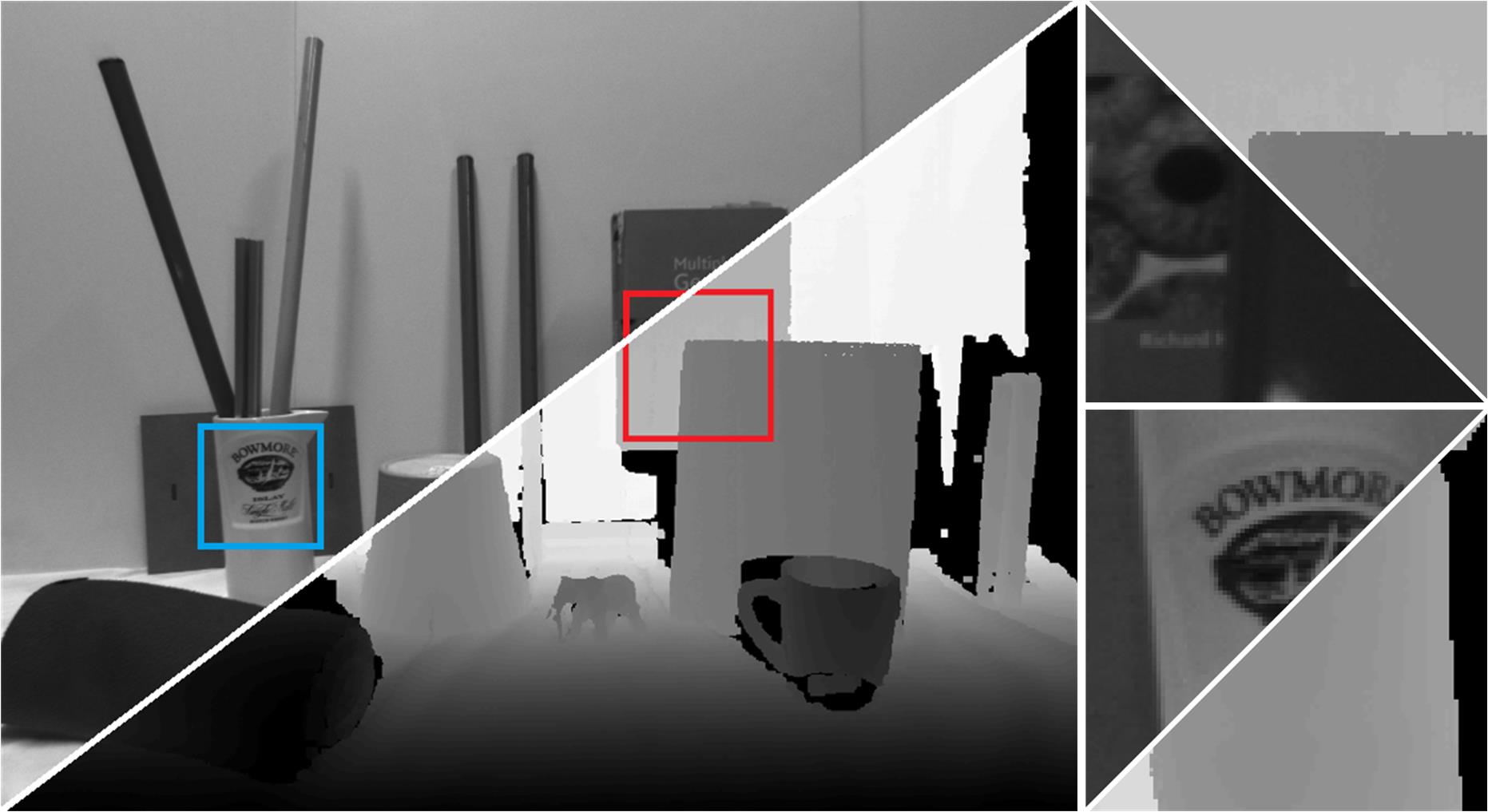} &
  \includegraphics[width=0.23\linewidth]{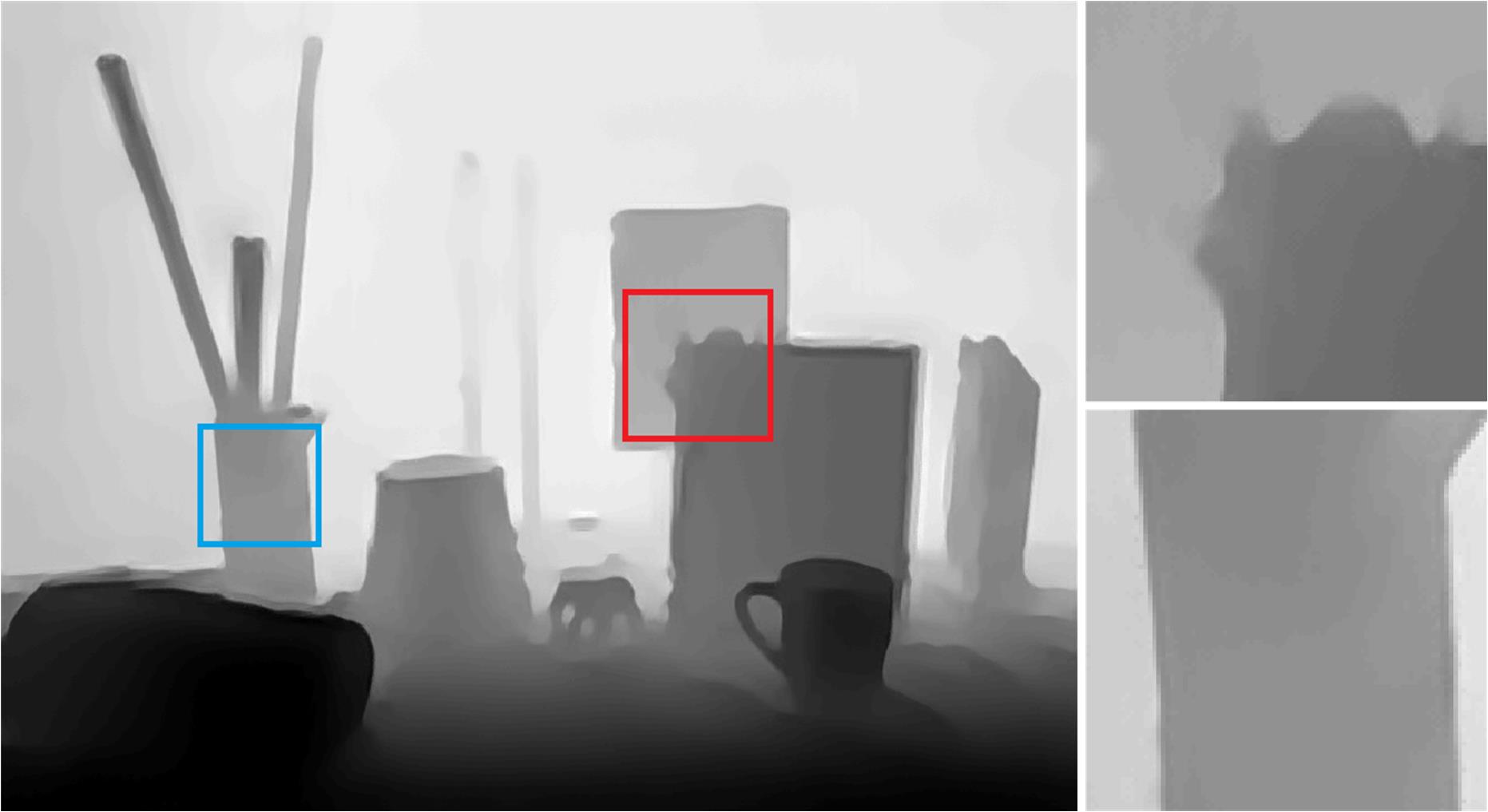} &
  \includegraphics[width=0.23\linewidth]{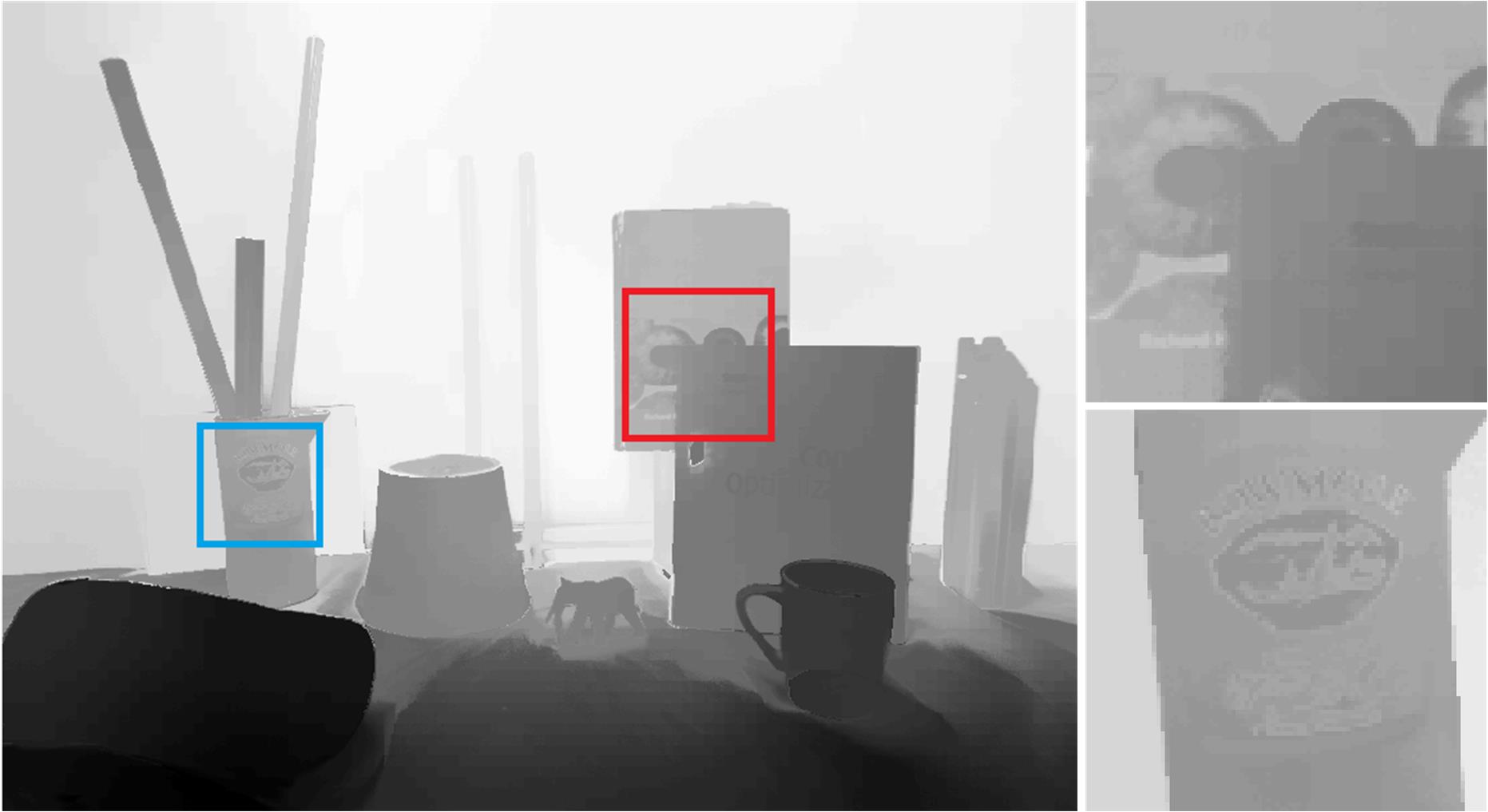} &
  \includegraphics[width=0.23\linewidth]{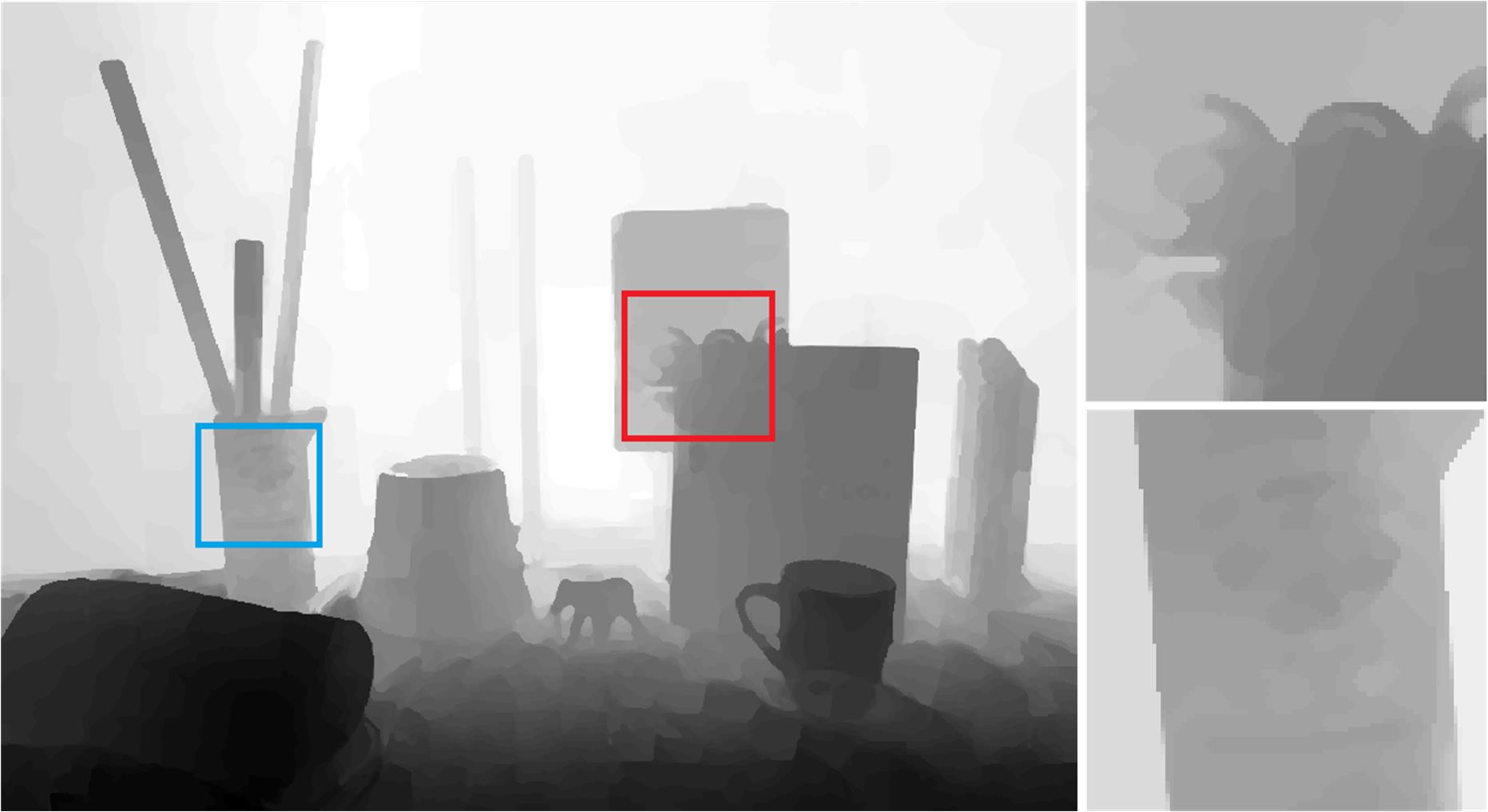} \\
  (a) intensity/ground-truth & (b) Gu et al. & (c) FBS & (d) SGF\\

  \includegraphics[width=0.23\linewidth]{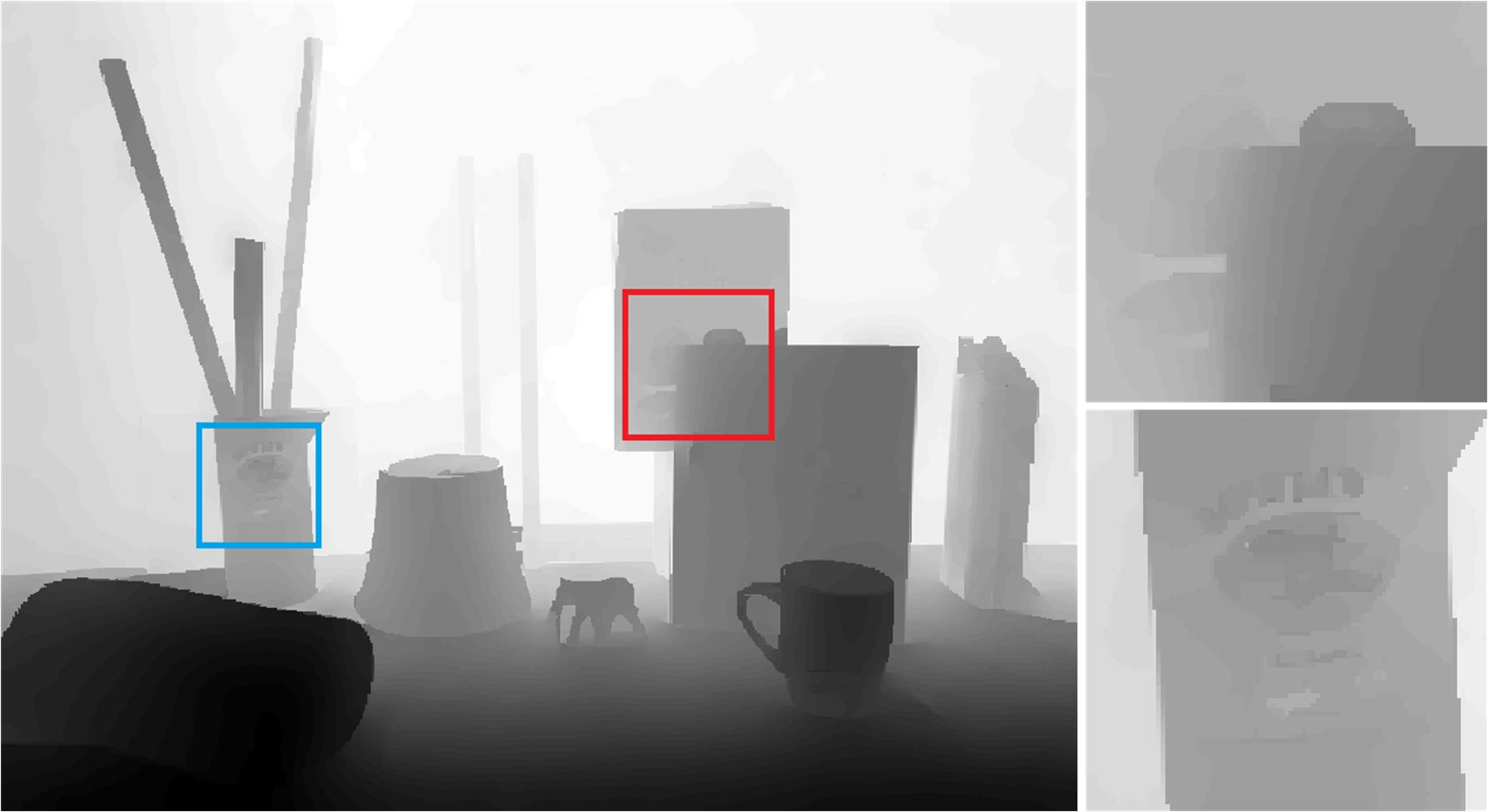} &
  \includegraphics[width=0.23\linewidth]{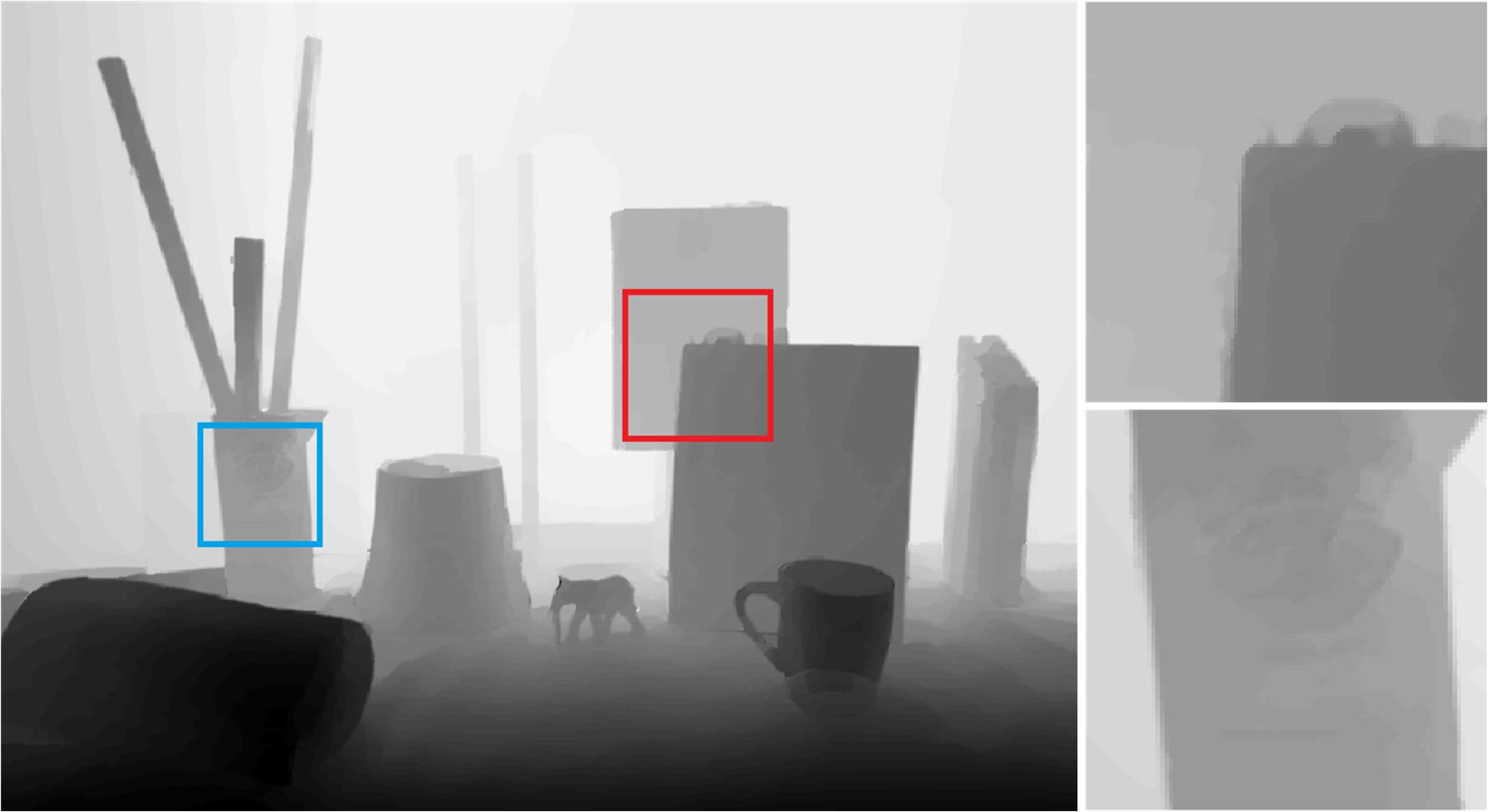} &
  \includegraphics[width=0.23\linewidth]{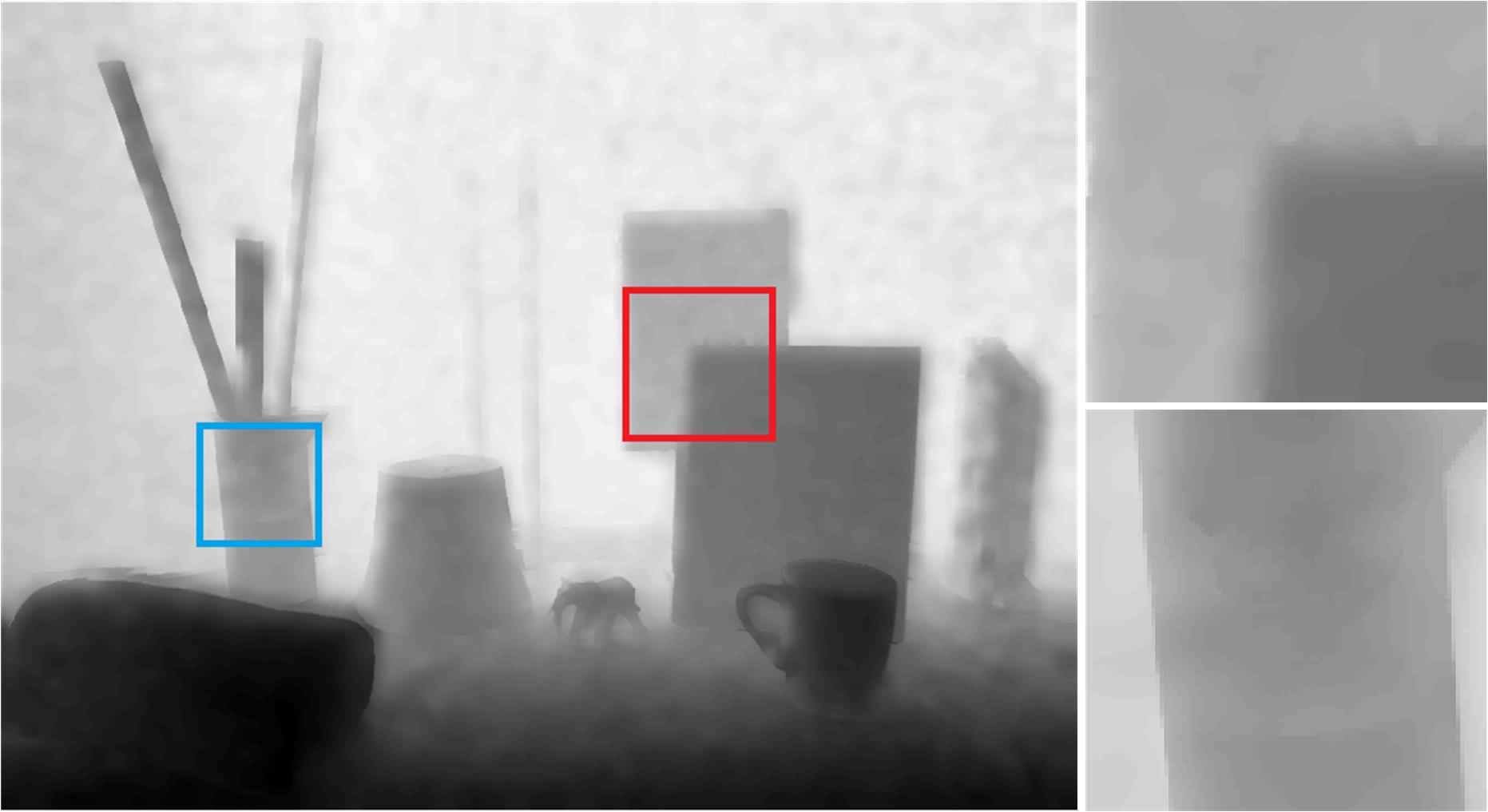} &
  \includegraphics[width=0.23\linewidth]{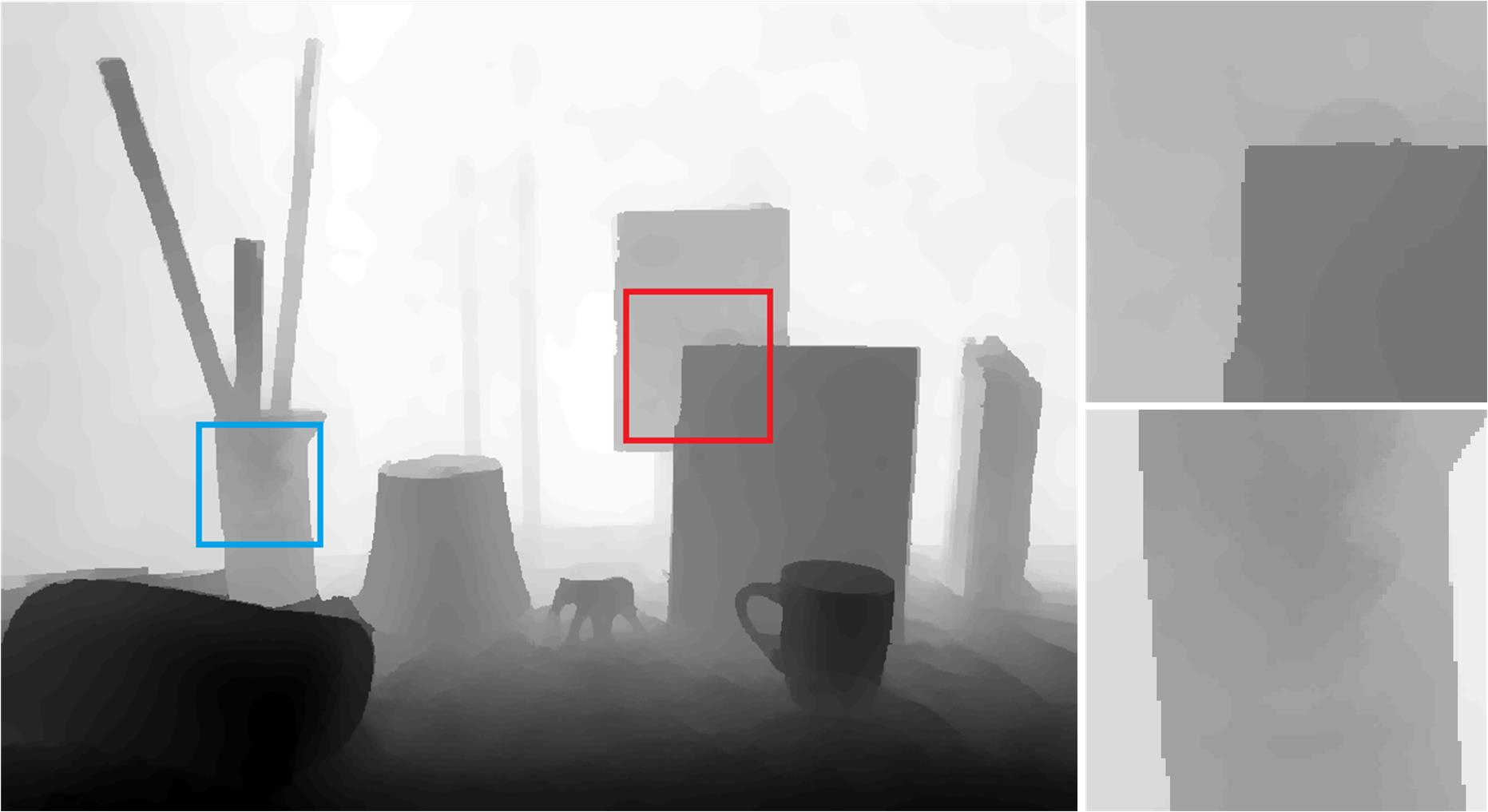} \\
  (e) SD filter & (f) TGV & (g) DJF & (h) ours(EP$\&$SP)\\
  \end{tabular}
  \caption{Guided depth upsampling results of real ToF data. (a) Guidance intensity image and the ground-truth depth map. Upsampling result of (b) the learning-based approach proposed by Gu et~al. \cite{gu2017learning}, (c) FBS ($\sigma_{xy}=8,\sigma_l=4,\sigma_{uv}=3, \sigma'_{xy}=\sigma'_{rgb}=16, \lambda=2.5$),  (d) SGF \cite{zhang2015segment} ($r=16, \sigma=0.075,\tau=30/225$), (e) SD filter \cite{ham2018robust}  ($\lambda=10,\mu=500,\nu=200,k=20$), (f) TGV \cite{ferstl2013image} ($\alpha_0=21.5,\alpha_1=0.75,w=2000, \beta=22, \gamma=0.8$),  (g) DJF \cite{li2019joint} and (h) our method of the EP$\&$SP mode ($r_d=r_s=3,s=1,\lambda=0.5,b_d=b_s=0.08$).}\label{FigToFReal}
\end{figure*}

 We also apply our method to clip-art compression artifacts removal, and the EP$\&$SP mode of our method is adopted. The input image is used as the guidance image in our method, i.e., $g=f$. Clip-art images are piecewise constant with sharp edges. When they are compressed in JPEG format with a low quality factor, there will be edge-related artifacts, and the edges are usually blurred as shown in Fig.~\ref{FigClipArt}(a). Therefore, when removing the compression artifacts, the edges should also be sharpened in the restored image. We evaluate all the compared methods with 30 different collected images. These images are first compressed in JPEG format with the quality factor ranging from 10 to 90. The parameters of our method are set as follows: $r_d=r_s=2$, $\lambda=0.4, b_d=b_s=0.15$ for the compression quality factor of 10, $\lambda$ is divided by 2 and $b_d=b_s$ is decreased by 0.01 when the quality factor is increased by 10. The values of the other parameters are fixed as those in Tab.~\ref{TabParameter}.

 Fig.~\ref{FigClipArt} shows the visual comparison of the results produced by different methods. The approach proposed by Wang et~al. \cite{wang2006deringing} can seldom handle heavy compression artifacts. Their result is shown in Fig.~\ref{FigClipArt}(c). Cho et~al. \cite{cho2014bilateral} also applied their bilateral texture filter (BTF) to compression artifacts removal, however, their method can blur edges instead of sharpening them, as shown in Fig.~\ref{FigClipArt}(d). The $L_0$ norm smoothing \cite{xu2011image} can eliminate most compression artifacts and properly sharpen salient edges, but it fails to preserve weak edges as shown in Fig.~\ref{FigClipArt}(e). The region fusion approach \cite{nguyen2015fast} is able to produce results with sharpened edges, however, it also enhances the blocky artifacts along strong edges as highlighted in Fig.~\ref{FigClipArt}(f). The recently proposed deep image prior method \cite{ulyanov2018deep} is a deep learning based approach which is applicable to clip-art compression artifacts removal, however, their method cannot sharpen edges and can also result in incorrect colors for small structures, as highlighted in Fig.~\ref{FigClipArt}(g). Our result is illustrated in Fig.~\ref{FigClipArt}(h) with edges sharpened and compression artifacts properly removed. Tab.~\ref{TabClipArt} further shows the quantitative evaluation of the results produced by different approaches. The evaluation is performed with 30 different collected images. The mean PSNR and SSIM between the restored image and the un-compressed image are adopted as evaluation metrics. As shown in the table, our method achieves the best performance in most cases.

 \subsection{Tasks in the Third Group}

Guided depth map upsampling belongs to the guided image filtering in the third group. Depth maps captured by modern depth cameras (e.g., ToF depth camera) are usually of low resolution and contain heavy noise. To boost the resolution and quality, one way is to upsample the depth map with the guidance of a high-resolution RGB image that captures the same scene. The RGB image is usually of high quality and can provide additional structural information to restore and sharpen the depth edges. The challenge of this task is the structure inconsistency between the depth map and the RGB guidance image, which can cause blurring depth edges and texture copy artifacts in the upsampled depth map.

\begin{figure*}
  \centering
  \setlength{\tabcolsep}{0.5mm}
  \begin{tabular}{cccc}
  \includegraphics[width=0.23\linewidth]{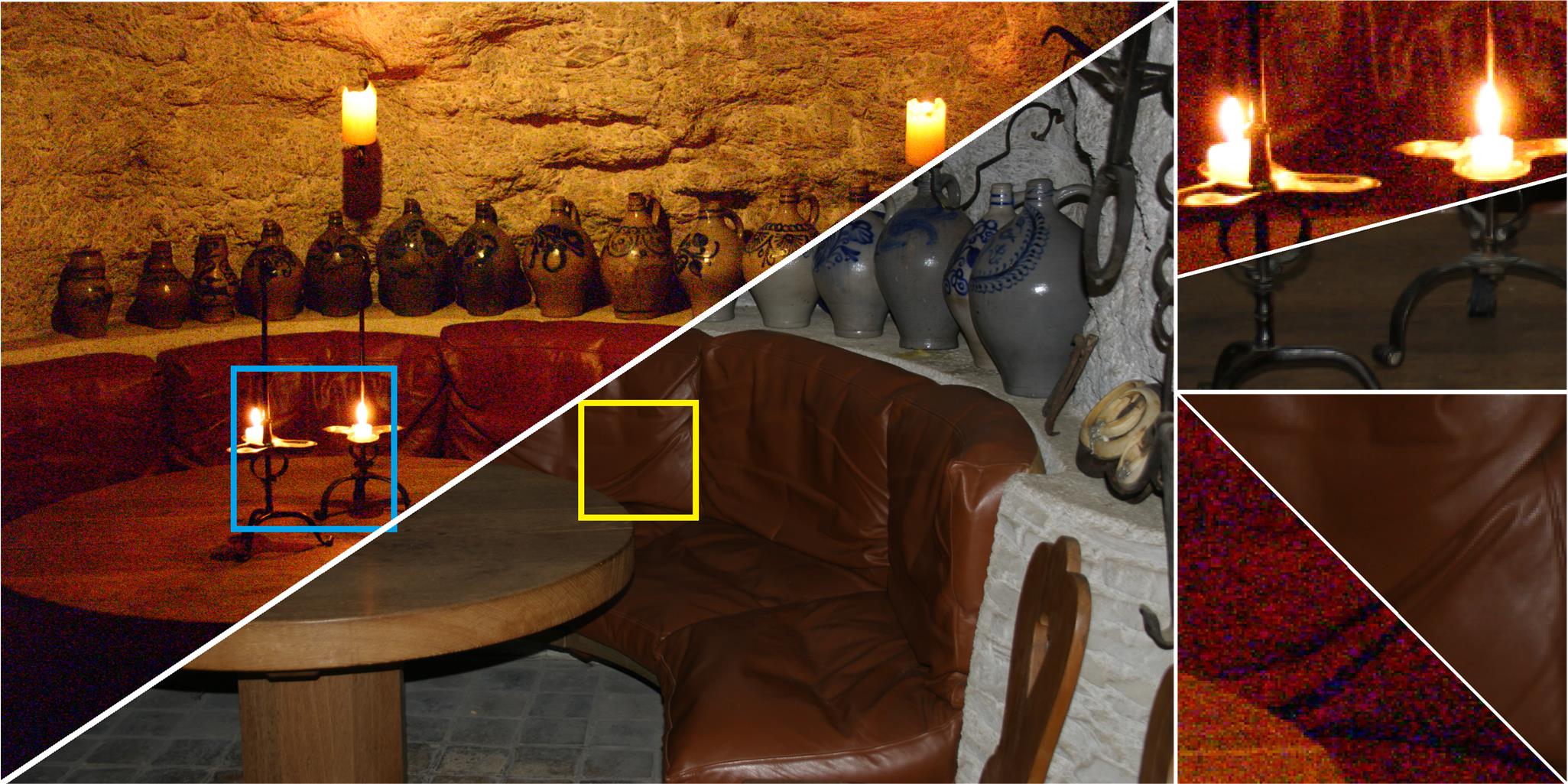} &
  \includegraphics[width=0.23\linewidth]{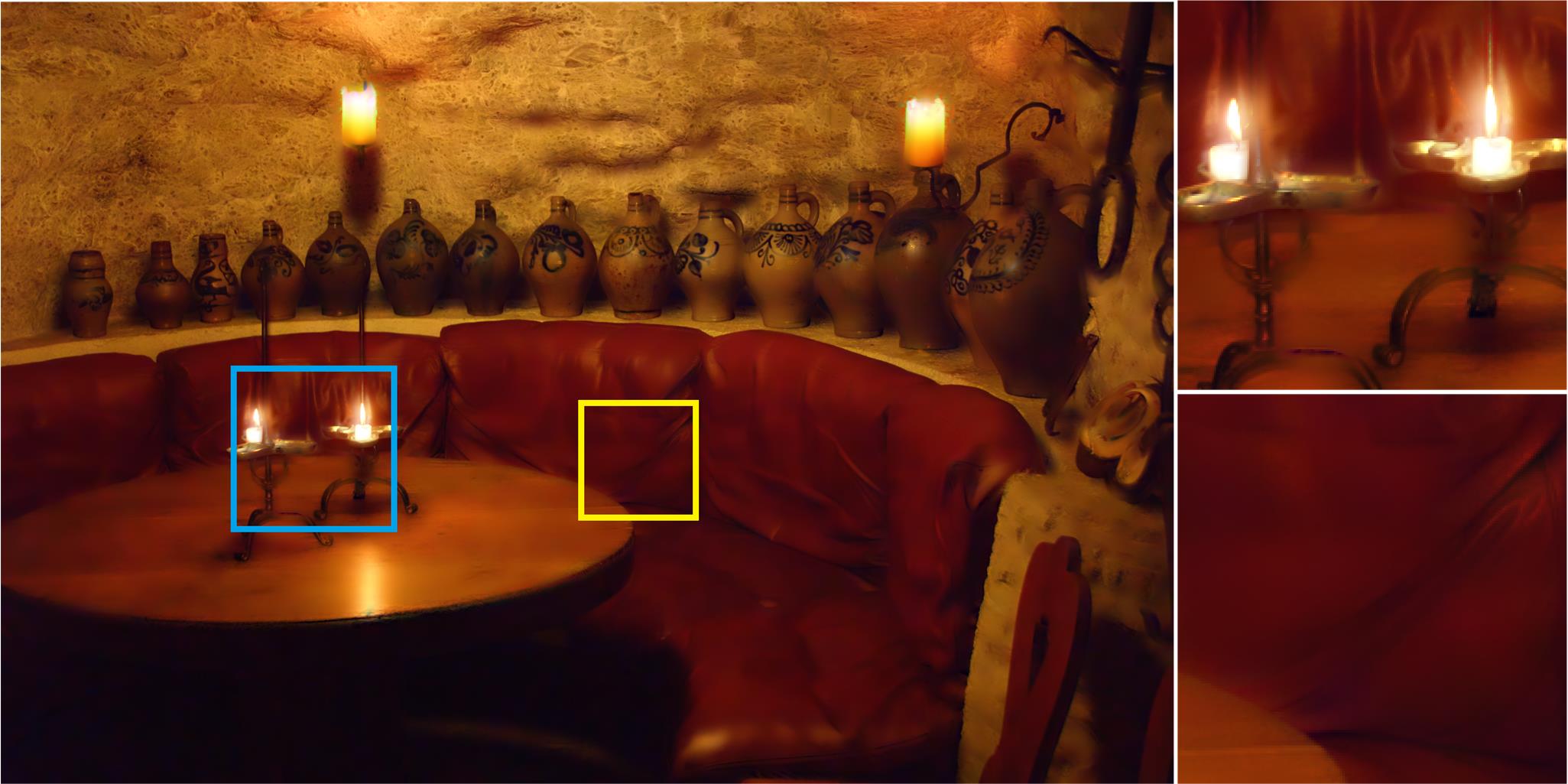} &
  \includegraphics[width=0.23\linewidth]{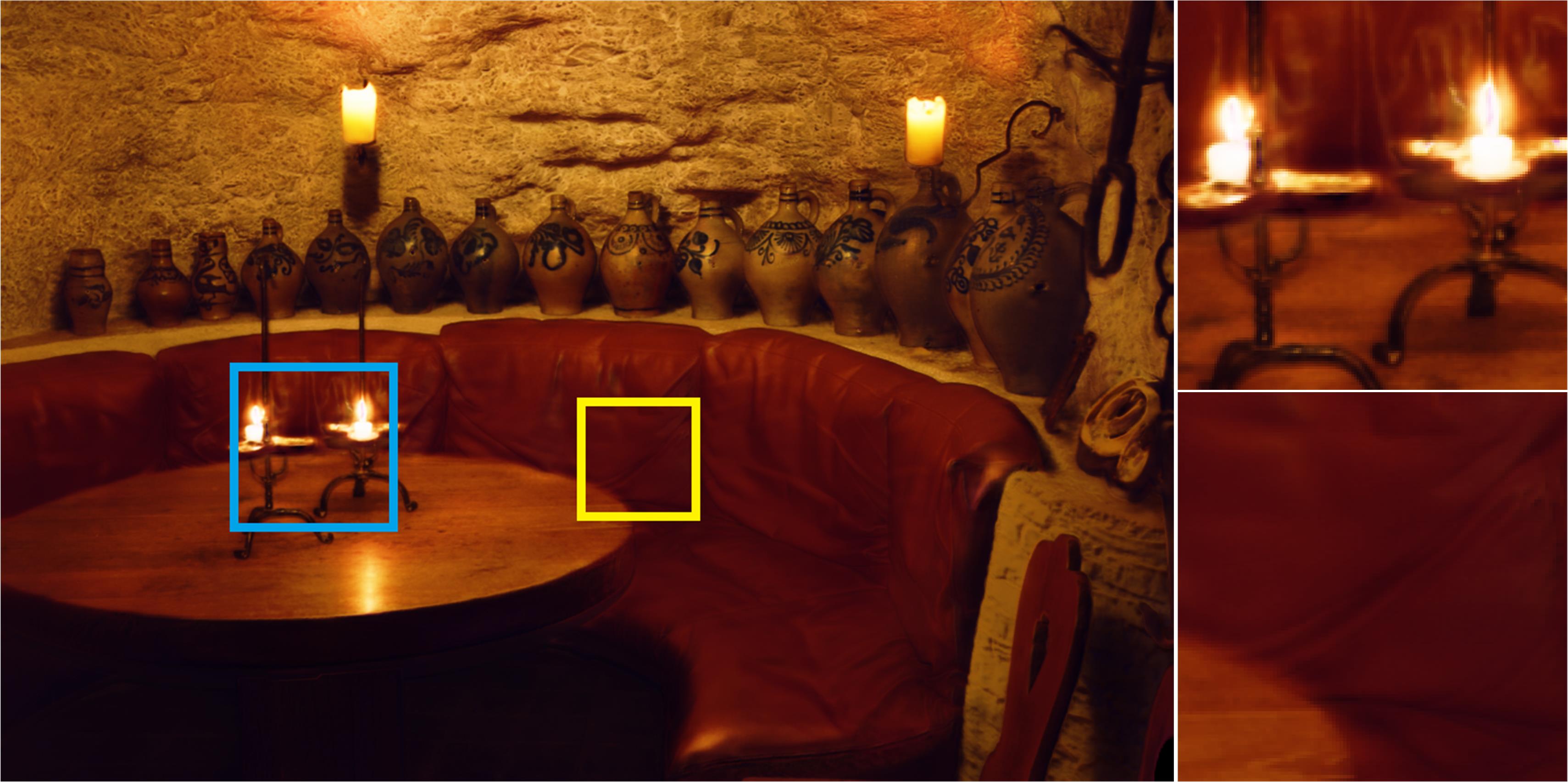} &
  \includegraphics[width=0.23\linewidth]{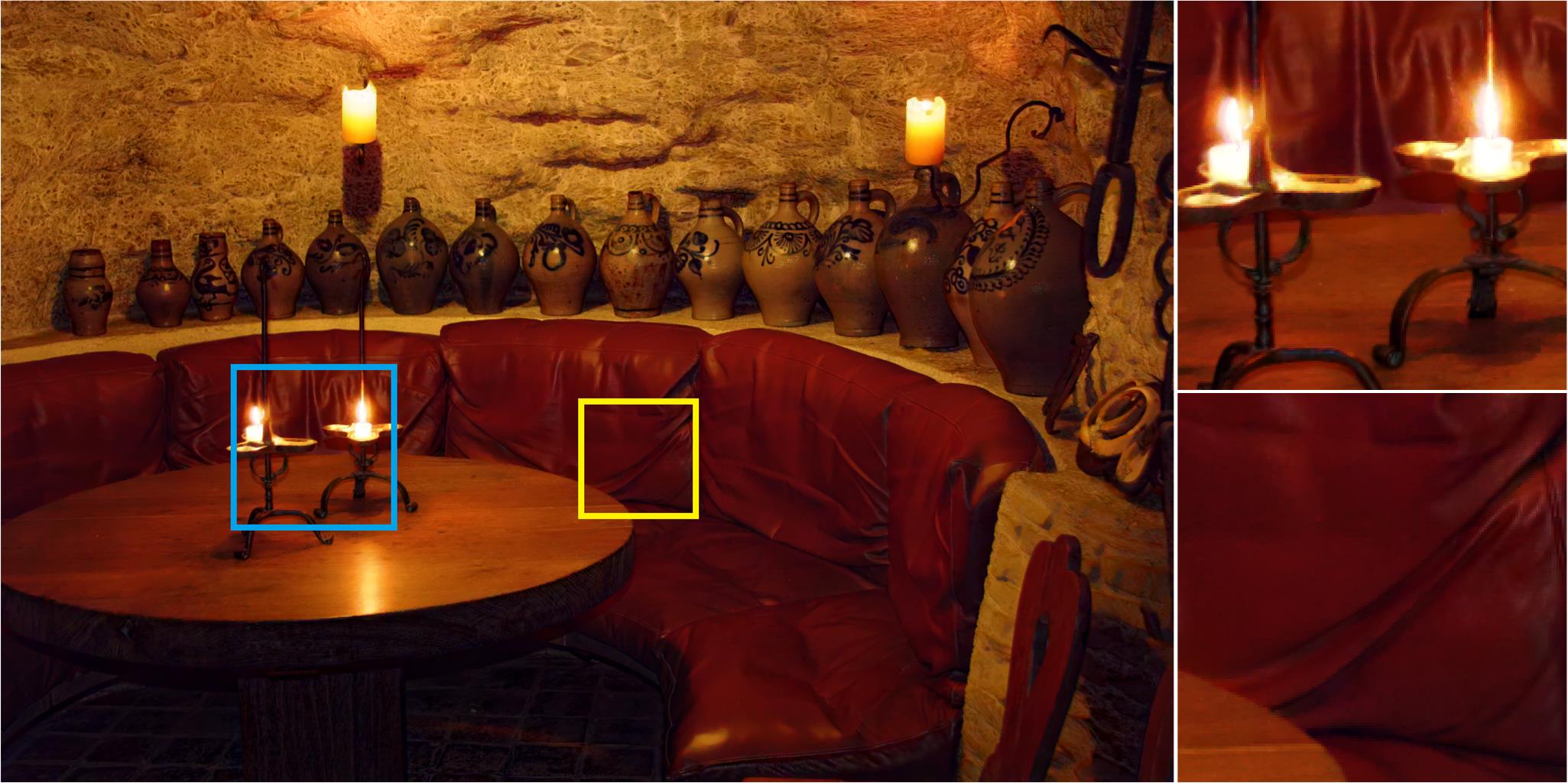} \\
  (a) flash/no flash & (b) GF & (c) deep image prior & (d) optimized scale map \\

  \includegraphics[width=0.23\linewidth]{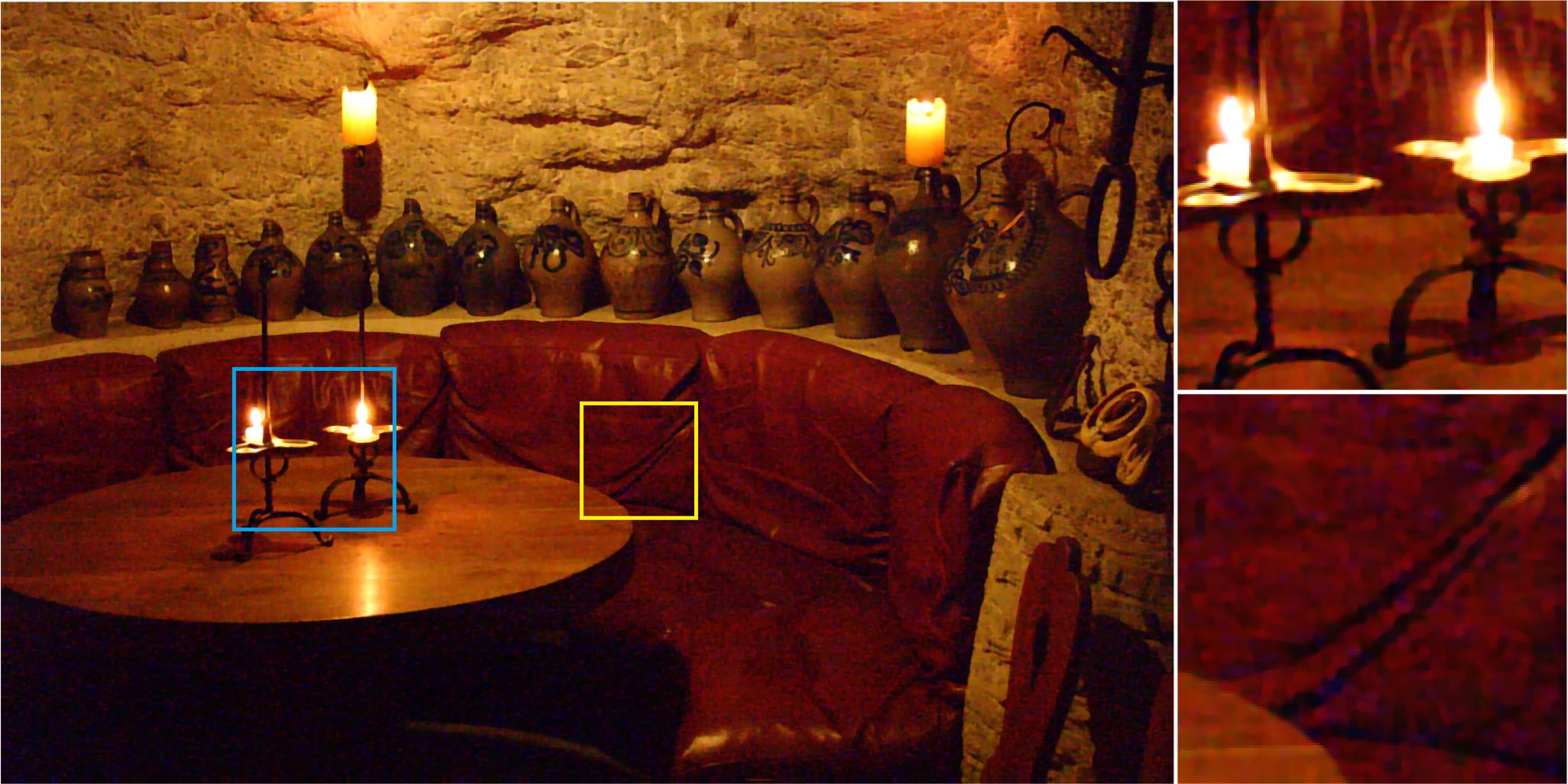} &
  \includegraphics[width=0.23\linewidth]{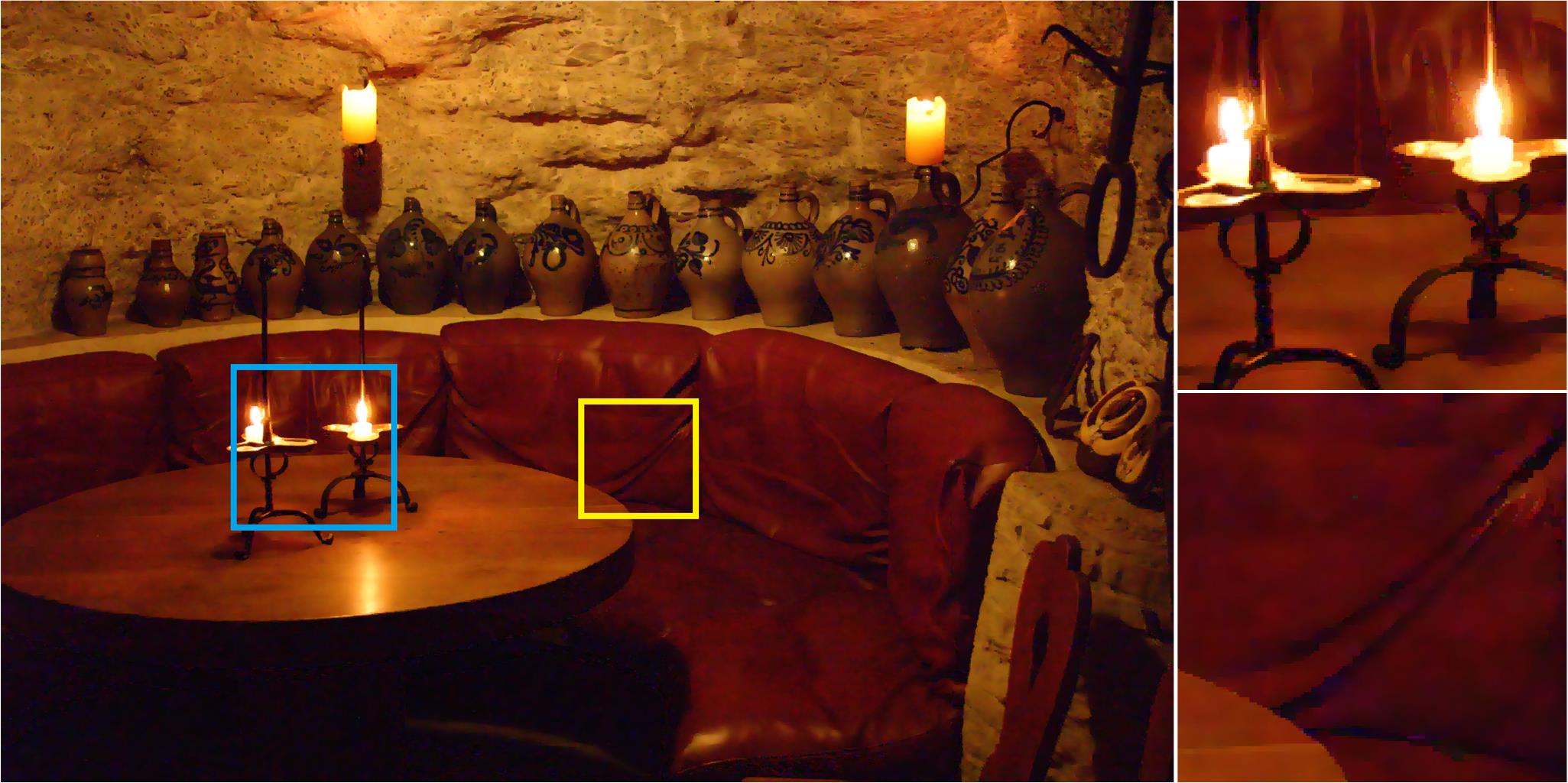} &
  \includegraphics[width=0.23\linewidth]{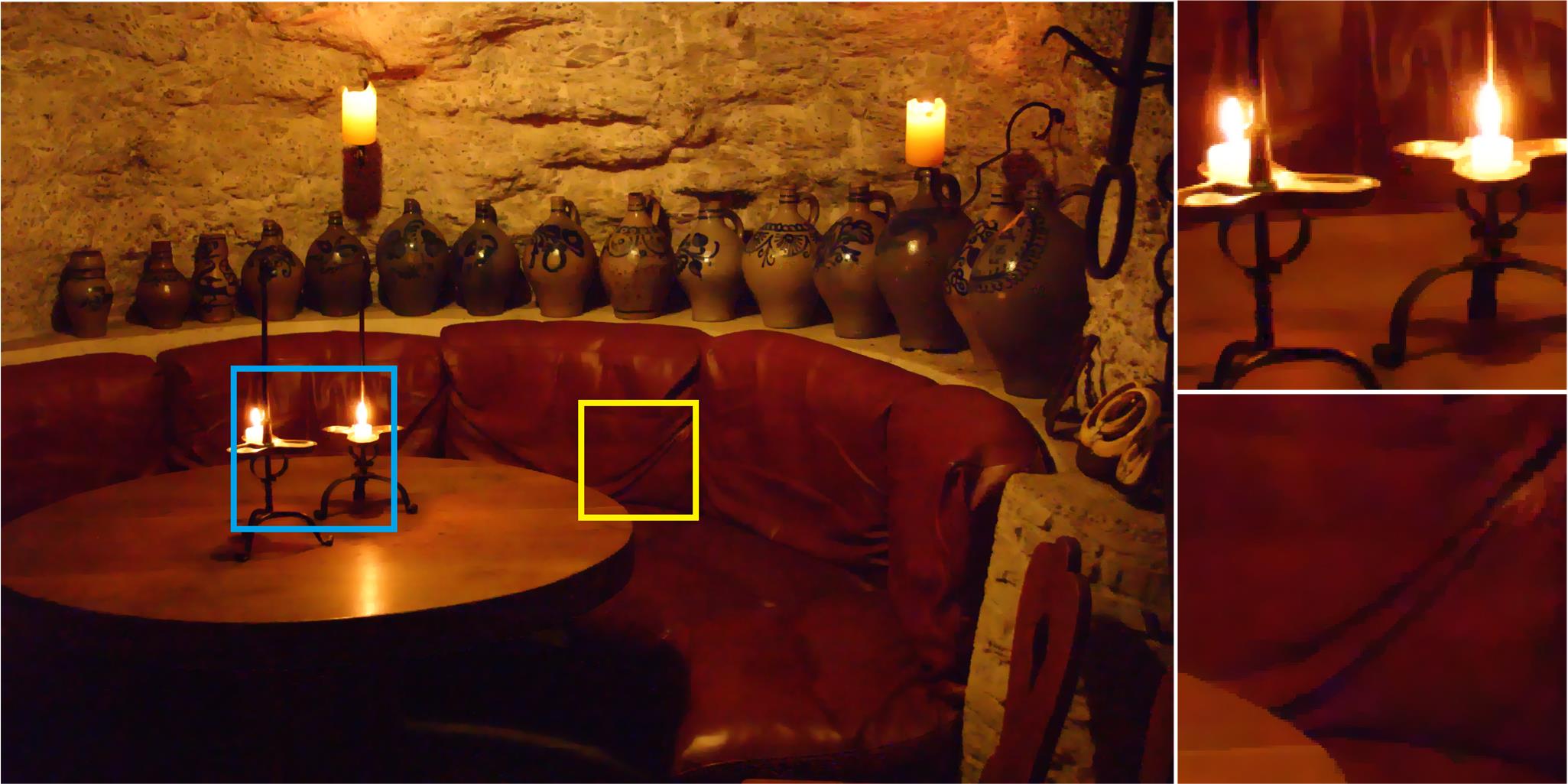} &
  \includegraphics[width=0.23\linewidth]{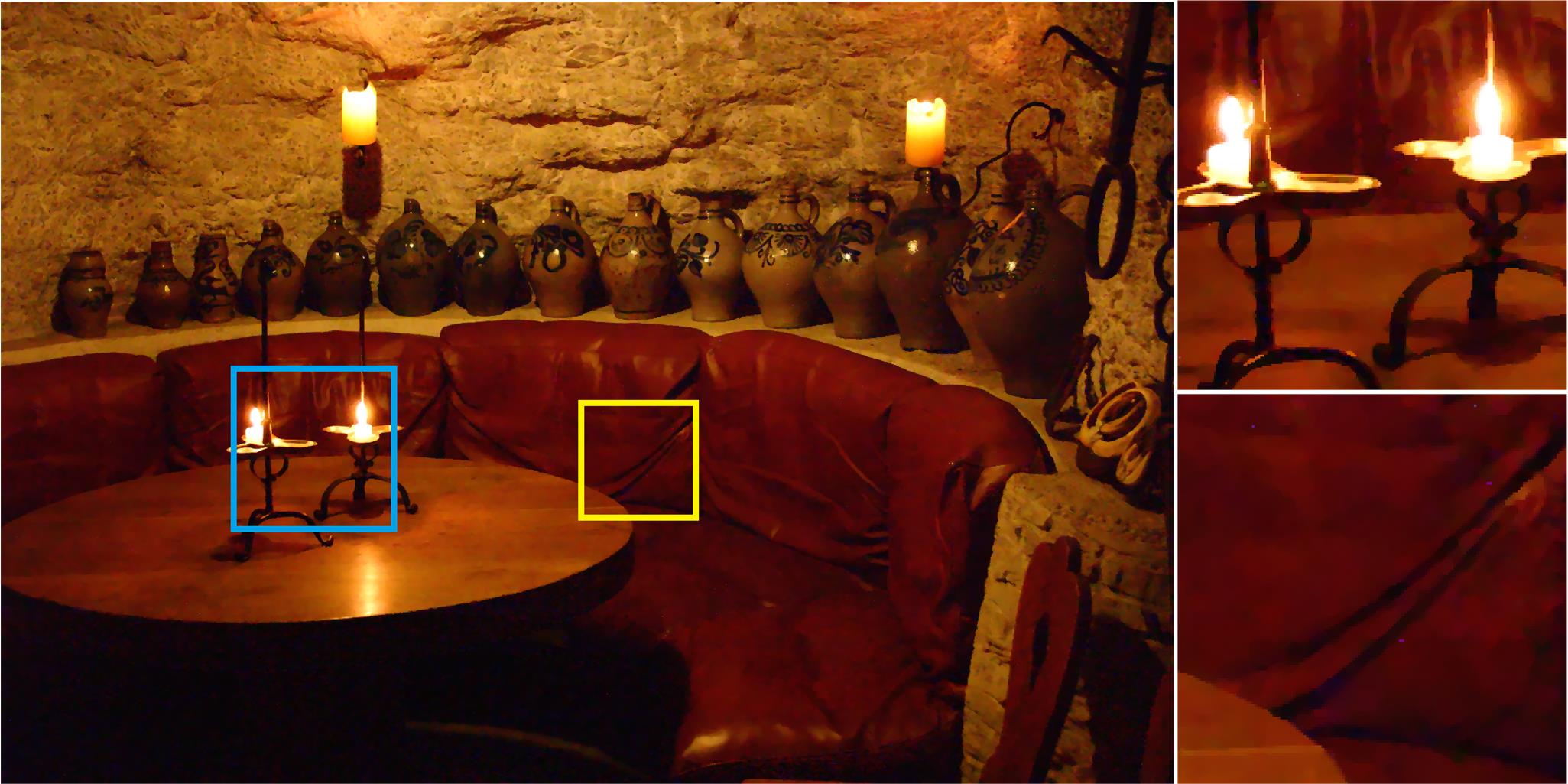} \\
  (e) DJF & (f) SD filter & (g) muGIF & (h) ours(EP$\&$SP)
  \end{tabular}
  \caption{Flash/no flash image filtering results of different methods. (a) Guidance flash image and no-flash image to be filtered. Result of (b) GF \cite{he2013guided} ($r=8, \epsilon=0.2^2$), (c) deep image prior \cite{ulyanov2018deep}, (d) optimal scale map filter \cite{shen2015multispectral} ($\lambda=6,\beta=0.6$), (e) DJF \cite{li2019joint}, (f) SD filter \cite{ham2018robust} ($\lambda=15,\mu=60,\nu=30,k=5$),  (g) muGIF \cite{guo2018mutually} ($\alpha_t=0.02, \alpha_r=0, N=10$) and (h) our method of the EP$\&$SP mode ($r_d=r_s=1,\lambda=0.1,b_d=b_s=0.15$).}\label{FigFlashNoFlash}
\end{figure*}

\begin{figure*}
  \centering
  \setlength{\tabcolsep}{0.5mm}
  \begin{tabular}{cccc}
  \includegraphics[width=0.23\linewidth]{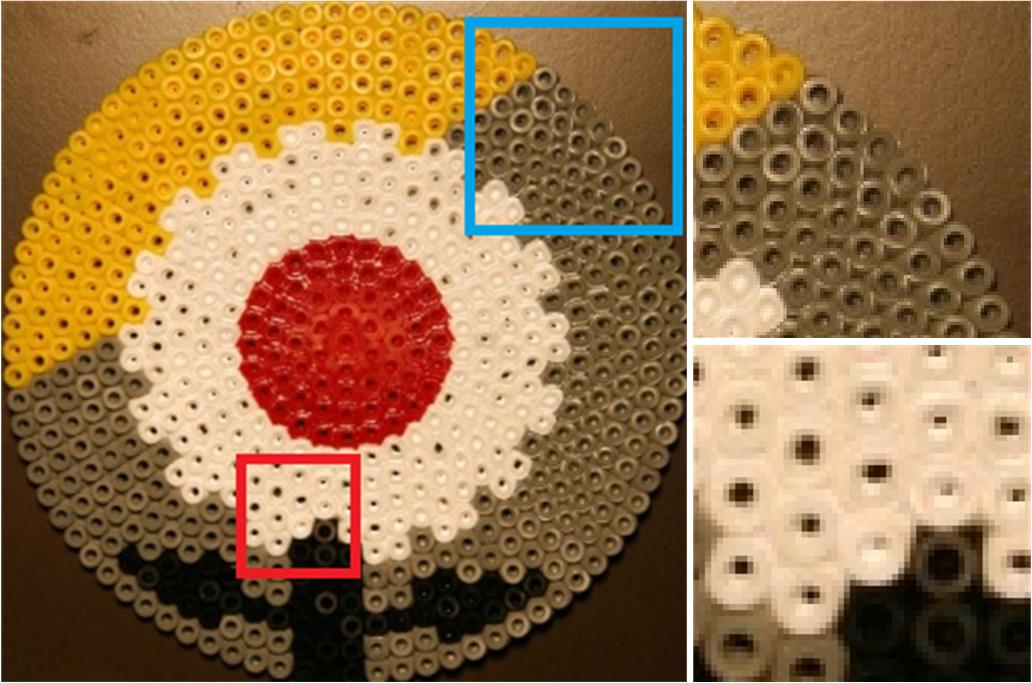} &
  \includegraphics[width=0.23\linewidth]{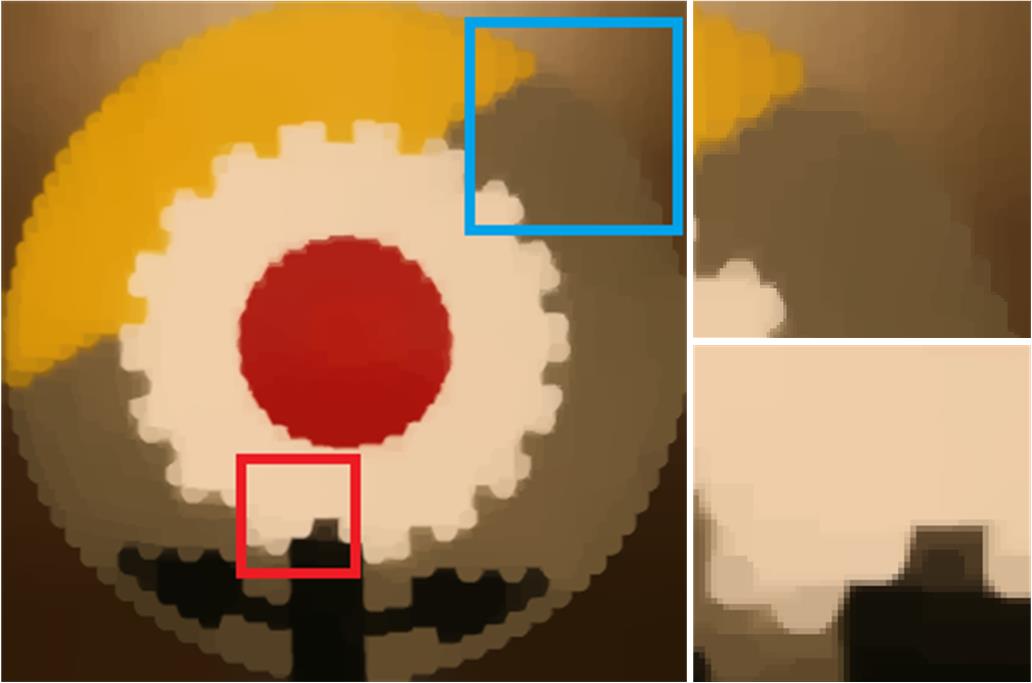} &
  \includegraphics[width=0.23\linewidth]{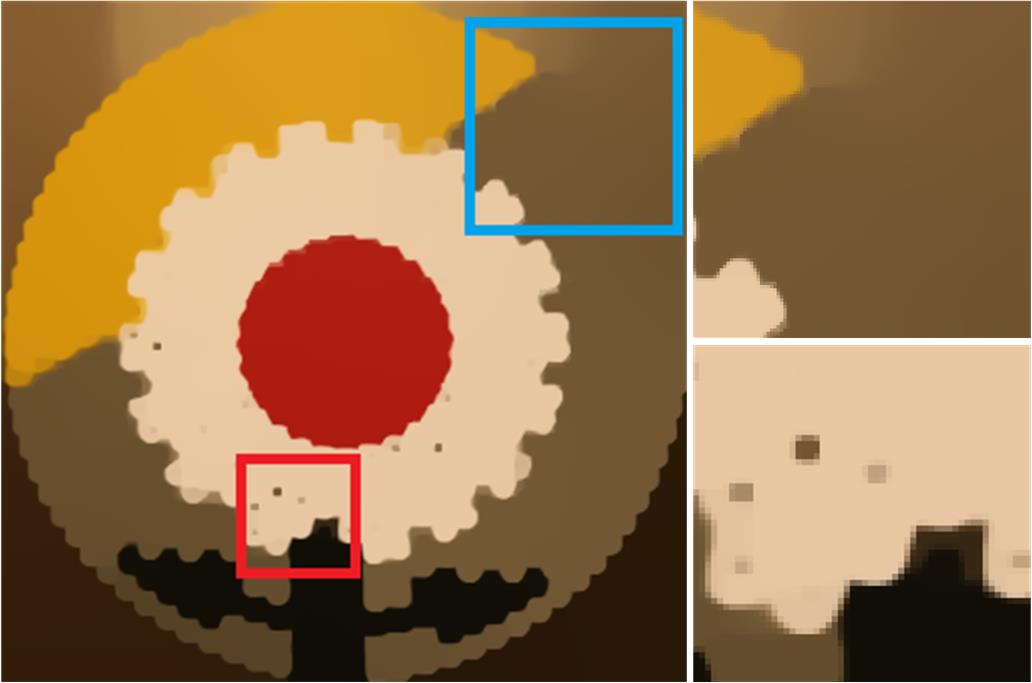} &
  \includegraphics[width=0.23\linewidth]{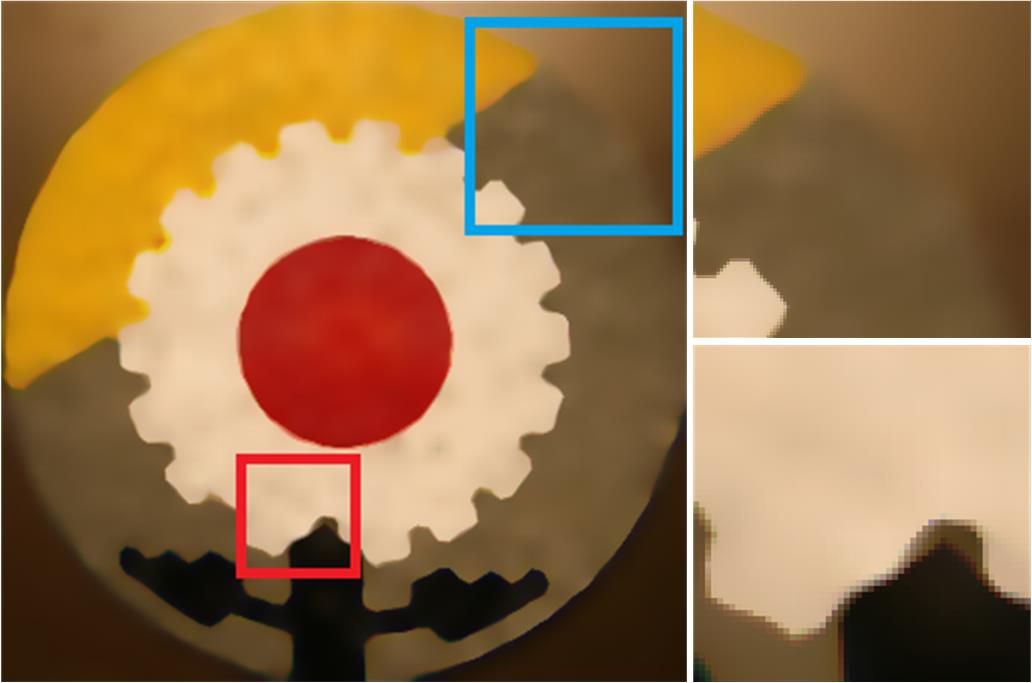} \\
  (a) input & (b) JCAS & (c) RTV & (d) RGF \\

  \includegraphics[width=0.23\linewidth]{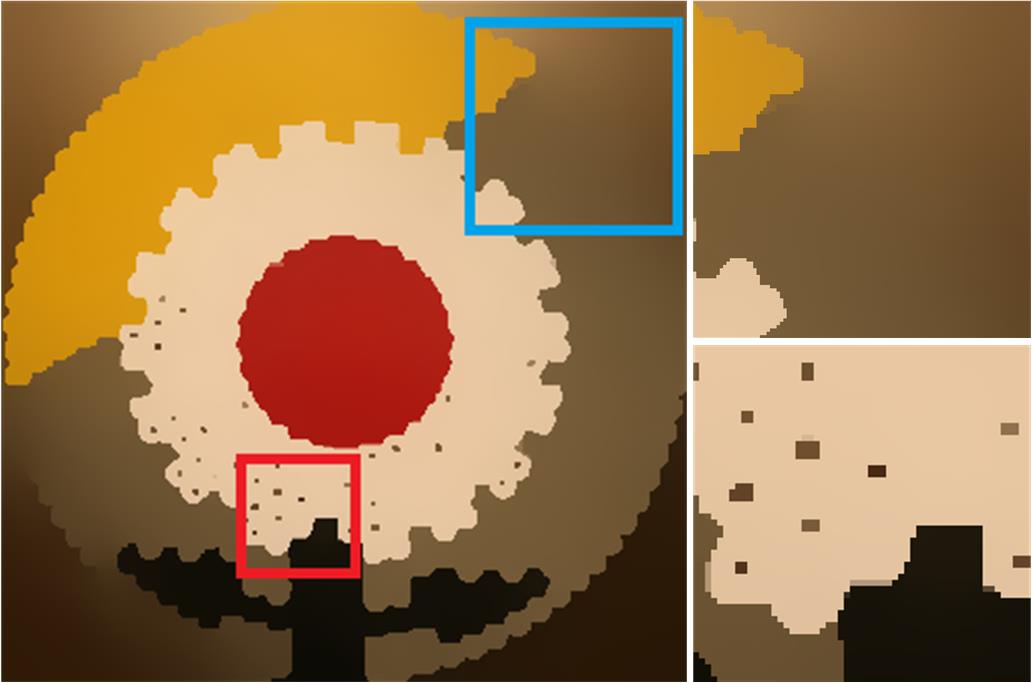} &
  \includegraphics[width=0.23\linewidth]{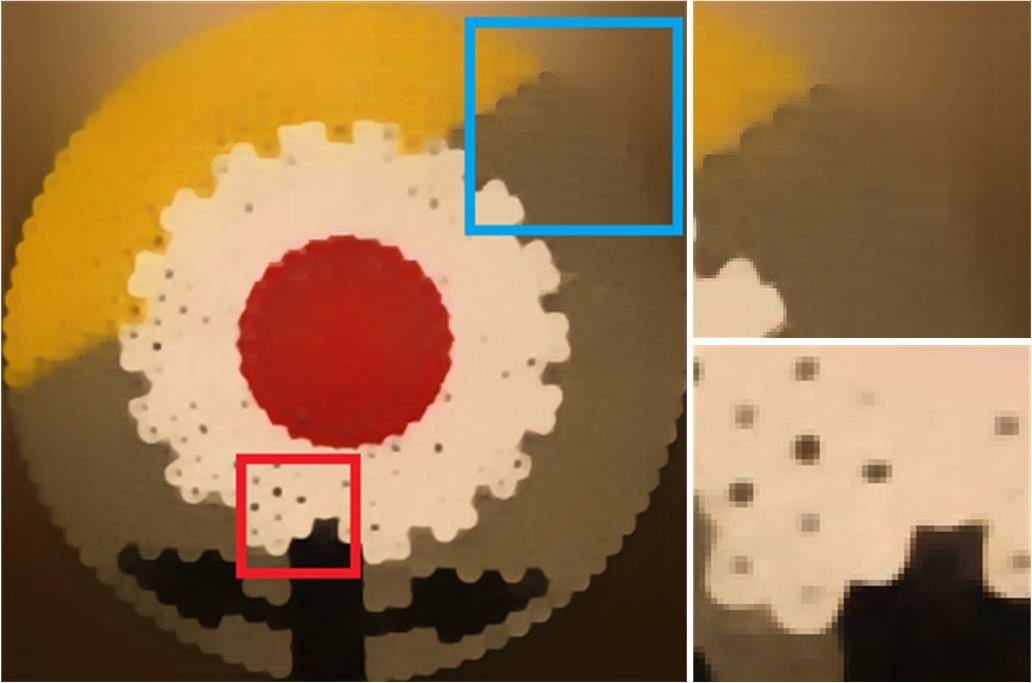} &
  \includegraphics[width=0.23\linewidth]{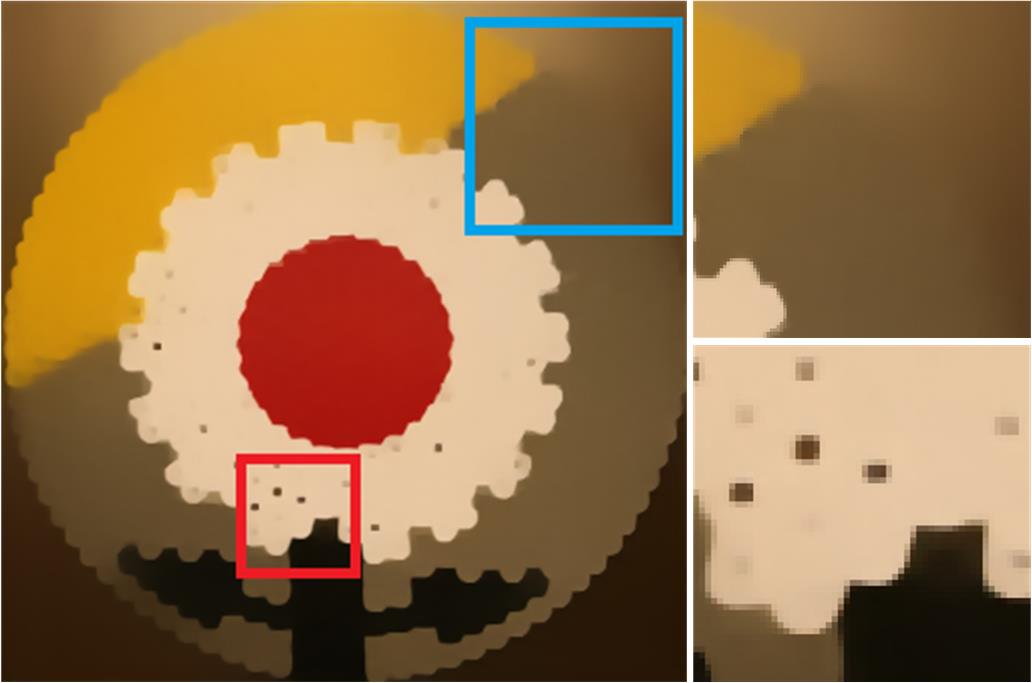} &
  \includegraphics[width=0.23\linewidth]{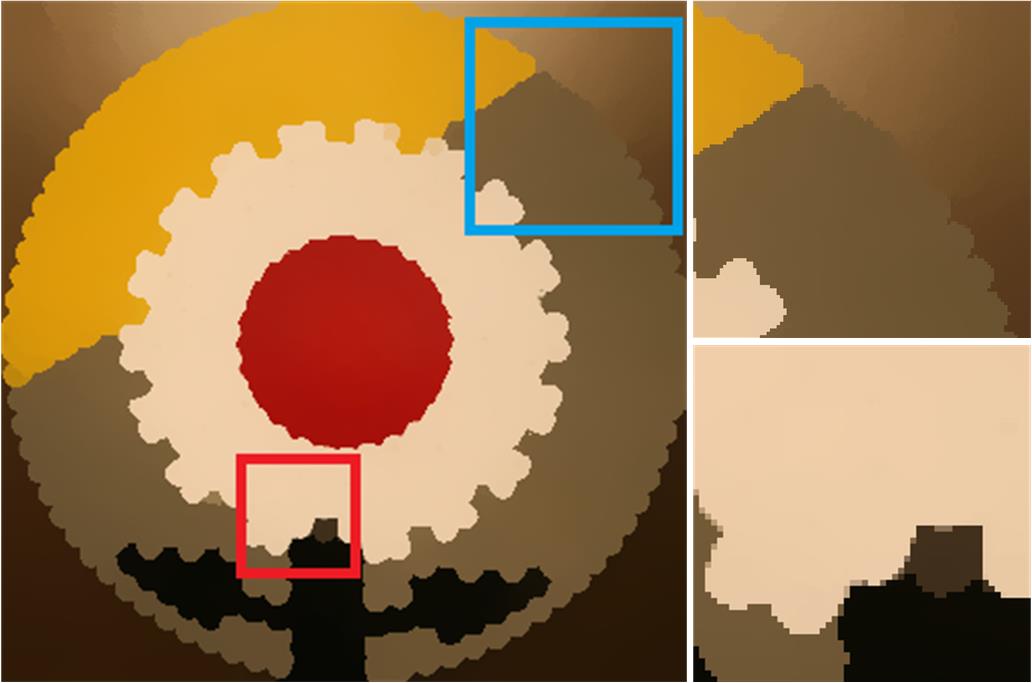} \\
   (e) muGIF & (f) FCN & (g) decouple learning & (h) ours(SP-1)\\

  \end{tabular}
  \caption{Image texture removal results of different methods. (a) Input image. Result of (b) JCAS \cite{gu2017joint} ($\lambda=0.05,\gamma=0.2$), (c) RTV \cite{xu2012structure} ($\lambda=0.015,\sigma=4$), (d) RGF \cite{zhang2014rolling} ($\sigma_r=0.075, \sigma_s=5, iter=5$), (e) muGIF \cite{guo2018mutually} ($\alpha_t=0.05, \alpha_r=0, N=10$), (f) FCN based approach \cite{chen2017fast}, (g) decouple learning \cite{fan2018decouple} and (f) our method of the SP-1 mode ($\lambda=1.25$).}\label{FigTextureSmooth}
\end{figure*}

Our method of the EP$\&$SP mode is applied to the guided depth map restoration. We first test our method on the simulated dataset constructed from the Middlebury dataset \cite{scharstein2007learning}. The simulated dataset contains six depth maps and four upsampling factors for each depth map, as listed in Tab.~\ref{TabToFSimulated}. We fix $r_d=r_s=5$ for all the upsampling factors, and $\lambda=0.1/0.25/0.5/0.95, b_d=b_d=0.1/0.1/0.08/0.07$ for $2\times/4\times/8\times/16\times$ upsampling. The other parameters are fixed as those in Tab.~\ref{TabParameter}. We compare our method against the state-of-the-art approaches including the very recently proposed deep learning based methods such as deep joint filtering (DJF) \cite{li2019joint} and deformable kernel network (DKN) \cite{kim2020deformable}. Note that these two approaches do not perform $2\times$ upsampling, the corresponding results are thus omitted in Tab.~\ref{TabToFSimulated}.  Fig.~\ref{FigToFSimulated} shows the visual comparison between our result and the results of the compared methods. Our method shows better performance in preserving sharp depth edges and avoiding texture copy artifacts, as illustrated in the highlighted regions of Fig.~\ref{FigToFSimulated}. Tab.~\ref{TabToFSimulated} also shows the quantitative evaluation of the results produced by different methods. Mean absolute error (MAE) between the upsampled depth map and the ground-truth one is adopted as the evaluation metric, which is also widely used in previous work \cite{guo2018mutually,li2016fast,liu2017semi,yang2014color}. As Tab.~\ref{TabToFSimulated} shows, our method can achieve the best or the second best performance among all the compared approaches.

We further validate our method on the real ToF data introduced by Ferstl et~al. \cite{ferstl2013image}. The real dataset contains three low-resolution depth maps captured by a ToF depth camera and the corresponding highly accurate ground-truth depth maps captured with structured light. The upsampling factor for the real dataset is $\sim6.25\times$. The parameters our method for this dataset are as follows: $r_d=r_s=3,b_d=b_d=0.08, \lambda=0.5$. The visual comparison in Fig.~\ref{FigToFReal} and the quantitative comparison in Tab.~\ref{TabToFReal} show that our method can outperform the compared methods and achieve state-of-the-art performance.

We also apply our method of the EP$\&$SP mode to flash/no flash image filtering. Images captured in low-light condition usually contain heavy noise. A high-quality image can be obtained with the help of flash. This flash image can be used as the guidance to smooth out the noise in the no-flash image. However, the flash can cause shadows in the image that do not exist in the no-flash image, as shown in the highlighted regions in Fig.~\ref{FigFlashNoFlash}(a). This structure inconsistency can lead to blurring edges/texture copy artifacts in the smoothed image if the smoothing procedure is not robust against the structure inconsistency. As the test images are collected from different datasets, the noise level and light condition are quite different. We thus only fix $r_d=r_s=1$ for different test images. The other parameters which are not fixed in Tab.~\ref{TabParameter} differs for different test images, and they are detailed in the figure caption. Fig.~\ref{FigFlashNoFlash} shows the visual comparison of the results produced by different approaches. The results of GF \cite{he2013guided}, deep image prior \cite{ulyanov2018deep} and the optimal scale map filter \cite{shen2015multispectral} suffer from blurring edges, as highlighted in Fig.~\ref{FigFlashNoFlash}(b), (c) and (d), respectively. The DJF \cite{li2019joint} can avoid these artifacts, however, the noise is less smoothed in its result, as shown in Fig.~\ref{FigFlashNoFlash}(e). The results of SD filter \cite{ham2018robust} in Fig.~\ref{FigFlashNoFlash}(f), mutually guided image filter (muGIF) \cite{guo2018mutually} in Fig.~\ref{FigFlashNoFlash}(g) and our result in Fig.~\ref{FigFlashNoFlash}(h) can properly preserve sharp edges and smooth the noise.

\begin{table*}
\centering
\caption{Running time (in seconds) of different methods for different image sizes. The value on the left of each cell is the running time for gray images and the right one is for color images. $r$ in our method refers to the $r_d$ and $r_s$ in Eq.~(\ref{EqObjFun}).}\label{TabTimeComp}
\resizebox{0.8\linewidth}{!}
{
\begin{tabular}{l|ccccc}
  \Xhline{1.2pt}
  & QVGA($320\times240$ ) & VGA($640\times480$)  & 720p($1280\times720$) & 1080p($1920\times1080$) & 2k($2048 \times 1080$) \\

  \hline

  FCN \cite{chen2017fast} (GPU, TenserFlow) & $\ast$ $|$ 0.016 & $\ast$ $|$ 0.19 & $\ast$ $|$ 0.47 & $\ast$ $|$ 0.87 & $\ast$ $|$ 1.12\\

  deep image prior \cite{ulyanov2018deep} (GPU, PyTorch) & $\ast$ $|$ 10.77 & $\ast$ $|$ 18.94 & $\ast$ $|$ 55.19 & $\ast$ $|$ 114.65 & $\ast$ $|$ 125.64\\

  decouple learning \cite{fan2018decouple} (GPU, PyTorch) & $\ast$ $|$ 0.0057 & $\ast$ $|$ 0.0109 & $\ast$ $|$ 0.0157 & $\ast$ $|$ 0.0204 & $\ast$ $|$ 0.0225\\

  DJF \cite{li2019joint} (GPU, MatConvNet) & 0.054 $|$ 0.094 & 0.097 $|$ 0.23 & 0.23 $|$ 0. 61 & 0.33 $|$ 0.85 & 0.35 $|$  0.91\\

  \hline
  AMF \cite{gastal2012adaptive} (CPU,  C++) & 0.011 $|$  0.028 & 0.043 $|$  0.11 & 0.12 $|$  0.29 & 0.28 $|$  0.68 & 0.30 $|$  0.71\\

  fast BLF \cite{paris2006fast}  (CPU,  C++)  & 0.0047 $|$  0.014 & 0.019 $|$  0.054 & 0.059 $|$  0.17 & 0.13 $|$  0.38 & 0.14 $|$  0.41\\

  GF \cite{he2013guided}  (CPU, C++)   & 0.0028 $|$  0.013 & 0.0079 $|$  0.058 & 0.026 $|$  0.16 & 0.064 $|$  0.35 & 0.066 $|$  0.37\\

  \hline

  $L_0$ norm \cite{xu2011image}  (CPU, MATLAB) & 0.12 $|$  0.33 & 0.47 $|$  1.37 & 1.53 $|$  4.40 & 3.59 $|$  10.75 & 4.09 $|$  11.72\\

  SD filter \cite{ham2018robust} (CPU, MATLAB) & 7.34 $|$ 8.95 & 25.35 $|$ 29.07 &	77.08 $|$	89.23 &	180.81 $|$ 211.67 & 197.73 $|$ 225.04\\

  RTV \cite{xu2012structure} (CPU, MATLAB) & 0.37 $|$ 0.73 & 0.97 $|$ 1.60 & 2.78 $|$ 5.17 & 6.75 $|$ 11.67 & 7.41 $|$ 12.82\\

  muGIF \cite{guo2018mutually}  (CPU, MATLAB) & 1.55 $|$ 4.23 & 5.62 $|$ 11.79 & 16.67 $|$ 32.84 & 40.01 $|$ 77.06 & 43.59 $|$ 84.42\\

  Ours (SP-1, CPU, MATLAB) & 1.03 $|$ 2.3 & 2.87 $|$ 5.70 & 9.89 $|$ 18.13 & 22.08 $|$ 40.53 & 23.91 $|$ 44.46\\

  Ours (SP-2, CPU, MATLAB) & 0.21 $|$ 0.43 & 0.45 $|$ 0.99 & 1.39 $|$ 2.96 & 3.24 $|$ 6.56 & 3.49 $|$ 7.07\\

  Ours (EP-1, CPU, MATLAB)/WLS \cite{farbman2008edge} &  0.22 $|$ 0.44 & 0.51 $|$ 1.04 & 1.49 $|$ 3.02 & 3.30 $|$ 6.65 & 3.56 $|$ 7.21\\

  Ours (EP-2, $r=1,s=1$, CPU, MATLAB) & 1.12 $|$ 2.34 & 3.14 $|$ 6.87 & 10.51 $|$ 19.82 & 23.53 $|$ 43.67 & 25.26 $|$ 47.06\\

  Ours (EP-2, $r=5,s=1$, CPU, MATLAB) & 8.63 $|$  15.18 & 33.01 $|$ 54.73 & 102.18$|$ 172.28 & 230.74 $|$  389.03 & 257.14 $|$  429.21\\
  Ours (EP-2, $r=5,s=2$, CPU, MATLAB) &  2.93 $|$ 4.98 & 12.42 $|$ 19.91 & 37.44 $|$ 59.86 & 83.91 $|$ 135.14 & 93.03 $|$ 147.75\\

  Ours (EP $\&$ SP, $r=1,s=1$, CPU, MATLAB) & 1.01 $|$ 2.21 & 2.86 $|$ 5.56 & 9.87 $|$ 18.04 & 22.32 $|$ 39.80 & 24.09 $|$ 43.67\\
  Ours (EP $\&$ SP, $r=5,s=1$, CPU, MATLAB) & 8.21 $|$ 14.92 & 32.61 $|$ 54.63 & 101.67 $|$ 174.83 & 229.64 $|$ 404.29 & 254.03 $|$ 445.42\\
  Ours (EP $\&$ SP, $r=5,s=2$, CPU, MATLAB) & 3.40 $|$ 6.39 & 14.91 $|$  24.68 & 44.71 $|$  78.60 & 99.56 $|$  179.18 & 108.91 $|$  198.77\\

  \Xhline{1.2pt}
\end{tabular}
}
\end{table*}

\subsection{Tasks in the Fourth Group}

We finally apply our method to image texture removal which belongs to the tasks in the fourth group. It aims at extracting salient meaningful structures while removing small complex texture patterns. Many meaningful structures can be formed by or appear over textured surfaces in natural images. Extracting these structures is challenging but is of great practical importance, which can benefit a number of applications, such as image vectorization, edge simplification and detection, content-aware image resizing \cite{avidan2007seam}, etc. The challenge of this task is that it requires structure-preserving smoothing rather than edge-preserving smoothing. Fig.~\ref{FigTextureSmooth}(a) shows a classical example of image texture removal: the small textures with strong edges should be smoothed out while the salient structures with weak edges should be preserved. The SP-1 mode is adopted in our method for this task. The value of $\lambda$ varies for different input images as the size of the structures to be removed can be quite different among different images. The other parameters are fixed as those in Tab.~\ref{TabParameter}.  Fig.~\ref{FigTextureSmooth} shows the results of the recent state-of-the-art approaches and ours. The joint convolutional analysis and synthesis sparse (JCAS) model \cite{gu2017joint} and rolling guidance filter (RGF) \cite{zhang2014rolling} can well remove the textures, but the resulting edges are also blurred. The relative total variation (RTV) method \cite{xu2012structure}, mutually guided image filtering (muGIF) \cite{guo2018mutually}, the deep learning approach based on fully-convolutional networks (FCN) \cite{chen2017fast} and decouple learning approach \cite{fan2018decouple} cannot completely remove the textures, in addition, the weak edges of the salient structures have also been smoothed out in their results. Our method can both preserve the weak edges of the salient structures and remove the small textures.

\subsection{Computation Efficiency Analysis}
\label{SecTimeAnalysis}

We also analyze the computation efficiency of our method. The test is performed on an i5 CPU with 32GB memory and a NVIDIA Titan RTX GPU. Both color images and gray images of 5 classical image resolutions are used for evaluation, which is detailed in Tab.~\ref{TabTimeComp}. We compare our method of different modes against some baselines including deep learning based methods \cite{chen2017fast,ulyanov2018deep,fan2018decouple,li2019joint} in the first group, local average based approaches \cite{gastal2012adaptive,paris2006fast,he2013guided} in the second group and global optimization based ones \cite{xu2011image,ham2018robust,xu2012structure,guo2018mutually} in the third group, as shown in Tab.~\ref{TabTimeComp}. All the deep learning based approaches are tested on the GPU while the rest compared methods are evaluated on the CPU.

Generally, most of the deep learning approaches and the local methods are much faster than ours and the other global methods. Our method of the EP-1 mode and the SP-2 mode also runs quite fast, which is the fastest among the compared global methods. When $r_d=1, r_s=1$ ($r=1$ in Tab.~\ref{TabTimeComp}), the computational cost of our method of the other modes (SP-1, EP-2 and EP$\&$SP) is also not quite high, and our method is slightly faster than muGIF \cite{guo2018mutually}. However, as the value of $r_d$ and $r_s$ increases, the computational cost of our method can increase greatly as shown in Tab.~\ref{TabTimeComp}. It drops when we adopt the dilated neighborhood with $s=2$. This can lead to a speedup of $\sim2.5\times$  to our method, which is now slightly faster than the SD filter \cite{ham2018robust}.

\section{Conclusion and Limitations}
\label{SecConclusion}

We propose a generalized framework for edge-preserving and structure-preserving image smoothing. We first introduce the truncated Huber penalty function which shows strong flexibility. Then a robust framework is presented. When combined with the flexibility of the truncated Huber penalty function, our framework is able to achieve different or even contradictive smoothing behaviors under different parameter settings. This is different from most previous approaches of which the inherent smoothing natures are usually fixed. Our method is also able to yield the smoothing behavior that is seldom achieved by previous approaches. It thus enables our method capable of more challenging cases which are not well handled by previous approaches. An efficient numerical solution to our model is proposed with the convergence theoretically guaranteed. We further provide a simple yet effective solution to reduce the computational cost of our method with the performance still maintained. The effectiveness of our method is demonstrated through comprehensive experimental results in a number of applications.

The limitations of our method are twofold. The first one is that our method has more parameters, which makes our method more complex than most existing approaches. This should be treated as the tradeoff between the complexity and the flexibility: it is these parameters that enable our model to enjoy the strong flexibility. In addition, we have fixed most of the parameters for different tasks as illustrated in Tab.~\ref{TabParameter}. The property analysis of our model in Sec.~\ref{SecPropertyAnalysis} can also work as the guideline for the choice of the rest un-fixed parameters. The second one is that our method is not time efficient as shown in Tab.~\ref{TabTimeComp}. Our method is thus not applicable to real-time image processing tasks.

\section*{Appendix A}
\label{SecAppendixA}
This appendix presents the proof of Eq.~(\ref{EqRelationWithHuberL0}) and Eq.~(\ref{EqTruncatedHuberMinCondition}) in Sec.~\ref{SecNumericalSolution}. Given the Huber penalty function $h(x)$ and truncated Huber penalty function $h_T(x)$ defined as Eq.~(\ref{EqHuber}) and Eq.~(\ref{EqTruncatedHuber}), then $h(x)$ and $h_T(x)$ are related through the following equation:

\begin{eqnarray}\label{EqHuberTruncatedHuberRelation}
{
  \begin{array}{c}
   h_T(x)=\min\limits_{y}g(x,y)\\
  \text{where} \ g(x,y)=h(x-y)+(b-\frac{a}{2})|y|_0,\ \text{s.t.} \ b\geq a,
  \end{array}
}
\end{eqnarray}
here $|y|_0$ is the $L_0$ norm of $y$: $|y|_0=0$ if $y=0$ and $|y|_0=1$ if $y\neq0$. The minimum of Eq.~(\ref{EqHuberTruncatedHuberRelation}) is achieved on the condition:

\begin{eqnarray}\label{EqTruncatedHuberCondition}
{
  y=\left\{\begin{array}{l}
   0, \ \ |x|\leq b\\
   x, \ \ |x|>b
  \end{array}\right..
}
\end{eqnarray}

\noindent\textbf{Proof:}

\noindent \textbf{Case 1: when $|x|<a$:}\\
\begin{itemize}
  \item when $y=0$, since $|x-0|<a$, we have:
\end{itemize}
\begin{equation}
  g(x,0) \triangleq g_1(x,y) = \frac{1}{2a}x^2,
\end{equation}
\begin{itemize}
  \item when $y\neq0$, we have:
\end{itemize}
\begin{footnotesize}
\begin{eqnarray}
{g(x,y)\triangleq\left\{
  \begin{array}{r}
  g_2(x,y)=\frac{1}{2a}(x-y)^2 + (b - \frac{a}{2}), \ \ |x-y|<a\\
  g_3(x,y)=|x-y| - \frac{a}{2} + (b - \frac{a}{2}), \ \ |x-y|\geq a\\
  \end{array}\right..
}
\end{eqnarray}
\end{footnotesize}
We have the following inequalities:
\begin{eqnarray}
\left\{
  \begin{array}{l}
     \ \ \ \ \ \ \ \ \ \ \ \ \ \  g_1(x,y)<\frac{a}{2}\\
     b - \frac{a}{2} \leq g_2(x,y) < b\\
     \ \ \ \ \ \ \ \ \ \ \ \ \ \  g_3(x,y)\geq b
  \end{array}\right.,
\end{eqnarray}
since $b\geq a$, then $b - \frac{a}{2}\geq \frac{a}{2}$, and thus $g_2(x,y)\geq b - \frac{a}{2}\geq\frac{a}{2}$. Finally, we have:
\begin{equation}
  g_1(x,y)<g_2(x,y)<g_3(x,y).
\end{equation}
Accordingly, the minimum of Eq.~(\ref{EqHuberTruncatedHuberRelation}) is $g_1(x,y)$ which is achieved when $y=0$:
\begin{equation}\label{EqCondition1}
  \min_{y}g(x,y)=g_1(x,y)=g(x,0)=\frac{1}{2a}x^2.
\end{equation}

\noindent \textbf{Case 2: when $a\leq|x|\leq b$:}\\
\begin{itemize}
  \item when $y=0$, since $|x-0|\geq a$, we have:
\end{itemize}
\begin{equation}
  g(x,0) \triangleq g_1(x,y) = |x| - \frac{a}{2},
\end{equation}
\begin{itemize}
  \item when $y\neq0$, we have:
\end{itemize}
\begin{footnotesize}
\begin{eqnarray}
{g(x,y)\triangleq\left\{
  \begin{array}{r}
  g_2(x,y)=\frac{1}{2a}(x-y)^2 + (b - \frac{a}{2}), \ \ |x-y|<a\\
  g_3(x,y)=|x-y| - \frac{a}{2} + (b - \frac{a}{2}), \ \ |x-y|\geq a\\
  \end{array}\right..
}
\end{eqnarray}
\end{footnotesize}
We have the following inequalities:
\begin{eqnarray}
\left\{
  \begin{array}{l}
     \ \ \ \ \ \ \frac{a}{2}\leq g_1(x,y)\leq b - \frac{a}{2}\\
     b - \frac{a}{2} \leq g_2(x,y) < b\\
     \ \ \ \ \ \ \ \ \ \ \ \ \ \  g_3(x,y)\geq b
  \end{array}\right.,
\end{eqnarray}
we then have:
\begin{equation}
  g_1(x,y)\leq g_2(x,y)<g_3(x,y).
\end{equation}
Accordingly, the minimum of Eq.~(\ref{EqHuberTruncatedHuberRelation}) is $g_1(x,y)$ which is achieved when $y=0$:
\begin{equation}\label{EqCondition2}
  \min_{y}g(x,y)=g_1(x,y)=g(x,0)=|x| - \frac{a}{2}.
\end{equation}

\noindent \textbf{Case 3: when $|x|> b$:}\\
\begin{itemize}
  \item when $y=0$, since $|x-0|>b$, we have:
\end{itemize}
\begin{equation}
  g(x,0) \triangleq g_1(x,y) = |x| - \frac{a}{2},
\end{equation}
\begin{itemize}
  \item when $y\neq0$, we have:
\end{itemize}
\begin{footnotesize}
\begin{eqnarray}
{g(x,y)\triangleq \left\{
  \begin{array}{r}
  g_2(x,y)=\frac{1}{2a}(x-y)^2 + (b - \frac{a}{2}), \ \ |x-y|<a\\
  g_3(x,y)=|x-y| - \frac{a}{2} + (b - \frac{a}{2}), \ \ |x-y|\geq a\\
  \end{array}\right..
}
\end{eqnarray}
\end{footnotesize}
We have the following inequalities:
\begin{eqnarray}\label{EqAppAInequality}
\left\{
  \begin{array}{l}
     \ \ \ \ \ \ \ \ \ \ \ \ \ \  g_1(x,y)> b - \frac{a}{2}\\
     b - \frac{a}{2} \leq g_2(x,y) < b\\
     \ \ \ \ \ \ \ \ \ \ \ \ \ \  g_3(x,y)\geq b
  \end{array}\right.,
\end{eqnarray}
from Eq.~(\ref{EqAppAInequality}) we can observe that the minimum value among $g_1(x,y), g_2(x,y)$ and $g_3(x,y)$ is $b - \frac{a}{2}$ which is achieved when $y=x$ in $g_2(x,y)$, i.e.:
\begin{equation}\label{EqCondition3}
  \min_{y}g(x,y)=g_2(x,y=x)=b - \frac{a}{2} .
\end{equation}

Finally, Eq.~(\ref{EqCondition1}), Eq.~(\ref{EqCondition2}) and Eq.~(\ref{EqCondition3}) consist with Eq.~(\ref{EqHuberTruncatedHuberRelation}) and the corresponding optimum conditions are equal to Eq.~(\ref{EqTruncatedHuberCondition}). By simply replacing $x$ with $\nabla^\ast_{i,j}$ and $y$ with $l^\ast_{i,j}$, we get Eq.~(\ref{EqRelationWithHuberL0}) and Eq.~(\ref{EqTruncatedHuberMinCondition}) in Sec.~\ref{SecNumericalSolution}.

\section*{Appendix B}

This appendix provides the proof of Eq.(\ref{EqMultHQ}) and Eq.~(\ref{EqMultHQCondition}) in Sec.~\ref{SecNumericalSolution}. The proof is based on the following theory in convex optimization \cite{boyd2004convex}: \emph{for a function $f(x)$, its conjugated function $g(y)$ is defined as $g(y)=\max\limits_x\{yx-f(x)\}$ which is convex. If $f(x)$ is continuous and convex, then we further have $f(x)=\max\limits_y\{yx-g(y)\}$.}

Given $h(x)$ defined as Eq.~(\ref{EqHuber}), it is clear that $h(x)$ is symmetric, i.e., $h(-x)=h(x)$. Thus, we only need to focus on its property for $x\geq0$. A new function $\theta(x)$ is first defined as:
\begin{eqnarray}
  \theta(x)=h(\sqrt{x}), \ x\geq0.
\end{eqnarray}
The function $-\theta(x)$ is convex. Defining its conjugated function as:
\begin{equation}\label{EqConjugateFun1}
  \psi(y)=\max_{x\geq0}\phi(x,y), \ \text{where}\ \phi(x,y)=(-y)x - (-\theta(x)),
\end{equation}
Then by using the property of the convexity of $-\theta(x)$, we have:
\begin{equation}\label{EqConjugateFun2}
  \theta(x)=\max_{y}\{(-y)x - \psi(y)\}=\min_{y}\{yx + \psi(y)\}
\end{equation}
Since we have $x\geq0$ in Eq.~(\ref{EqConjugateFun1}) and Eq.~(\ref{EqConjugateFun2}), then $x$ can be replaced with $x^2$. By using $\theta(x^2)=h(x)$, we have:
\begin{equation}
  h(x)=\min_{y}\{yx^2 + \psi(y)\}
\end{equation}
By setting $x=\nabla_{i,j}^\ast-l_{i,j}^\ast$ and $y=\mu^\ast_{i,j}$, we get Eq.~(\ref{EqMultHQ}) in Sec.~\ref{SecNumericalSolution}.

The optimum condition of $\psi(x,y)$ in Eq.~(\ref{EqConjugateFun1}) can be obtained by setting $\frac{\partial \psi(x,\hat{y})}{\partial x}=-\hat{y}+\theta'(x)=0$, i.e.:
\begin{equation}\label{EqConjugateFun3}
  y=\theta'(x)=\frac{h'(\sqrt{x})}{2\sqrt{x}}=\frac{h'(x)}{2x},
\end{equation}
By setting $x=\nabla_{i,j}^\ast-l_{i,j}^\ast$,  $y=\mu^\ast_{i,j}$ and substituting the expression of $h'(x)$ into Eq.~(\ref{EqConjugateFun3}), we get Eq.~(\ref{EqMultHQCondition}) in Sec.~\ref{SecNumericalSolution}.

\bibliographystyle{IEEEtran}
\bibliography{egbib}

\begin{IEEEbiography}[{\includegraphics[width=1in,height=1.25in,clip,keepaspectratio]{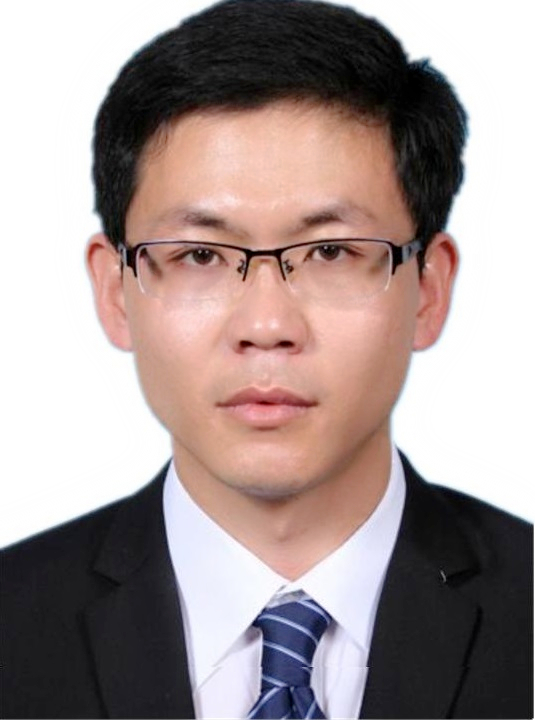}}]{Wei Liu}
received the B.S. degree in control science and engineering from Xi'an Jiaotong University, Xi'an, China, in 2012.  He received the Ph.D. degree in control science and engineering from Shanghai Jiao Tong University, Shanghai, China in 2019. He was a research fellow in The University of Adelaide from 2018 to 2021. He has been working as a research fellow in The University of Hong Kong since 2021. His current research areas include low-level computer vision and graphics, especially in the field of image filtering.
\end{IEEEbiography}
\vspace{-1em}

\begin{IEEEbiography}[{\includegraphics[width=1in,height=1.25in,clip,keepaspectratio]{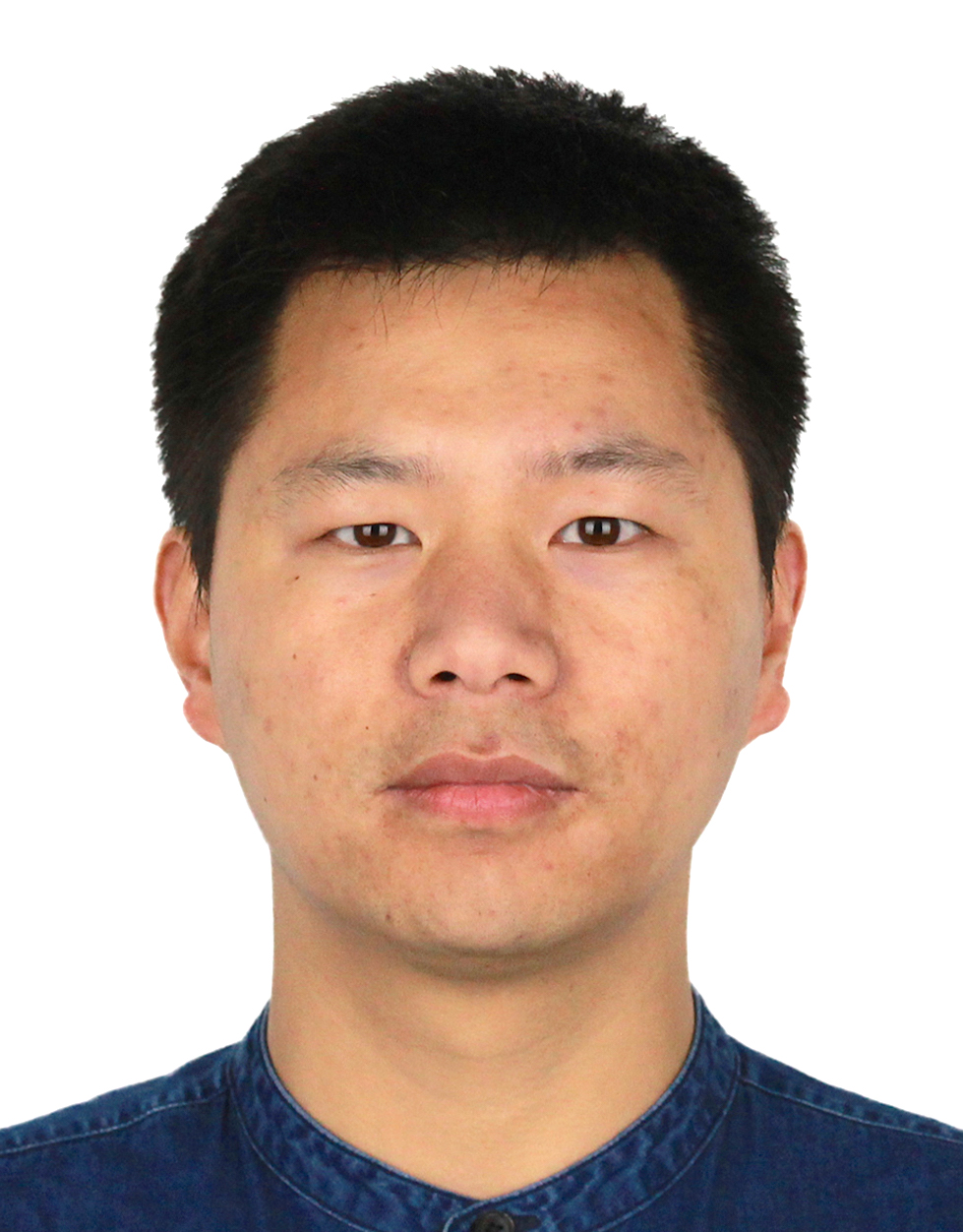}}]{Pingping Zhang}
received the B.E. degree in mathematics and applied mathematics from Henan Normal University (HNU), Xinxiang, China, in 2012, and the Ph.D. degree in signal and information processing from the Dalian University of Technology (DUT),  Dalian, China, in 2020. He is currently an Associate Professor with the School of Artificial Intelligence, DUT. His research interests include deep learning, saliency detection, object tracking, and semantic segmentation.
\end{IEEEbiography}
\vspace{-1em}

\begin{IEEEbiography}[{\includegraphics[width=1in,height=1.25in,clip,keepaspectratio]{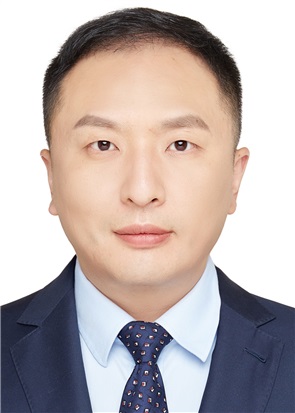}}]{Yinjie Lei}
(Member, IEEE) received the M.S. degree in image processing from Sichuan University (SCU), China, in 2009, and the Ph.D. degree in computer vision from The University of Western Australia (UWA), Australia, in 2013. Since 2017, he has been serving as the Vice Dean of the College of Electronics and Information Engineering, SCU, where he is currently an Associate Professor. He has authored over 60 journals/conference papers. His main research interests include deep learning, 3D vision, and semantic segmentation.
\end{IEEEbiography}
\vspace{-1em}

\begin{IEEEbiography}[{\includegraphics[width=1in,height=1.25in,clip,keepaspectratio]{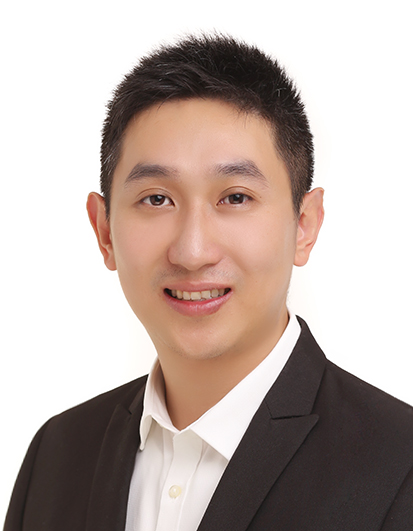}}]{Xiaolin Huang}
received the B.S. degree in control science and engineering, and the B.S. degree in applied mathematics from Xi'an Jiaotong University, Xi'an, China in 2006. In 2012, he received the Ph.D. degree in control science and engineering from Tsinghua University, Beijing, China. From 2012 to 2015, he worked as a postdoctoral researcher in ESAT-STADIUS, KU Leuven, Leuven, Belgium. After that he was selected as an Alexander von Humboldt Fellow and working in Pattern Recognition Lab, the Friedrich-Alexander-Universit\"{a}t Erlangen-N\"{u}rnberg, Erlangen, Germany. From 2016, he has been an Associate Professor at Institute of Image Processing and Pattern Recognition, Shanghai Jiao Tong University, Shanghai, China. In 2017, he was awarded by "1000-Talent Plan" (Young Program). His current research areas include machine learning and optimization.
\end{IEEEbiography}
\vspace{-1em}

\begin{IEEEbiography}[{\includegraphics[width=1in,height=1.25in,clip,keepaspectratio]{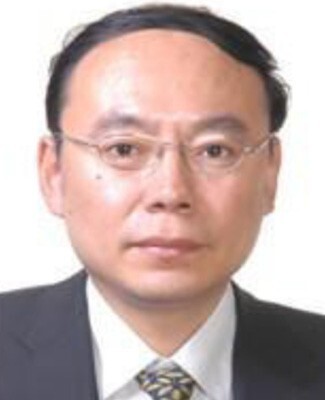}}]{Jie Yang}
received his Ph.D. from the Department of Computer Science, Hamburg University, Germany, in 1994. Currently, he is a professor at the Institute of Image Processing and Pattern Recognition, Shanghai Jiao Tong University, China. He has led
many research projects (e.g.,National Science Foundation, 863 National High Tech. Plan). His major research interests are object detection and recognition, data fusion and data mining, and medical image processing.
\end{IEEEbiography}

\begin{IEEEbiography}[{\includegraphics[width=1in,height=1.25in,clip,keepaspectratio]{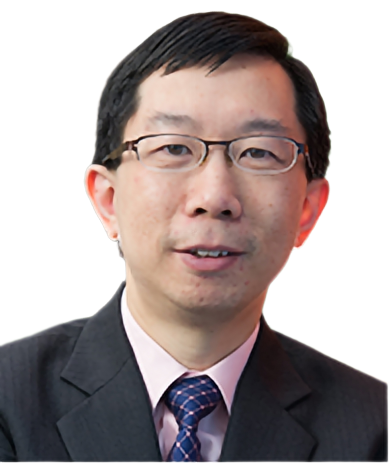}}]{Michael Ng} received the B.S. and M.Phil. degrees from the University of Hong Kong in 1990 and 1992, respectively, and the Ph.D. degree from the Chinese University of Hong Kong in 1995. He was a Chair Professor in Department of Mathematics at Hong Kong Baptist University from 2006 to 2019. He is currently a Chair Professor in Research Division of Mathematical and Statistical Science at The University of Hong Kong. His research interests include bioinformatics, image processing, scientific computing, and data mining. He was selected for the 2017 Class of Fellows of the Society for Industrial and Applied Mathematics. He obtained the Feng Kang Prize for his significant contributions in scientific computing. He serves on the Editorial Board members of several international journals.
\end{IEEEbiography}
\end{document}